\documentclass[PHD, final]{macro/neu_msthesis}

\usepackage[acronym]{glossaries}
\usepackage{acronym}
\newacronym{gnn}{GNN}{graphical neural network}
\newacronym{fg}{FG}{IEEE International Conference on Automatic Face and Gesture Recognition}

\newacronym{nue}{NU}{northeastern.edu}

\newacronym{ml}{ML}{machine learning}
\newacronym{sota}{SOTA}{state-of-the-art}

\newacronym{hog}{HOG}{histogram of gradients}
\newacronym{ss}{SS}{sister-sister}
\newacronym{bb}{BB}{brother-brother}
\newacronym{sibs}{SIBS}{brother-sister}

\newacronym{m}{M}{\textit{Male}}
\newacronym{f}{F}{\textit{Female}}
\newacronym{a}{A}{\textit{Asian}}
\newacronym{b}{B}{\textit{Black}}
\newacronym{i}{I}{\textit{Indian}}
\newacronym{w}{W}{\textit{White}}
\newacronym{af}{AF}{\textit{Asian}-\textit{Female}}
\newacronym{am}{AM}{\textit{Asian}-\textit{Male}}
\newacronym{bf}{BF}{\textit{Black}-\textit{Female}}
\newacronym{bm}{BM}{\textit{Black}-\textit{Male}}
\newacronym{if}{IF}{\textit{Indian}-\textit{Female}}
\newacronym{im}{IM}{\textit{Indian}-\textit{Male}}
\newacronym{wf}{WF}{\textit{White}-\textit{Female}}
\newacronym{wm}{WM}{\textit{White}-\textit{Male}}

\newacronym{fs}{FS}{father-son}
\newacronym{ms}{MS}{mother-son}
\newacronym{fd}{FD}{father-daughter}

\newacronym{md}{MD}{mother-daughter}
\newacronym{gfgs}{GFGS}{grandfather-grandson}
\newacronym{gmgs}{GMGS}{grandmother-grandson}
\newacronym{gfgd}{GFGD}{grandfather-granddaughter}

\newacronym{gmgd}{GMGD}{grandmother-granddaughter}

\newacronym{sdm}{SDM}{signal detection model}
\newacronym{roc}{ROC}{receiver operating characteristic}
\newacronym{nmse}{NMSE}{Normalized Mean Square Error}
\newacronym{mlp}{MLP}{multi-layered perception}

\newacronym{tp}{TP}{true-positive}
\newacronym{tn}{TN}{true-negative}
\newacronym{ap}{AP}{average precision}
\newacronym{ae}{AE}{autoencoder}

\newacronym{bce}{BCE}{Binary Cross Entropy}

\newacronym{fr}{FR}{face recognition}
\newacronym{fv}{FV}{facial verification}

\newacronym{tar}{TAR}{True Acceptance Rate}
\newacronym{far}{FAR}{False Acceptance Rate}
\newacronym{eer}{EER}{Equal Error Rate}
\newacronym{ta}{TA}{True Acceptance}
\newacronym{tnr}{TNR}{true-negative rate}
\newacronym{tpir}{TPIR}{true-positive identification rate}
\newacronym{frir}{FRIR}{false-reject identification rate}
\newacronym{fpir}{FRIR}{false-positive identification rate}
\newacronym{fp}{FP}{false-positive}
\newacronym{fn}{FN}{false-negative}
\newacronym{frr}{FRR}{false-reject rate}
\newacronym{fnr}{FNR}{false-negative rate}
\newacronym{fpr}{FPR}{false-positive rate}
\newacronym{tpr}{TPR}{true-positive rate}

\newacronym{det}{DET}{Detection error trade-off}

\newacronym{fiw}{FIW}{\textit{Families In the Wild}}
\newacronym{fiwmm}{FIW-MM}{\textit{FIW in Multimedia}}

\newacronym{tsk}{TSKIN}{\textit{Tri-Subject Kinship}}
\newacronym{kfw}{KinFaceW}{\textit{Kin-Faces in the Wild}}
\newacronym{kfvw}{KFVW}{\textit{KinFaceW Videos}}
\newacronym{rfiw}{RFIW}{\textit{Recognizing Families In the Wild}}
\newacronym{talkin}{TALKIN}{TALking KINship}

\newacronym{d}{$D$}{discriminator}
\newacronym{g}{$G$}{generator}
\newacronym{nn}{NN}{neural network}
\newacronym{cnn}{CNN}{Convolutional Neural Network}
\newacronym{caae}{CAEE}{conditional adversarial \gls{ae}}
\newacronym{gan}{GAN}{generative adversarial network}
\newacronym{dae}{DAE}{denoising \gls{ae}}

\newacronym{lut}{LUT}{Look-Up-Table}
\newacronym{mid}{MID}{Member ID}
\newacronym{fid}{FID}{Family ID}
\newacronym{pid}{PID}{Photo ID}

\newacronym{svm}{SVM}{Support Vector Machine}

\newacronym{nrml}{NRML}{Neighborhood Repulsed Metric Learning}

\newacronym{fsp}{FSP}{`From same photograph'}

\newacronym{tfidf}{TF-IDF}{term frequency-inverse document frequency learning}

\newacronym{dkmr}{DKMR}{Deep Kinship Matching and Recognition}

\newacronym{lflkin}{LFL-KIN}{Linear Feature Learning}
\newacronym{hsl}{HSL}{Heterogeneous Similarity Learning}

\newacronym{sml}{SML}{Support Vector Data Description-based metric learning}
\newacronym{msml}{MSML}{multi-view SML}

\newacronym{lm3l}{LM$^3$L}{large-margin multi-metric learning}

\newacronym{svdd}{SVDD}{Support Vector Data Description}

\newacronym{nlp}{NLP}{Natural Language Processing}
\newacronym{map}{MAP}{mean average precision}

\newacronym{aflw}{AFLW}{Annotated Facial Landmarks in the Wild}

\newacronym{mtcnn}{MTCNN}{\emph{multi-task \gls{cnn}}}

\newacronym{mm}{MM}{multimedia}

\newacronym{pca}{PCA}{Principle Component Analysis}
\newacronym{lbp}{PCA}{Local Binary Patterns}
\newacronym{lbph}{PCA}{Local Binary Patterns Histograms}

\newacronym{se}{SE}{\emph{Squeeze-and-Excitation}}

\newacronym{dbvae}{DB-VAE}{Debiasing Variational Autoencoder}
\newacronym{rnn}{RNN}{recurrent neural network}
\newacronym{soa}{SOTA}{state-of-the-art}

\newacronym{lfw}{LFW}{Labeled Faces in the Wild}
\newacronym{bfw}{BFW}{Balanced Faces In the Wild}
\newacronym{rfw}{RFW}{Racial Faces in-the-Wild:}
\newacronym{dp}{DemogPairs}{Demographic Pairs}
\newacronym{itwcc}{ITWCC}{Wild Child Celebrity}

\newacronym{ge2e}{GE2E}{generalised end-to-end}
\newacronym{cmc}{CMC}{Cumulative Matching Characteristic}

\newacronym{nist}{NIST}{National Institute of Standards and Technology}

\newacronym{bbox}{BB}{bounding box}
\newacronym{cs}{CS}{Cosine Similarity}

\newacronym{lmcl}{LMCL}{Large Margin Cosine Loss}

\newacronym{lime}{LIME}{Local Interpretable Model-Agnostic Explanations}
\newacronym{nas}{NAS}{Neural Architecture Search}
\newacronym{gapf}{GAPF}{Generative Adversarial Privacy and Fairness}
\newacronym{vid}{VID}{Video ID}  

\usepackage{pdflscape}
\usepackage{enumerate}

\usepackage{graphicx}
\graphicspath{{fig/}}
\usepackage{amsmath}
\usepackage{amssymb}
\usepackage{tabularx, booktabs, ragged2e}

\usepackage{multicol}
\usepackage{multirow}
\usepackage{adjustbox}
\usepackage{rotating}
\usepackage{caption}
\usepackage{subcaption}

\usepackage{ltablex, enumitem, makecell, float, textcomp}
\usepackage{array,xcolor,colortbl}
\usepackage{comment}
\usepackage{xspace}
\usepackage{pifont}
\usepackage{tikz}
\usetikzlibrary{calc,shapes.callouts,shapes.arrows}

\newcounter{mycallout}

\usepackage[ruled,vlined]{algorithm2e}

\SetCommentSty{mycommfont}

\usepackage{color,soul}

\usepackage[utf8]{inputenc}
\usepackage{bbm}
\usepackage{subfloat}
\usepackage{epigraph}

\newcommand{\etc}{etc.\@\xspace}
\newcommand{\etal}{\textit{et al}. }
\newcommand{\ie}{\textit{i}.\textit{e}., }
\newcommand{\eg}{\textit{e}.\textit{g}., }

\definecolor{ao(english)}{rgb}{0.0, 0.5, 0.0}
\newcommand{\image}{\mathbf{d}}

\newcommand{\lkl}{\textrm{LaplaceKL}}
\newcommand{\sam}{\textrm{softargmax}}
\newcommand{\Sam}{\textrm{Softargmax}}

\newcommand{\ra}[1]{\renewcommand{\arraystretch}{#1}}

\newcommand{\xmark}{\ding{56}}%
\newcommand{\checkc}{\ding{51}}%

\usepackage{amsfonts,amssymb,amsmath}			
\usepackage{times}		
\usepackage{bm}

\newcommand{\figref}[1]{Figure~\ref{#1}}
\newcommand{\tabref}[1]{Table~\ref{#1}}
\newcommand{\chapref}[1]{Chapter~\ref{#1}}
\newcommand{\secref}[1]{Section~\ref{#1}}

\newcommand{\ifno}[1]{}

\usepackage{multirow}
\makeindex
\usepackage{makeidx}
\usepackage{acronym}

\usepackage{url}
\ifnum\pdfoutput>0
\usepackage[pdftex,
bookmarks=true,
bookmarksnumbered=true,
hypertexnames=false,
breaklinks=true,           
citecolor=blue,     
filecolor=magenta,          
urlcolor=cyan
]{hyperref}
\else
\usepackage[hypertex,
bookmarks=true,
bookmarksnumbered=true,
hypertexnames=false,
breaklinks=true,           
citecolor=green,     
filecolor=magenta,          
urlcolor=cyan
]{hyperref}
\fi

\hypersetup{				
pdfauthor = {\authorRef},
pdftitle = {\titleRef},
pdfsubject = {\expandafter{\degreeRef} thesis submitted to Northeastern University},
pdfkeywords = {add keywords here}
pdfcreator = {LaTeX with hyperref package},
colorlinks=true,
linkcolor=red,
filecolor=magenta,      
urlcolor=blue,
citecolor=green,
pdftitle={Sharelatex Example},
pdfpagemode=FullScreen,
pdfproducer = {dvips + ps2pdf}}

\usepackage{microtype}

\definecolor{mygray2}{gray}{.8}
\usepackage{amsthm}
\newtheorem{lemma}{Lemma}

\usepackage{tcolorbox}

\definecolor{Gray}{gray}{0.85}
\definecolor{LightCyan}{rgb}{0.88,1,1}

\newcolumntype{a}{>{\columncolor{Gray}}c}
\renewcommand{\arraystretch}{1.0}

\usepackage{amssymb}

\let\emptyset\varnothing

\title{Automatic Face Understanding: Recognizing Families in Photos}

\author{Joseph P Robinson}

\nuid{000549887}

\dept{Electrical and Computer Engineering}

\degree{"Doctor of Philosophy"}

\degreename{Computer Engineering}

\field{Computer Engineering}

\submitdate{\today}

\numberofmembers{3}
\principaladviser{Dr. Yun Fu}
\firstreader{Dr. First Member}
\secondreader{Dr. Octavia Camps}
\thirdreader{Dr. Sarah O}
\chairman{Dr. Miriam Leeser}
\dean{Dr. Thomas C. Sheahan}

\begin{document}

\pdfbookmark[1]{Cover}{cover}

\titlepage

\begin{frontmatter}


\begin{dedication}
To my mom, who has always shown me the world as a world of opportunities.
\end{dedication}

\hypersetup{linkcolor=black}

\pdfbookmark[1]{Table of Contents}{contents}
\tableofcontents
\listoffigures
\newpage\ssp
\listoftables
\hypersetup{linkcolor=red}



\chapter*{List of Acronyms}
\addcontentsline{toc}{chapter}{List of Acronyms}

\begin{acronym}
\acro{AE}{Auto Encoder}.
A model consisting of an encoder and a decoder module, where the encoder projects input signal to a latent, hidden state and the decoder reconstructs the original input.
\acro{AP}{Average Precision}.
A measure of the number of true-positives per total number of positives.
\acro{BB}{Brother-brother}.
Relationship type (pairwise).
\acro{BFW}{Balanced Faces in the Wild}.
 A facial recognition dataset for measuring bias across different demographics (\ie ethnicity and gender).
\acro{CMC}{Cumulative matching characteristic [curve]}.
\acro{DL}{Deep learning}.
\acro{DET}{Detection error trade-off [curve]}.
\acro{FAR}{False-acceptance rate}.
\acro{FD}{Father-daughter}.
Relationship type (pairwise).
\acro{FID}{Family ID}.
Unique identifier assigned to each family of FIW.
\acro{FIW}{Families In the Wild}.
 A large-scale dataset for recognizing family members in photos.
\acro{FIW-MM}{Families In the Wild in Multimedia}.
 A large-scale dataset for recognizing family members in multimedia (\ie photos, video, audio, and text transcripts).
\acro{FMD}{father/mother-daughter}.
Relationship type (triplet).
\acro{FMS}{father/mother-son}.
Relationship type (triplet).
\acro{FR}{Facial Recognition}.
 Machinery that recognizes identities from facial imagery or videos.
\acro{FS}{Father-son}.
Relationship type (pairwise).
\acro{GFGD}{Grandfather - granddaughter}.
Relationship type (pairwise).
\acro{GFGS}{Grandfather - grandson}.
Relationship type (pairwise).
\acro{GGFGGD}{Great GFGD}.
Relationship type (pairwise).
\acro{GGFGGS}{Great GFGS}.
Relationship type (pairwise).
\acro{GMGD}{Grandmother - granddaughter}.
Relationship type (pairwise).
\acro{GMGS}{Grandmother - grandson}.
Relationship type (pairwise).
\acro{GGMGGD}{Great GMGD}.
Relationship type (pairwise).
\acro{GGMGGS}{Great GMGS}.
Relationship type (pairwise).
\acro{MAP}{Mean Average Precision}.
The average AP across various instances.
\acro{MD}{Mother-daughter}.
Relationship type (pairwise).
\acro{MID}{Member ID}.
Unique identifier assigned to each family member of FIW.
\acro{MS}{Mother-son}.
Relationship type (pairwise).
\acro{PID}{Picture ID}.
Unique identifier assigned to each photo of FIW.
\acro{RFIW}{Recognizing FIW}.
Annual data challenge for recognizing kinship in visual media.
\acro{ROC}{Receiver operating characteristic [curve]}.
\acro{SAM}{Soft-argmax}.
2D softmax function.
\acro{SIBS}{Brother-sister}.
Relationship type (pairwise).
\acro{SS}{Sister-sister}.
Relationship type (pairwise).
\acro{TAR}{True-acceptance rate}.
\acro{VID}{Video ID}.
Unique identifier assigned to each video of FIW-MM.
\end{acronym}


\begin{acknowledgements}
    With the highest regards, and the upmost appreciation, are the incredible people in my life. 
    My life has been blessed with diverse network of brilliant, passionate, and sincere people from all over the globe. A sub-population of which are also a part of my professional network; with many more personal. Let me now take a moment for me to express the deep appreciation I have for all of those that have had an impact on my life as far as the pages of this thesis goes. Hence, I owe many thanks to many that helped with completion of my dissertation, for they had involvement that directly relates to the success of this thesis. Whether they realized it or not, I would not be in the position I currently find myself as I prepare this report in exchange for a PhD in Computer Engineering.

\section*{MENTORS}
During this lifetime, there have been many who inspired, influenced, and impressed me. However, few, if any, come close to matching the level of impact on me as my PhD advisor, Dr. Yun Raymond Fu. For starters, he convinced me to get a PhD, helped get me admitted and registered past deadlines (\ie allowed since an undergraduate husky with the support of my advisor to be).  Now nearly five-years later, and with the knowledge acquired from this decision, there succeeded an extent never imagined.  That was hope for a prospective graduate student to gain research experience from Raymond, renowned in research communities and proven very clearly why.  Others and I learn under his remarkable insights and knowledge of research.  He enables us at high and low-levels: the high-level being the motivation of research, while the low-level is the technical novelty in the mathematical models and algorithms. Nearly every week for the past 5 years, Raymond set his sight in the lab on each PhD (\ie an average of over ten students at any given time) allowing us to learn from each other and provide time to review progress. Raymond’s expertise generally just knows the best route to take (\eg the best conference to target for paper submission, the proper way to reach out to other research groups to inquire about a need of ours, even who we should try to speak with at an upcoming conference). At the same time, Raymond trusts us, so he often reaches to one of his students for the most accurate responses and feedback. A big emphasis is put on humility and aggressiveness such as teaching us to work for what we want, appreciate it when there, but readily move onward to the next task.  Beyond research, Raymond remains an incredible mentor. Ask anyone in our department– Raymond is a tough advisor with a high expectation. Over the years, I have come to realize the amount of extra effort that is for him (\ie it is much easier and quicker to say little, but Raymond very rarely cuts it cheaply for us). Week-in, week-out, Raymond devotes great efforts into us: pushing us to be the best possible by recognizing our strengths and weaknesses to help us leverage one while improving the other. He is a trustworthy mentor, and a great friend. I have learned so many different facets from him (\ie research, professional, social). I witnessed countless alumni go through the process under his continuous guidance, advice, effort, patience, and encouragement.


I would also like to thank the rest of my PhD dissertation committee, Professor Sarah Ostadabbas and Professor Octavia Camps. Both provided constant support in preparing and delivering my dissertation, while getting opportunities to get to know them through independent means. Specifically, the first time I was exposed to a computer vision topic it was per one of Professor Camp's courses during undergrad. From the get go, Professor Camps was completely approachable. Furthermore, her knowledge in machine vision had inspired me to push myself to learn all that is possible. Along with several vision and related course, Professor Camps also oversaw the oral portion of my qualifying exam completed second year of graduate. Furthermore, at the start and until today I regularly see Professor Camps at vision conferences-- it is always a pleasure to catch up with Professor Camps, whether we are on campus or a venue in South Korea. There is never a time she waves me away: I appreciate deeply her expressing interest in my research and status as a graduate student, for she will always check in and have discussions when we bump in to one another. The same for Professor Ostadabbas: another faculty active at conferences and events. Sara Ostadabbas always shares useful bits of information, whether through formal or informal discussions-- for instance, during a conference workshop we both chaired, we were sitting together up front taking turns introducing the next speaker, at which time Professor Ostadabbas kept elaborating on several points she raised during the PhD dissertation proposal presentation I had recently delivered. This really meant a lot: several weeks after my proposal, and Professor Ostadabbas still could recall specifics to further elaborate on advise she had given. Along with a good memory, she clearly cared. Both Professor Camps and Professor Ostadabbas clearly have exceptional amounts of care for us students, and the studies as a whole. The two of them alongside my advisor Raymond made for an all-star PhD committee that really made a difference for me in the end.

Professor Ming Shao, although not a part of my PhD committee, has been an outstanding mentor of mine for many years. I was fortunate that the year I joined the lab was about the year Professor Shao was graduating: when we started working together he was a senior student of SMILE. Thankfully, he remained in academia as a professor at University of Dartmouth. Ming and I have collaborated on many works and efforts: from the first kinship paper on FIW to the most recent challenge. Especially in cases the task is new to me (\eg first journal rebuttal or organized workshop), Ming has been outstanding at teaching me to understand (\ie not to just complete this time, but to get every time). Lucky for me: Ming has grown more of a friend than a colleague.

\section*{COLLABORATORS AND COLLEAGUES}
I would like to thank all the members of the SMILE Lab whom I have had the pleasure of working directly with, such as Dr. Handong Zhao, Yu Yin, Can Qin, Yulun Zhang, Professor Sheng Li, Professor Zhengming Ding, Lichen Wang, Kunpeng Li, and Zaid Khan-- many of whom I had great moments traveling with on behalf of the work. Also, the many other members of SMILE Lab for which I spent endless hours alongside working independently like Dr. Kang Li, Dr. Chengcheng Jia, Professor Hongfu Liu, Dr. Shuyang Wang, Dr. Shuhui Jiang, Dr. Jun Li, Professor Yu Kong, Dr. Zhiqiang Tao, Songyao Jiang, Bin Sun, Professor Yi Tian, Professor Qianqian Wang, Gan Sun, Kunpeng Li, Kai Li,  Huixian Zhang, Zhenglun Kong, and Chang Liu. Grouped together by Dr. Fu into SMILE Lab, this hard-working, highly-expecting individuals, yielded a closely knit, synergistic nature in an inspiring environment and allowed many to successfully prosper.  

I was extremely lucky to get matched up with kind and intelligent people bosses on interns-- mentors that taught me vast types of knowledge. Acknowledging the most recent, graduate-level advisors: Samson Timenor (ISM Connect), Sergey Tulyakov (Snap Inc.), Jeffrey Byrne (STR). Having interned with ISMConnect as a senor grad student, the time could not have been more perfect for the many lessons learned from Samson: beyond technical, and full of life-long concepts and snippets of wisdom to better shape me for the professional (and personal) life to come. While at Snapchat, Sergey and I spent lots of time together to draft a paper, end-to-end, in the limited time of summer. Sergey had endless lessons and technical critique that helped form my way of thinking as research scientists (\ie every experiment needs hypothesis, should be well thought out and with consideration to variables under investigation, and pre-notions for the next steps whether a null or alternative hypothesis results). As simple of a concept, and as obvious as it seems when speaking of, us researchers often find ourselves overwhelmed in thoughts and with a medley of ideas that it is not uncommon to windup saturated, where remaining grounded yields higher quality output of our work-- the emphasis Sergey put on this, whether he realized it or not, will forever remain at the forefront of my thought process as a professional and, sometimes, even beyond. Finally, Jeffrey at STR took me on as an intern for consecutive summers my earliest years of graduate school. Even with many projects being restricted to me as a student (\ie government classified), Jeffrey was able to carve out meaningful projects based on technical concepts that are regularly found useful today (\eg algorithms to sort and search, along with entire topics such as clustering, adversarial ML, advanced code development and API design in C++ and Python, and more). Furthermore, Jeffrey included me in many meetings-- the lessons learned here were exceptional: the ways Jeffrey hosted, conducted, and led team meetings in ways that motivated, inspired, and included everyone. Jeffrey's tendency to build strong teams-- the language used with those who are the best in the respective topic at hand; the organization and communication that tied everything together-- it was later clear to me how Jeffrey led the top team in many competitive government programs (\eg JANUS).

Even before the PhD, but still at Northeastern as an undergraduate, there had been countless scenarios that helped shape me as a whole. Too many to accurately pin-point every instance in a quick writing piece. Nonetheless, few had such outstanding contributions to me as a person that lessons and other impressions from them will likely continue to propagate for many more years to come. Specifically, those that had a direct hand in helping me find and establish myself both personally and professionally had me realizing I could not be more fortunate in my network-- Dr. Charles DiMarzio (and his wife Sheila and friend Maureen) by welcoming me into his lab, taking on the role as a mentor and friend, while providing guidance to my young professional-self; Kristin Hicks, having been the very first person I had contact with from Northeastern, with a phone-call welcoming me to a summer research program as a visiting community college student, which winded up being the first of many opportunities with Kristin (including graduate school fellowship); Dr. David Kaeli, with his advice that helped me transfer to NEU, and then later land my first co-op at Analogic; Dr. Richard Harris, and just being an incredible role model for me to watch, learn from, and often talk to (there would be nights in his office after we organized a workshop with him on his computer guiding me in more ways than he probably even imagined), and providing me the opportunity to grow as a public speaker. There are so many others that I hesitated in naming any. However, the ones mentioned have been there from start, and are still there today. Just incredible and so fortunate. 
\section*{FAMILY AND FRIENDS}

My mother has had such profound impact on me as a whole; had I not been blessed enough to have my mother chances of me aiming for the highest sights, while remaining as happy and healthy as possible, would likely not be at the levels for which it is. Whether it be encouraging me when down, challenging me at times of comfort, or teaching me at times of growth, my mother has always been there-- as the gym teacher when in kindergarten, the driver for doctor's appointments well before the age I could drive, or one of the many times moving while furthering my education-- my mother has always there, she always showed up, and she has always loved myself and siblings. Words cannot express to what extent having a mother like mine has had on me as a person-- the explanation is infinite, while words are discrete, so I would argue it is impossible for me to express the true extent for which this is meant (physically, emotionally, spiritually, and many other facets encompassed in the nature of my very existence). Beyond my mother, Lisa Robinson, taking on the role of \emph{super mom}, she has also been the greatest teacher of mine in the broad subject of \emph{Life}. Not only has she helped me financially when in need, but she was also there in spirit when I needed a boost; on the other hand, she would also be the first to take me down a peg or further challenge me at times of over comfort. It is almost like she indirectly trained me for Dr. Fu, for Dr. Fu would often remind us to remain ``humble, but aggressive.'' Looking back at the big picture, such a mindset needed to live by these words lie at foundation of many of the lessons and experiences posed by mother. 

Alongside my mother were my siblings Stacey (Robinson) McGuire, Thomas Robinson, Brendan Crocker, and Briar Crocker, whom are also amongst my best friends. Also, my fathers Peter Robinson and Paul Crocker, my Aunt Theresa Robinson, my cousins John Robinson and Lorri Robinson, and my grandparents Tom and Patricia Floramo who helped make this dream a reality. Also, my girlfriend through the completion of my dissertation and beyond, Briana, who has been by my side through the entire PhD process: dealt with me being consumed by deadlines, and been there as my support at moments of feeling overwhelmed or burned out.

Many family members and friends have supported me by peer reviewing papers coherency and language-- my brother Thomas Robinson and my friends Laura Rose, Maureen Hawe, and Bruce Collins. All of whom taught me indirectly by providing feedback on my writing pieces. Especially towards the end, when Laura Rose went over and above helping me polish up language in the final papers published as a graduate student, including this work (\ie this thesis, end-to-end).

Study buddies friends who I spent countless hours with helped motivate and push me to higher limits: Jordan Kiellach, Brian Toner, Juan Ramirez, Ryan Snyder, Joshua Mcdougall, Robert Watson, and many, many more.        
Last, but certainly not least, are those that supported me via letters of recommendation and as references, both personal and professional, for one or more of the many endeavors that led me to today: Bonnie-Jeanne Toner, Richard Harrison, Chris McQuire, Samson Timoner, Yun (Raymond) Fu, David R. Kaeli, Charles A. DiMarzio, Kristin Hicks, Paul Chandley, Clair Duggan, Nancy Nickerson, and certainly others over many years of doing co-ops, research, and other rich experiences made possible as a member of the NEU community. Additionally, are those who helped me move many times between undergraduate and graduate tracks at NEU: Paul Crocker, Lisa Robinson, Bruce Collins, Theresa Robinson, and John Robinson.

I thank you all for your love, patience, and support. I must acknowledge it, for it was by you the seemingly impossible goal of yesterday (\ie a doctor of engineering) has become the reality of today. I hope to return the favor in years of research to come, \ie let me apply the many years of training and life experiences to making the world a better place come tomorrow.
\end{acknowledgements}


\begin{abstract}
Visual kinship recognition has an abundance of practical uses. For this, we built the largest database for kinship recognition, \gls{fiw}. Built entirely in-house with no cost using a semi-automatic labeling scheme. Specifically, we first aligned faces detected in family photos with names in the corresponding text metadata to mine the label proposals with high confidence. The remaining data were labeled using a novel clustering algorithm that used label proposals as side information to guide more accurate clusters. Great savings in time and human input was had. Statistically, \gls{fiw} shows enormous gains over its predecessors. We have several benchmarks in kinship verification, family classification, tri-subject verification, and large-scale search \& retrieval. We also trained CNNs on \gls{fiw} and deployed the model on the renowned KinWild I and II to gain \gls{soa}. Most recently, we further augmented \gls{fiw} with multimedia (MM) for 200 of its 1,000 families- a labeled collection we dubbed FIW-MM. Now, video dynamics, audio, and text captions can be used in the decision making of kinship recognition systems.

\gls{fiw} continues to pave the way for this research track: (1) advanced \gls{soa} (\eg marginalized denoising auto-encoder based on metric learning that preserves intrinsic structures of kin-data and encapsulates discriminating information in learned features); (2) introduced generative models to predict a child’s appearance from a parent pair (\ie proposed an adversarial autoencoder conditioned on age and gender to map between facial appearance and these higher-level features for control of age and gender); (3) designed evaluations with benchmarks to support challenges, workshops, and tutorials at top tier conferences (\eg CVPR, MM, FG, ICME), and a premiere Kaggle Competition. We expect \gls{fiw} will significantly impact research and reality.

Additionally, we tackled the classic problem of facial landmark localization in images. This is a task that has been in focus for decades, and many solutions have been proposed. However, there are revamped interests in pushing facial landmark detection technologies to handle more challenging data with deep networks now prevailing throughout machine learning. A majority of these networks have objectives based on L1 or L2 norms, which inherit several disadvantages. First of all, the locations of landmarks are determined from generated heatmaps (\ie confidence maps) from which predicted landmark locations (\ie the means) get penalized without accounting for the spread: a high scatter corresponds to low confidence and vice-versa. To address this, we introduced a LaplaceKL objective that penalizes for low confidence. Another issue is a dependency on labeled data, which is expensive to collect and susceptible to error. We addressed both issues by proposing an adversarial training framework that leverages unlabeled data to improve model performance. Our method claims \gls{soa} on renowned benchmarks. Furthermore, our model is robust with a reduced size: 1/8 the number of channels (\ie 0.0398 MB) is comparable to state-of-that-art in real-time on a CPU. Thus, our method is of high practical value to real-life applications.

Finally, we built the Balanced Faces in the Wild (BFW) dataset to serve as a proxy to measure bias across ethnicity and gender subgroups, allowing us to characterize FR performances per subgroup. We show performances are non-optimal when a single score threshold is used to determine whether sample pairs are genuine or imposter. Furthermore, actual performance ratings vary greatly from the reported across subgroups. Thus, claims of specific error rates only hold for populations matching that of the validation data. We mitigate the imbalanced performances using a novel domain adaptation learning scheme on the facial encodings extracted using SOTA deep nets. Not only does this technique balance performance, but it also boosts the overall performance. A benefit of the proposed is to preserve identity information in facial features while removing demographic knowledge in the lower dimensional features. The removal of demographic knowledge prevents future potential biases from being injected into decision making. Additionally, privacy concerns are satisfied by this removal. We explore why this works qualitatively with hard samples. We also show quantitatively that subgroup classifiers can no longer learn from the encodings mapped by the proposed.

\end{abstract}

\end{frontmatter}

\pagestyle{headings}



\chapter{Introduction}\label{chap:intro}

\epigraph{When your face says it all, your mouth waits its turn.}{\textit{Anthony T. Hincks}.}


As known by many, and often in the form of common sense, facial cues hold an abundance of information-- whether it be the identity of the subject, their age, or even the way they feel in the moment. Hence, the human face, as a biometric, holds high potential in its relevance in vast practical use-cases. For starters, automatic face understanding makes up an exceptionally large problem space: face-based problems can be as specific as identity classification, or no interest in the identity but more broadly as it as an object (\ie face detection). Now, beyond the more traditional problems of identification and detection exist a slew of attribute-based tasks. As mentioned, lots can be learned from facial cues, which range from the measure of the presence or absence of an emotion, comfort or pain, honest or adversary, focused or distracted, and even rested or exhausted. Furthermore, facial cues (\ie faces captured in imagery or video data) encapsulate knowledge of intrinsic characteristics or attributes, such as gender and age; extrinsically faces can relate to one another by grouping. For instance, determine whether two or more subjects are blood relatives by comparing face data. 
 
Nowadays, many use-cases for face data in machine vision have been explored-- face-based technology can be seen throughout society, and in various forms. To name a few: used to unlock mobile device, provide access control in a security sensitive setting, and automatically tag on a social media platform. For any of these to have been possible, and to match the high capacity set by the modern-day, data-driven deep learning models, many have spent effort and resources to acquire and share large-scale benchmark facial image collections. Hence, some of the great advances in automatic \gls{fr} technology would never have been achieved without the large-scale datasets of labeled faces: of the many face-based problems there are opportunities in research and venture capital that can then be achieved. The same holds for the face-based attributes: due to the fine-grained nature of face data, along with a large number of samples readily available to scrape from the web, there exists a large variance inherent naturally by the data. As we will discuss in the later chapters on bias, consideration for the large variance, along with controlled variables that allow us to minimize bias, is critical. Thus, the same factor that motivates us upfront relates to a problem that stems up thereafter (\ie the need of big data and the effects of working with big data in face-based problems).

When drafting up the research questions (or topic), we reviewed the state of various face-based problems. When drafting a research plan the winter of 2015, kinship recognition from faces, in particular, caught my attention-- automatic \gls{fr} of nearly every type is used in main-stream solutions. For instance, identification system for recommending tags on social media platforms, or a face tracking and landmark detection system for apps like Snapchat. A \gls{fr}-based view that received minimal attention was kinship recognition. A problem statement we next will define.

\section{Objective and Significance}\label{chap:intro:objective}
Our goal was to acquire a framework to detect pairs of kinship from facial images. Furthermore, we intend to bridge the gap separating research-from-reality by working to develop machinery to automatically detect kinship via face data in visual media. Put differently: we set out to develop a system that accepts two or more face images as input, and outputs the class of KIN if blood relatives and NON-KIN otherwise. 
We focus on an analysis that  highlights different cases brought on by the evaluation metrics discussed as a part of Part I (\chapref{chap:facialunderstanding}). As part of this dissertation, we aim to determine how well kinship can be recognized from facial cues. As we will see, a richer resource (\ie paired data) was required to sufficiently model kinship from sets of faces from different subjects.



\subsection{Scope}
There are various views of the problem space for visual kinship recognition. In essence, the different views are organized using different data splits, label types, and metrics. Typically, the problem was either crafted to fit the data, or the data was shaped for the problem statement. Also, existing settings of related tasks for similar problems are often borrowed or used for inspiration (\eg LFW identification benchmark inspired data splits for our \gls{fiw} image collection).

\section{Contributions}
Most of this dissertation covers work previously published in peer reviewed journals and conference proceedings. For convenience, all work published as a part of this effort are listed at the end of this section. Besides in the following subsection, citations used throughout the report are in reference to the main bibliography at the end of the document-- the list of papers provided here is for quick reference as we review contributions and describe organization. Again, it is important to note that this section is the only part of the manuscript that uses the paper list at the end of this section.
 
\subsection{Organization}
 The dissertation consists of three parts: \emph{Preliminary}, \emph{Processing}, and \emph{Post-processing}. Let us now explain the contents of each. Note that references in the following paragraph refer to the list provided in the previous section. This is the last time the list of personal publications is referred to, and the bibliography contains the only citations used from here onward.

\vspace{3mm}
\noindent\textbf{\emph{Preliminary (Part I)}}. We start by reviewing \gls{fr} as a whole in \chapref{chap:facialunderstanding}. There are many topics common between traditional and kinship-based face recognition, all of which we introduce first and foremost. This leads to the pre-processing of faces (\ie face detection and alignment) that is required at the beginning of a \gls{fr} system (\chapref{chap:facedetection}). For the discussion we hone-in on our work in [11].

\vspace{3mm}
\noindent\textbf{\emph{Process} (Part II)}. Covers kinship recognition. Specifically, we review existing work in visual kinship recognition (\chapref{chap:vkr}), and then our contributions with \gls{fiw} dataset and \gls{sota} benchmarks. A number of our works fall in  (\chapref{chap:fiw}), \ie [2], [12], [13], [14], [15], [16], [17], [18]. Then, in \chapref{chap:fiwmm}, we discuss the recent release of the labeled multimedia for subjects of \gls{fiw}: \gls{fiwmm}, along with the semi-automatic machinery used to label the multi-modal data with minimum human inputs and no financial costs~[3].

\vspace{3mm}
\noindent\textbf{\emph{Post-processing} (Part III)}. We do qualitative studies on the different results in kinship recognition in \chapref{chap:krstateoftechnology}. This naturally leads to a discussion on the limitations of \gls{sota}, along with the technical challenges currently at the forefront. The aforementioned were findings from~[2], [7], [14]. 

Finally, our studies on bias in \gls{fr} systems is reviewed (\chapref{chap:bias}). For this, we built and shared the novel Balanced Faces in the Wild (BFW) face dataset to benchmark facial recognition systems with balanced data [8]. In addition to the machinery being bias, we also measure bias inherent in humans as well-- we conducted a human survey to measure and analyze the human bias. Furthermore, we describe a novel feature adaptation technique we proposed to mitigate issues from unbalanced performances across subgroups [1]. Additionally, our debiasing technique also benefits in areas of privacy and protection-- the objective used to adapt features involves a penalty for recognizing the subgroup.  Thus, the resulting mapping of the debiases and the features removes knowledge of the particular subgroup as well.  

We conclude with a discussion on next steps while concluding the various works represented and discussed in this thesis (\chapref{chapter:discussion}). 

\subsection{Publications}\label{chap:intro:subsec:publications}
Publications are listed in reverse-chronological order. Each item available online has `[\textcolor{blue}{paper}]' appended, which provides a direct link to the respective paper along with references to the main bibliography.

\begin{enumerate}
     \item \textbf{Joseph P. Robinson}, Can Qin, Yann Henon, Samson Timoner, and Yun Fu. ``Balancing Biases and Preserving Privacy on Balanced Faces in the Wild,'' in \emph{IEEE Transactions on Pattern Analysis and Machine Intelligence (TPAMI)}, 2020 (Under review). 
     
     \item \textbf{Joseph P. Robinson}, Ming Shao, and Yun Fu. ``Visual Kinship Recognition: A Decade in the Making,'' CoRR arXiv:2006.16033, 2020. (Under review, Trans. on PAMI). [\href{https://arxiv.org/pdf/2006.16033}{paper}]~\cite{robinsonKinsurvey2020}
    
     \item \textbf{Joseph P. Robinson}, Z. Khan, Y. Yin, M. Shao, and Yun Fu, ``Families in wild multimedia (FIW-MM): A multi-modal database for recognizing kinship,'' \emph{CoRR arXiv:2007.14509}, 2020. (Under review, Trans. on MM). [\href{https://arxiv.org/pdf/2007.14509.pdf?fbclid=IwAR1NW_0IpxOLjQNPCQ6Jgfhl9M3S-t3ZyGJQPwmzgVPfR1xFFPTuecuUGuU}{paper}]~\cite{robinson2020familiesinMM}
     
     \item Yu Yin, \textbf{Joseph P. Robinson}, and Yun Fu, ``Multimodal In-bed Pose and Shape Estimation Under the Blankets,'' CoRR arXiv:2012.06735, 2020. [\href{https://arxiv.org/pdf/2012.06735.pdf}{paper}]~\cite{yin2020multimodal}
     
     \item Yu Yin, \textbf{Joseph P. Robinson}, Songyao Jiang, Yue Bai, Qin Can, and Yun Fu, ``SuperFront: From Low-resolution to High-resolution Frontal Face Synthesis,'' CoRR arXiv:2012.04111, 2020. [\href{https://arxiv.org/pdf/2012.04111.pdf}{paper}]~\cite{yin2020superfront}
    
    \item Chengyao Zheng, Siyu Xia, \textbf{Joseph P. Robinson}, Changsheng Lu, Wayne Wu, Chen Qian, and Ming Shao. ``Localin Reshuffle Net: Toward Naturally and Efficiently Facial Image Blending,'' in 15-th Asian Conference on Computer Vision (ACCV), 2020. [\href{http://accv2020.kyoto/programs/sessions/}{paper}]~\cite{Zheng_2020_ACCV}
    
    \item Yu Yin, Songyao Jiang, \textbf{Joseph P. Robinson}, and Yun Fu ``Dual-attention GAN for large-pose face frontalization,'' in \emph{IEEE International Conference on Automatic Face and Gesture Recognition (FG)}, 2020. [\href{https://www-computer-org.ezproxy.neu.edu/csdl/proceedings-article/fg/2020/307900a024/1kecHPwIBLa}{paper}]~\cite{yin2020dual}
    
     \item Lichen Wang, Bin Sun, \textbf{Joseph P. Robinson}, T. Jing, and Yun Fu,``Ev-action: Electromyography-vision multi-modal action dataset,'' in \emph{IEEE International Conference on Automatic Face and Gesture Recognition (FG)}, 2020. [\href{https://www-computer-org.ezproxy.neu.edu/csdl/proceedings-article/fg/2020/307900a129/1kecHWmNk5y}{paper}]~\cite{wang2019ev}
     
     \item \textbf{Joseph P. Robinson}, Yu Yin, Zaid Khan, Ming Shao, Siyu Xia, Michael Stopa, Samson Timoner, Matthew A. Turk, Rama Chellappa, and Yun Fu,``Recognizing Families In the Wild (RFIW): The 4th Edition,'' in \emph{IEEE International Conference on Automatic Face and Gesture Recognition (FG)}, 2020. [\href{https://www-computer-org.ezproxy.neu.edu/csdl/proceedings-article/fg/2020/307900a877/1k}{paper}]~\cite{robinson2020recognizing}
     
    \item \textbf{Joseph P. Robinson}, G. Livitz, Y. Henon, C. Qin, Yun Fu, and S. Timoner, ``Face recognition: too bias, or not too bias?'' in \emph{Conference on Computer Vision and Pattern Recognition Workshop}, 2020. [\href{https://openaccess.thecvf.com/content_CVPRW_2020/papers/w1/Robinson_Face_Recognition_Too_Bias_or_Not_Too_Bias_CVPRW_2020_paper.pdf}{paper}]~\cite{robinson2020face}
    
     \item Yu Yin, \textbf{Joseph P. Robinson}, Zhang, Y., and Yun Fu. ``Joint super-resolution and alignment of tiny faces,'' in \emph{Proceedings of the AAAI Conference on Artificial Intelligence}, 2020. [\href{https://arxiv.org/pdf/1911.08566.pdf}{paper}]~\cite{yin2020joint}
    
     \item W. Zhuang, Y. Wang, \textbf{Joseph P. Robinson}, C. Wang, M. Shao, Y. Fu, S. Xia. ``Towards 3D Dance Motion Synthesis and Control,'' in \emph{CoRR  arXiv preprint arXiv:2006.05743}, 2020. [\href{https://arxiv.org/pdf/2006.05743.pdf}{paper}]~\cite{zhuang2020towards}
     
     \item \textbf{Joseph P. Robinson}, Yuncheng Li, Ning Zhang, Yun Fu, and Sergey Tulyakov, “Laplace landmark localization,'' in \emph{IEEE International Conference on Computer Vision (ICCV)}, 2019. [\href{https://arxiv.org/pdf/1903.11633.pdf}{paper}, \href{https://web.northeastern.edu/smilelab/misc//iccv_poster_landmarks.pdf}{poster}]~\cite{robinson2019laplace}
     
    \item P. Gao, S. Xia, \textbf{Joseph P. Robinson}, J. Zhang, C. Xia, M. Shao, and Yun Fu, ``What will your child look like? DNA-net: Age and gender aware kin face synthesizer,'' \emph{CoRR arXiv:1911.07014}, 2019. [\href{https://arxiv.org/pdf/1911.07014.pdf}{paper}]~\cite{gao2019will}
    
    \item Yue Wu, Z. Ding, H. Liu, \textbf{Joseph P. Robinson}, and Yun Fu, ``Kinship classification through latent adaptive subspace,'' in \emph{IEEE International Conference on Automatic Face and Gesture Recognition (FG)}, 2018. [\href{https://web.northeastern.edu/smilelab/papers/fiw_fg2018.pdf}{paper}]~\cite{wu2018kinship}
    
     \item \textbf{Joseph P. Robinson}, Ming Shao, and Yun Fu, ``To recognize families in the wild: A machine vision tutorial,'' in \emph{ACM Conference on Multimedia (ACMMM)}, 2018. [\href{https://dl.acm.org/doi/pdf/10.1145/3240508.3241471}{paper}]~\cite{robinson2018recognize}
     
     \item \textbf{Joseph P. Robinson}, Ming Shao, Yue Wu, Hongfu Liu, Timothy Gillis, and Yun Fu, ``Visual kinship recognition of families in the wild,'' in \emph{IEEE Transactions on Pattern Analysis and Machine Intelligence (TPAMI)}, 2018. [\href{https://web.northeastern.edu/smilelab/papers/fiw_pami.pdf}{paper}]~\cite{robinson2018visual}
     
    \item \textbf{Joseph P. Robinson}, M. Shao, H. Zhao, Y. Wu, T. Gillis, and Yun Fu, ``Recognizing families in the wild (RFIW),'' in \emph{RFIW at ACM MM}, 2017. [\href{https://www.dropbox.com/s/ya7ji5za3ljd8ge/main.pdf?dl=0}{paper}]~\cite{robinson2017recognizing}
    
    \item S. Wang, \textbf{Joseph P. Robinson}, and Yun Fu, ``Kinship verification on families in the wild with marginalized denoising metric learning,'' in \emph{IEEE International Conference on Automatic Face and Gesture Recognition (FG)}, 2017. [\href{https://web.northeastern.edu/smilelab/papers/fiw_fg2017.pdf}{paper}]~\cite{kinFG2017}
    
    \item \textbf{Joseph P. Robinson}, M. Shao, Y. Wu, and Yun Fu, ``Families in the wild (FIW): Large-scale kinship image database and benchmarks,'' in \emph{ACM Conference on Multimedia (ACMMM)}, 2016. [\href{https://arxiv.org/pdf/1604.02182.pdf}{paper}]~\cite{robinson2017recognizing}
    
    \item \textbf{Joseph P. Robinson}, Edward Scott, and Yun Fu, ``Pre-trained D-CNN Models for Detecting Complex Events in Unconstrained Videos'' in \emph{SPIE Commercial and Scientific Sensing \& Imaging}, 2016. [\href{https://www.dropbox.com/s/5sov5fgd5366yaj/98710O.pdf?dl=0}{paper}]~\cite{robinson2016pre}
    
    \item \textbf{Joseph P. Robinson}, M. Shao, Y. Wu, and Yun Fu, ``NEU- MITLL @ TRECVid 2015: Multimedia Event Detection by Deep Feature Learning,'' in \emph{Proceedings of TRECVID, NIST, USA}, 2015. [\href{https://www.jrobsvision.com/uploads/4/5/1/8/45182585/med-2015.pdf}{paper}]~\cite{robinson2015neu}
    
    
     
 \end{enumerate}

\subsection{Service}\label{chap:intro:subsec:services}
Here are selective services completed as part of this dissertation. Having served as a reviewer on several journals (\eg IEEE Trans. on PAMI, TIP, many others), PC or SPC for many conferences (\eg CVPR, AAAI, IJCAI, ICCV, ECCV, FG (3x awarded \emph{Outstanding Reviewer}, and many others), and for many years, these listings were omitted and are, thus, not explicitly listed. Instead, workshops, challenges, and tutorials at top tier conferences, for which my contributions were critical to the success of the event are listed as follows.
\subsubsection{Workshops}
 \begin{enumerate}
 \item 2021 Workshop Chair, \emph{10th Workshop on the Analysis \& Modeling Faces \& Gestures (AMFG)}, CVPR (online). [\href{https://web.northeastern.edu/smilelab/amfg2021/}{web}]
    \item 2020 Challenge Chair and Organizer, \emph{4th Recognize Families In Wild (RFIW) Challenge}, IEEE FG Argentina. [\href{https://web.northeastern.edu/smilelab/rfiw2020/}{web}]
\item 2019 Host \& Organizer, \emph{Recognizing Families in the Wild}, CVPR Long Island, CA. [\href{https://web.northeastern.edu/smilelab/fiw/cvpr19_tutorial/}{web}]
\item 2019 Tutorial Host \& Organizer, \emph{Recognize Families: A Machine Vision Tutorial (II)}, IEEE FG Lille, France. [\href{http://fg2019.org/participate/workshops-and-tutorials/visual-recognition-of-families-in-the-wild/}{web}]
\item 2019 Workshop Chair, \emph{9th Workshop on AMFG}, CVPR Long Island, CA. [\href{https://web.northeastern.edu/smilelab/amfg2019/}{web}]
\item 2019 Workshop Chair, \emph{2nd Workshop on Faces in Multimedia (FacesMM)}, ICME Shanghai, China. [\href{https://web.northeastern.edu/smilelab/facesmm19/}{web}]
\item 2019 Challenge \& Workshop Chair, \emph{3rd RFIW Challenge}, IEEE FG Lille, France. [\href{https://web.northeastern.edu/smilelab/RFIW2019/}{web}]
\item 2018 Host \& Organizer, \emph{RFIW: Machine Vision Tutorial}, ACM MM Seoul, S. Korea. \cite{robinson2018recognize}
\item 2018 Host \& Organizing Chair, \emph{1st Workshop on FacesMM}, ICME San Diego, CA. [\href{https://web.northeastern.edu/smilelab/FacesMM2018/}{web}]
\item 2019 Challenge \& Workshop Chair, \emph{2nd RFIW Challenge}, IEEE FG China. [\href{https://web.northeastern.edu/smilelab/RFIW2018/}{web}]
\item 2018 Workshop Host \& Organizing Chair, \emph{8th Workshop on AMFG}, CVPR SLC, Utah. [\href{https://web.northeastern.edu/smilelab/AMFG2018/}{web}]
\item 2017 Host \& Organizing Chair, \emph{New England CV Workshop}, NEU Boston, MA. [\href{https://web.northeastern.edu/smilelab/necv2017/}{web}]
\item 2017 Host \& Organizer, \emph{RFIW Data Challenge Workshop}, ACM MM Mountain View. [\href{https://web.northeastern.edu/smilelab/RFIW2017/}{web}]
 \end{enumerate}

\part{Preliminaries}\label{part:preliminaries}

\chapter{Automatic Facial Recognition}\label{chap:facialunderstanding}
\section{Overview}
To describe our contributions in automatic face recognition and understanding technology, an overview of fundamentals, such as problem statements, related face-based systems and efforts, along with data preparation and evaluation concepts that relate to the work done for this dissertation.

We start by reviewing major milestones in conventional \gls{fr}. A basic understanding in concepts pertaining to conventional \gls{fr} is imperative for understanding kinship recognition from image data (\ie facial images). 
For this, a brief look at traditional systems before those that are more popular nowadays. As we will discuss, deep learning-based approaches are mostly data-driven: feature learning, a large model complexity that demands more data to avoid over-fitting (\eg \glspl{cnn}, \glspl{gan}, and much much more). As a part of the basic depiction of the aforementioned is a section of the different loss functions that had previously claimed \gls{sota} and had later been employed in a face-based kinship recognition system. Performance ratings for both the traditional and the modern-day \gls{fr} methods are reported for face identification task to provide insight in our expectations later when used in kinship recognition-- this is especially true for loss functions, which has been pivotal in \gls{fr}.




\begin{figure}
    \centering
    \includegraphics[width=.75\linewidth]{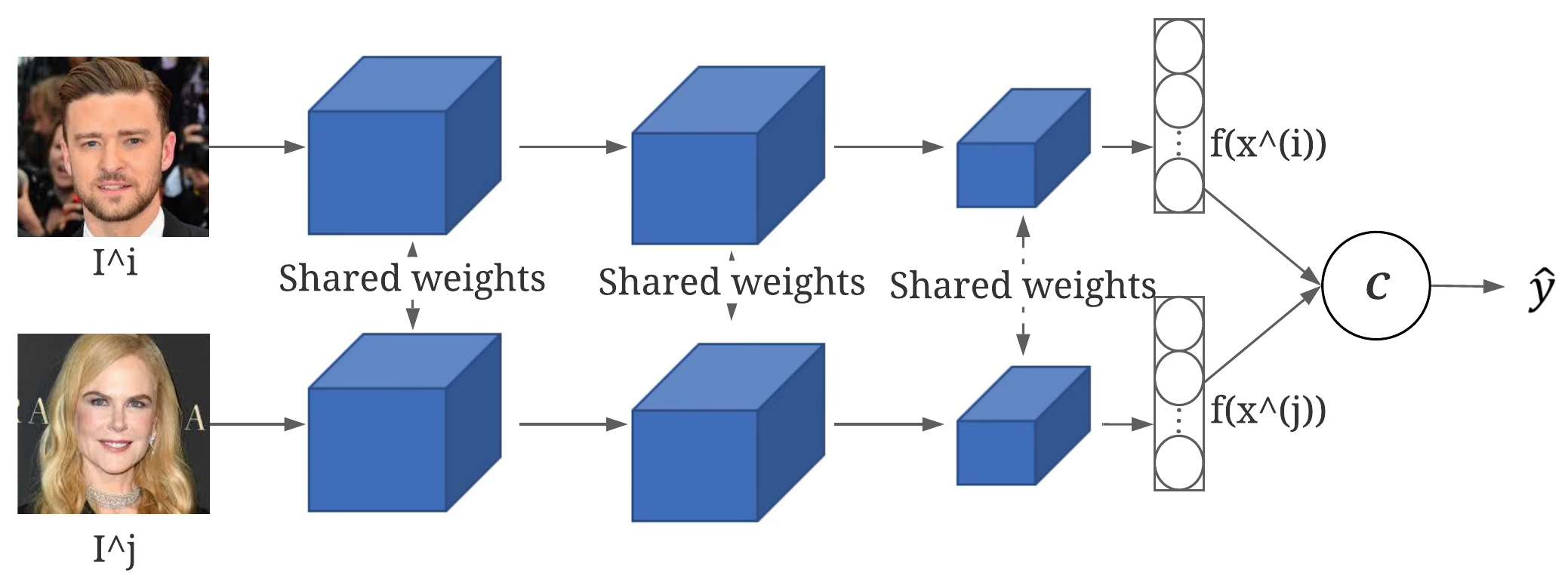}
    \caption{\textbf{A generic verification system.} Regardless of the network specifications (\ie independent of layer counts, layer sizes, type of metric set topmost). The aim is to map two image inputs to a single logistic value. Input images (\ie $I^i$ and $I^j$) must then assume the same size as that of the input layer of the model. Then, the learned representation of the $i-{th}$ and $j-{th}$ faces are output from the last layer of the network prior to the classification layer. An aspect of the model shown is the mechanism used to fuse (\ie single logit value output from $\mathcal{C}$, which was fed the features of the $i-{th}$ and $j-{th}$ face encoding). The framework is inherently boolean, as the task is to map a sample pair (\ie pair of faces) to a boolean class label (\ie 1 if \emph{genuine} and 0 for \emph{imposter}).}
    \label{chap:preprocessing:fig:generic-fv}
\end{figure}

We then step back a couple of modules in the ML pipeline to preprocessing. Most face-based systems depend on a face detector (\ie object detector, with face as a boolean target for whether it is present or not). Furthermore, specific landmarks of the face are too detected: facial landmarks are often used to crop \glspl{bb} more consistently across a large set; also, landmarks are treated as points of reference to frontalize (\ie align) faces in images prior to feeding to model. Considering the preprocessing of faces (\ie detection, aligning, \etc). is conducted on nearly all of our work, while a majority of all others (\ie for conventional face recognition, along with nearly all other discriminative tasks like kinship recognition using facial cues). Finally, we built several large-scale face imagesets-- an effective preprocessing setup shows to be exceptionally important in problem spaces of increasing difficulty. 

All topics covered in this chapter are meant to be span out broadly. In other words, this preliminary information is not the core (\ie meat) of this dissertation. However, the basic understanding we hope to provide the reader is believed to be essential for the topics covered in the later chapters. Furthermore, several efforts spent on this dissertation fall in the realm of face-based preprocessing. On the one hand, our papers that relate to the topics of this chapter are included. Still, the thesis is not on these works and will not tailor the upcoming discussion.

Having recently joined the researchers of the Machine Vision community, I have already been exposed to a wide range of contemporary problems tackled by both classic and modern methods. However, up to this past month I have had no experience working with faces. That was until a recent research assignment involving kinship recognition from digital photos came about. Hence, understanding faces in the eyes of computer vision is essential here. Plus, as a young researcher in the field, I felt it was essential to gain some understanding of face detection and facial recognition.

\begin{figure}
    \centering
    \includegraphics[width=.4\linewidth]{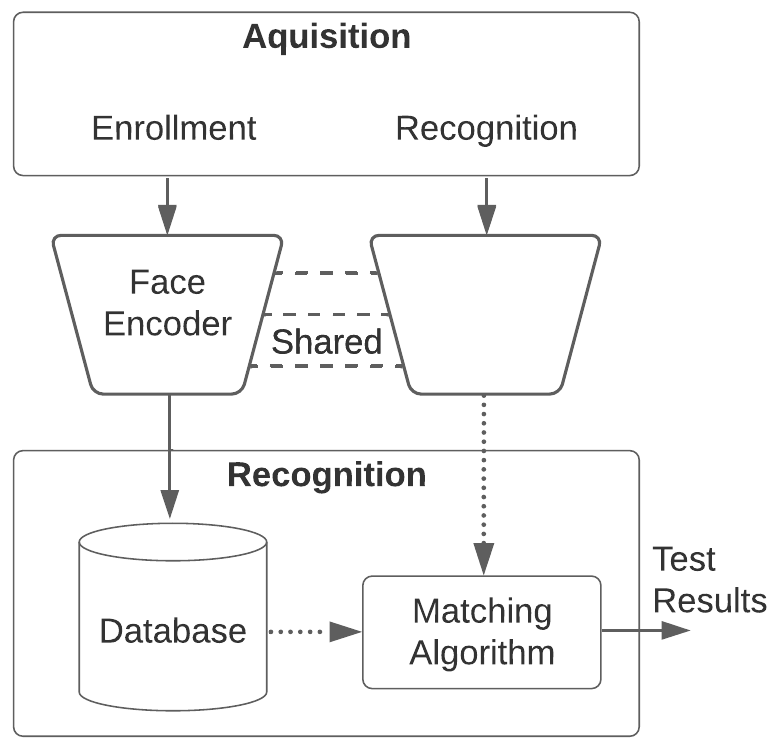}
    \caption{\textbf{Generic recognition system.} Faces are often stored as encodings in a database through \emph{enrollment}. \emph{Recognition} is then to compare the encoding of an input face to those in the database.}
    \label{chap:preprocessing:fig:generic-fr}
\end{figure}

Due to both space and time constraints and, moreover, relevance to the contributions of the dissertation, only a few \gls{fr} et \emph{Eigenfaces}, \emph{Fisherfaces}, Local Binary Pattern Histograms.
Modern Facial Recognition
The goal of human facial recognition is to automatically verify or identify an individual from digital data, \ie images or videos (image stacks). Facial recognition, in itself, is applicable in a wide range of technologies. Much of this applies to security-based applications (\eg biometric identification systems, human tracking, and surveillance systems of all sorts). Nonetheless, its uses span well beyond the scope of just security-related problems. For example, Facebook uses face verification to suggest its users when a photo is uploaded. In addition, facial verification is largely used in search and retrieval-based tasks, involving images, videos, or even both. Facial recognition is also directly applicable in applications involving kinship (\eg sorting a family photo album, determining ethnicity, and such)~\cite{fang2010towards}. A purpose for facial recognition is common in areas involving facial images, which have and continue to increase rapidly in this ''mobile age''~\cite{MillerElements}.

Common challenges faced with facial recognition can be generalized into two groups, external disturbances (\eg changes in illumination, occlusions from glasses, beards, makeup, \etc) and internal influences (\eg facial expressions, head rotations, human aging, \etc). It are these challenges that decipher the pros and cons of any given facial recognition approach, as will be exemplified through analysis of the methods covered in this report.

The following sections introduce three modern approaches to facial recognition, which are \emph{Eigenfaces}, \emph{Fisherfaces} and LBPH. Following this, experimental results running these methods on two face databases are shared and further analyzed.

\begin{figure}[!t]
    \centering
    \includegraphics[width=.35\textwidth]{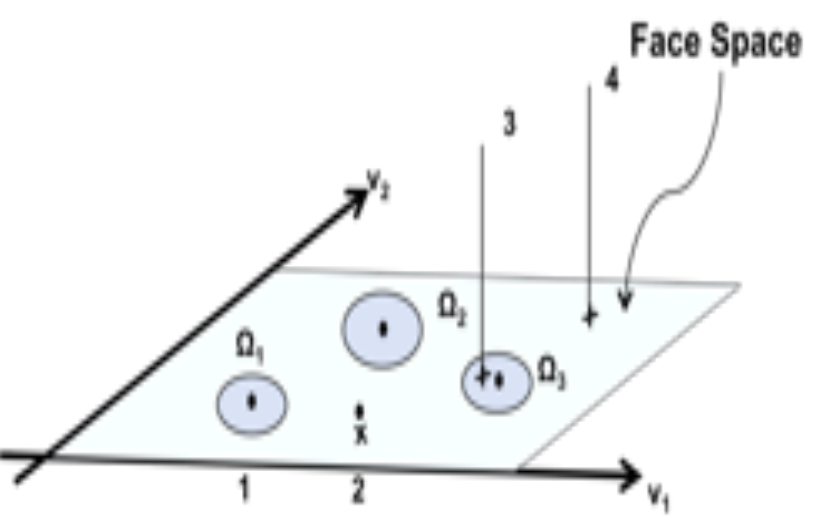}
    \caption{\textbf{Illustration  of the Face Space, which spans the area of the light-blue rectangle.} Note that the blue spheres in the Face Space ($\Omega_1$, $\Omega_2$, $\Omega_3$) describe a face of a particular person, \ie used to identify or verify individual instances. Beyond that, but within the bounds of the Face Space, are varying faces of unknown type, \ie could be used, for instance, to detect any face.}
    \label{chap:preproc:facespace}
\end{figure}

\section{Traditional Methods}
\subsection{Eigenfaces}
\subsubsection{Overview}
\emph{Eigenfaces}: the eigenvectors of face images introduced in the early 1990s in research as a \gls{fr} system. \emph{Eigenface}-related concepts are still widely used today. It is a simple scheme that drastically reduces the search space. Moreover, it performs reasonably well with non-complex face data-- a fair assumption to make in many real-world applications based on facial images.

In short, \emph{Eigenfaces} was motivated by the idea that highly dimensional face images are highly correlated, from which the size of the image space can be projected to a space that ignores common features, \ie preserve the feature dimensions that have less correlation, resulting in a higher variance. In essence, \emph{Eigenface} are based on a dimensionality reduction that selects the features of greatest variance from all features. In turn, preserve fewer features with most information. 

\gls{pca}, a dimensionality reduction technique, finds the subspace, projects a feature encoding of an image, and compares to other images via the Euclidean distance or another similarity or distance metric. In other words, the difference between \emph{Eigenfaces} is found and compared to a threshold to determine a final decision, match or no match.
This \gls{pca}-based dimensionality reduction scheme, generally speaking, uses lower-dimensional vectors to represent some high-dimensional data. In our case, the high-dimensional data are the original images, and the low-dimensional vectors are the \emph{Eigenfaces}. These feature vectors are eigenvectors and the face images projected into this lower-dimensional face-space have been appropriately named \emph{Eigenfaces}, \ie  given a collection of facial images, \gls{pca} returns a set of corresponding basis vectors that are used to describe the faces in a lower-dimensional subspace.

\begin{figure}[!t]
    \centering
    \includegraphics[width=.4\textwidth]{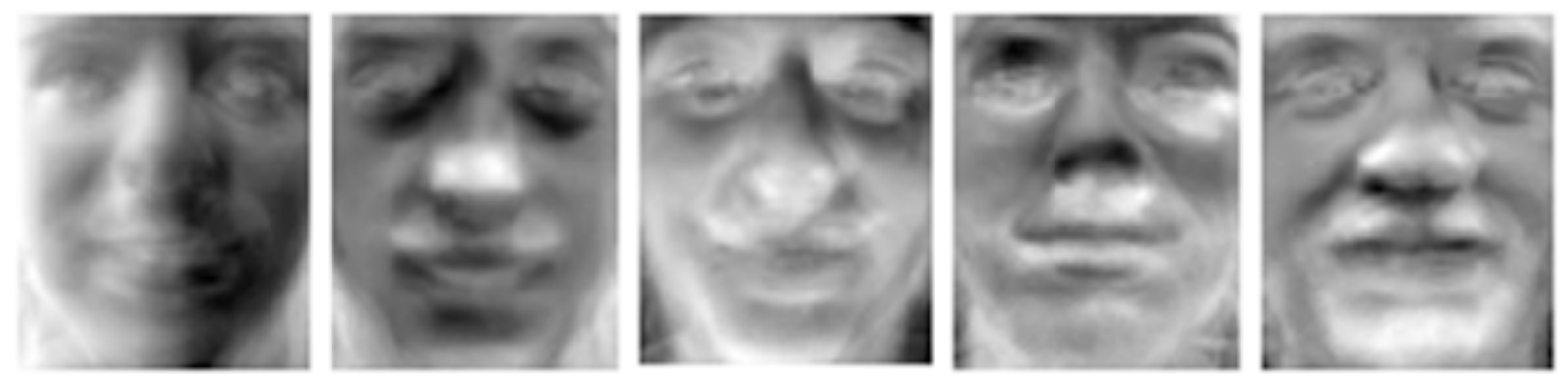}
    \caption{\textbf{By taking the difference between each face and the mean of all faces \emph{Eigenfaces} are normalized.}  Results in images of facial structure, and resembling a ghost-- some call these Ghostfaces, it is most common, and agreeably more appropriate, to refer to them as \emph{Eigenfaces}.}
    \label{chap:preproc:ghostfaces}
\end{figure}

In summary, given a face image of size N x N (pixels) in its vector exists as a single point in N2-dimensional image space. Considering that facial images contain correlated features, these projections shall not be distributed in the large image space, but yet a smaller, greatly reduced space—Such to capture the variations and remove the redundancy of the set of facial images.

In essence, it is a process of minimizing the non-diagonal elements of the covariance matrix, while maximizing the diagonal, which results in a form similar to the following:

\begin{equation}
    W = PX = Cov(W)=\frac{1}{n-1}WW^T=\begin{bmatrix} 
* & \dots & 0 \\
 \vdots & \ddots &  \vdots\\
0 & \dots & * \\
\end{bmatrix},
\end{equation}

\noindent where the rows of $P$ are the principal components of data $X$, which preserves most of the information of $X$, \ie  $W$ is the diagonal, also known as the \emph{Eigenfaces}.

\begin{figure}
    \centering
    \includegraphics[width=.45\textwidth]{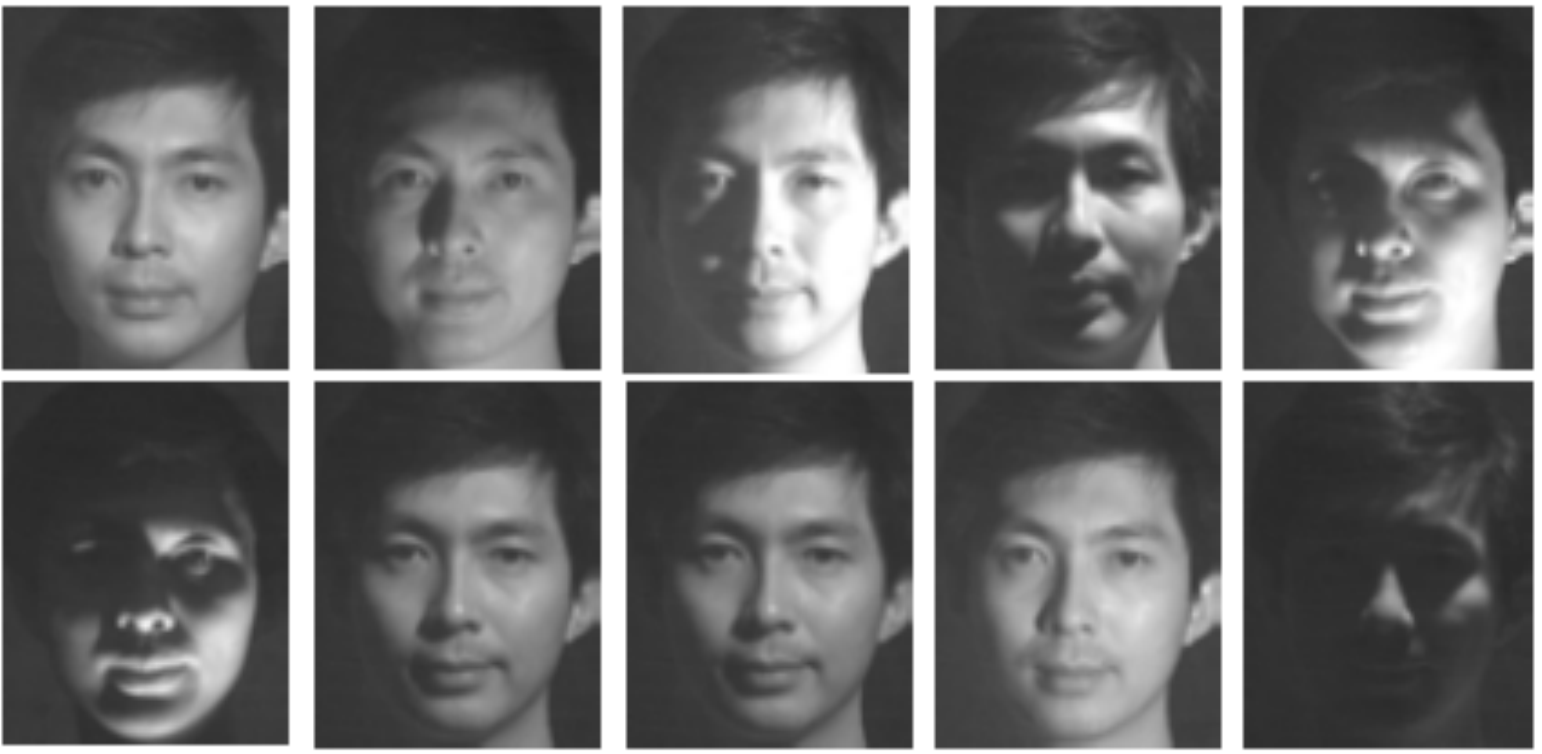}
    \caption{\textbf{Eigenfaces for the same face under different lighting conditions.} The original face images and their \emph{Eigenface} equivalent are shown on top row and bottom row, respectively. Notice the variation between \emph{Eigenfaces}.}
    \label{chap:preproc:eigenfaces}
\end{figure}

\subsubsection{Learning face space}
\emph{Eigenfaces}— or as some call it, Ghostfaces (\figref{chap:preproc:ghostfaces})--maps a facial image to its \emph{Eigenface} as follows:

\begin{figure}[h!]
    \centering
    \includegraphics[width=.7\textwidth]{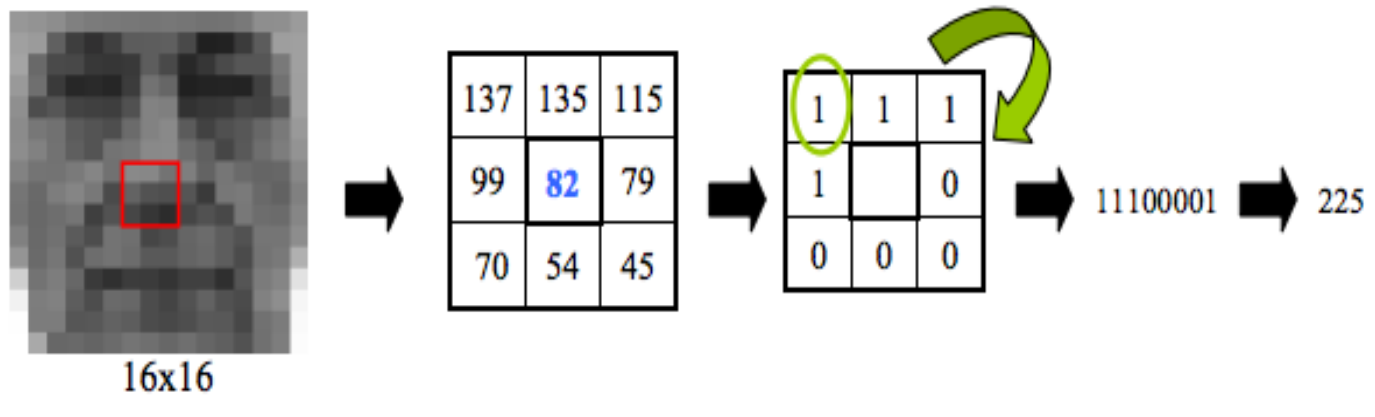}
    \caption{\textbf{Illustration depicting the process of obtaining LBH features from an image for a single pixel.} With every pixel  processed, a histogram is generated to represent each image.}
    \label{chap:preproc:lbpfeatures}
\end{figure}

 \begin{enumerate}
 \item 	Collect/ obtain a set of $M$ face images to use for training, $I_1, I_2, \dots, I_M$.
 
 \item Vectorize each $N\times N$ image ($I_i$) to span $N^2\times1$ dimensional space ($\Gamma_i$).\footnote{Note that all images are required to have the same resolution, \ie  a preliminary step would be to resize all images to the same size.}  

  \item Calculate the average face ($\Psi$) of all $M$ images.
   \begin{equation}
     \Psi = \frac{1}{M} \sum_{i=1}^M\Gamma_i.
 \end{equation}
 
 \item Subtract the mean face from all facial images, \ie  normalize feature vector.
\begin{equation}
    \Phi_i=\Gamma_i-\Psi.
\end{equation}
 
 \item Find the covariance matrix of all mean-shifted facial vectors.

\begin{equation}
    S = \frac{1}{M}\sum_{n=1}^M\phi_n\phi_n^T=\Phi_n\Phi^T,
\end{equation}

where $\Phi=[\phi_1, \phi_2, \dots, \phi_M$ is a complete set of orthonormal eigenvectors   spanning $N^2\times M$ and $S$ is size $N^2\times N^2$.

The mean-shifted image features of each    is a linear combination of the eigenvectors, \ie
\begin{equation}
    \hat{\Phi} - \Psi = \mathbf{\omega}_1\mathbf{u}_1+\omega_2\mathbf{u}_2+\dots+\omega_{N^2}\mathbf{u}_{N^2}.
\end{equation}

Each face can be represented using just the top $K$ eigenvectors, reducing the dimension of the problem from $N^2$ to $K$, where $k\ll N^2$ and, hence, is approximated as follows:

\begin{equation}
    \hat{\Phi} - \Psi = \mathbf{\omega}_1\mathbf{u}_1+\omega_2\mathbf{u}_2+\dots+\omega_{N^2}\mathbf{u}_{N^2},
\end{equation}

\noindent or equivalently $\hat{\Phi} - \Psi = \sum_{j=1}^K\mathbf{\omega}_j\mathbf{u}_j$, and where the $j^{th}$ $u$ and $\omega$ is the \emph{Eigenface} and eigenvalue, respectively: the $k$ eigenvector corresponding to the largest eigenvalues are kept.

\begin{figure}[h!]
    \centering
    \includegraphics[width=.5\textwidth]{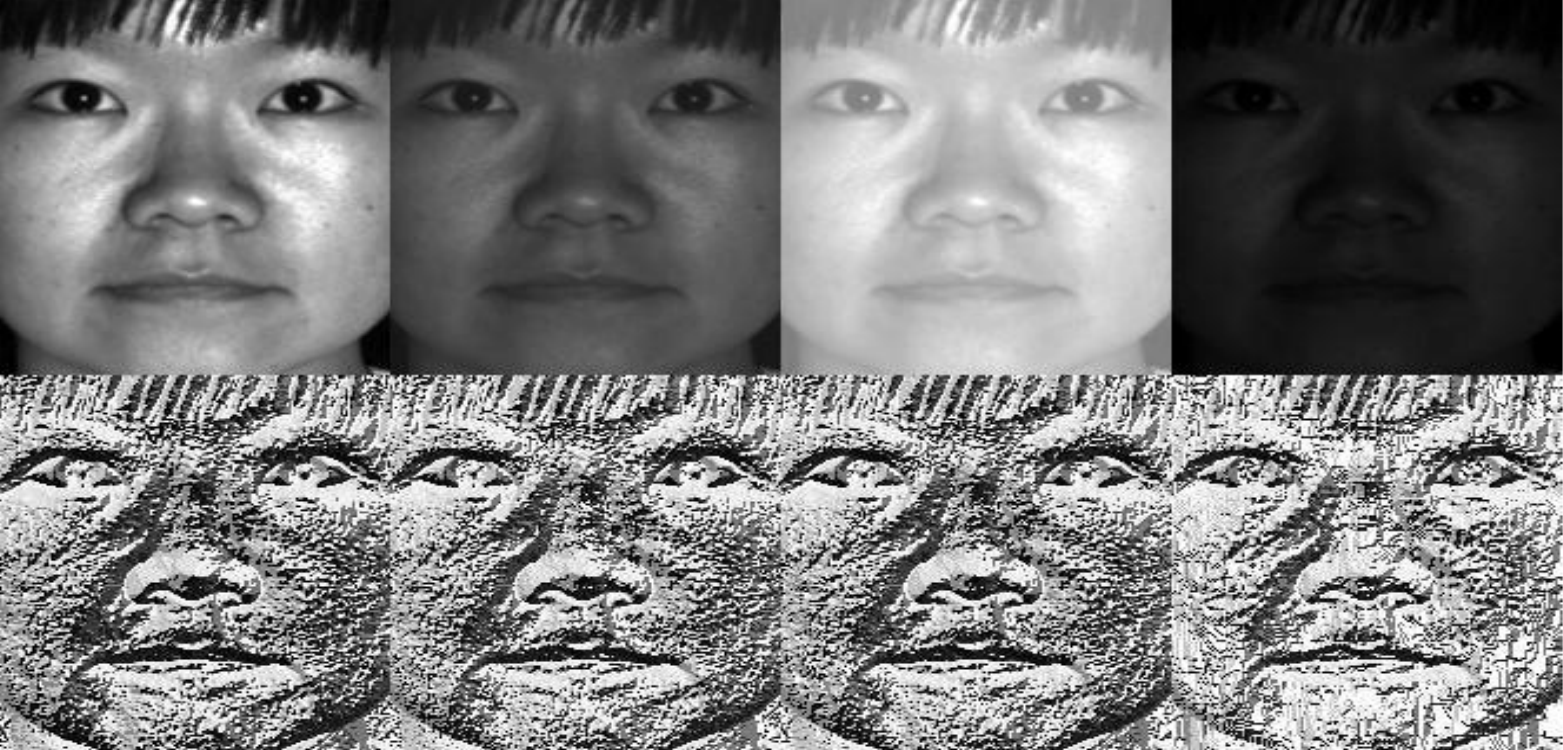}
    \caption{\textbf{LBPH of the same face under different lighting conditions.} Notice the light invariance that is inherited with this feature.}
    \label{chap:preproc:lpbh}
\end{figure}

 \end{enumerate}

\subsubsection{Facial Recognition}
Provided an unseen image, and a subspace $\Omega$ that was found during training, images are projected to the Face Space by the formula:

$$
Y=\Omega^T(X-\Psi),
$$

\noindent where

$$
\Omega_i = \begin{bmatrix} 
\omega_1 \\
 \omega_2\\
\vdots \\
\omega_K
\end{bmatrix}, \text{ for } i \in [1, 2, \dots, M].
$$

The distance between $y$ and each face class is found as the Euclidean distance. 

$$
\mathcal{E}_k^2=||y-y_k||^2,
$$
where $k = 1, 2, \dots, M$.

A distance threshold $\Theta_c$, is half the largest distance between any two face images: 
$$
  \Theta_c=\frac{1}{2}\max_{j,k}, 
$$
\noindent where $j = 1, 2, \dots, M$ and  $k = 1, 2, \dots, M$.

The distance $\epsilon$ is found between the original image $x$ and its reconstructed image \emph{Eigenface} from Face space $x_f$:
$$
\epsilon^2=||x-x_f||^2,
$$
where $x_f=Wx+\mu$.

\begin{figure}[t!]
    \centering
    \includegraphics[width=.65\textwidth]{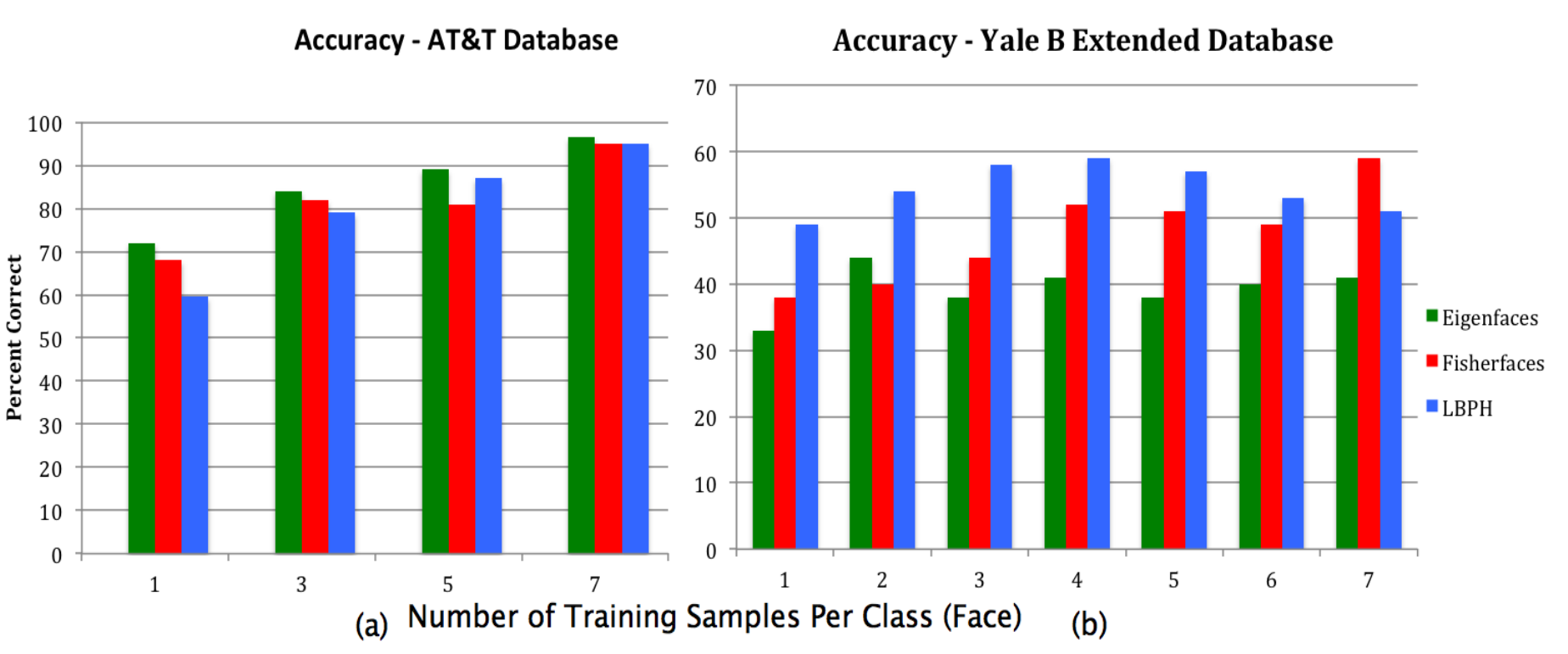}
    \caption{\textbf{Accuracy measure.} Correct (\%) as function of training samples count per class for the AT\&T dataset (a). Correct (\%) as function of training samples count per class for the Yale B Extended dataset (b).}
    \label{chap:preproc:accuracy}
\end{figure}

\noindent\textbf{Recognition summarized.}
\begin{itemize}
    \item IF $\epsilon\geq \Theta_c\Rightarrow$ input image is not a face image; 
    \item IF $\epsilon< \Theta_c \&\& \epsilon_k\geq \Theta_c\text{ for all }k\Rightarrow$ input image contains an unknown face; 
    \item IF $\epsilon< \Theta_c \&\& \epsilon_k^{*}=\min_{k}(\epsilon_k)<\Theta\Rightarrow$ input image contains the face of individual $k*$. 
\end{itemize}
The main idea is that the $K$ dimensions embody the most variant aspects across the face images. By this approach, the size reduction of search space greatly outweighs the amount of information compromised. Hence, enough information is preserved in \emph{Eigenfaces} to reconstruct any of the faces used to find the mean-face by doing the above steps in reverse, with, of course, the same mean face that was used to initially shift (normalize) the facial images.

Although \emph{Eigenfaces} work well, there are drawbacks. Considering that we are compressing the information according to the $K$ eigenvectors with largest variations, \emph{Eigenfaces} are prone to capturing variations caused by external distortions, making it non-invariant to various external sources. An example of this shown in \figref{chap:preproc:eigenfaces}, as the same face in image space is drastically changed in Face Space by changes in lighting. 

\begin{figure}[t!]
    \centering
    \includegraphics[width=.75\textwidth]{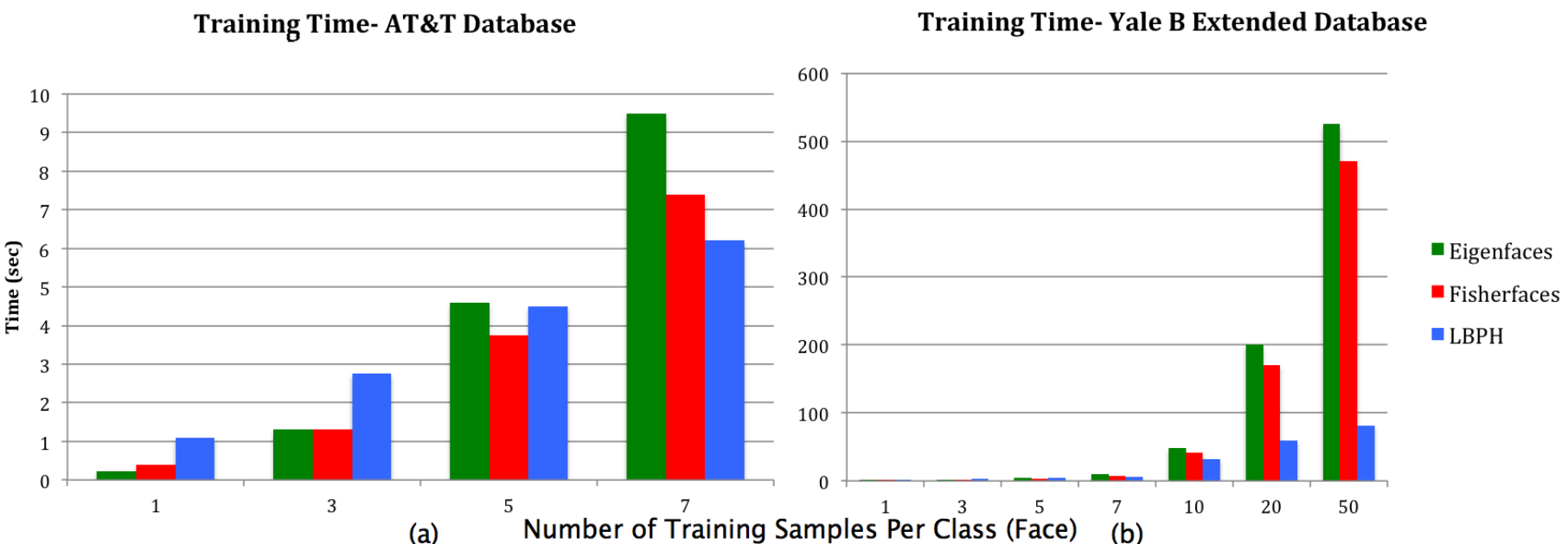}
    \caption{\textbf{Performance measure.} Training time as function of training samples count per class for the AT\&T dataset (Left). Training time as function of training samples count per class for the Yale B dataset (Right).}
    \label{chap:preproc:trainspeed}
\end{figure}

The following section introduces a method that addressed the drawbacks of \emph{Eigenfaces}— An approach called \emph{Fisherfaces}.

\subsection{Fisherfaces}
\emph{Fisherfaces}, like \emph{Eigenfaces}, seek to find a low-dimensional subspace to facilitate the facial recognition task from; \emph{Fisherfaces}, unlike \emph{Eigenfaces}, search for a subspace in a discriminant manner. It overtakes the shortcomings of \emph{Eigenfaces}, while its simple principles are still preserved. It does so by taking the Linear Discriminant Analysis approach~\cite{belhumeur1997eigenfaces}.

\subsubsection{Linear Discriminant Analysis (LDA)}
LDA finds a basis for projection such that the intra-class variation is minimized, while the inter-class variation is maximized. Rather than explicitly modeling its deviation, we linearly project the high-dimensional data into a subspace that discounts those regions of the face with large deviation. When this approach is used for facial recognition, we refer to it as using \emph{Fisherfaces}. Unlike \emph{Eigenfaces}, and due to the nature of LDA, \emph{Fisherfaces} contain class-information that can be used for classification of the images provided.

LDA class-specific dimensionality approach is summarized as follows:
\begin{itemize}
    \item Obtain features that best separate between classes, opposed to global scatter.
    \item Clusters the same classes tightly; Separates between different classes.
    \item Reduces search space size (\eg \emph{Eigenfaces}).
    \item Immune to external agents (unlike \emph{Eigenfaces}).
\end{itemize}
Given random vector X made up of samples from $c$-classes.

$$
X={X_1, X_2, \dots, X_c}; X_i={x_1, x_2, \dots, x_n}.
$$

\subsubsection{Within (W)/ Between (B) class scatters}

Between-class covaraince:
$$
SS_B=\sum_{i=1}^cN_i(\mu_i-\mu)(\mu_i-\mu)^T,
$$

where $\mu=\frac{i}{n}\sum_{i=1}^nx_i$.

Within-class covaraince:
$$
SS_W=\sum_{i=1}^cS_i,
$$

\noindent where $S_i=\sum_{x\in D}(x-m_i)(x-m_i)^T$ and $m$ coordinates retained.

Searches for $W$ that maximizes class separability criterion:

$$
W_{opt} = arg\max_{\omega}\frac{|W_TS_BW|}{|W_TS_WW|}.
$$

\subsection{Local Binary Pattern Histograms (LBPH)}
\begin{itemize}
    \item Uses local (texture) descriptors.
    \item Thresholds each pixel with neighboring pixels to generate a binary pattern.
    \item Bins all features for same class into a histogram to obtain a single representation.
\end{itemize}

The process of obtaining \gls{lbph} involves a simple calculation, one involving the summation of the differences between the central pixel and all of its neighbors~\cite{georghiades2001few}.

Unlike \emph{Eigenfaces}, \gls{lbph} is invariant to changes in lighting (Fig~\ref{chap:preproc:lbpfeatures}). The figure shows four images of the same face under different lighting conditions (top row) and the corresponding \gls{lbph} representation (bottom row). Notice the histograms are nearly identical, regardless of the large variations in light.

\subsection{Results and analysis}

\subsubsection{Experimental Setup}
\begin{enumerate}
    \item Implemented using many MATLAB built-in functions.
    \item Built program with two modes:
    \begin{enumerate}
        \item Performance Mode: Experiment varies number of training \& test samples.
        \item Test Mode: User specifies algorithm, dataset, and number of training samples.
    \end{enumerate}
    \item Time is measured (seconds) for all calls to training and testing functions.
    \item Accuracy measures are based on the number of testing samples.
\end{enumerate}
\subsubsection{AT\&T Database}
AT\&T is a simple dataset containing mild variations in facial expressions, small rotational shifts, and minor facial obstructions (\eg glasses). There are a total of 40 subjects, each with 10 face images. The images are 8bit (gray scale) with a size of 92$\times$112 pixel, formatted as PGM image files. For every experiment the number of training images is specified, and the remaining (of the 10) are used for testing.
\subsubsection{Extended Yale B Face Database}
Yale B is a more Complex dataset that contains high variation in facial expressions, image rotations, facial obstacles (\eg glasses and beard), and changes in illumination. There are a total of 38 subjects, each with 64 face images. The images are 8bit (gray scale), 168$\times$192, and PGM formatted. And again, for each experiment the number of training images is specified, and the remaining (of the 10) are used for testing.
\paragraph{Accuracy Metric}

\subsubsection{Discussion}
In its presence, the variations caused by these external distortions are amongst the largest of the variations between the faces themselves. In such situations, such as with the Yale B extended Database, the \gls{pca}-based approach acquires a noisy mean-face and hence, \emph{Eigenfaces} that captures unwanted distortion in its description. Since the images of a particular face, under varying illumination but fixed pose, lie in a 3D linear subspace of the high-dimensional image space (without shadow), \emph{Fisherfaces} work well with the more complex face images, so does \gls{lbph}.

As expected, overall higher accuracy were obtained on the simpler AT\&T dataset. \emph{Eigenfaces} led the 3 algorithms slightly on AT\&T with 7 training samples, which was interesting to see. This shows the true power of the \gls{pca}-based approach in the ideal case; the lack of unwanted variations in the simple AT\&T dataset makes \emph{Eigenfaces} the perfect candidate for it.
Eigenfaces score lowest on more complex dataset; \gls{lbph} does the best, especially with illumination variations; \emph{Fisherfaces} is the runner-up.

For the average time to train, \emph{Eigenfaces} are consistently the slowest; \gls{lbph} is the quickest overall, as it steadily increases in time with increasing number of training samples; and \emph{Fisherfaces} are consistently quicker than \emph{Eigenfaces}, but always slower than \gls{lbph}.

Provided an analysis and comparison between a few modern facial recognition techniques. \emph{Fisherfaces} demonstrated top performance in terms of accuracy and performance. \gls{lbph} achieves a high accuracy at the price of an increased testing time.

\section{Modern-day, data driven deep learning}
 Hence, the traditional workflow of a machine learning system have the feature extraction and modeling as independent modules (\eg a system extracts color histograms as the feature and trains a nearest neighbor model on top). In other words, as a schematic or flowchart drawing, the features and the model are separate entities, while deep learning encapsulates the two steps into one-- a known benefit of deep learning technology is that the models find the most interesting features at the bottom (\ie beginning) of the network, that minimize the loss using the model making up the top. The upside here is a built-in mechanism that determines the features of highest interest, opposed to a human having to determine the optimal feature type for a problem, making it such that an expert of a specific domain was typically required for each data type. Largely, it was typically the feature extraction step that demanded specialists with years of knowledge of the quarks and in depth to acquire a means of hand-crafting features. Nowadays, the feature extractor and trained model are one and the same. Of course, domain specific knowledge is still essential in some instances and tasks. However, it is from this that researchers were then enabled to design multimodal systems more frequently; like no other time were there as many \gls{sota} approaches shared across problem domains-- attention for text~\cite{bahdanau2014neural}, vision~\cite{ba2014multiple}, speech~\cite{chorowski2015attention}, graph structures~\cite{velivckovic2017graph}, and others. The same even holds true for major topics such as \glspl{rnn} and \glspl{cnn}, along with fundamental concepts like back-propagation~\cite{lecun1989backpropagation} and drop-out~\cite{krizhevsky2012imagenet}. Note that this is not a claim that knowledge has only recently begun to transfer between problem domains in such an explicit manner, as that would be by no means accurate (\eg bag of words models~\cite{yang2007evaluating}). However, the lines separating experts that specialize in specific data types seems to be fading away. Furthermore, provided scripts to preprocess the inputs to a \gls{nn}, the workings of deep learning technology tends to remain across different data types. Hence, the \emph{black box} encapsulated within a deep model shares tendencies for various signal types.
 
Modern-day deep models have worked with prominence in automatic face understanding problems. In 2014 Taigman~\etal of Facebook first proposed using a deep \gls{cnn} for \gls{fr}~\cite{taigman2014deepface}. Over a half of a decade later there has been at least one major contribution in conventional \gls{fr} each year ever since (\ie 2015~\cite{Parkhi15}, 2016~\cite{wen2016a}, 2017~\cite{liu2017sphereface}, 2018~\cite{deng2019arcface}, 2019~\cite{duan2019uniformface}, and even 2020, the year of facial masks~\cite{wang2020masked}). Details on deep learning advances in \gls{fr} technology are provided as a part of recent surveys~\cite{guo2019survey, masi2018deep}. 

To briefly summarize, and to a degree needed for a more complete understanding of the work in using facial cues to detect family members, let us better understand the uniqueness to face-based problems. In conventional object recognition, the task to to identify a predefined object in imagery (\eg a cat). For this, one could perceive the problem as boolean (\ie \emph{is} or \emph{is not} a cat). Alternatively, in a closed-set problem, in a multi-class task, the approach can be founded on determining which class is the instance of (\eg trained on a set of classes representing \emph{pets}). Regardless, in a supervised setting, the model has access to labeled instances for the object(s) of interest during training. Now, in \gls{fr}, the multi-class setting for this fine-grained classification problem takes on a different form: given a set of face images for $C$ classes (\ie identities), the goal is to train a model to verify a pair of faces for being of the same subject or not. However, the subject in question is not in the \emph{train set}. In fact, the subjects in all \emph{test} photos were never seen by the model prior. By this, \gls{fr} is inherently a \emph{one shot} learning problem-- only one sample per class is assumed, which is the very sample given at inference when asked to determine whether or not it is a match.

Why is it important to understand specifics in settings followed for \gls{fr} evaluations? How does it being a \emph{one shot} problem change anything? Well, the answers directly relate to overarching goal of a \gls{fr} system-- to learn a mapping that encodes face images as features in such a way that best separates samples of different classes (\ie identities), while bringing those of the same identity closer together. Hence, a deep model for \gls{fr} is typically referred to as a face encoder, for the model serves as a feature extractor, and with the encoding as a representation of the face. A pair of faces is encoded and compared to one another via a distance or similarity metric, where the output of the metric determines whether or not the pair is a true match or not (\ie does the similarity score surpass a threshold that acts as a decision boundary between \emph{genuine} and \emph{imposter} pairs). This is also the reason many propose Siamese networks~\cite{koch2015siamese} as a solution--\glspl{nn} that share weights for a pair of inputs, which, like in our case, can later be compared in the setting of a verification task. Furthermore, we want a deep face encoder that projects high dimensional images to a highly discriminate feature space (\ie compactness intra-class and separation inter-class).


\begin{figure}[t!]
    \centering
    \includegraphics[width=.85\linewidth]{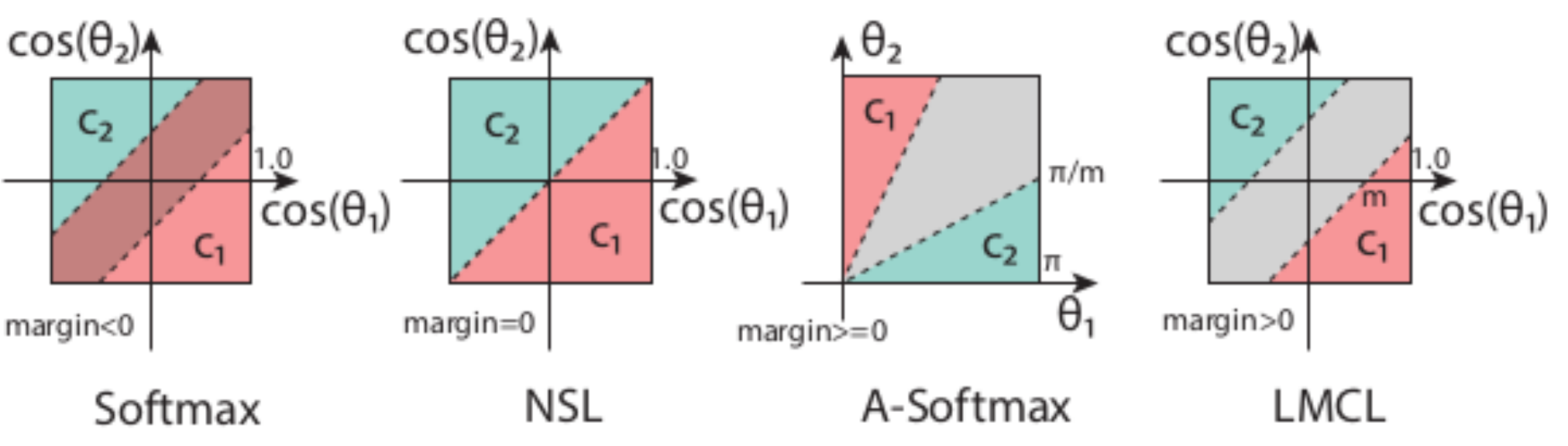}
    \caption{\textbf{Learned spaces (visualization from~\cite{wang2018cosface}).} These schematics are derived from the respective loss function-- \emph{left-to-right}: traditional softmax, NSL, ArcFace, and LMCL.}
\end{figure}

\subsection{Loss functions}

Like in many \gls{ml}-based solutions over the past several years is the unquestioned successes with deep learning in automatic \gls{fr} systems. Inherently, much of the success owes itself to other machine vision solution spaces-- network backbones often founded in a more generic problem statement and traditional object were mostly adapted in \gls{sota} \gls{fr} systems. All the while, it was the loss function that had evolved for face-based model training.

\subsubsection{Softmax}
Softmax is amongst the most popular losses used in deep learning-- map the output of the final, topmost layer maps the signal to a score-space by activating the \emph{logits scores} to produce a vector in probability space (\ie per mathematical axioms that formalize probability theory). 
Hence, it is here in the deep network that the classification layers are set (\ie deep learning encapsulates feature extraction and modeling training as a single module, with the features are derived from the input signal at the bottom of the network to then classify from the topmost layers). The output of the layer just below the softmax passes the raw scores, then the vector is mapped as a probability vector, which is then compared to the input during training via cross-entropy to produce a loss to back-propagate. Specifically, cross-entropy compares the predicted to the true in classification by comparing the normalized vector to a one-hot encoding representative of the class instance for the sample. Mathematically, given $N$ training samples of paired data (\ie image $x$ and label $y$),                                                                                        
\begin{equation}
L_s=\frac{1}{N}\sum_{i=1}^{N}-log(p_i)=\frac{1}{N}\sum_{i=1}^{N}-log(\frac{e^{f_{y_{i}}}}{\sum_{j=1}^Ce^{f_{y_{j}}}}),
\end{equation}

where the posterior probability $p$ is conditioned on the predicted class $c_j$ defined by $C=\mathrm{R}^k$ for $|C|=K$. Then, $f$, the activation of the topmost fully-connected layer (\ie the raw score). Hence, the purpose of the normalized exponential mapping is the softmax operation, while a comparison of bits with that of the true label vector (\ie one-hot encoding) and the predicted (\ie normalized between zero and one via softmax) determines the loss that the model parameters are adjusted to minimize.

As emphasized earlier, the underlying goal of encoding faces is to do so such that faces of the same person are close, while faces of different folks are spread far apart. For this, a fair expectation would be an objective that explicitly pushes samples of different classes apart while pulling those of the same closer together-- neither is done with the softmax loss. Also, it is susceptible to favoring instances that closely mimics the classes with a majority during training. In other words, imbalanced classes could yield a bias system-- a phenomena that is especially of concern in biometrics (more on this in \chapref{chap:bias}).

Scaled by a weight vector $W$ and without a bias term (\ie bias set to zero),

\begin{equation}
    \hat{y}_j = W_j^Tx=||W_j||||x||\cos\theta_j.
\end{equation}

Hence, the posterior $p$ is effected by the angle $\theta$ (\ie angle between $W$ and $x$). For encodings that have a invariant to the norm of $W$ so the norm is set constant via $L^1$ and $L^2$ normalization-- during inference, the resulting similarity score of a pair of features only depend on cosine similarity, so with the norm fixed during training the following loss can be determined:

$$
L_{ns}=\frac{1}{N}\sum_{i}-log(\frac{e^{s\cos(\theta_{{y_i},i})}}{\sum_{j}e^{s\cos(\theta_{j,i})}}.
$$

Now, with the fixed norm $s$ yields normalized features separable that are separable in angular space--a loss known as the Normalized Version of Softmax Loss (NSL).





\chapter{Face Detection}
\label{chap:facedetection}
\glsreset{aflw}
\section{Overview}\label{chap:facedetection:overview}
In landmark detection the task is to find the pixel locations in visual media corresponding to points of interest. In face alignment, these points correspond to face parts. For bodies and hands, landmarks correspond to projections of joints on the camera plane~\cite{supancic2015depth, wang2019ev}. Historically, landmark detection and shape analysis tasks date back decades: from Active Shape Models~\cite{cootes1992active} to Active Appearance Models~\cite{cootes2001active}, with the latter proposed to analyze and detect facial landmarks.

Like many other \gls{ml}-based problems, the task of landmark detection was regained attention with the advancement of deep learning-- models capable of encapsulating increasingly tricky views. In other words, the high capacity of deep learning models revamped interest in facial landmark localization-- one of the older problems researched in \gls{fr}~\cite{chellappa1995human}. As a result in came a wave of different types of deep neural architectures that pushed \gls{sota} on more challenging datasets. These modern-day networks are trained end-to-end on paired labeled data $(\image, \mathbf{s})$, where $\image$ is the image and $\mathbf{s}$ are the actual landmark coordinates. Many of these used encoder-decoder style networks to generate feature maps (\ie heatmaps) to transform into pixel coordinates~\cite{newell2016stacked,peng2018red, yang2017stacked}. The network must be entirely differentiable to train end-to-end. Hence, the layer (or operation) for transforming the $K$ heatmaps to pixel coordinates must be differentiable~\cite{honari2018improving}. Note that each of the $K$ heatmaps corresponds to the coordinates of a landmark. Typically, the $\sam$ operation determines the location of a landmark as the expectation over the generated 2D heatmaps. Thus, metrics like L1 or L2 determine the distance between the actual and predicted coordinates $\tilde{\mathbf{s}}$, \ie $\mathbf{e}=\tilde{\mathbf{s}} - \mathbf{s}$.

There are two critical shortcomings of the methodology discussed above. (1) These losses only penalize for differences in mean values in coordinate space, and with no explicit penalty for the variance of heatmaps. Thus, the generated heatmaps are highly scattered: high variance means low confidence. (2) This family of objectives is entirely dependent on paired training samples ($(\image, \mathbf{s})$). However, obtaining high-quality data for this is expensive and challenging. Not only does each sample require several marks, but unintentional, and often unavoidable, labels are of pixel-level marks subject to human error (\ie inaccurate and imprecise ground-truth labels). All the while, plenty of unlabeled face data are available for free. 

\begin{figure}[t!]
    \centering
    \includegraphics[width=.58\linewidth]{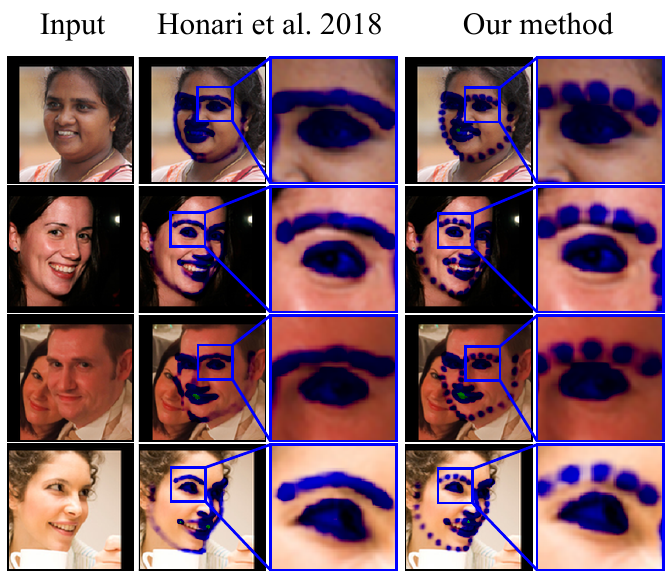}
    \caption{\textbf{Problem statement.} Heatmaps generated by SAM-based models (middle block) and the proposed $\lkl$ (right block), each with heatmaps on the input images (left) and a zoomed-in view of an eye region (right). These heatmaps are confidence scores (\ie probabilities) that a pixel is a landmark. Softargmax-based methods generate highly scattered mappings (low certainty), while the same network trained with our loss is concentrated (\ie high certainty). We further validate the importance of minimizing scatter experimentally (\tabref{chap:facedetection:tab:face-comparisons}). Best if viewed electronically.} 
    \label{chap:facedetection:fig:kl-versus-softargmax}
\end{figure}

\section{Research Contributions}\label{chap:facedetection:contributions}
We proposed a practical framework to satisfy the two shortcomings~\cite{robinson2019laplace}. Specifically, our first contribution alleviates the problem of an unaccounted for spread . For this, we introduce a new loss function that penalizes for the difference in distribution defined by location and scatter (\figref{chap:facedetection:fig:kl-versus-softargmax}). Independently, we treat landmarks as random variables with $\textrm{Laplace}(\mathbf{s}, 1)$ distributions, from which the KL-divergence between the predicted and ground-truth distributions defines the loss. Hence, the goal is to match distributions, parameterized by both a mean and variance, to yield heatmaps of less scatter (\ie higher confidence). We call this objective the $\lkl$ loss.

Our second contribution in landmark detection was an adversarial training framework. We proposed a method that tackles the problem of paired data requirements by leveraging unlabeled data accessed for free. We treat our landmark detection network as a \gls{g} of normalized heatmaps (\ie probability maps) that pass to the \gls{d} to learn to distinguish between the real and fake heatmaps. We could then add large amounts of unlabeled data to further boost the performance of our $\lkl$-based models. In the end, \gls{d} improved the predictive power of the $\lkl$-based model by injecting unlabeled data into the pipeline during training. Experiments demonstrate that our adversarial training framework complements the proposed $\lkl$ loss (\ie an increase in unlabeled data results in a decrease in error). We first show the effectiveness of the proposed $\lkl$ loss by claiming \gls{sota} (\ie first-place) on labeled set and second-to-best (\ie second-place) without the adversarial training. We then record results for the adversarial training scheme (\ie leveraging increasing amounts of unlabeled data).Finally, we further improve results using our $\lkl$ loss with more unlabeled data added during training!

Furthermore, we reduced the size of the model by using $\frac{1}{16}$, $\frac{1}{8}$, $\frac{1}{4}$, and $\frac{1}{2}$ the original number of convolution filters, with the smallest costing only 79 Kb on disk. We show an accuracy drop for models trained with the proposed $\lkl$ as far less than the others trained with a $\sam$-based loss. So again, it is the case that more unlabeled training data results in less of a performance drop at reduced sizes. It is essential to highlight that variants of our model at or of larger size than $1/8$ the original size compare well to the existing \gls{sota}. We claim that the proposed contributions are instrumental for landmark detection models used in real-time production, mobile devices, and other practical purposes.

Our contributions are three-fold: (1) A novel Laplace KL-divergence objective to train landmark localization models that are more certain about predictions; (2) An adversarial training framework that leverages large amounts of unlabeled data during training; (3) Experiments that show our model outperforms recent works in face landmark detection, along with ablation studies that, most notably, reveal our model compares well to \gls{sota} at $1/8$ its original size (\ie $<$160 Kb) and in real-time (\ie$>$20 fps).

\section{Background Information}
\label{chap3:sec:relatedwork}
In this section, we review relevant works on landmark localization and \gls{gan}.

\subsection{Landmark localization} 
As mentioned, landmark localization (or detection) has been of interest to researchers for decades. At first, most methods were based on Active Shape Models~\cite{cootes1992active} and Active Appearance Models~\cite{cootes2001active}. Then, Cascaded Regression Methods (CRMs) were introduced, which operate sequentially; starting with the average shape, then incrementally shifting the shape closer to the target shape. CRMs offer high speed and accuracy (\ie $>$1,000 fps on CPU~\cite{ren2014face,kazemi2014one}). 

More recently, deep-learning-based approaches have prevailed in the community due to end-to-end learning and improved accuracy. Initial works mimicked the iterative nature of cascaded methods using recurrent convolutional neural networks~\cite{peng2018red, trigeorgis2016mnemonic,wang2016recurrentaccv,wang2018recurrentpami}. Besides, there have been several methods for dense landmark localization~\cite{guler2017densereg,jeni2015dense} and 3D face alignment~\cite{tulyakov2018consistent, zhu2016face} proposed: all of which are fully-supervised and, thus, require labels for each image.

Nowadays, there is an increasing interest in semi-supervised methods for landmark localization. Recent work used a sequential multitasking method which was capable of injecting labels of two types into the training pipeline, with one type constituting the annotated landmarks and the other type consisting of facial expressions (or hand-gestures)~\cite{honari2018improving}. The authors argued that the latter label type was more easily obtainable, and showed the benefits of using both types of annotations by claiming \gls{sota} on several tasks. Additionally, they explore other semi-supervised techniques (\eg equivariance loss). In~\cite{dong2018supervision}, a supervision-by-registration method was proposed, which significantly utilized unlabeled videos for training a landmark detector. The fundamental assumption was that the neighboring frames of the detected landmarks should be consistent with the optical flow computed between the frames. This approach demonstrated a more stable detector for videos, and improved accuracy on public benchmarks. 

Landmark localization data resources have significantly evolved as well, with the 68-point mark-up scheme of the MultiPIE dataset~\cite{gross2010multi} widely adopted. Despite the initial excitement for MultiPIE throughout the landmark localization community~\cite{zhu2012face}, it is now considered one of the easy datasets captured entirely in a controlled lab setting. A more challenging dataset, \gls{aflw}~\cite{koestinger2011annotated}, was then released with up to 21 facial landmarks per face (\ie occluded or ``invisible'' landmarks were not marked). Finally, came the 300W dataset made-up of face images from the internet, labeled with the same 68-point mark-up scheme as MultiPIE, and promoted as a data challenge~\cite{sagonas2013300}. Currently, 300W is among the most widely used benchmarks for facial landmark localization. In addition to 2D datasets, the community created several datasets annotated with 3D keypoints~\cite{bulat2017far}.

\subsection{GANs} were recently introduced~\cite{goodfellow2014generative}, quickly becoming popular in research and practice. \Glspl{gan} have been used to generate images~\cite{radford2015unsupervised} and videos~\cite{saito2017temporal,tulyakov2017mocogan}, and to do image manipulation~\cite{geng20193d}, text-to-image\cite{han2017stackgan}, image-to-image~\cite{zhu2017unpaired}, video-to-video~\cite{wang2018vid2vid} translation and re-targeting~\cite{siarohin2018animating}.

An exciting feature of \Glspl{gan} is the ability to transfer visual media across different domains. Thus, various semi-supervised and domain-adaptation tasks adopted \Glspl{gan}~\cite{ding2018one, hoffman2017cycada, shrivastava2017learning, yang20183d}. Many have leveraged synthetic data to improve model performance on real data. For example, a \gls{gan} transferred images of human eyes from the real domain to bootstrap training data~\cite{shrivastava2017learning}. Other researchers used them to synthetically generate photo-realistic images of outdoor scenes, which also aided in bettering performance in image segmentation~\cite{hoffman2017cycada}. Sometimes, labeling images captured in a controlled setting is manageable (\ie versus an uncontrolled setting). For instance, 2D body pose annotations were available \textit{in-the-wild}, while 3D annotations mostly were for images captured in a lab setting. Therefore, images with 3D annotations were used in adversarial training to predict 3D human body poses as seen \textit{in-the-wild}~\cite{yang20183d}. \cite{ding2018one} formulated one-shot recognition as a data imbalance problem and augmented additional samples in the form of synthetic embeddings.

Our work differs from these others in several ways. Firstly, a majority, if not all, used a training objective that only accounts for the location of landmarks~\cite{honari2018improving,trigeorgis2016mnemonic,wang2018recurrentpami}, \ie no consideration for variance (\ie confidence). Thus, landmarks distributions have been assumed to be describable with a single parameter (\ie a mean). Networks trained this way yield an uncertainty about the prediction, while still providing a reasonable location estimate. To mitigate this, we explicitly parametrize the distribution of landmarks using location and scale. For this, we propose a KL-divergence based loss to train the network end-to-end. Secondly, previous works used \Glspl{gan} for domain adaptation in some fashion. In this work, we do not perform any adaptation between domains as in~\cite{hoffman2017cycada, shrivastava2017learning}, nor do we use any additional training labels as in~\cite{honari2018improving}. Specifically, we have \gls{d} do the quality assessment on the predicted heatmaps for a given image. The resulting gradients are used to improve the ability of the generator to detect landmarks. We show that both contributions improve accuracy when used separately. Then, the two contributions together boost \gls{sota} results.

\begin{figure}
    \centering
    \includegraphics[width=\textwidth]{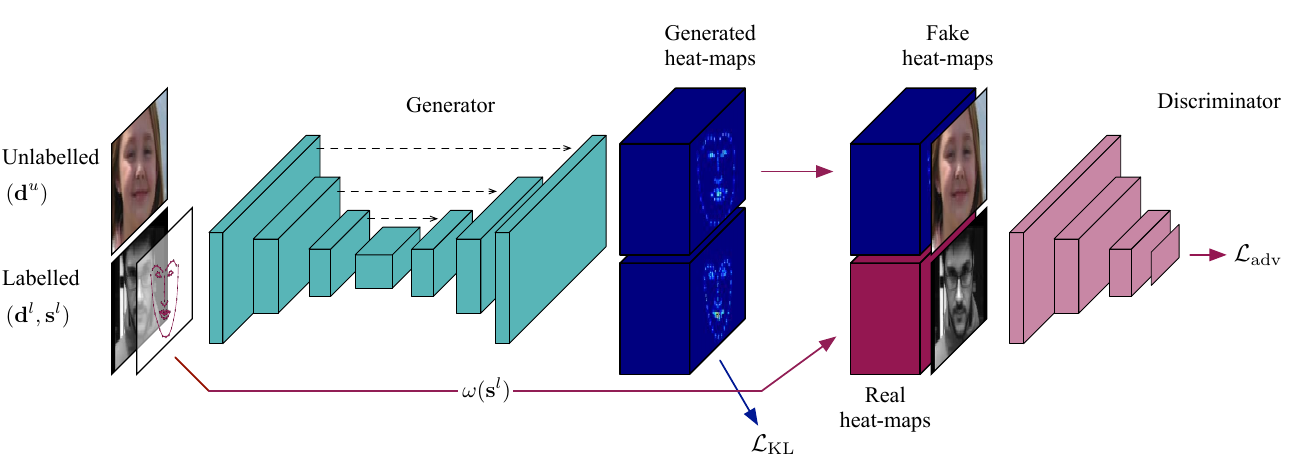}
    \caption{\textbf{Our semi-supervised framework for landmark detection.} The labeled and unlabeled branches are marked with \textcolor{blue}{blue} and \textcolor{red}{red} arrows, respectfully. Given an input image, \gls{g} produces $K$ heatmaps, one for each landmark. Labels are used to generate real heatmaps as~$\omega(\mathbf{s}^l)$. \gls{g} produces fake samples from unlabeled data. Source images are concatenated on heatmaps and passed to \gls{d}.}
    \label{chap:facedetection:fig:framework}
\end{figure}

\section{Laplace Landmark Localizer}
\label{chap3:sec:method}
Our training framework utilizes both labeled and unlabeled data during training. Shown in \figref{chap:facedetection:fig:framework} are the high-level graphical depiction of cases where labels are available (blue arrows) and unavailable (red arrows). Notice the framework has two branches, supervised (Eq.~\ref{eq:kl-loss}) and unsupervised (Eq. \ref{eq:kl-gan}), where only the supervised (blue arrow) uses labels to train. Next, are details for both branches.

\subsection{Fully Supervised Branch}
\label{sec:with-labels}
We define the joint distribution of the image $\image \in \mathbb{R}^{h \times w \times 3}$ and landmarks $\mathbf{s} \in \mathbb{R}^{K\times2}$ as $p(\image, \mathbf{s})$, where $K$ is the total number of landmarks. The form of the distribution $p(\image, \mathbf{s})$ is unknown; however, joint samples are available when labels are present (\ie $(\image, \mathbf{s}) \sim p(\image, \mathbf{s})$). During training, we aim to learn a conditional distribution $q_{\theta}(\mathbf{s} | \image)$ modeled by a neural network with parameters $\theta$. Landmarks are then detected done by sampling $\tilde{\mathbf{s}} \sim q_\theta(\mathbf{s}|\image)$. We now omit parameters $\theta$ from notation for cleaner expressions. The parameter values are learned by maximizing the likelihood that the process described by the model did indeed produce the data that was observed, \ie trained by minimizing the following loss function w.r.t. its parameters: 

\begin{equation}
    \mathcal{L}(\theta) = \mathbb{E}_{(\image, \mathbf{s}) \sim p(\image, \mathbf{s})} \| \tilde{\mathbf{s}} - \mathbf{s} \|_2.
    \label{eq:l2-loss}
\end{equation}
Alternatively, it is possible to train a neural network to predict normalized probability maps(\ie heatmaps): $\tilde{\mathbf{h}} \sim q(\mathbf{h} | \image)$, where $\mathbf{h} \in \mathbb{R}^{K \times h \times w}$ and each $\mathbf{h}_k \in \mathbb{R}^{h\times w}$ represents a normalized probability map for landmark $k$, where $k=1\dots K$. To get the pixel locations, one could perform the argmax operation over the heatmaps by setting 
$\tilde{\mathbf{s}}= \textrm{argmax}(\tilde{\mathbf{h}})$). However, this operation is not differentiable and, therefore, unable to be trained end-to-end.

A differentiable variant of argmax (\ie $\sam$~\cite{chapelle2010gradient}) was recently used to localize landmarks~\cite{honari2018improving}. For the 1D case, the $\sam$ operation is expressed 

\begin{equation}
    \begin{aligned}
        \textrm{softargmax}(\beta\mathbf{h})
             & =  \sum_x \textrm{softmax}(\beta  \mathbf{h}_x) \cdot x  \\
             & =  \sum_x \frac{e^{\beta  \mathbf{h}_x}}{\sum_j e^{\beta  \mathbf{h}_j}} \cdot x \\
             & =  \sum_x p(x) \cdot x = \mathbb{E}_\mathbf{h}[x],
        \label{eq:softargmax-loss}
    \end{aligned}
\end{equation}

\noindent where $\mathbf{h}_x$ is the predicted probability mass at location $x$, $\sum_j e^{\beta  \mathbf{h}_j}$ is the normalization factor, and $\beta$ is the temperature factor controlling the predicted distribution~\cite{chapelle2010gradient}. Coordinates are in boldface (\ie $\mathbf{x}=(x_1, x_2)$), and write 2D $\sam$ operation as $\tilde{\mathbf{s}}=\mathbb{E}_\mathbf{h}[\mathbf{x}]$ with $\mathcal{L}_\mathrm{SAM}=\mathcal{L}(\theta)$.

Essentially, the $\sam$ operation is the expectation of the pixel coordinate over the selected dimension. Hence, the $\sam$-based loss assumes the underlying distribution is describable by just its mean (\ie location), regardless of how sure a prediction, the objective then is to match mean values. To avoid cases in which the trained model is uncertain about the predicted mean, while still yielding a low error, we parameterize the distribution using $\{\mu, \sigma\}$, where $\mu$ is the mean or the location and $\sigma$ is the variance or the scale, respectfully, for the selected distribution.

We want the model to be certain about the predictions (\ie a small variance or scale). We consider two parametric distributions~$\textrm{Gaussian}(\mu, \sigma)$ and $\textrm{Laplace}(\mu, b)$ with $\sigma^2=\mathbb{E}_\mathbf{h}[(\mathbf{x} - \mathbb{E}_\mathbf{h}[\mathbf{x}])^2]$ and $b=\mathbb{E}_\mathbf{h}[|\mathbf{x} - \mathbb{E}_\mathbf{h}[\mathbf{x}]|]$. We define a function $\tau(\tilde{\mathbf{h}})$ to compute the scale (or variance) of the predicted heatmaps $\tilde{\mathbf{h}}$ using the location, where the locations are now the expectation of being a landmark in the heatmap space. Thus, $\tau(\tilde{\mathbf{h}}) = \sum p(\mathbf{x}) ||\mathbf{x} - \tilde{\mathbf{s}}||^\alpha_\alpha$, where $\tilde{\mathbf{s}}=\mathbb{E}_\mathbf{h}[\mathbf{x}]$, $\alpha=1$ for Laplacian, and $\alpha=2$ for Gaussian. Thus, $\tilde{\mathbf{s}}$ and $\tau(\tilde{\mathbf{h}})$) are used to parameterize a Laplace (or Gaussian) distribution for the predicted landmarks $q(\mathbf{h} | \image)$.

\begin{algorithm}[t]
\SetAlgoLined
\KwData{${\{ (\mathbf{d}_i^l, \mathbf{s}_i^l) \}_{i=1,...,n}}$,
        ${\{ (\mathbf{d}_i^u) \}_{i=1,...,m}}$

}
$\mathbf{\theta}_D, \mathbf{\theta}_G \leftarrow$ initialize network parameters\

\While{$t \leq T$}{
  $(\mathbf{D}_t^l, \mathbf{S}_t^l) \leftarrow$  sample mini-batch from labeled data\
  
  $(\mathbf{D}_t^u) \leftarrow$ sample mini-batch from unlabeled data\
  
  $\mathbf{H}_\mathrm{fake} \leftarrow G(\mathbf{D}_t^u)$\
  
  $\mathbf{H}_\mathrm{real} \leftarrow \omega(\mathbf{S}_t^l)$\
  
  $\mathcal{L}_\mathrm{adv} \leftarrow \log D([\mathbf{D}_t^l, \mathbf{H}_\mathrm{real}]) + \log( 1 - D([\mathbf{D}_t^u, \mathbf{H}_\mathrm{fake}])$
  
  $\mathcal{L}_\mathrm{G} \leftarrow$ compute loss using Eq.~\ref{eq:softargmax-loss} or Eq.~\ref{eq:kl-loss} 
  
  \vspace{0.2cm}
  
  \tcp{update model parameters}
  
  $\mathbf{\theta}_D \xleftarrow{+} - \nabla_{\mathbf{\theta}_D} \mathcal{L}_\mathrm{adv} $
  
  $\mathbf{\theta}_G \xleftarrow{+} - \nabla_{\mathbf{\theta}_G} ( \mathcal{L}_\mathrm{G} - \lambda \mathcal{L}_\mathrm{adv} )$
 }
 \caption{Training the proposed model.}
 \label{alg:training-kl}
\end{algorithm}

With the true conditional distribution of the landmarks as $p(\mathbf{s}|\image)$, the objective is
\begin{equation}
    \begin{aligned}
        \mathcal{L}_\mathrm{KL} = \mathbb{E}_{(\image, \mathbf{s}) \sim p(\image, \mathbf{s})} \Big [ \mathrm{D}_{KL} (q(\mathbf{s} | \image) || p(\mathbf{s}| \image)) \Big ],
        \label{eq:kl-loss}
    \end{aligned}
\end{equation}
where $\mathrm{D_{KL}}$ is the KL-divergence. We assumed a true distribution for the case of Gaussian (\ie $\textrm{Gaussian}(\mathbf{\mu} , 1)$, where $\mu$ is the ground-truth locations of the landmarks). For the case with Laplace, we sought $\textrm{Laplace}(\mathbf{\mu}, 1)$. KL-divergence conveniently has a closed-form solution for this family of exponential distributions~\cite{hoffman2013stochastic}. Alternatively, sampling yields an approximation. The blue arrow in \figref{chap:facedetection:fig:framework} represent the labeled branch of the framework.

\begin{figure}[t!]
    \centering
    \includegraphics[width=2.5in, trim={0in 9.65in 0in 0in},clip]{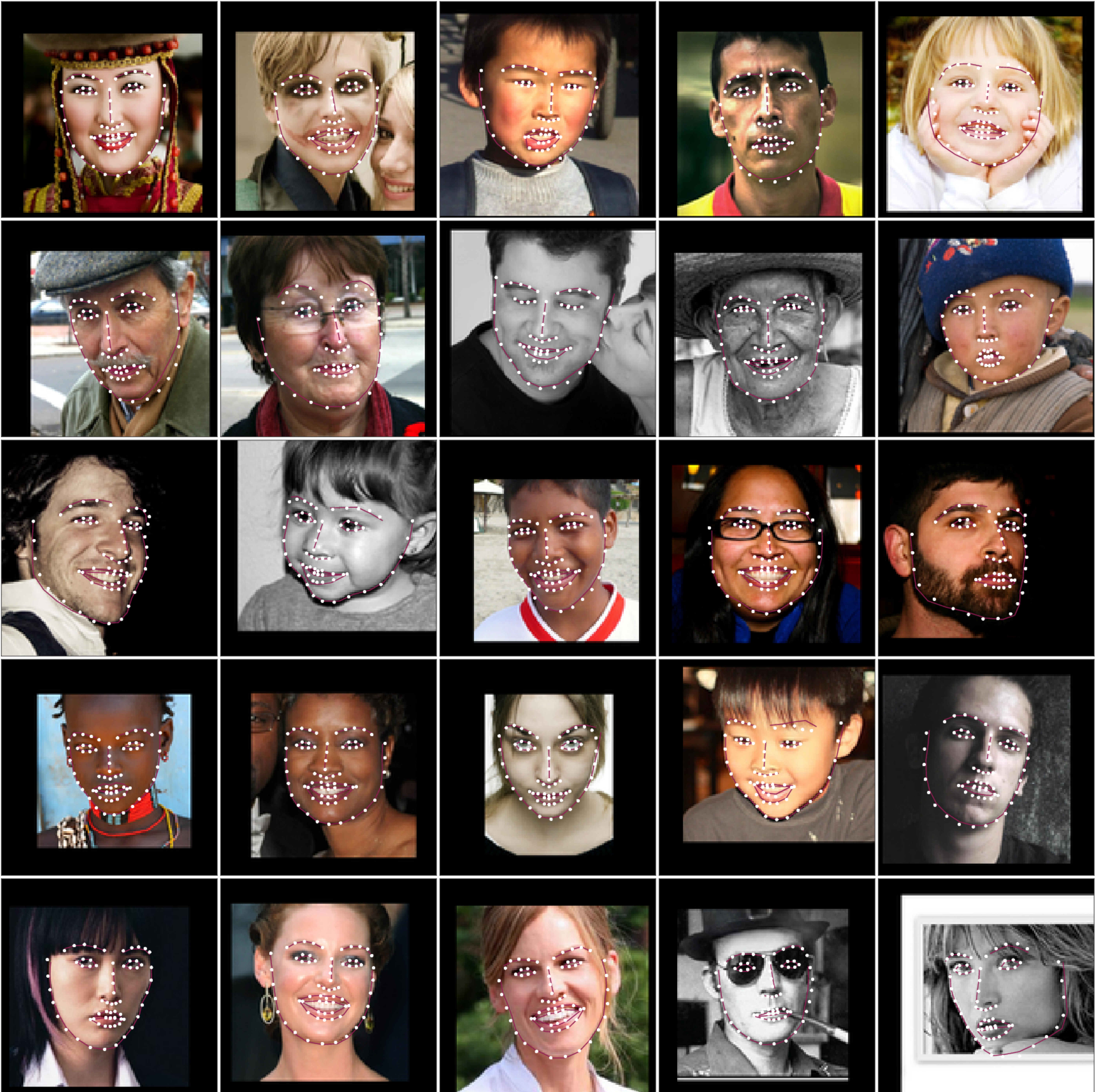}
    \caption{\textbf{Qualitative results.} Random samples of landmarks predicted using $\lkl$ (white), with the ground truth drawn as line segments (\textcolor{red}{red}). Notice the predicted points tend to overlap with the ground-truth. Best viewed in color. Zoom-in for greater detail.}
    \label{chap:facedetection:fig:faces}
\end{figure}

Statistically speaking, given two estimators with different variances, we would prefer one that has a smaller variance (see \cite{domingos2000unified} for an analysis of the bias-variance trade-off). A lower variance implies higher confidence in the prediction. To this end, we found an objective measuring distance between distributions is accurate and robust. The neural network must satisfy an extra constraint on variance and, thus, yields predictions of higher certainty. See higher confident heatmaps in \figref{chap:facedetection:fig:kl-versus-softargmax} and \figref{chap:facedetection:fig:qualitative-figure}. The experimental evaluation further validates this (Table \ref{chap:facedetection:tab:face-comparisons} and Table \ref{chap:facedetection:tab:model-sizes}). Also, \figref{chap:facedetection:fig:faces} shows sample results.

\subsection{Unsupervised Branch}\glsreset{d}\glsreset{g}
\label{sec:no-labels}
The previous section discusses several objectives to train the neural network with the available paired or fully labeled data (\ie $(\image^l, \mathbf{s}^l)$). We denote data samples with the superscript $l$ to distinguish them from unpaired or unlabeled data $(\image^u)$. In general, it is difficult for a human to label many images with landmarks. Hence, unlabeled data are abundant and easier to obtain, which calls for capitalizing on this abundant data to improve training. In order to do so, we adapt the adversarial learning framework for landmark localization. We treat our landmarks predicting network as a \gls{g}, $G=q(\mathbf{h} | \image)$; \gls{d} takes the form $D([ \image, \mathbf{h} ])$, where $[\cdot, \cdot]$ is a tensor concatenation operation . We define the real samples for \gls{d} as $\{\image^l, \mathbf{h}= \omega(\mathbf{s}^l)\}$, where $\omega(\cdot)$ generates the true heatmaps given the locations of the ground-truth landmarks. Fake samples are given by $\{ \image^u, \tilde{\mathbf{h}} \sim q(\mathbf{h} | \image^u) \}$. We then define the min-max objective for landmark detection as:

\begin{eqnarray}
    \min_{G} \max_{D} \mathcal{L}_\mathrm{adv}(D,G),
    \label{eq::gan-problem}
\end{eqnarray}
where $\mathcal{L}_\mathrm{adv}(D,G)$ writes as:

\begin{eqnarray}
        \!\!\!&  &\!\!\! \mathbb{E}_{(\image^l, \mathbf{s}^l) \sim p(\image, \mathbf{s})} \Big [\log D([\image^l, \omega(\mathbf{s}^l)]) \Big ] + \nonumber\\
        \!\!\!&   &\!\!\!\mathbb{E}_{(\image^u) \sim p(\image)} \Big [\log( 1 - D([\image^u, G(\image^u))]) \Big ].
\end{eqnarray}

With this, provided an input image, the goal of \gls{d} is to learn to decipher between the real and fake heatmaps from appearance. The goal of \gls{g} is to produce fake heatmaps that closely resemble the real. Within this framework, \gls{d} intends to provide additional guidance for \gls{g} by learning from labeled and unlabeled data. The objective, Eq.~\ref{eq::gan-problem}, is solved using alternating updates.

\begin{table}
\glsreset{g}
\centering
	\caption{\small\textbf{Architecture of \gls{g}.} Layers with size and number of filters (\ie $h$ $\times$ $w$ $\times$ $n$). DROP, MAX, and UP are dropout (probability 0.2), max-pooling (stride 2), and bilinear upsampling (2$x$), respectively. Note, skip connections about the bottleneck: coarse-to-fine, connecting encoder ($E_{ID}$) to decoder ($D_{ID}$) by concatenating feature channels and fusing. Number of feature preserved at all but top two layers (\ie transform $\rightarrow$ features $\rightarrow$ $K$ heatmaps). Padded to match sizes listed.}

    \tiny
    \begin{tabular}{lll}\toprule

		{} & \textbf{Layers} & \textbf{Tensor Size}\\\midrule
	    Input          & RGB image, no data augmentation                & 80 x 80 x 3 \\
		Conv($E_1$)    & 3 $\times$ 3 $\times$ 64, LReLU, DROP, MAX                   & 40 $\times$ 40 $\times$ 64\\
        Conv($E_2$)    & 3 $\times$ 3 $\times$ 64, LReLU, DROP, MAX                   & 20 $\times$ 20 $\times$ 64\\
        Conv($E_3$)    & 3 $\times$ 3 $\times$ 64, LReLU, DROP, MAX                   & 10 $\times$ 10 $\times$ 64\\
        Conv($E_4$)    & 3 $\times$ 3 $\times$ 64, LReLU, DROP, MAX                   & 5 $\times$ 5 $\times$ 64\\
        Conv($D_4$)    & 1 $\times$ 1 $\times$ 64 $+E_4$, LReLU, DROP, UP      & 10 $\times$ 10 $\times$ 128\\
        Conv($D_F$)    & 5 $\times$ 5 $\times$ 128, LReLU                             & 20 $\times$ 20 $\times$ 128\\
        Conv($D_3$)     & 1 $\times$ 1 $\times$ 64 $+E_3$, LReLU, DROP, UP       & 20 $\times$ 20 $\times$ 128\\
        Conv($D_F$)    & 5 $\times$ 5 $\times$ 128, LReLU, DROP                       & 40 $\times$ 40 $\times$ 128\\
        Conv($D_2$)    & 1 $\times$ 1 $\times$ 64 $+E_2$, LReLU, DROP,  UP      & 40 $\times$ 40 $\times$ 128\\
        Conv($D_F$)    & 5 $\times$ 5 $\times$ 128, LReLU, DROP                       & 80 $\times$ 80 $\times$ 128 \\
        Conv($D_1$)    & 1 $\times$ 1 $\times$ 64 $+E_1$, LReLU, DROP, UP     & 80 $\times$ 80 $\times$ 128\\
        Conv($D_F$)    & 5 $\times$ 5 $\times$ 128, LReLU, DROP                       & 80 $\times$ 80 $\times$ 128\\
        Conv($D_F$)    & 1 $\times$ 1 $\times$ 68, LReLU, DROP                        & 80 $\times$ 80 $\times$ 68\\
        Output         & 1 $\times$ 1 $\times$ 68                                     & 80 $\times$ 80 $\times$ 68\\
\bottomrule
    \end{tabular}
\label{chap:facedetection:tab:arch}
\end{table}
\subsection{Training}
\label{sec:training}
We fused the $\sam$-based and adversarial losses as 
\begin{eqnarray}
       \min_G \Big ( \max_D \big ( 
                \lambda \cdot \mathcal{L}_\mathrm{adv}(G, D)
            \big )
            + \mathcal{L}_\mathrm{SAM}(G)
        \Big ),
\label{eq:sam-gan}
\end{eqnarray}
with the KL-divergence version of the objective defined as:
\begin{eqnarray}
       \min_G \Big ( \max_D \big ( 
                \lambda \cdot \mathcal{L}_\mathrm{adv}(G, D)
            \big )
            + \mathcal{L}_\mathrm{KL}(G)
        \Big ),
\label{eq:kl-gan}
\end{eqnarray}
with the weight for the adversarial loss $\lambda=0.001$. This training objective includes both labeled and unlabeled data in the formulation. In the experiments, we show that this combination significantly improves the accuracy of our approach. We also argue that the $\sam$-based version cannot fully utilize the unlabeled data since the predicted heatmaps differ too much from the \textit{real} heatmaps. See Algorithm~\ref{alg:training-kl} for the training procedure for $T$ steps of the proposed model. We show the unlabeled branch of the framework is shown graphically in red arrows (\figref{chap:facedetection:fig:framework}).

\subsection{Implementation}
\label{sec:implementation}
We follow the ReCombinator network (RCN) initially proposed in~\cite{honari2016recombinator}. Specifically, we use a 4-branch RCN as our base model, with input images and output heatmaps of size 80$\times$80. Convolutional layers of the encoder consist of 64 channels, while the convolutional layers of the decoder output 64 channels out of the 128 channels at its input (\ie 64 channels from the previous layer concatenated with the 64 channels skipped over the bottleneck via branching). We applied Leaky-ReLU, with a negative slope of 0.2, on all but the last convolution layer. See \tabref{chap:facedetection:tab:arch} for details on the generator architecture. Drop-out followed this, after all but the first and last activation. We use Adam optimizer with a learning rate of 0.001 and weight decay of $10^{-5}$. In all cases, networks were trained from scratch, with no data augmentation or 'training tricks.' 

\gls{d} was a 4-layered PatchGAN~\cite{isola2017image}. Before each convolution layer Gaussian noise ($\sigma=0.2$) was added~\cite{tulyakov2017mocogan}, and then batch-normalization (all but the top and bottom layers) and Leaky-ReLU with a negative slope of 0.2 (all but the top layer). The original RGB image was stacked on top of the $K$ heatmaps from \gls{g} and fed as the input of \gls{d} (\figref{chap:facedetection:fig:framework}). Thus, \gls{d} takes in ($K$ + 3) channels. We set $\beta=1$ for \ref{eq:softargmax-loss}. Pytorch was used to implement the entire framework. An important note to make is that models optimized with Laplace distribution consistently outperformed the Gaussian-based. For instance, our $\lkl$ baseline had a \gls{nmse} of 4.01 on 300W, while Gaussian-based got 4.71. Thus, the sharper,``peakier'' Laplace distribution proved to be more numerically stable under current network configuration, as Gaussian required a learning rate of a magnitude smaller to avoid vanishing gradients. Indeed, we used Laplace.

\begin{table}[!t]
\centering

\caption{\textbf{Quantitative results}. \gls{nmse} on \gls{aflw} and 300W normalized by the square root of BB area and interocular distance, respectfully.}
\label{chap:facedetection:tab:face-comparisons}
    \tiny
 \begin{tabular}{rccccc}
     &\textbf{\gls{aflw}}&& \multicolumn{3}{c}{\textbf{300W}} \\
      &                 &            & \textbf{Common}        & \textbf{Challenge}   & \textbf{Full}              \\
SDM~\cite{xiong2013supervised}            & 5.43      &                  & 5.57          & 15.40         & 7.52              \tabularnewline
LBF~\cite{ren2014face}                    & 4.25       &                 & 4.95          & 11.98         & 6.32         		\tabularnewline
MDM~\cite{trigeorgis2016mnemonic}         & -           &                & 4.83          & 10.14         & 5.88         		\tabularnewline
TCDCN~\cite{zhang2014facial}                & -          &                 & 4.80          & 8.60          & 5.54        		\tabularnewline
CFSS~\cite{zhu2015face}                    & 3.92         &               & 4.73          & 9.98          & 5.76         		\tabularnewline
CFSS~\cite{lv2017deep}                     & 2.17          &              & 4.36          & 7.56          & 4.99         		\tabularnewline
RCSR~\cite{wang2018recurrentpami}          & -             &              & 4.01          & 8.58          & 4.90         		\tabularnewline
RCN+ (L$+$ELT)~\cite{honari2018improving}            & \textbf{1.59}    &           & 4.20          & 7.78          & 4.90         		\tabularnewline
CPM $+$ SBR~\cite{dong2018supervision}            & 2.14        &                & 3.28          & 7.58          & 4.10     	    	\tabularnewline\midrule
$\Sam$                                & 2.26                &        & 3.48          & 7.39          & 4.25    		    \tabularnewline
$\Sam$+D(10K)                         & -                    &       & 3.34          & 7.90          & 4.23     		    \tabularnewline
$\Sam$+D(30K)                         & -                   &        & 3.41          & 7.99          & 4.31     		    \tabularnewline
$\Sam$+D(50K)                         & -               &            & 3.41          & 8.06          & 4.32     		    \tabularnewline
$\Sam$+D(70K)                         & -               &            & 3.34          & 8.17          &  4.29     		    \tabularnewline\midrule
$\lkl$~\cite{robinson2019laplace}                                & 1.97            &            & 3.28          & 7.01          & 4.01     			\tabularnewline
$\lkl$+D(10K)                         & -               &            & 3.26          & 6.96          & 3.99            \tabularnewline
$\lkl$+D(30K)                         & -               &            & 3.29          & 6.74          & 3.96              \tabularnewline
$\lkl$+D(50K)                         & -               &            & 3.26          & \textbf{6.71} & 3.94              \\
$\lkl$+D(70K)                         & -               &            & \textbf{3.19} & 6.87          & \textbf{3.91}     \\\bottomrule
\end{tabular}
\end{table}

\section{Experiments}

\label{sec:experiments}
We evaluated the proposed on two widely used benchmark datasets for face alignment. No data augmentation techniques used when training our models nor was the learning rate dropped: this leaves no ambiguity into whether or not the improved performance came from training tricks or the learning component itself. All results for the proposed were from models trained for 200 epochs.

We next discuss the metric used to evaluate performance, \gls{nmse}, with differences between datasets in the normalization factor. Then, the experimental settings, results, and analysis for each dataset are covered separately. Finally, ablation studies show characterizations of critical hyper-parameters and, furthermore, the robustness of the proposed $\lkl$+D(70K) with a comparable performance with just $1/8$ the number of feature channels and $>$20 fps.

\subsection{Metric}
Per convention~\cite{sagonas2013300, bulat2017far, cristinacce2006feature}, \gls{nmse}, a normalized average of euclidean distances, was used. Mathematically speaking: 

\begin{equation}
    \text{NMSE} = \sum_{k=1}^{K} \frac{\|s_k - \tilde{s}_k\|_2}{K\times d}, 
    \label{eqn:NMSE}
\end{equation}
where the number of visible landmarks set as $K$, $k=\{1,2,...,K\}$ are the indices of the visible landmark, the normalization factor $d$ depends on the face size, and $s_k \in \mathbb{R}^2$ and $\tilde{s}_k \in \mathbb{R}^2$ are the ground-truth and predicted coordinates, respectfully. The face size $d$ ensured that the \gls{nmse} scores across faces of different size were fairly weighted. Following predecessors, \gls{nmse} was used to evaluate both datasets, except with different points referenced to calculate $d$. The following subsections provide details for finding $d$.

\begin{figure}
\centering
    \begin{tabular}{p{.28in}p{.27in}p{.75in}p{.3in}p{.27in}p{.3in}p{.1in}}
        &\scriptsize\rotatebox{15}{L-KL+D(70K)}&\scriptsize\rotatebox{15}{SAM+D(80K)}&\scriptsize\rotatebox{15}{L-KL+D(70K)}&\scriptsize\rotatebox{15}{SAM+D(80K)}&  \\
    \multicolumn{7}{c}{
    \includegraphics[width=.85\linewidth, trim={1.75mm 0mm 0.0mm 0mm},clip]{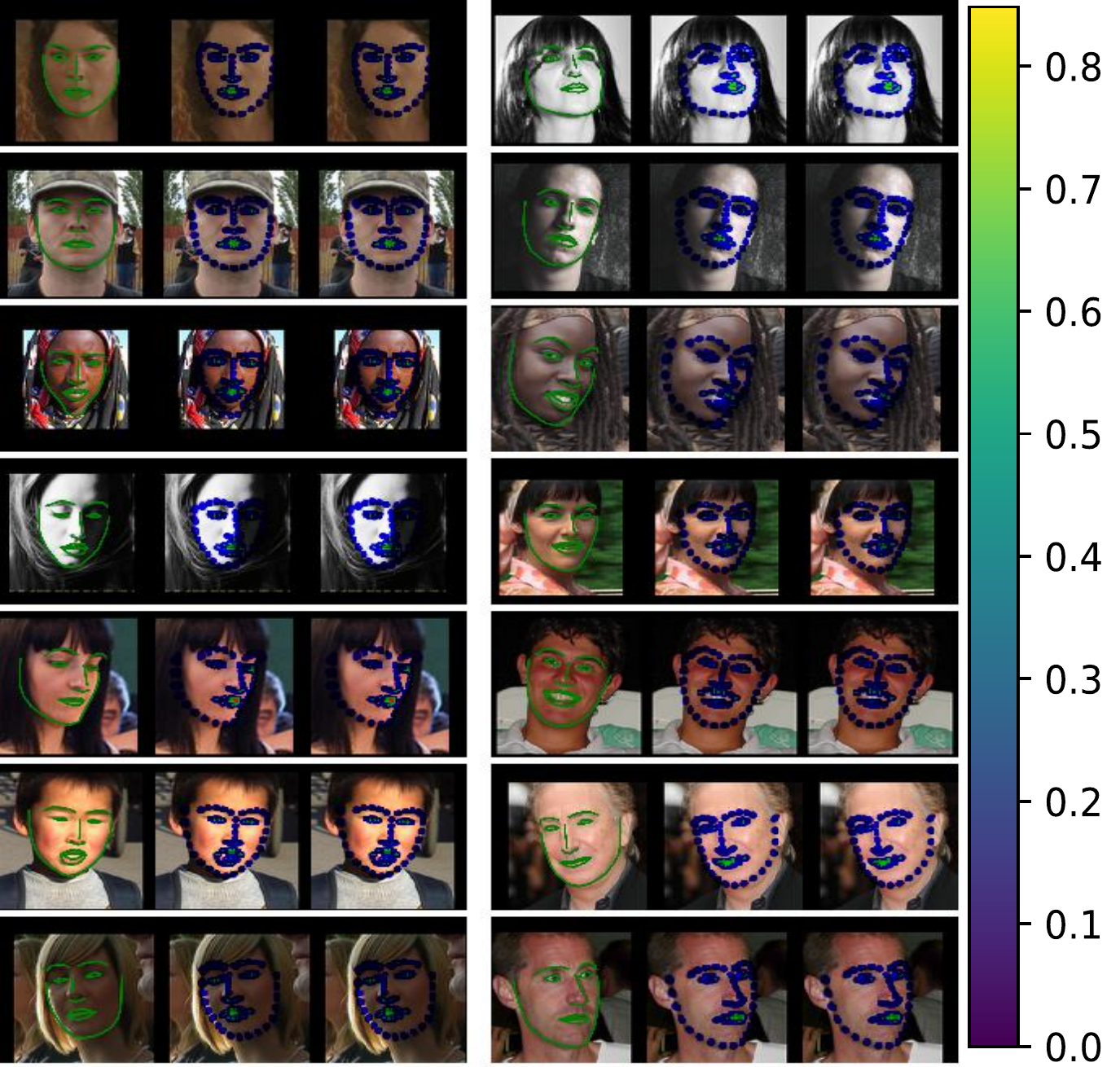}
    }
    \end{tabular}
    \caption{\textbf{Random samples (300W).} Heatmaps predicted by our $\lkl$+D(70K) (middle, \ie L-KL+D(70K)) and $\sam$+D(70K) (right, \ie SAM+D(70K)) alongside face images with ground-truth sketched on the face (left). For this, colors were set by value for the $K$ heatmaps generated for each landmark (\ie range of [0, 1] as shown in color bar), and then were superimposed on the original face. Note that the KL-divergence loss yields predictions of much greater confidence and, hence, produced separated landmarks when visualized heatmap space. In other words, the proposed has minimal spread about the mean, as opposed to the $\sam$-based model with heatmaps with individual landmarks smudged together. Best viewed electronically. }\label{chap:facedetection:fig:qualitative-figure}
\end{figure}
\subsection{300W + MegaFace}
The 300W dataset is amongst the most popular datasets for face alignment. It has 68 visible landmarks (\ie $K=68$) for 3,837 images (\ie 3,148 training and 689 test). We followed the protocol of the 300W challenge~\cite{sagonas2013300} and evaluated using \gls{nmse} (Eq.~\ref{eqn:NMSE}), where $d$ is set as the interocular distance (\ie distance between outer corners of the eyes). Per convention, we evaluated different subsets of 300W (\ie \textit{common} and \textit{challenge}, which together form \textit{full}).

We compared the performance of the proposed objective trained in a semi-supervised fashion. During training, 300W dataset made-up the labeled data (\ie \textit{real}), and a random selection from MegaFace provided the unlabeled data (\ie \textit{fake})~\cite{nech2017level}. MTCNN\footnote{\url{https://github.com/davidsandberg/facenet}} was used to detect five landmarks (\ie eye pupils, corners of the mouth, and middle of nose and chin)~\cite{zhang2016joint}, which allowed for similar face crops from either dataset. Specifically, we extended the square hull that enclosed the five landmarks by 2$\times$ the radii in each direction. In other words, the smallest bounding box spanning the 5 points (\ie the outermost points lied on the parameter), and then transformed from rectangles-to-squares with sides of length 2$\times\max(height, width)$. Note that the midpoint of the original rectangle was held constant to avoid shift translations (\ie rounded up a pixel if the radius was even and extended in all directions).

The $\lkl$+D(70K) model obtained \gls{sota} on 300W, yielding the lowest error on 300W (\tabref{chap:facedetection:tab:face-comparisons}~(300W columns)). $\lkl$ and $\sam$ with $N$ unlabeled faces, s+D($N$). denote the models trained with unlabeled data, where $N$ representing the number of unlabeled images added from MegaFace.

First, notice that $\lkl$ trained without unlabeled data still achieved \gls{sota}. The $\lkl$-based models then showed relative improvements with more unlabeled data added. The $\sam$-based models cannot fully take advantage of the unlabeled data without minimizing for variance (\ie generates heatmaps of less confidence and, thus, more spread). Our $\lkl$, on the other hand, penalizes for spread (\ie scale), making the job of \gls{d} more challenging. As such, $\lkl$-based models benefit from increasing amounts of unlabeled data.

Also, notice the largest gap between the baseline models~\cite{dong2018supervision} and our $\lkl$+D(70K) model on the different sets of 300W. Adding more unlabeled helps more (\ie $\lkl$ versus $\lkl$+D(70K) improvement is about 2.53\%). However, it is essential to use samples not covered in the labeled set. To demonstrate this, we set the \textit{real} and \textit{fake} sets to 300W (\ie $\image^l = \image^u$ in the second term of Eq.~\ref{eq:kl-gan}). \gls{nmse} results for this experiment are listed as follows: $\lkl$+D(300W) 4.06 (baseline-- 4.01) and $\sam$+D(300W) 4.26 (baseline-- 4.24). As hypothesized, all the information from the labeled set had already been extracted in the supervised branch, leaving no benefit of using the same set in the unsupervised branch. Therefore, more unlabeled data yields more hard negatives to train with, which improves the accuracy of the rarely seen samples (\tabref{chap:facedetection:tab:face-comparisons}~(300W \textit{challenge} set)). Our best model was $\approx$2.7\% better than~\cite{dong2018supervision} on easier samples (\ie \textit{common}), $\approx$4.7\% better on average (\ie \textit{full}), and, moreover, $\approx$9.8\% better on the more difficult (\ie \textit{challenge}), $\approx$4.7\% better on average (full), and, moreover, $\approx$9.8\% better on the more difficult (challenge). These results further highlight the advantages of training with the proposed $\lkl$ loss, along with the adversarial training framework.

Additionally, the adversarial framework boosted our 300W baseline (\ie more unlabeled data yields a lower \gls{nmse}). Specifically, we demonstrated this by pushing \gls{sota} of the proposed on 300W from a \gls{nmse} of 4.01 to 3.91 (\ie no unlabeled data to 70K unlabeled pairs, respectfully). There were boosts at each step size of \textit{full} (\ie larger $N$ $\rightarrow$ \gls{nmse}).

\begin{table}[t!]
\centering
\caption{\textbf{Ablation.} \gls{nmse} on 300W (full set) for networks trained with fewer channels in each convolutional layer by $1/16$, $1/8$, $1/4$, $1/2$, and unmodified in size (\ie the original) listed from left-to-right. We measured performance with a 2.8GHz Intel Core i7 CPU.}
\label{chap:facedetection:tab:model-sizes}
\centering
    \begin{tabular}{lccccc}\toprule

            & \multicolumn{5}{c}{Number of parameters, millions}\vspace{.2mm}\tabularnewline 
            \cline{2-6}
                                      &      $0.0174$                        & $0.0389$ 
                                         & $0.1281$
                                         & $0.4781$   & $1.8724$
                                      \tabularnewline\midrule
$\Sam$                                &         9.79               &             6.86 &  4.83      & 4.35 & 4.25     		    \tabularnewline
$\Sam$+D(70K)                         &    9.02                        &           6.84  & 4.85  &   4.38  & 4.29      	\tabularnewline	    
$\lkl$                                &          7.38            &      5.09    &  4.39    & 4.04 & 4.01     			\tabularnewline
$\lkl$+D(70K)                         &                    \textbf{7.01}       & \textbf{4.85} &       \textbf{4.30}  &  \textbf{3.98} & \textbf{3.91}   
\tabularnewline\midrule\midrule
Storage (MB) &0.076 & 0.162 & 0.507& 1.919&7.496 \tabularnewline
Speed (fps) & 26.51 &21.38 &16.77& 11.92 & 4.92
\tabularnewline\bottomrule
    \end{tabular}
\end{table}

We randomly sampled the unlabeled for $\lkl$+D(70K) and $\sam$+D(70K) to visualize predicted heatmaps (\figref{chap:facedetection:fig:qualitative-figure}). In each case, the heatmaps produced by the softargmax-based models spread wider, explaining the worsened quantitative scores (\tabref{chap:facedetection:tab:face-comparisons}). Our loss and adversarial learning scheme yield higher probable pixel location (\ie a more concentrated predicted heatmaps). For most images, the heatmaps generated by models trained with the $\lkl$ loss have distributions for landmarks of more confidence and properly distributed: $\lkl$+D(70K) yielded heatmaps that vary $\pm$1.02 pixels from the mean, while $\sam$+D(70K) has a variation of $\pm$2.59. Learning the landmark distributions with our $\lkl$ loss is conceptually and theoretically intuitive (\figref{chap:facedetection:fig:kl-versus-softargmax}), and experimentally proven (\tabref{chap:facedetection:tab:face-comparisons}).

\subsection{The Annotated Facial Landmarks in the Wild (AFLW) dataset}
We evaluated the $\lkl$ loss on the \gls{aflw} dataset~\cite{koestinger2011annotated}. \gls{aflw} contains 24,386 faces with up to 21 landmarks annotations and 3D head pose labels. Following~\cite{honari2018improving}, 20,000 faces were used for training with the other 4,386 for testing. We ignored the two landmarks for the left and right earlobes, leaving up to 19 landmarks per face~\cite{dong2018supervision}.

Since faces of \gls{aflw} have such variety head poses, most faces have landmarks out of view (\ie missing). Thus, most samples were not annotated with the complete 19 landmarks, meaning that it does not allow for a constant sized tensor (\ie \textit{real} heatmaps) for the adversarial training. Therefore, we compared the $\sam$ and KL-based objectives with existing \gls{sota}. The face size $d$ for the \gls{nmse} was the square root of the bounding box hull~\cite{bulat2017far}.

Our $\lkl$-based model was comparable to \gls{sota} (\ie RCN+ (L$+$ELT)~\cite{honari2018improving}) on the larger, more challenging \gls{aflw} dataset while outperforming all others. It is essential to highlight here that ~\cite{honari2018improving} puts great emphasis on data augmentation, while we do not apply any. Also, since landmarks are missing in some samples (\ie no common reference points exist across all samples), we were unable to prepare faces for our semi-supervised component-- a subject for future work.

\begin{figure}[t!]
    \centering
    \includegraphics[width=0.38\linewidth, trim={1.0mm 0mm .5mm 0mm},clip]{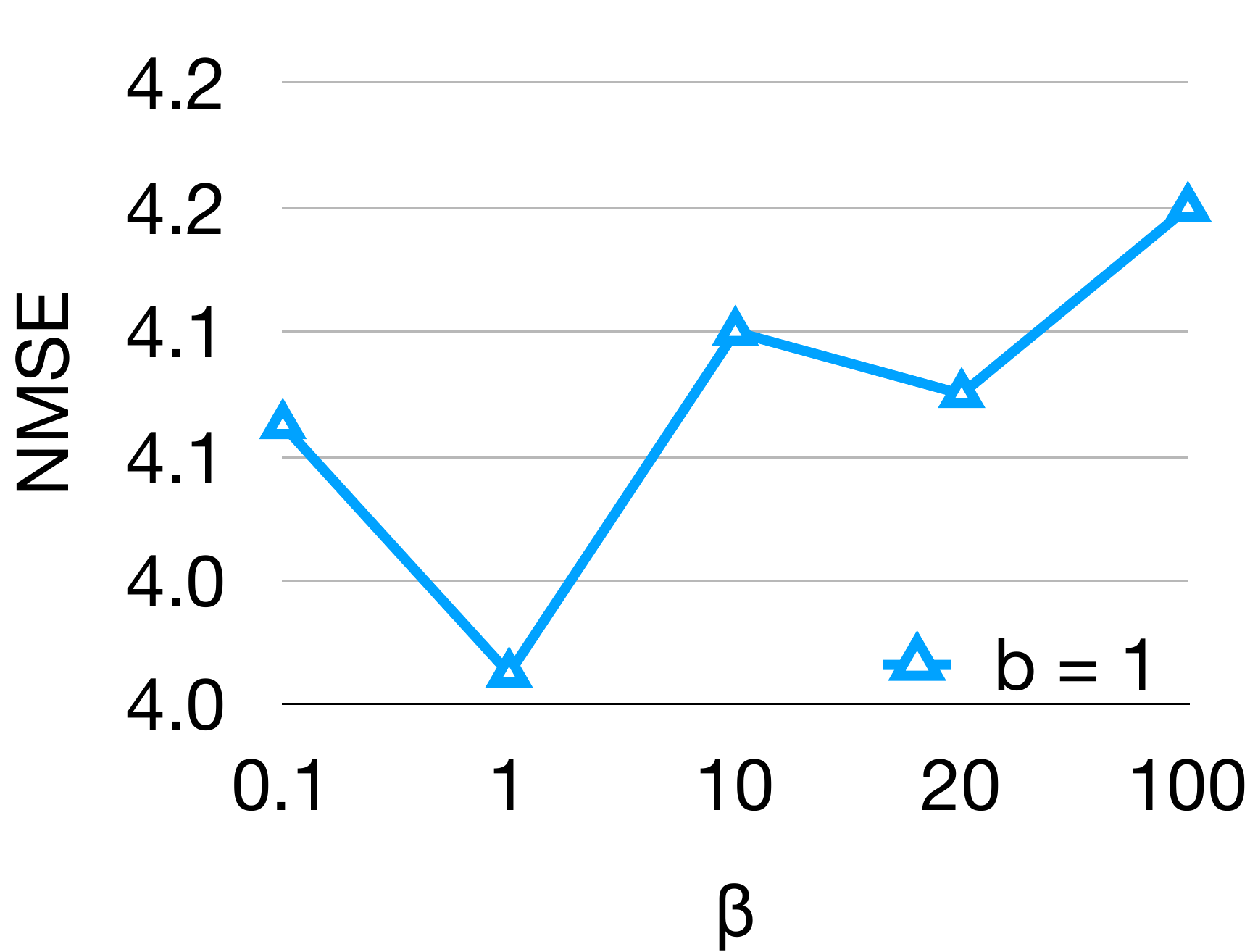}
    \includegraphics[width=0.38\linewidth, trim={1.0mm 0mm 0.5mm 0mm},clip]{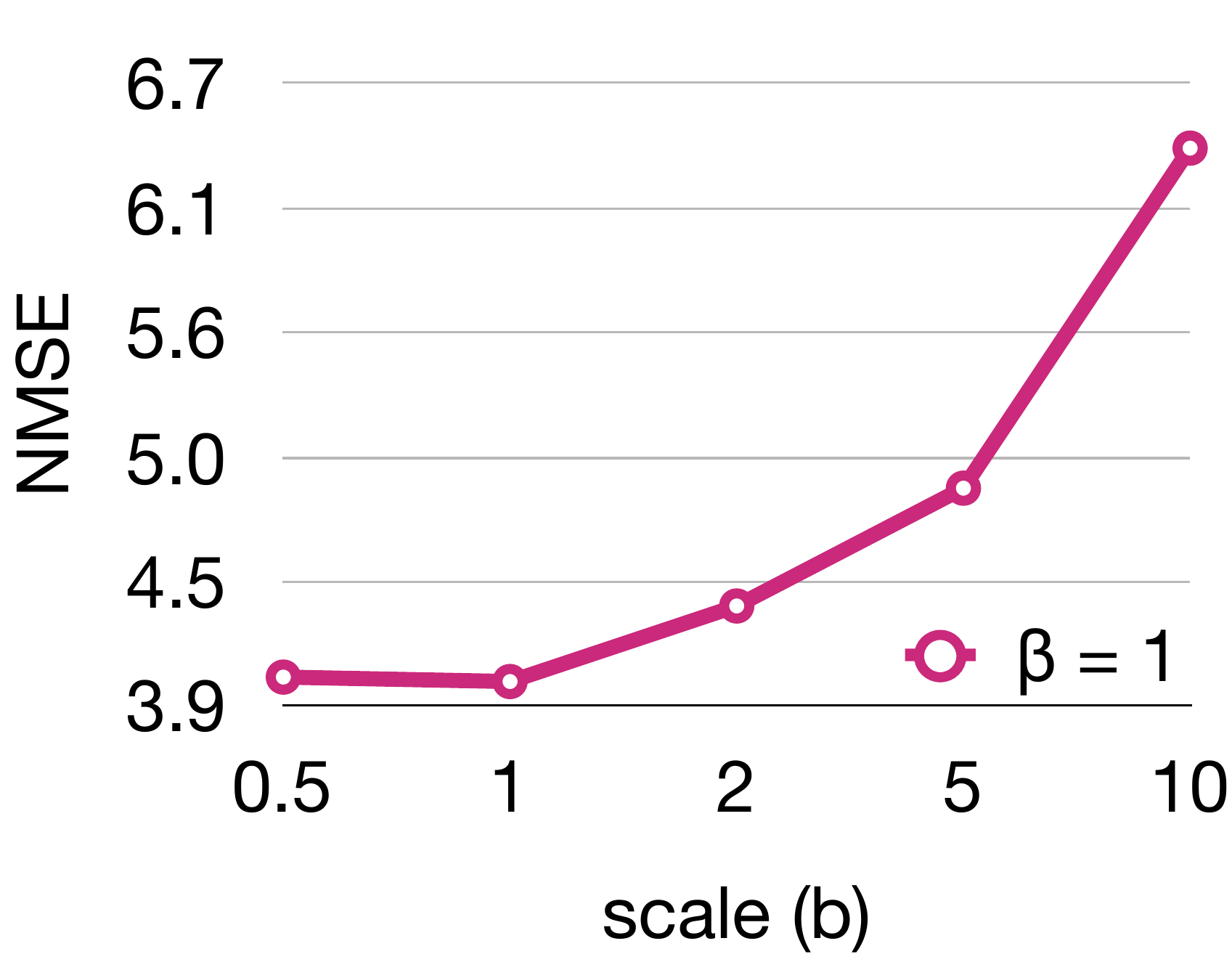}

    \caption{\textbf{Ablation.} Results of ablation study on $\lkl$.}
    \label{chap3:fig:ablation}
\end{figure}

\subsection{Ablation Study}
The error is next measured as a function of model size (\tabref{chap:facedetection:tab:model-sizes}), along with different $\beta$ values (Eq.~\ref{eq:softargmax-loss}) and scales $b$ used to parameterize the Laplacian (\figref{chap3:fig:ablation}). The latter characterizes the baseline and supports the values used for these hyper-parameters, while the former reveals a critical characteristic for the practicality of the proposed. 

Specifically, we decreased the model size by reducing the number of channels at each convolutional layer by factors of 2. The $\sam$-based model worsened by about 13\% and 59\% in \gls{nmse} at a $\frac{1}{4}$  and $\frac{1}{8}$ the channel count, respectfully (\ie 4.25 $\rightarrow$ 6.86 and 9.79). $\lkl$, on the other hand, decreased by about 24\% with an $8^{th}$ and 59\% with a $16^{th}$ the number of channels (\ie 4.01 $\rightarrow$ 5.09 and 7.38, respectfully). Our model trained with unlabeled data (\ie $\lkl$+D(70K)) dropped just about 21\% and 57\% at factors of 8 and 16, respectfully (\ie 3.91 $\rightarrow$ 4.85 and 7.01). $\lkl$+D(70K) proved best with reduced sizes: with $<$0.040M parameters, it still compares to previous \gls{sota}~\cite{honari2018improving,wang2018recurrentpami, lv2017deep}, which is a clear advantage. For instance, SDM~\cite{xiong2013supervised}, requires 1.693M parameters (25.17MB)\footnote{\url{https://github.com/tntrung/sdm\_face\_alignment}} for 7.52 in \gls{nmse} (300W \textit{full}); our smallest and next-to-smallest got 7.01 and 4.85 with 0.174M (0.076 MB) and 0.340M (0.166 MB) weights.

The model also speeds up with fewer channels (\ie to train and at inference). For instance, the model reduced by a factor of 16 processes 26.51 frames per second (fps) on a CPU of Macbook Pro (\ie 2.8GHz Intel Core i7), with the original running at 4.92 fps. Our best $\lkl$-based model proved robust to size reduction: 4.85 \gls{nmse} at 21.38 fps when reduced by $1/8$.

\section{Discussion}
\label{chap3:sec:conclusion}
We demonstrated the benefits of the proposed $\lkl$ loss and leveraging unlabeled data in an adversarial training framework. Hypothetically and empirically, we showed the importance of penalizing a landmark predictor's uncertainty. Thus, training with the proposed objective yields predictions of higher confidence, outperforming previous \gls{sota} methods. We also revealed the benefits of adding unlabeled training data to boost performance via adversarial training. In the end, our model performs \gls{sota} on all three splits of the renown 300W (\ie \textit{common}, \textit{challenge}, and \textit{full}), and second-to-best on the \gls{aflw} benchmark. Also, we demonstrate the robustness of the proposed by significantly reducing the number of parameters. Specifically, with $1/8$ the number of channels (\ie $<$170Kb on disk), the proposed still yields an accuracy comparable to the previous \gls{sota} in real-time (\ie 21.38 fps). Thus, the contributions of the proposed framework are instrumental for models intended for use in real-world production.
\part{Visual Kinship Recognition of Families In the Wild}\label{part:rfiw}
\chapter{Visual Kinship Recognition}\label{chap:vkr}
\section{Overview}  
About a decade ago, pioneers in visual kinship recognition research published~the seminar work in detecting family relationships with face images~\cite{fang2010towards}. Before discussing our specific research contributions in kinship recognition technology (\chapref{chap:fiw}), we first look back at the last decade of research done by the community as a whole: review the key milestones that brought us to the state in technology for where we are today. Furthermore, let us highlight the key challenges, practical use-cases, and promising future directions for research. By doing so, we will have paved the way to diving into the details of the many related efforts:
\begin{itemize}
\item Our purpose and motivations;
\item The technical novelty and the ways at which it fits into the bigger picture;
\item Our hopes and beliefs for which the work we had done as part of this dissertation could and should be considered by other researchers;
\item Whether a junior scholar looking for a problem to hone in on as part of a dissertation;
\item Experts that have published in the automatic kinship recognition problem space.
\end{itemize}

In any case, there are great benefits to reap from the advancement of kinship recognition-- it has a multitude of practical and scholarly uses. Relationships provide rich information in sociology, anthropology, and genetics;  privacy protections and concerns, along with potential use-cases that can be found in social media, personal discovery, entertainment, and more. Besides its entrepreneurial value, visual kinship recognition has significant non-commercial (or humane) value as well. For instance, in cases of missing children, reconnecting families split across refugee camps, border control and customs, criminal investigations, ancestral-based studies, and even genome-based research. Socially, family gives a sense of belonging (\ie membership, connection). Per Furstenberg,

\begin{quote}
    ... important function of family systems receives far less attention in the literature than it merits: The family ... social arrangement responsible for giving its members a sense of identity and shared belonging ... not only those inside the natal family household but also among relations living elsewhere as well~\cite{furstenberg2020kinship}.
\end{quote}

Hence, a recent surge in many seeking out their pedigree. With an abundance of visual data online, familial resources can benefit.

\begin{sidewaysfigure}
\centering
    \includegraphics[width=.65\textwidth]{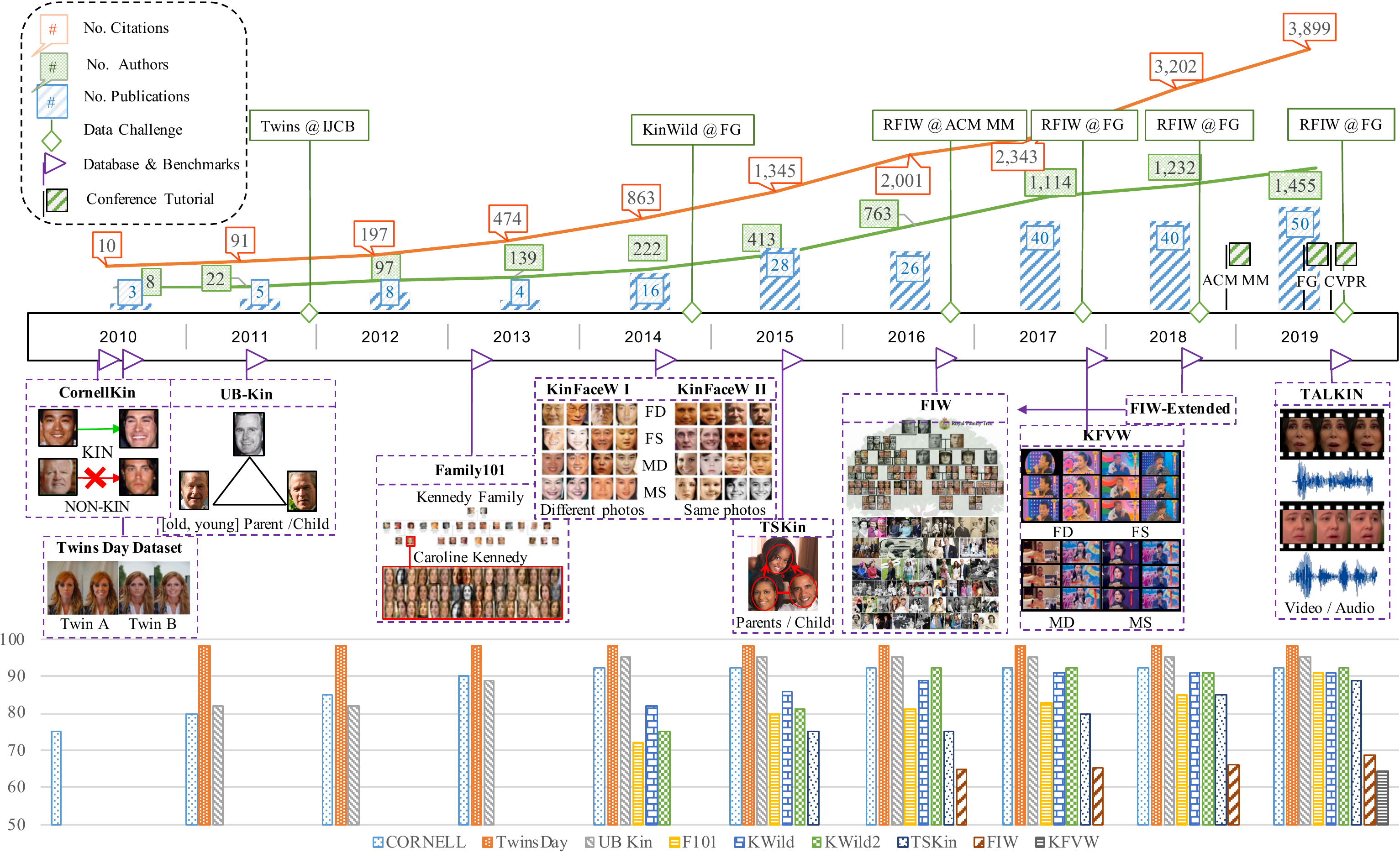}
    \caption{{\textbf{A decade of research in visual kinship recognition.} The timeline shows correlations between the data resources (\emph{below timeline}) and citation metrics and events indicating the amount of research impact (\emph{above timeline}). We built a pipeline to scrape the data needed for the plots above: (1) \emph{Publish or Perish}~\cite{publishperish} was installed on a Mac Book Pro to gather metadata for publications from various sources (\ie Google Scholar, Cross Ref, and Scorpus) into a CSV file; (2) metadata in CSV was parsed into a BIB file using Python; (3) \emph{Mendeley Reference Manager} was used to automatically detect duplicates while keeping as much information as possible by merging reference listings; (4) queried Google Scholar for all \emph{Related Works} and \emph{Cited By} using PyPi's scholarly (\href{https://pypi.org/project/scholarly/}{https://pypi.org/project/scholarly/}), which extended the paper-pile and increased the amount of metadata available from the richer metadata accessible using scholarly (\eg paper abstracts); (5) we clustered the documents by abstract via \gls{tfidf}~\cite{ramos2003using}. The clusters were high in recall, as true clusters were a majority of papers on kinship recognition in multimedia: this reduced the burden of manual inspection of hundreds of thousands to thousands. It is important to note that only citation metrics were considered, leaving out other factors of impact like the \emph{number of times tweeted}, \emph{Github stars}, and other indicators of impacting research.}}
    \label{chap:fiwmm:fig:timeline}
\end{sidewaysfigure}

\begin{figure}[!t]
    \centering
    \includegraphics[width=.8\linewidth]{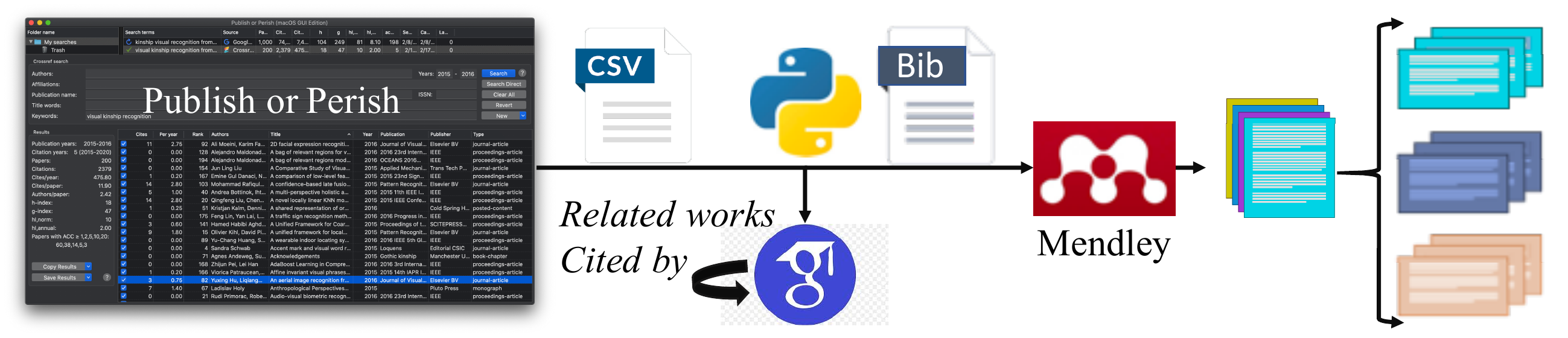}
    \caption{\textbf{Workflow to scrape publication metadata for \figref{chap:fiwmm:fig:timeline}.} From \emph{Publish or Perish}~\cite{publishperish}, we queried Scholar for \emph{Related works} and \emph{Cited by}, increasing the size of our list nearly 20-fold. Mendeley merged duplicates, while keeping as much information as possible. Applied \gls{nlp} to cluster relevant documents.}
    \label{chap:fiwmm:fig:citations}
\end{figure}

We now review the state of automatic kinship recognition after the first decade of research - with emphasis on the milestones that led us up to now (Fig.~\figref{chap:fiwmm:fig:citations}). Furthermore, we reflect on the problem statements of the different tasks to establish clear definitions and an understanding across the domain in a consistent manner. For this, we aim to use consistent terminology, assess the practical usefulness (or lack thereof), and highlight any outstanding challenges and obstacles that prevent the transition of visual kinship recognition technology from research-to-reality. With clear problem statements, and an established measure for the practical significance, we compile a list of the \gls{sota} scores and methods of the main tasks with emphasis on our large-scale \gls{fiw}, which will be revisited in the next chapter with great detail. 

In light of recent advances in our annual \gls{rfiw} data challenge, along with the new task evaluations added as part of the most recent 2020 edition, we review the details for the different paradigms of kin-based problems in the visual domain as formulated for our large-scale, multi-task \gls{fiw} database. We also look back at the existing datasets for visual kin-based problems motivated by different real-world scenarios (\tabref{chap:fiwmm:tab:datasets}). In summary, our overarching goal is to pour the foundation for a deeper understanding of visual kinship recognition problem domain. 

Even still, there are many unanswered questions that are high in potential for future efforts in machine vision research, like studies on familial and inheritance (\ie nature-based), and beyond. We take a glimpse at these promising next steps, while highlighting key challenges that we must overcome - both intrinsic to the image and inherent to the problem. We strongly urge there be attempts in research to establish cross-discipline studies-- we are so ever ready to form such synergy.

\begin{sidewaystable}
 \centering
    \caption{\textbf{Publicly available datasets for kinship recognition}. Each listed by the original name per reference. Kin-based image (or video) stats, which include the label types that support a specific evaluation metric and the respective \gls{sota} score. URLs to the project page of each data resource are included.  Abbreviations used for \emph{Stats} are for the family count (\textbf{F}), face count (\textbf{f}), number of unique people (\textbf{P}), sample count (\textbf{S}), image count (\textbf{I}), video count (\textbf{V}), and multimedia (\textbf{MM}).}\label{chap:fiwmm:tab:datasets}
    \tiny
         \resizebox{\textwidth}{!}{%
    \begin{tabular}{c|>{\centering\arraybackslash}p{.3in}|>{\centering\arraybackslash}p{1.3in}|>{\centering\arraybackslash}p{1.5in}|>{\centering\arraybackslash}p{1.1in}|>{\noindent\justifying\arraybackslash}p{1.12in}}
\toprule
        DB& Ref(s)& Stats & Label types & Metric, performance, \gls{sota} & Web\tabularnewline\midrule
        CornellKin
        &\cite{fang2010towards}& \textbullet~150~\textbf{F} \textbullet~300~\textbf{S} \textbullet~300~\textbf{f} &parent-child &verification accuracy  94.4\%~\cite{kohli2018supervised} & \href{http://chenlab.ece.cornell.edu/projects/KinshipVerification/}{chenlab.ece.cornell.edu}\tabularnewline \midrule
        
       UB Face &\cite{shao2011genealogical}~\cite{Xia201144}& \textbullet~200~\textbf{F} \textbullet~250~\textbf{P} \textbullet~600~\textbf{f} \textbullet~400~\textbf{S}& ([young, old] parent)-child &accuracy, 95.3\%~\cite{kohli2018supervised} &  \href{http://www1.ece.neu.edu/~yunfu/research/Kinface/Kinface.htm}{www1.ece.neu.edu/yunfu/}\tabularnewline
        \midrule
        
        \multirow{2}{*}{Twins Day} &  \multirow{2}{*}{\cite{vijayan2011twins}}& \textbullet~1,736~(finger, 3D face, iris, DNA)~\textbf{S} \textbullet~197~\textbf{I} & \multirow{2}{*}{twin pairs}& \multirow{2}{*}{accuracy, 98.8\%~\cite{vijayan2011twins}}& \href{https://biic.wvu.edu/data-sets/twins-day-dataset-2010-1015}{/twins-day-dataset-2010-1015}\tabularnewline\midrule
    
        SibFace & \cite{bottino2012new} &\textbullet~184~\textbf{S} \textbullet~78~\textbf{P} \textbullet~78~\textbf{F} \textbullet~184~\textbf{f}& siblings (brothers, sisters, mixed)&accuracy, 52.5\%~\cite{guo2014graph}&
       \href{https://areeweb.polito.it/ricerca/cgvg/siblingsDB.html}{areeweb.polito.it/ricerca/}\tabularnewline\midrule

       UvA-NEMO Smile  &\cite{dibeklioglu2013like} &\textbullet~162~\textbf{P} \textbullet~515~\textbf{V} \textbullet~512~\textbf{S}& 7 relationships (core family)&  accuracy, 88.16\%~\cite{boutellaa2017kinship}& \href{https://www.uva-nemo.org/}{https://www.uva-nemo.org/}\tabularnewline\midrule
        Family101 &\cite{fang2013kinship} &\textbullet~101~\textbf{F} \textbullet~607~\textbf{S}&family-tree structure&  rank@10, 70.1\%~\cite{wu2018kinship}& \href{http://chenlab.ece.cornell.edu/projects/KinshipClassification/index.html}{chenlab.ece.cornell.edu/}\tabularnewline\midrule

        KFW I + II &\cite{lu2015fg}&\textbullet~533 + 1,000~\textbf{P} \textbullet~1,066+2,000~\textbf{f} & parent-child; same + different photo& accuracy, 96.9\% + 97.1\%~\cite{kohli2018supervised} &\href{http://www.kinfacew.com/}{http://www.kinfacew.com/}\tabularnewline\midrule

        \multirow{1}{*}{TSKin} &  \multirow{1}{*}{\cite{qin2015tri}}&\multirow{1}{*}{\textbullet~787~\textbf{F} \textbullet~2,589~\textbf{S}} &\multirow{1}{*}{(father \& mother)-child}& \multirow{1}{*}{accuracy, 91.4\%~\cite{liang2018weighted}}&
        \href{http://parnec.nuaa.edu.cn/xtan/data/TSKinFace.html}{parnec.nuaa.edu.cn/TSKinFace} \tabularnewline\midrule

        \multirow{2}{*}{FIW}& \cite{ robinson2018visual, robinson2020recognizing} &\textbullet~1,000~\textbf{F} \textbullet~33,000~\textbf{f} \textbullet~1-M~\textbf{P} \textbullet12,000~\textbf{S} \textbullet~13,000~\textbf{I} &large-scale; person-, family-, and image-level& accuracy, 78\%; tri-subject accuracy, 79\%; mAP 18\% \& rank@5 60\%~\cite{robinson2020recognizing} &\href{https://web.northeastern.edu/smilelab/fiw/}{https://web.northeastern.edu/
        smilelab/fiw/}\tabularnewline\midrule
                
        KFVW &\cite{yan2018video} & \textbullet~418 (video) \textbf{P} (100-500 \textbf{f} per video);&  parent-child &
         accuracy, 61.8\%~\cite{yan2018video}& \href{https://www.kinfacew.com/index.html}{https://www.kinfacew.com}\tabularnewline\midrule

        KIVI &  \cite{kohli2018supervised} & \textbullet~503 (video) \textbf{S} \textbullet~503 \textbf{I} & 7 relationships (core family) &   accuracy, 83.2\%~\cite{kohli2018supervised} & \href{http://iab-rubric.org/resources/KIVI.html}{http://iab-rubric.org} \tabularnewline\midrule

         \multirow{2}{*}{FIW-MM} &    \multirow{2}{*}{\cite{robinson2020familiesinMM}} & \multirow{2}{*}{FIW + 937 \textbf{MM}} & {FIW + multimedia for $\approx$~200~\textbf{F} (\ie video, audio, and contextual data)} &   \multirow{2}{*}{EER,  89.8\%; mAP, 0.24~\cite{robinson2020familiesinMM}} & \href{https://web.northeastern.edu/smilelab/fiw/}{https://web.northeastern.edu/
        smilelab/fiw/} \tabularnewline
        
    \end{tabular}}
\end{sidewaystable}

The current chapter is of the form of a literature review in kinship recognition. There are a few related manuscripts that, together, span the material presented here. 

Firstly, is the first survey on visual kinship recognition published in 2016, which was an extensive overview of the \gls{sota} methods and data resources of the time~\cite{wu2016kinship}. The authors proposed future directions with great emphasis on the lack of labeled data both in sample counts and relationship label types. Hence, Wu~\etal claimed that traditional metric-based solutions were inferior to deep learning models (\eg \gls{cnn}), which has since shown to be true-- evidence in the current \gls{sota}, which we will introduce in the current chapter, with support and additional details provided in the chapters that proceed. Hence, the release of our large-scale \gls{fiw} dataset to support modern-day, data-driven solutions was introduced at the end of that very same year~\cite{FIW}. 

In 2018m Georgopoulos~\etal surveyed kinship and age in \gls{fr}~\cite{georgopoulos2018modeling}. Although a comprehensive piece, kin-based problems ought to be surveyed independently. Nonetheless, prior knowledge of one could benefit the other; knowledge of various soft biometrics tends to complement and are beneficial (\eg gender and emotion). On the one hand, a survey on age or kinship should mention the other; however, the directed graphs and concepts of inheritance make kin-based studies worthy of surveying as an independent topic. Finally, looking at the problem from the view of understanding age, we make a similar claim-- knowledge of kinship could most certainly help an age estimate. For instance, we have photos of a father at one or more known ages, while we are tasked to predict the age of the son. The knowledge available in the set of faces of known age is available as a prior and be modeled accordingly. Point is, we do not mean that the different attribute-based face understanding tasks ought to be treated as independent. Note, we do claim a study on the modeling and analysis of kin-based media should be surveyed alone. Still, as we cover later, age, gender, and variations of in other attributes do provide additional challenges in kin-based tasks.

More recently, Qin~\etal surveyed kinship recognition methods as being founded on \emph{a measure of kinship traits} or \emph{statistical learning}~\cite{qin2019literature}. Furthermore, the groups were characterized for being \emph{low}\ or \emph{mid-level features}, \emph{metric learning}, or \emph{transfer learning}. The authors reported scores for several kin-based datasets. Additionally, \emph{human} performance compared to machines was included. As part of their work, the authors proposed a standardized vision system based on four-steps to provide a generic, modular solution. To complement this, we define the problems consistently for the many kin-based tasks, and with details on the \gls{sota} for each.

Most recently, we published an extensive survey on the topic of automatic kinship understanding~\cite{robinsonKinsurvey2020}. In part, our survey was motivated by it being the 10 year anniversary since the first work in machine vision was proposed with several benchmarks and labeled data made available to the community for research purposes. On the other hand, our \gls{fiw} dataset attracted lots of attention the past few years for solution, and we review the major-milestones in research of automatic visual kinship understanding over the first decade. The discussion is supported by a detailed illustration to assess the problem as it evolved over time, along with the public data supporting the progress, and with data statistics and web links of the source. Furthermore, we look at kinship recognition research that compares humans to machines, showing resemblance is detectable via the human eye (Section~\ref{sec:background}). Next, we introduce kin-based tasks by discussing the different problem statements (Section~\ref{sec:data:benchmarks:resources}). 
Following this, we discuss experimental details for each of the tasks-- summarize the protocols of the laboratory-based evaluations, including the data splits, metrics, and baseline results for each (Section~\ref{sec:sota:visualkin}).  Then, we cover methodologies, both traditional and deep learning based for both discriminative and generative (Section~\ref{sec:methods}).

Then, we discuss technical challenges preventing this technology from working reliably in real-world applications. Specifically, we cover the current limitations of \gls{sota}-- raise the discussion on a more broad perspective of the impact from kin-based technologies (\ie in our everyday lives). This is supported by a rigorous analysis on the edge cases and commonalities of falsely predicted samples. We highlight challenges posed by nature and the environment, and then shine a light on the inherent difficulties of obtaining sufficient data for kin-based problem (Section~\ref{sec:challenges}). This leads to the applications that line up with specific task-evaluations, both existing (\ie practically existing) and high in potential (\ie hypothetically possible). Emphasis is especially placed on the more robust models-- typically, assuming we can improve the performance of the current \gls{sota} (Section~\ref{sec:applications}). 

\section{Background Information}\label{sec:background}
The story of visual kinship recognition research can be told through the data. Therefore, we speak of the progress through the first decade from the perspective of the resources available (\figref{chap:fiwmm:fig:timeline}), and it is shown that interest has been contingent on the amount and quality of labeled data. We end by discussing the data challenges, workshops, and tutorials used to motivate researchers.

\subsection{The evolution of the problem}
An increasing number of researchers have focused their attention on the problem of learning families in photos since the seminal paper was published in 2010~\cite{fang2010towards}. The research progress had the past decade coincided with the supporting labeled data released in part to it. Following \figref{chap:fiwmm:fig:timeline}, we will next look back at the problem.

A trend observed in the progress in visual kinship recognition over the past decade is its correlation with the respective data resources released for public use. Critical points in the research stemmed from the respective problem statements supported by data labeled for the task. Hence, to review the problem statements and protocols as the problem evolved over time.

Fang \etal proposed training machinery to visually discriminate between \emph{KIN} and \emph{NON-KIN} using various facial cues~\cite{fang2010towards}. Specifically, the authors demonstrated an ability to verify kinship given a face pair. To support this, they built and released the first facial image-based kinship database called Cornell Kin. Cornell Kin consisted of 150 face pairs of type \emph{parent}-\emph{child} from the web (\ie public figures, politicians, and other famous persons). Next came biometric data of twins collected at an annual event called~\emph{Twins' Day}~\cite{vijayan2011twins}. This effort yielded in a collection of 197 individuals of multiple modalities (\ie finger prints, 3D face scans, images of irises, and DNA samples) spanning multiple years (\ie from 2011 onward, each year new samples for subjects were added). Shao~\etal then proposed UB Face made-up of 250 parent-child, each supported by three samples (\ie child, parent at a younger age, and parent at an older)~\cite{shao2011genealogical}. The motivation for the pairs having a sample of each parent at a young and old age was directly spawned up from consideration for the difficulty imposed by large age gaps~\cite{xia2012understanding}. Soon thereafter came Family101~\cite{fang2013kinship}-- the first image collection with knowledge of family tree information, with 101 trees and multiple samples per subject. In 2014, Kin-wild I \& II then provided a rich collection of 2,000 parent-child pairs~\cite{lu2014kinship}-- will be discussed in the following section, along with Section~\ref{subsec:verification}, describing this database had significant impact for many expert researchers who proposed clever metric learning methods. Following this came the \gls{tsk} dataset, which structured the problem differently: given a parents-child pair (\ie both parents and a child), determine \emph{KIN} or \emph{NON-KIN}. Then, came \gls{fiw}, which remains the largest kin-based image collection up to today. \gls{fiw} is the main data used for experiments in this survey-- more information provided in detail in the sections to come. Finally, and most recently, was the release of multimedia collections in support of kin-based tasks. First of these was released in 2017 - \gls{kfw} released video data for parent-child pairs, which then allowed for richer, dynamic models to be trained across video frames. Last year, in 2019, came the release of kin-based data that also leveraged audio media, \ie \gls{talkin}. Lastly, \gls{fiw} was extended with multimedia data added to over 200 of its 1,000 families~\cite{robinson2020familiesinMM}. Specs of the aforementioned data (\eg label types, \gls{sota}, reference links) are in \tabref{chap:fiwmm:tab:datasets}. Furthermore, the advancements in methodologies are later covered in detail.

To build \figref{chap:fiwmm:fig:timeline},~\emph{Publish and Perish} was used to acquire the paper-pile for the analysis (Fig~\ref{chap:fiwmm:fig:citations}). For this, a series of queries was executed, each using \emph{visual kinship recognition} as the keywords: Google Scholar, limited to 1,000 search results per query, was run two times (\ie 2010-2020 and 2015-2020); Scorpus was queried from 2010-2020, as only 48 items were found; CrossRef, with a limit of 200 items per search, was queried by year of publication (\ie 2010-11, 2011-12, \dots, 2015-16, 2016, 2017, \dots, 2020). Notice that the years were set such that fewest were expected the first year, more in the first half of the decade, and the most in the latter half. Many papers returned were not on automatic kinship recognition in visual media. However, using the \gls{tfidf} representation, we were able to quickly filter out irrelevant papers by semi-supervised clustering (\ie side-information-based) cosine-similarity k-means~\cite{robinson2018visual} with labels assumed positive for the papers with keywords or titles that contain \emph{visual kinship recognition}.~A rise in the number of annual publications indicates an increase in interest of researchers; the impact on the research community, as a whole, nearly grows exponentially (\ie citation count). The incentive provided by data challenges, along with the increase in labeled data, influence the attention given to kin-based problems in multimedia (MM).

\subsection{Humans recognizing kinship in photos}
Several works in vision research evaluate the ability of humans to detect kinship given a face pair. In the seminar work, Fang~\etal evaluated humans using a subset of their Cornell Kin dataset and found that, on average, humans were about 4\% worse than the machinery: average human performance  was 67.19\%, with the top performance reaching 90\% and the worst at just 50\%~\cite{fang2010towards}. The authors also found supporting results of a previous cognitive study on the perceivable similarity of offspring of different genders (\ie sons tend to be more recognizable as kin than daugthers). Provided the time of this work- a time when it was still unclear whether \emph{KIN} from \emph{NON-KIN} signals are detectable via facial cues- this contribution was not only to compare humans and machines but was to get a sense if even possible for humans.

\gls{fiw} was later used to evaluate humans at a greater scale, and with more diverse data~\cite{robinson2018visual}. Specifically, eleven relationship types, opposed to the four from Cornell Kin. Furthermore, pair counts were ten-times the prior. As described in the following subsection, the findings on \gls{fiw} were that the machines outperform humans with the unconstrained data; additionally, and surprisingly, it was shown that it makes no difference, on average, if the human has prior knowledge of the relationship type in question.

More recently, a study focused on comparing human performance verifying kinship on a grander scale. Furthermore, the study presented extended coverage in the topic of human performance from the view of psychology. That is the work done by Lopez~\etal~\cite{lopez2018kinship}. Their evaluation displayed a face pair and prompted over 300 individuals connected through crowd sourcing services to answer: \emph{Are these two people related (\ie part of the same family)}, with possible responses set as \emph{Yes} or \emph{No}. Face pairs were made up of possible and negatives from both Kin-Wild~\cite{lu2015fg} (\ie image set) and Uva Nemo Smile~\cite{dibeklioglu2013like} (\ie video set)~\cite{dibekliouglu2012you}. The machine again outperformed the human. Taking it a step further, Hettiachchi~\etal analyzed the effects of gender and race, in both the human and data, finding that both genders are similar in ability to recognize kin, while both tend to perform best on same gender pairs (\eg brother-brother, mother-daughter, \etc)~\cite{hettiachchi2020augmenting}. Furthermore, the authors validated previous findings in own-race bias for humans recognizing kinship.


\subsubsection{Results}
\begin{figure}[!t] 
	\centering    
	\begin{subfigure}[t]{\textwidth}
        \centering
        \includegraphics[width=.6\linewidth, trim={0mm 0mm 0mm 0cm},clip]{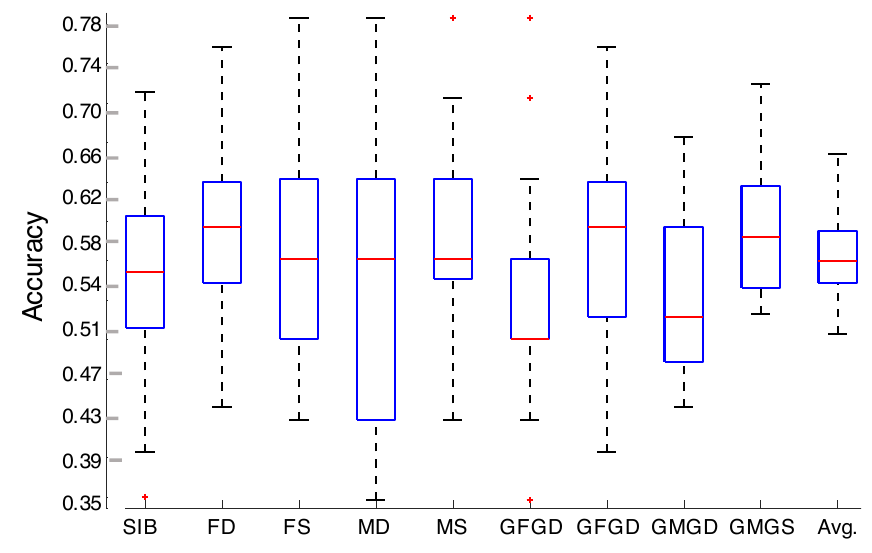}
        \caption{By Type (\textit{Case 1})}\label{chapter:fiw:subfig:box1}
    \end{subfigure}%
    	\begin{subfigure}[t]{\textwidth}
        \centering
        \includegraphics[angle=270, width=.6\linewidth, trim={.5mm 1mm 1mm 0mm},clip]{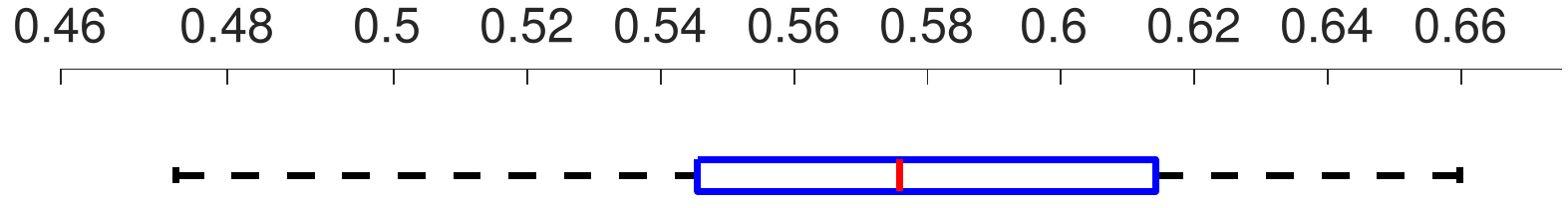}
        \caption{Boolean Case (\textit{Case 2})}\label{chapter:fiw:subfig:box2}
    \end{subfigure}%
	\caption{\textbf{Box plot for humans on kinship verification.} \textit{Case 1:} Relationship type dependent evaluations. \textit{Case 2:} Evaluations with type unspecified.} \label{chapter:fiw:fig:box_chart} 
\end{figure}

We assess both human evaluations via box plots (see Fig. \ref{chapter:fiw:fig:box_chart}).

In \textit{Case 1}, the minimum scores across most categories are below random (\ie  $<$ \%50). In response, we confirmed that no single person scored lowest in more than 1 of the 9 categories. Another observation is the distribution of averages, and its mean of 57.5\%, had the smallest variance-- no average below 50\% or above 67.5\%, which indicates that no single, or more than a few subjects, dominated the average scores for the better or the worse. Examining the pairs where errors were made, three conclusions can be made: (1) especially for relationship types spanning 1 or more generations (\ie parents and grandparents), the common pairs consistently marked incorrectly are cases where the face of the expected elder is at a younger age or the face of the descendant appears older (\eg grandfather in his thirties and grandson in his fifties); (2) different ethnic groups typically made common mistakes on face pairs of different backgrounds; (3) females often deviated from males on the mistakes made that are common and across different ethnic groups-- varying females were always the top scorer, but never the same twice. Apparently, nature and nurture can play a role in humans' ability to do kinship verification as well. There are many interesting directions for future work (\eg even larger and more diverse subject pool, or samples with added semantics like full body views or entire photos with background context).

\begin{figure}[t!] 
\centering
	\includegraphics[width=.75\linewidth, trim={0cm 9mm 0 0},clip]{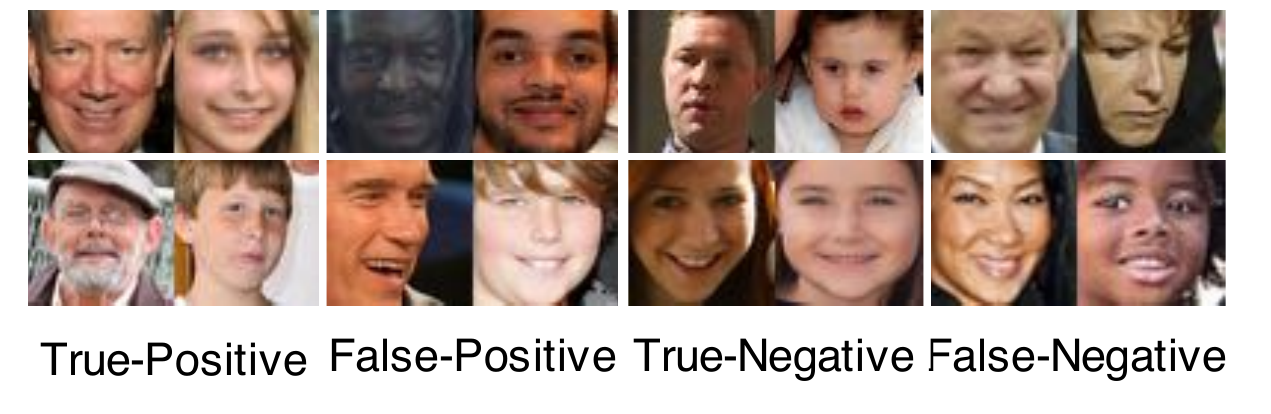}	
	\caption{\textbf{Samples used for human evaluation.} Each column displays pairs most commonly marked correctly and incorrectly, and in cases where the correct answers were true and false. Specifically, \gls{tp}, \gls{fp}, \gls{tn}, and \gls{fn} are displayed, respectively. Each of these pairs was properly classified by the fine-tuned CNN.} 
	\label{chapter:fiw:fig:facepairs}
\end{figure}

For \textit{Case 2}, we evaluated humans' ability to recognize kinship in faces, but, this time, without specifying the relationship. From this, we were aiming to determine whether the relationship direction and face age impacted human responses. Overall, the mean values barely changed, however, the set of pairs commonly marked wrong did-- relationship direction does seem to worsen human ability to recognize kinship when the direction of the relationship contradicts with the age appearance of face pairs; however, in cases without the age contradiction, knowledge of the relationship type helps humans to determine whether or not the face pairs are of that type (\ie even though the set of common pairs incorrectly classified changed, the overall mean did not, as the average fell between 57-58\% in both cases). \figref{chapter:fiw:fig:facepairs} shows face pairs most commonly classified correctly or incorrectly considering both cases. 

Quantitatively, human performers scored an average of ~57.5\%. This is comparable to hand-crafted features such as LBP and SIFT, but nearly 15\% lower than our fine-tuned CNN (\ie the SphereFace CNN fine-tuned for this experiment scores 72.15\%).

\section{Visual Kinship Problems}\label{sec:data:benchmarks:resources}
\begin{figure}[t!] 
	\centering
	\includegraphics[width=\textwidth]{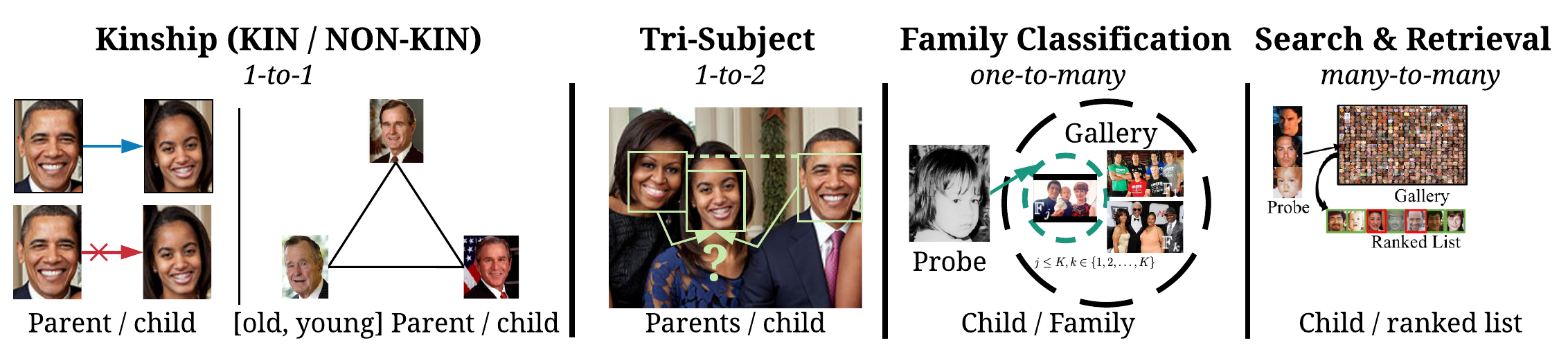}
	\caption{\textbf{Visual kin-based discriminate tasks for the \gls{fiw} dataset.} Robinson \etal posed problems of verification (\ie one-to-one)~\cite{FIW} and family classification (\ie one-to-many)~\cite{robinson2017recognizing, robinson2018visual}, along with more recently supporting tri-subject verification  (\ie one-to-two) and search \& retrieval for ``missing'' children (\ie many-to-many)~\cite{robinson2020recognizing}: the \gls{fiw} database supported the aforementioned tasks, while the annual data challenge \gls{rfiw} was, and continues to be, motivated by promotional purposes. The most recent data challenge supported include three of the four shown here, as family classification was found to carry less potential for practical use-cases, while the others were done using three data splits disjoint in terms of family labels (\tabref{chap:fiwmm:tab:track1:counts},~\ref{tbl:track2:counts},~\ref{tbl:track3:counts}). Protocols and benchmarks for each view are described in~\cite{robinson2020recognizing}. Best viewed electronically.}
	\label{chap:fiwmm:fig:tasks}
\end{figure}

As mentioned, there are several kin-based tasks, each defined by specific protocols to best help control the experiment while simulating the use-case. Before we introduce the experimental details in the proceeding section, let us first introduce the various views of kin-based problems by the motivation and, thus, the problem statements for which they were found. Specifically, we review the discriminatory tasks (\figref{chap:fiwmm:fig:tasks}), along with the generative.

\subsection{Kinship verification}\label{subsec:verification}
The goal of the most popular kin-based task is to determine whether a face pair are blood relatives (\ie \emph{KIN} or \emph{NON-KIN}). Scholarly findings in the fields of psychology and computer vision revealed that different types of kinship share different familial features. From this, the task has evolved into verification over a broad range of relationship types, like \emph{parent}-\emph{child} (\ie \gls{fd}, \gls{fs}, \gls{md}, \gls{ms}) or siblings (\ie \gls{bb}, \gls{ss}, \gls{sibs}). Typically, we assume prior knowledge of the relationship type, both during training and testing. Hence, it is typical to train separate models or learn different metrics. Until the release of \gls{fiw}~\cite{FIW, robinson2017recognizing}, small sample sets limited experiments. Thus, the larger data-pool of \gls{fiw} resulted in larger-scale evaluations that better mimicked true distributions of diverse families globally. With it, also came additional relationship types that span multiple generations (\ie \gls{gfgd}, \gls{gfgs}, \gls{gmgd}, \gls{gmgs}). \gls{fiw} is made-up of 1,000 disjoint family trees of various structures (\ie the number of family members range from five to forty-four). Furthermore, subject nodes making up the trees typically contain multiple face samples-- often samples that span over time, with face shots of most family members at different times in their lives. The families are split into five-folds with no overlap between folds.  The trees are converted to pairs of various relationship types, with an average of five face samples per family member. The pairs present a variety of additional challenges, as, for instance, a \gls{gmgs} pair may or may not be with faces of similar age. It could be an image of a younger grandmother, the GS as an adult, or maybe even as a GF himself.

Another flavor of kinship verification is best explained by the motivation behind UB Face: using knowledge of age as a prior and conditioned on whether or not \emph{KIN} is true~\cite{shao2011genealogical}. The idea was founded on analyzing the type of paired data frequently in the set of \gls{fp}. Specifically, facial pairs of relatives separated by larger age gaps. Thus, based on perceived hard positives, the UB Face dataset provided a pair of images per parent-- one at a younger and the other at an older age. In the end, Shao~\etal supported their hypothesis experimentally by showing a pair of true \emph{KIN} in \emph{parent}-\emph{child} relationships were closest when the parent was at a younger age. Then, Xia~\etal used this to formally claim \gls{sota} on UB Face by treating it as a transfer learning problem, with the target being that of the older parent and child, and the source being younger version of the respective parent and child~\cite{Xia201144}. Many have shown that paired data with greater age gaps are a challenge, and regardless of the level to which older children (\ie an elderly aged \emph{child}) compares to older \emph{parent}.
\begin{figure}[t!]
    \centering
    \includegraphics[height=1.75in,angle=-90,origin=c]{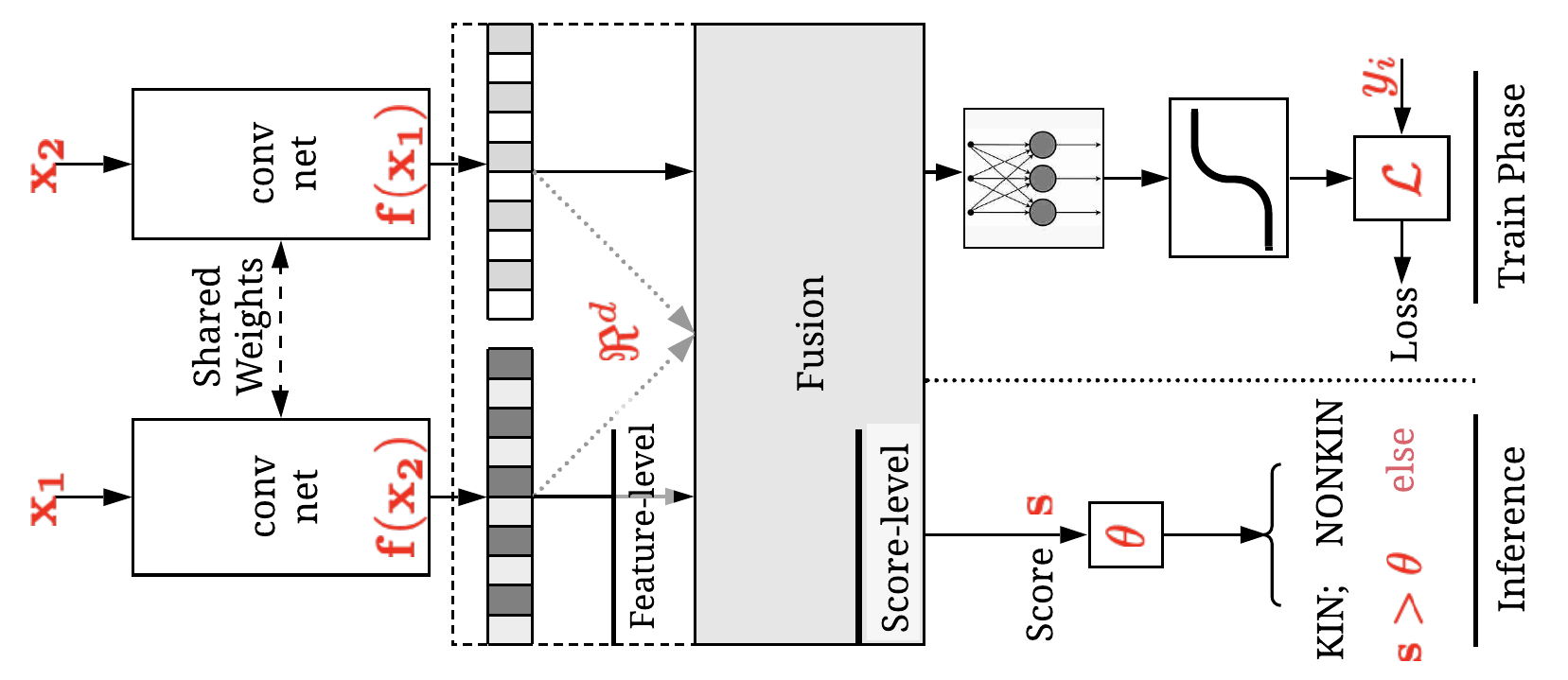}
    \caption{\textbf{Generic Siamese network.} Approaches tend to follow the Siamese model, differing in method of fusion. Specifically (from \emph{top}-\emph{to}-\emph{bottom}), an image pair shot $x_1$ and $x_2$.}\label{fig:siamese}
\end{figure}

\subsection{Family classification} Family classification, the problem where one family member is set aside, and all other members are used to model the classes (\ie family), is reviewed next. Hence, the task is to determine the family that the unknown subject belongs to, which is formulated as a closed-form, multi-class classification problem. This one-to-many problem is challenging, and only increases in difficulty with more families. The challenge stems from the large intra-class variations, which was revealed by a performance drop with an increasing number of families. Fang~\etal~\cite{fang2013kinship} was first to demonstrate this on Family101. Specifically, the authors showed a drop in performance from ten-to-fifty families (increments of 10); opposite to this, the performance improved with one-to-four (increment of one) family member during training. Robinson \etal included 316 families originally~\cite{FIW}, then 512~\cite{robinson2018visual}, and finally 564~\cite{wu2018kinship}. After being supported as part of the \gls{rfiw} annual data challenge the first three consecutive years (\ie 2017, 18, and 19), the overview of the latest \gls{rfiw} mentioned the unrealistic setting of the problem, as families to employ on must be \emph{a priori} knowledge (\ie unable to generalize well). Thus, Robinson~\etal~\cite{robinson2020recognizing} omitted family classification from the latest challenge and substituted in two, more realistic views explained next. Nonetheless, as of the 2019 \gls{rfiw},  17.1\% (accuracy) is \gls{sota}~\cite{aspandi2019heatmap}.

\subsection{Tri-subject verification}
Tri-subject verification, first introduced in~\cite{qin2015tri} (\ie \gls{tsk}), focused on a slightly different view-- predict whether a child is related to a pair of parents. In practice, this setting makes the most sense, as knowledge of one parent means the other can likely be easily inferred. Recently, as part of the \gls{rfiw} data challenge, \gls{fiw} was used to organize and benchmark tri-subject verification on scales much larger than ever before~\cite{robinson2020recognizing}.

\begin{table}[t!]
\centering
\caption{\textbf{KinWild benchmarks.} Results for KinWild I and II.}
\label{chap:fiwmm:tab:kinwild}
\centering
    \begin{tabular}{rccccc|ccccc}\toprule
     &\multicolumn{5}{c}{KinWild I}& \multicolumn{5}{c}{KinWild II} \\
     & \textbf{FD}& \textbf{FS}  & \textbf{MD} & \textbf{MS}   & \textbf{Avg.}& \textbf{FD}& \textbf{FS}  & \textbf{MD} & \textbf{MS}   & \textbf{Avg.} \\\midrule
     
 MNRML~\cite{lu2014neighborhood}   & 72.5&66.5& 66.2 & 72.0   & 69.9  & 76.9 & 74.3 & 77.4 & 77.0 & 76.4 \\
   PDFL~\cite{yan2014prototype} & 73.5& 67.5 &66.1 &73.1&70.0& 77.3& 74.7& 77.8 &78.0 & 77.0 \\
   DMML~\cite{yan2014discriminative} & 74.5 & 69.5& 69.5& 75.5&72.3& 78.5& 76.5& 78.5& 79.5 & 78.3\\
   multiviewSSL~\cite{ZHOU2016136} & 82.8 & 75.4 & 72.6 & 81.3 &78.0 & 81.8 & 74.0 & 75.3 & 72.5&75.9 \\
   SSML~\cite{fang2016sparse}& 81.7 &75.3 &71.4 &77.9&79.6& 82.4& 78.6& 79.8& 77.9 & 79.7\\
   SPML-P~\cite{8019375}&75.4& 84.3& 81.1 &72.4&78.3& 82.4& 77.6& 76.6& 76.2&78.2\\
   ELM~\cite{wuX2018kinship} & 70.0& 64.2& 73.0& 77.2&71.1& 78.6 &73.6 &81.0 &79.6 &78.2\\
   KVRL-fcDBN~\cite{kohli2016hierarchical} & \textbf{96.3}&\textbf{98.1} & \textbf{98.4}& \textbf{90.5} & \textbf{96.1} & \textbf{94.0} & \textbf{96.0} & \textbf{96.8} & \textbf{97.2} & \textbf{96.2}\\
   MvGMML~\cite{hu2019multi}&69.3& 73.1& 69.4 &72.8 &71.1& 70.4 &73.4 &65.8& 69.2 &69.7\\
   DDMML~\cite{lu2017discriminative}&79.1&86.4&87.0&81.4 & 83.5 & 83.8 & 87.4 & 83.0 & 83.2 & 84.3\\
   KML~\cite{zhou2019learning}&-&-&-&-&82.8 &- &- &- &-& 85.7\\
   MSIDA+WCCN~\cite{laiadi2020multiview} & 86.0 & 85.93& 90.1 &88.6 &87.7 &89.4 &82.8& 87.8 &88.0 &87.0\\
   \bottomrule
    \end{tabular}
\end{table}

\subsection{Search and retrieval} 
This view, the most recent to be introduced~\cite{wu2018kinship}, formulates the problem of missing (\ie unknown) children. A search \emph{gallery} made up of all faces of \gls{fiw}, but those of the single child held out as the \emph{probes} for $F$ families. Thus, the input is visual media of an individual, and the output is a ranked list of all subjects in the \emph{gallery}. This \textit{many-to-many} task is staged as a \textit{closed set} problem. Thus, the number of \gls{tp} varies for each subject, ranging anywhere from $[1, k]$ relatives present in the \emph{gallery}. In other words, there are always relatives present.

\subsection{Multi-modal data}
Additional modalities (\eg video~\cite{kohli2018supervised, yan2018video}, audio~\cite{wu2019audio}, multimedia~\cite{robinson2020familiesinMM}), although limited attempts and fairly new in literature, have proven quite effective. \gls{kfvw}, spawned out of the same group as \gls{kfw}, meaning notable contributions by these authors at about the halfway point and towards the end of the decade (\figref{chap:fiwmm:fig:timeline}).  Wu~\etal demonstrated that speech can be modeled to detect kinship~\cite{wu2019audio}. In particularly, audio has shown promising, but through minimal attempts. To better understand the patterns that allow for speech to work-- whether that be jargon used, accents shared, or other acoustical features-- we have seen that a kinship detection system can be improved with audio; however, an in-depth look at the model and the salient components of highly matched signals is subject to future work.

\subsection{Kin-based facial synthesis}\label{subsec:synthesis}
Technology to post-process images  (or even curate in real-time, \ie Snapchat filters) have grown popular in the modern-day main-stream. 
From this alone-- kin-based face synthesis for entertainment and digital art is employable. As a concrete example, Snap Inc. introduced the ability to predict the offspring from a face pair of faces in their app mid-2019. Surely, a natural curiosity. Furthermore, studies support links between DNA and appearance~\cite{walsh2016predicting}, meaning it possible.

Another, nearly default use-case for synthesizing faces based on kinship is in law enforcement to predict the appearance of an unknown perpetrator provided knowledge of kin.  Furthermore, missing family members (\eg a kidnapped child) with face images of years prior (\ie images of adolescence) could be used as prior knowledge, along with the appearances of family in their adulthood, to predict the face of that missing family member as an adult. Also, nature-based studies where latent variables control the appearance of an offspring in a manner that allows for the analyzer to explore. And, projecting further in time, presumably, is its place in genetics. If genetics allows for tweaking the fusion of male and female chromosomes to avoid face deformities of an offspring, the ability to visualize changes in appearance as a function of changes in latency, would likely be needed. 

Some have attempted to predict the appearance of offspring in research (Section~\ref{sec:synthesis}, while others seek a way to commercialize (Section~\ref{sec:applications}): the former (\ie laboratory-style experimentation) and the latter (\ie applied in practical use-cases) are revisited in greater detail later.

\chapter{Families In the Wild (FIW)}\label{chap:fiw}
\section{Overview}
Visual kinship recognition has an abundance of practical uses, such as issues of human trafficking and in missing children, problems from the modern-day refugee crises, and social media platforms. Use cases exist for academia as well, whether for machine vision (\eg reducing the search space in large-scale face retrieval) or a different field entirely (\eg historical \& genealogical lineage studies). However, before the release of \gls{fiw} in 2016~\cite{FIW}, no reliable system existed in practice. This is certainly not due to a lack of effort by researchers, as many works focused on kinship. 

We identified two reasons that clearly slowed the rate at which visual kinship recognition technology evolved:

\begin{enumerate}
\item Data resources for visual kinship were too small to capture true data distributions.
\item Hidden factors in visual appearance shared by blood relatives are complex and less discriminant than in more conventional problems (\eg object classification or face identification).
\end{enumerate}

\begin{figure}[!t]%
	\centering
	\includegraphics[width=.65\linewidth, trim={0cm 4mm 0 0},clip]{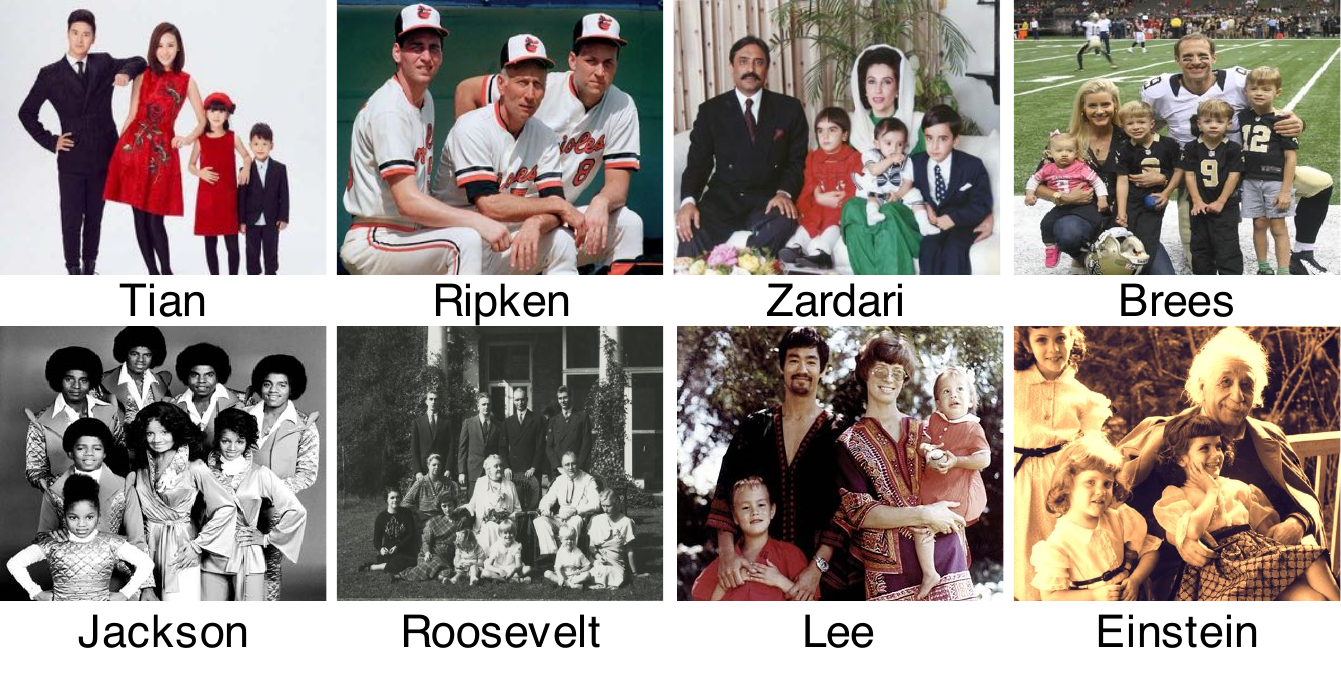}
	\caption{\textbf{Sample family photos from FIW.} Randomly picked 8 / 1,000 families.}
 	\label{chapter:fiw:fig:family_montage} 
\end{figure}

\begin{figure}[t!]
    \centering
    \begin{tabular}{c}
    \includegraphics[width=.5\linewidth]{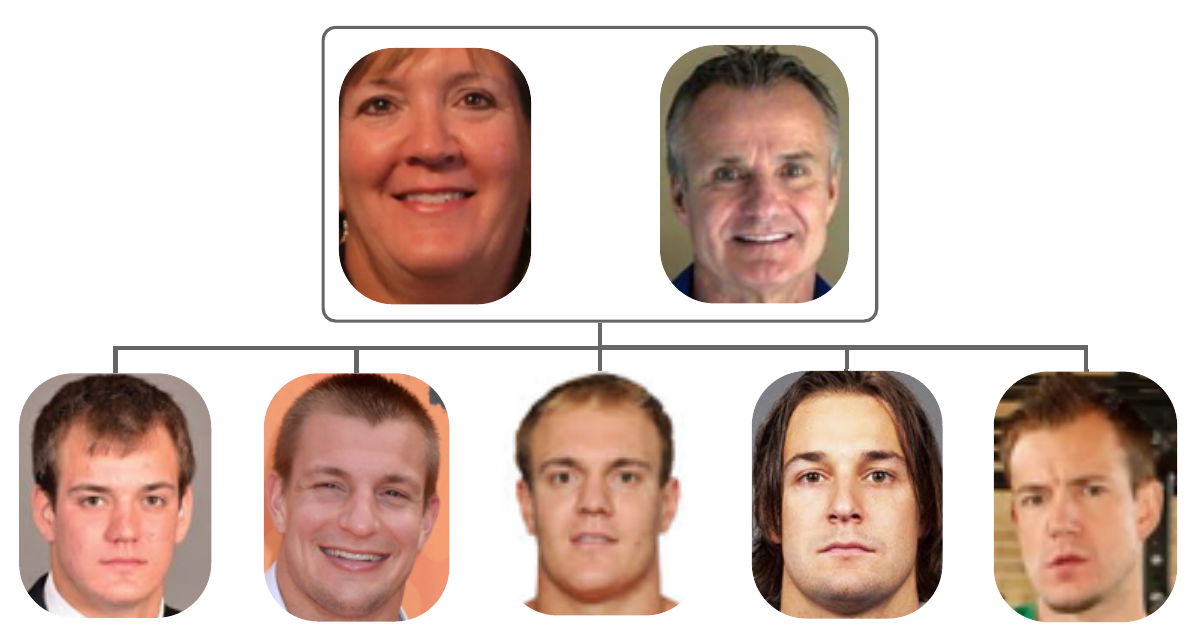}\\
    \includegraphics[trim=0 1mm 0 0, clip, width=.9\linewidth]{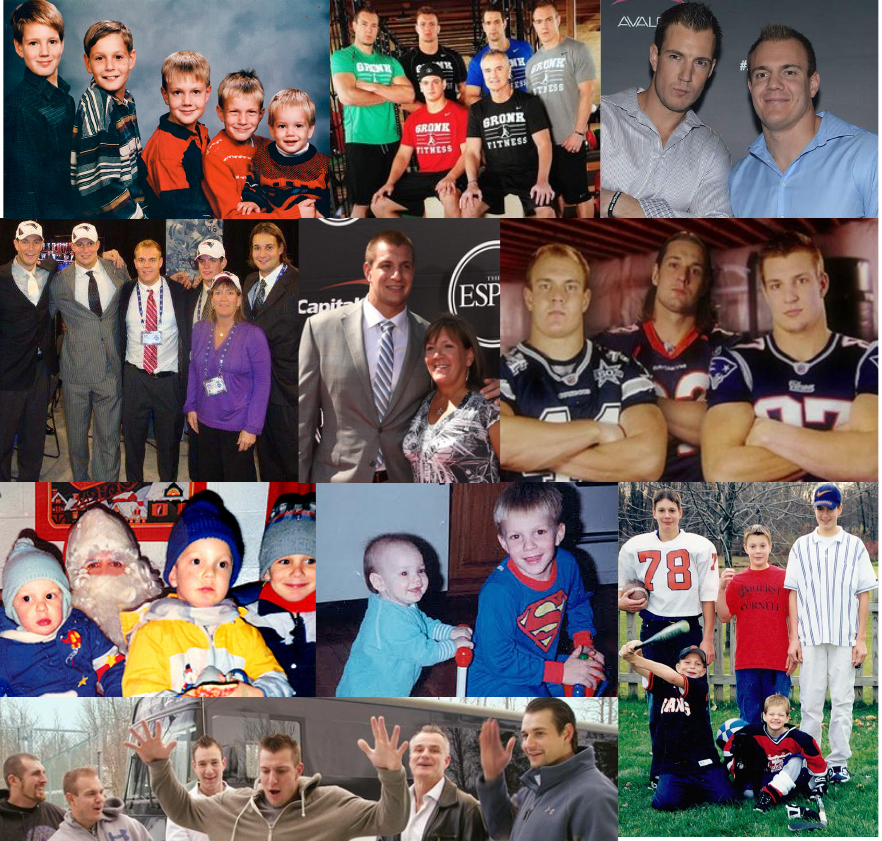}
     \end{tabular}
    \caption{\textbf{Sample family of \gls{fiw}~\cite{robinson2018visual}.} Faces and relationships of the American Football family, the Gronkowski's (\emph{Top}). The montage shows less than half of all photos for respective family. Photo types are various, spanning profile faces (top) to images of different subgroups of family members. Furthermore, samples capture different times of life. Note, crops were made to fit montage (\emph{Bottom}).}
    \label{chap:fiwmm:fig:gronk:montage}
\end{figure}

A large image-set that properly represents families worldwide was needed. Firstly, the distribution one that also could meet the capacity of more complex, data-driven models (\ie deep learning). This was the primary motivation for us to build the first large-scale image database for kinship recognition, \gls{fiw}. \gls{fiw} is made-up of rich label information that captures the complex, hierarchical structures of 1,000 unique family trees. Families have an average of 13 family photos each (\ie $>$13,000 family photos), and with family sizes that range from 3-38 members. Furthermore, most subjects have multiple samples across various ages (see \figref{chapter:fiw:fig:family_montage}). \gls{fiw} remains the \textbf{largest} and \textbf{most comprehensive} database of its kind.\footnote{Visit the \gls{fiw} project page, \url{https://web.northeastern.edu/smilelab/fiw/}.} Samples of a family from \gls{fiw} is shown in \figref{chap:fiwmm:fig:gronk:montage}.

We proposed a multi-modal labeling scheme to minimize the amount of human input needed during the annotation process behind acquiring ground-truth for all the faces in \gls{fiw}. As part of the process is a novel semi-supervised clustering method that proves to work effectively in practice (\ie generating label proposals for new data using existing labeled data as side information). Hence, after labeling about half of the data for half of the families via JAVA GUI, we could use these labels as the side information. Furthermore, we inferred face labels by comparing names in text metadata and the labeled faces and keeping the proposals of high confidence. A simple validation stage at the end takes minutes per a few set of families, opposed to hours. In the end, we show a significant reduction in manual labor and time spent on labeling new data.

Deep learning can now be applied to the problem, as we demonstrate on two benchmarks, kinship verification and family classification. We fine-tune deep models to improve all benchmarks, and provide details on the training procedure. We also measure human performance on verification and compare with benchmarks.

\gls{fiw} was first introduced in~\cite{FIW}, and later extended in~\cite{robinson2018recognize}, has constituted a number of contributions. These are listed as follows.
\begin{enumerate}
\item Added additional faces for verification and complete families for classification (\secref{chapter:fiw:sec:fw}). 
\item Improved the labeling process with novel semi-supervised clustering method (\secref{chapter:fiw:sec:autolabels}).
\item Boosted baseline scores using up-to-date deep learning approaches (\secref{chapter:fiw:sec:experimental}).
\item Obtained SOTA on smaller datasets via transferring CNN fine-tuned on \gls{fiw}  (\secref{subsec:transfer}).
\item Compared human performance with algorithms (\secref{chap:fiw:sec:humanassess}).
\end{enumerate}

\section{Related Works}
\subsection{Related Databases}

The story of visual kinship recognition begins in 2010, at which time the first kin-based image collection (\ie CornellKin) was made public~\cite{fang2010towards}. CornellKin included 150 \textit{parent}-\textit{child} face pairs (\ie celebrities and their parents). Next, UB KinFace-I \& II~\cite{xia2012understanding, Xia201144, Ming_CVPR11_Genealogical} were introduced to address a different view of kinship recognition-- compare parent faces when young and old faces of parents were paired with a child, with a total of 600 face photos of 400 unique subjects (\ie celebrities and politicians). Then, KinWild I-II~\cite{lu2014neighborhood} was released and used in a 2015 FG Challenge~\cite{lu2015fg}, which too focused on \textit{parent}-\textit{child} pairs. Shortly thereafter, Family101~\cite{fang2013kinship} was introduced as the first attempt of multi-class classification (\ie \textit{one-to-many}) for kinship recognition. Thus, it is an organized set of structured families~\cite{fang2013kinship}, including 206 sets of parents and their children (\ie \textit{core families}) that make up 101 unique family trees. In 2015, TSKinFace~\cite{qin2015tri} was built to support yet another view of kinship recognition, tri-subject verification, where both parents and a child were used-- 513 Father/Mother-Daughter pairs and 502 Father/Mother-Son pairs (\ie \textit{two-to-one} verification).

However, even after all these contributions, there existed no single resource that satisfied the concerns of insufficient data. A single resource with the features of previous works, but in a more complete and abundant manner, was the underlying vision for \gls{fiw}. As shown in Tables \ref{chapter:fiw:tab:pair_count1} \& \ref{chapter:fiw:tab:compared}, and discussed in later sections, \gls{fiw} far exceeds others in terms of number of families, face pairs, and relationship types.

\subsection{Automatic Kinship Recognition}
As mentioned, Fang~\etal first attempted kinship verification on \textit{parent}-\textit{child} face pairs~\cite{fang2010towards}. They proposed selecting the 14 (of 44) most effective hand-crafted features. Following this, researchers recognized that a child's face more closely resembles their parents at younger ages~\cite{xia2012understanding, Xia201144, Ming_CVPR11_Genealogical}. In response, they used transfer subspace learning methods that uses the younger faces of parents to help fill the appearance gap between their older faces and that of their children. To benchmark the KinWild dataset, Lu~\etal proposed a metric learning method used in Euclidean space called \gls{nrml} and its multi-view counterpart (MNRML) that learns a common distance metric for multiple feature types~\cite{lu2014kinship}. Fang~\etal focused on \textit{one-to-many} (\ie family classification) by representing faces as a linear combination of sparse features (\ie feature selection via lasso) of 12 facial parts encoded via a learned dictionary~\cite{fang2013kinship}.

Progress made in kinship recognition, along with release of varying task protocols, coincides with an increasing availability of structured and labeled data. Although there have been several significant contributions, none have overcome the challenges posed earlier.

\subsection{Deep Kinship Recognition} \label{subsec:deeplearnkinship}
Since the AlexNet CNN~\cite{krizhevsky2012imagenet} won the 2012 ImageNet Challenge~\cite{russakovsky2015imagenet}, deep learning has achieved \gls{sota} in a wide range of machine learning tasks. Central to this frenzy has been facial recognition~\cite{wolf2014deepface, Parkhi15, schroff2015facenet}. In spite of this, there are only a few works that use deep learning for kinship recognition~\cite{zhang12kinship,dehghan2014look,xiong2015convolutional, kinFG2017}. 

Deep learning has yet to show an advantage for visual kinship recognition, with metric learning seeming more promising. As mentioned in a recent literature review~\cite{wu2016kinship}, the reason for this stems from an insufficient amount of data. In this work, we include several benchmarks on \gls{fiw} using deep learning, obtaining a clear advantage in both tasks.
\subsection{Semi-Automatic Image Tagging \& Data Exploration}
\label{subsec:intro-imagetagging}
Automatic image tagging was recently done by first labeling a small amount of the data, and then using it as side information to help guide the clustering process in a semi-supervised manner~\cite{Liu15ICDM}. Following this, we take advantage of side information from labeled \gls{fiw}.

Previous works used image captions, whether from Flickr or other sources of images tagged by users, to discover labels and annotate images in an automatic fashion~\cite{Leong:2010:TMA:1944566.1944640}. Generally, methods mining text for image tags treat it as a problem of noisy labels~\cite{law2010learning, wang2014mining}. CASIA-WebFace~\cite{yi2014learning}, a large-scale dataset for facial recognition, successfully extended the scale of the renowned LFW~\cite{LFWTech}. By crawling the web, and leveraging knowledge from IMDB, multiple face samples for 10,000 unique subjects were collected. Although related in the sense of automatic labeling, these problems are very different from the one we present here. We aim to add more data to underrepresented families of the \gls{fiw} database, and doing so by using the existing labels for each family as side information to guide our semi-supervised clustering method. We wish to maximize the number of labeled faces available to facilitate the clustering in order to generate label proposals. For this, we use the existing \gls{fiw} labeled faces and the text metadata of the unlabeled data to automatically tag faces using an iterative process governed by both visual and contextual evidence. As discussed in Section \ref{subsec:semi}, our method consistently improves with increasing amounts of side information.

 \begin{figure}[t!]
        \centering
        \begin{subfigure}[t]{.25\linewidth}\centering
            \includegraphics[height=2.4in]{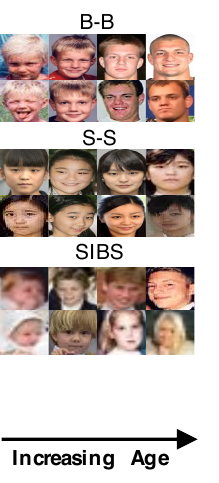}
             \caption{Same Generation}\label{chapter:fiw:subfig:sib}
             \end{subfigure}\hfill
             \begin{subfigure}[t]{.25\linewidth}\centering
            \includegraphics[height=2.3in]{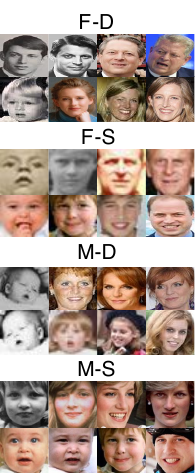}
              \caption{$1^{st}$ Generation}\label{chapter:fiw:subfig:p-c}
             \end{subfigure}\hfill
            \begin{subfigure}[t]{.25\linewidth}\centering
            \includegraphics[height=2.3in]{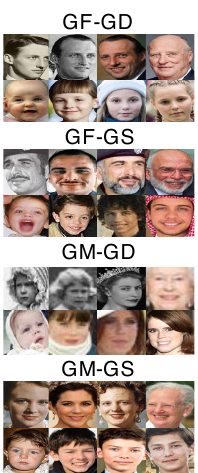}
             \caption{$2^{nd}$ Generation}\label{chapter:fiw:subfig:gp-gc}
             \end{subfigure}
            \caption{\textbf{Samples of eleven pair types of FIW.} Each type is of a unique pair randomly selected from a set of diverse families, while four faces of each individual depict age variations.} \label{chapter:fiw:fig:allpairs}
        \end{figure}     

\section{Families in the Wild (FIW) Database}\label{chapter:fiw:sec:fw}
We next cover the \gls{fiw} database by first recalling the original \gls{fiw} dataset and old labeling scheme~\cite{FIW}. Then, we describe the improved semi-automatic labeling process that enabled the collection to grow as large as it did. Finally, we compare the two.

\subsection{FIW v0.1}\label{subsec:existingfw}
Our goal for \gls{fiw} was to collect about 10 family photos for 1,000 unique families and support with 2 types ground-truth labels, photo-level (\ie who and where in the image) and family-level (\ie all members and the relationships between them). Fig. \ref{chapter:fiw:fig:label_types} depicts the 2 label types. \gls{fiw} was organized as follows: each family was assigned a unique ID (\ie FID), and pictures collected were also assigned a unique ID (\ie PID). Finally, members added were assigned their own unique ID (\ie MID). For instance, $FID_1\rightarrow MID_1$ in $PID_1$ refers to the first member of the first family in the first photo collected. The order of IDs was arbitrary, as assignments were made in the order that the family, member, and photo were added. Before introducing the new and improved semi-automatic process, we briefly review the process used initially in \cite{FIW}, which involved 3 steps: (1) \textit{Data Collection}, (2) \textit{Data Labeling}, and (3) \textit{Data Parsing}.

For \textit{Data Collection}, a team of 8 students from different parts of the world, and with vast knowledge of famous persons, compiled a list of families with a primary focus on their place of origin (\ie an attempt to compile a diverse family list). Table \ref{chapter:fiw:tab:ethnicity_distribution} lists the ethnicity distributions of the 1,000 families. Note that this is not the exact distribution, as each family was counted once according to the \textit{root} member for which the search was based (\ie not per member, but per family). For instance, for Spielberg's family we consider just Stephen. Future work could entail adding more families from underrepresented ethnic groups, as the distribution still favors Caucasians. 

For \textit{Data Preparation}, we built a labeling tool to guide the process of generating the two label types. Annotators work through all family photos on a family-by-family basis, specifying who was in each photo by clicking member faces and choosing their names from a drop-down menu. Names, genders, and relationships for members were only entered on the first instance in an image--once added to the family then only must select their names upon clicking where in photo. 

For \textit{Data Parsing}, all family photos were detected using classic HOG features trained on top of a linear classifier using image pyramids and sliding windows via DLIB~\cite{dlib09}. Faces were cropped and normalized as done in~\cite{kazemi2014one}, and then resized  to 224$\times$224. Finally, the structure of the database was organized into a hierarchy of directories, FID$\rightarrow$MID$\rightarrow$Face-ID (\ie 1,000 folders, $F0001$-$F1000$, containing family labels and folders for MIDs with face samples of that member).

Even though it only took a small team to label 10,676 family photos and 1,000 families, the process relied heavily on human input. Plus, in the end, many families were not properly represented (\ie either too few members, face samples, or family photos). Thus, we aim to reduce the manual labor and overall time requirements to add additional data provided various amounts of labels existed for each (\ie 61 existing families and 4 replacement). We added replacement families (\ie newly added families) to make up for cases of overlapping families or an insufficient online presence when searching for photos (\ie unable to locate family photos for 2 of the under-represented families). Before we propose the semi-automatic labeling model, we first review the two benchmarks included in this work, along with the related statistics of each. We then present the new labeling process that enabled us to add additional data with far less manual labor and in just a fraction of the time.

\begin{figure}[!t]
\centering
	\includegraphics[width=.5\linewidth]{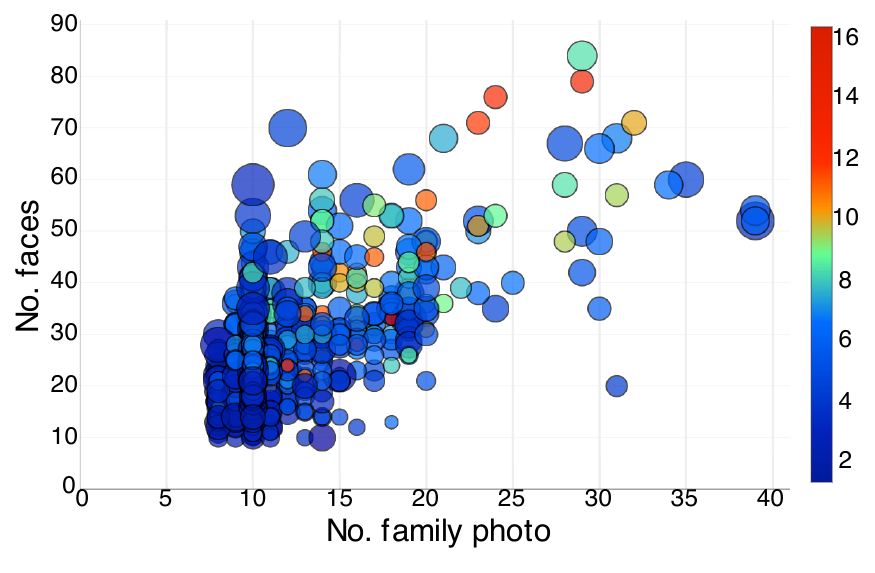}
	\caption{\textbf{Database statistics.} Horizontal and vertical axes represent counts for photos and faces per family, respectively. Bubble size and color represent counts for members and average faces per member, respectively.}
	\label{chapter:fiw:fig:fiwstats}
\end{figure}

\subsection{Data Preparation}
Due to the nature of the label structure, \gls{fiw} can serve as a resource for various types of vision tasks. For this, we benchmark the two popular tracks, kinship verification and family classification. Next, we introduce both of these tasks and the means of preparing the data.

\subsubsection{Kinship verification} Kinship verification aims to determine whether two faces are blood relatives (\ie kin or non-kin). Prior research mainly focused on \textit{parent}-\textit{child} pairs (\ie father-daughter (F-D), father-son (F-S), mother-daughter (M-D), and mother-son (M-S)); some considered sibling pairs (\ie brother-brother (B-B), sister-sister (S-S), and brother-sister/mixed gender siblings (SIBS)). However, research in both psychology and computer vision revealed that different kin relations render different familial features, which motivated researchers to model different relationship types independently. With the existing image datasets used for kinship verification limited to, at most, 1,000 faces and typically only 4 relationship types, we believe such minimal data leads to overfitting and, hence, models that do not generalize well to unseen data captured \textit{in the wild}. \gls{fiw} currently supports 11 relationship types (see \figref{chapter:fiw:fig:allpairs}), 4 being introduced to the research community for the first-time (\ie \textit{grandparent}-\textit{grandchild}) and, most importantly, each category contains many more pairs-- 418,000 face pairs in~\cite{FIW} has increased to 656,954 after extending \gls{fiw} via the proposed semi-supervised approach. 

The 11 relationship types provide a more accurate representation for real-world scenarios. As mentioned, \gls{fiw} was structured such that the labels can be parsed for different types of tasks and experiments, and additional kinship types can easily be inferred.

\begin{figure}[!t]
\centering
	\includegraphics[width=.55\linewidth, trim={1mm 2.5mm .5mm 0mm},clip]{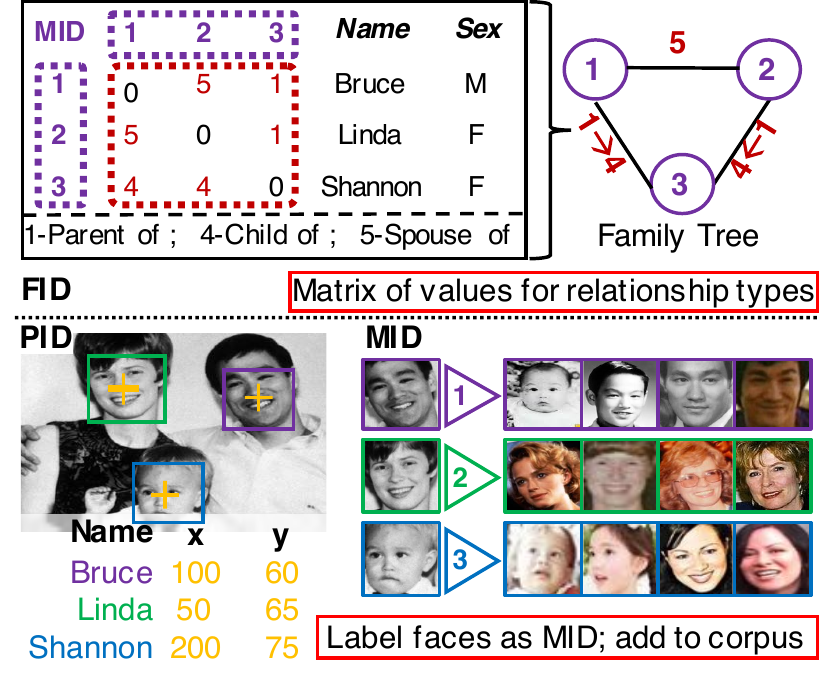}
	\caption{\textbf{Visual of the label types of \gls{fiw}, \textit{Family}-\textit{level} (FID) and \textit{Photo}-\textit{level} (PID).} FID has individual family member (MID) and relationship information. PIDs contain information of MIDs + their locations in photos.}
	\label{chapter:fiw:fig:label_types}
\end{figure}

\begin{figure}[t!]
\centering
\includegraphics[width=1\linewidth]{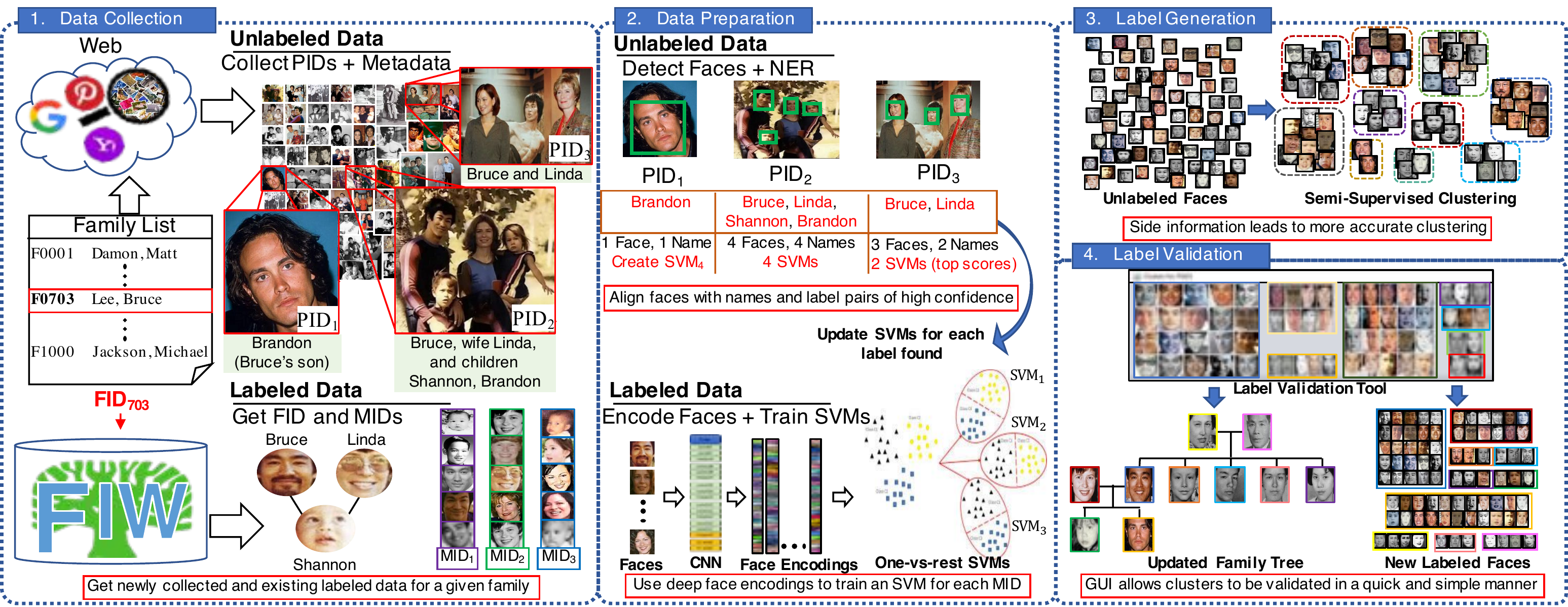}
   \caption{\textbf{Semi-automatic labeling pipeline.} \textit{Data Collection.} Photos and text metadata were collected for underrepresented families in \gls{fiw} and assigned unique IDs (\ie PIDs). Each new member requires at least 1 profile picture (\eg Brandon in $PID_1$) to add to known labels. \textit{Data Preparation.} With the existing \gls{fiw} labels, we next aim to increase the amount, both in labeled faces and member labels, using multiple modalities-- names in metadata and scores of \glspl{svm} were used to automatically label some unlabeled data-- face-name pairs were assumed labeled for cases of high confidence. Starting from profile pictures (\ie 1 face, 1 name) and working towards less trivial scenarios (\eg 3 faces and 2 names, with 2 faces from 1 member at different ages, like in $PID_3$). This step adds to the amount of side information used for clustering. \textit{Label Generation.}  Label proposals for remaining unlabeled faces were generated using the proposed semi-supervised clustering model that leverages labeled data as side information to better guide the process. \textit{Label Validation.} A GUI designed to validate clusters and ensure clusters matched the proper labels.} 
\label{chapter:fiw:fig:labeling_process}
\end{figure}
\subsubsection{Family classification} Family classification aims to determine the family an unknown subject belongs to. Families are modeled using the faces of all but one family member, with the member left out used for testing. This \textit{one-to-many} classification problem is a challenging problem that gets more challenging with more families. This is because families contain large intra-class variations that typically fool the feature extractors and classifiers, and each additional family further adds to the complexity of the problem. Additionally, and like conventional facial recognition, when the target is unconstrained faces \textit{in the wild} \cite{LFWTech} (\eg the variation in pose, illumination, expression, \etc), the problem continues to become more difficult. In~\cite{FIW}, the experiment included only 316 families (\ie families with $>$5 members). In this extended version, we now can include 524 families with the added data. We next present the process followed to extend \gls{fiw}. 

\subsection{Extending FIW}\label{subsec:extendingfw}
The proposed semi-supervised model was used to generate label proposals, using existing and newly labeled data as side information for clustering-- More side information consistently yields better performance (see Section \ref{subsec:semi} \& Fig. \ref{chapter:fiw:fig:nmiplot}). Thus, we set out to maximize the amount of side information (\ie labeled faces) by inferring labels with high confidence by aligning faces and names from the unlabeled photos and metadata. Additionally, we modeled labeled data to discriminate between different family members in a photo. A single family was processed at a time to reduce both the search and label spaces. We aimed to discover labels using evidence from multiple modalities (\ie visual and contextual). This increased the amount of side information for clustering, and also the sample count modeled and used to label more faces. Resulting clusters were then set as ground truth upon human verification in \textit{Step 4}.

We demonstrate the effectiveness of the new labeling scheme by comparing the number of user inputs (\ie mouse clicks and keystrokes) and overall time with the process followed in \cite{FIW}. It took just a few inputs and a few minutes on average per family, opposed to hundreds of inputs and several minutes to over an hour (see Table \ref{chapter:fiw:tab:compared}).

\begin{table}\centering
	\caption {\textbf{Pairwise counts of \gls{fiw}.} Notice \gls{fiw} is first to provide Grandparent-Grandchild pairs. \tabref{chapter:fiw:tab:compared} further characterizes that data, and \figref{chapter:fiw:fig:allpairs} shows samples from it.}
     \resizebox{\textwidth}{!}{%
	\begin{tabular}{@{}rccccccccccccccl@{}}\toprule
		& \multicolumn{3}{c}{{siblings}} & \phantom{a}& \multicolumn{4}{c}{{parent-child}} &
		\phantom{b} & \multicolumn{4}{c}{{grandparent-grandchild}}&
		\phantom{c} & \multicolumn{1}{c}{Total}\\
		\cmidrule{2-4} \cmidrule{6-9} \cmidrule{11-14} 
		&B-B & S-S & SIBS&& F-D & F-S & M-D & M-S && GF-GD & GF-GS & GM-GD & GM-GS & \\\midrule
		KinWild I~\cite{lu2014kinship}&0&0&0&&134&156 & 127 &116 &&0&0&0&0 && 533 \\
			KinWild II~\cite{lu2014kinship}&0&0&0&&250&250 & 250 &250 &&0&0&0&0 && 1,000 \\
	Sibling Face~\cite{bottino2012new}& 232 & 211 &  277 &&0& 0&0&0&&0&0&0&0&&720\\
	Group Face~\cite{guo2014graph}&40 &32 & 53 && 69&69 & 62 & 70 && 0 &0&0&0&&395\\
FIW(Ours)~\cite{FIW} &\textbf{103,724} &\textbf{39,978} & \textbf{73,506} &&\textbf{92,088} &\textbf{129,846}&\textbf{82,160} &\textbf{112,618}&&\textbf{7,078}&\textbf{4,830}&\textbf{6,512}& \textbf{4,614}&&\textbf{656,954}\\
		\bottomrule
\end{tabular}}
\label{chapter:fiw:tab:pair_count1}
\end{table}
We next explain the improved multi-modal scheme made-up of 4 steps: (1) \textit{Data Collection}, (2) \textit{Data Preparation}, (3) \textit{Label Generation}, and (4) \textit{Label Validation}. The goal of (1) and (2) is to gather and increase the amount of side information available for (3), while (4) is to ensure correct labels for all new data. In other words, we set out to increase the labeled sample pool (\ie side information) by inferring labels for unlabeled faces, which adds to the set of training exemplars. The faces that were still unlabeled in (3) were clustered using all labels as side information. All newly added data is then verified by a human. The process is illustrated in \figref{chapter:fiw:fig:labeling_process} and described next.

\subsubsection{Step 1: Data Collection}\label{subsubsec:collecting}
The goal was to collect additional data for under-represented families of \gls{fiw} (\ie families lacking in number of members, faces, and/or family photos). There were 65 families extended in total, with 1 family replaced due to a lack of available data and 3 merging together due to family overlap (\ie Catherine, Duchess of Cambridge, and her immediate family merged with the \textit{Royal} family, as her and Prince William have 2 kids and, thus, bridging the two families). Several new labels and relationships resulted from this merge, with the \textit{Royal} family growing from 29 to 38 members, which is now the largest tree of \gls{fiw}. See \figref{chapter:fiw:fig:fiwstats} and \tabref{chapter:fiw:tab:ethnicity_distribution} for \gls{fiw} statistics.

\begin{figure}[!t] 
	\centering    %
	\begin{subfigure}[t]{0.5\textwidth}
        \centering
        \includegraphics[width=\linewidth]{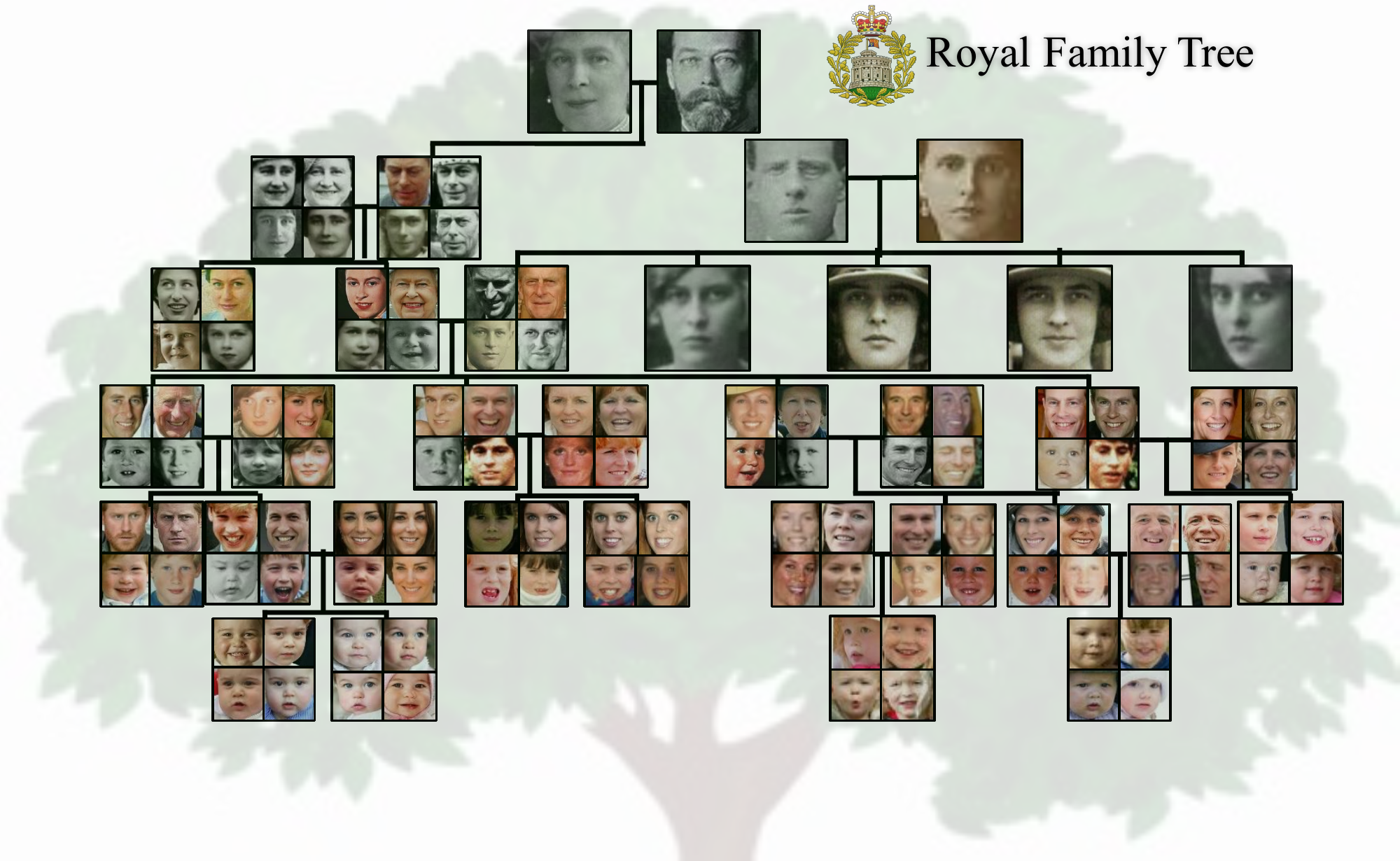}
       
    \end{subfigure}
    
	\begin{subfigure}[t]{0.5\textwidth}
        \centering
        \includegraphics[width=\linewidth]{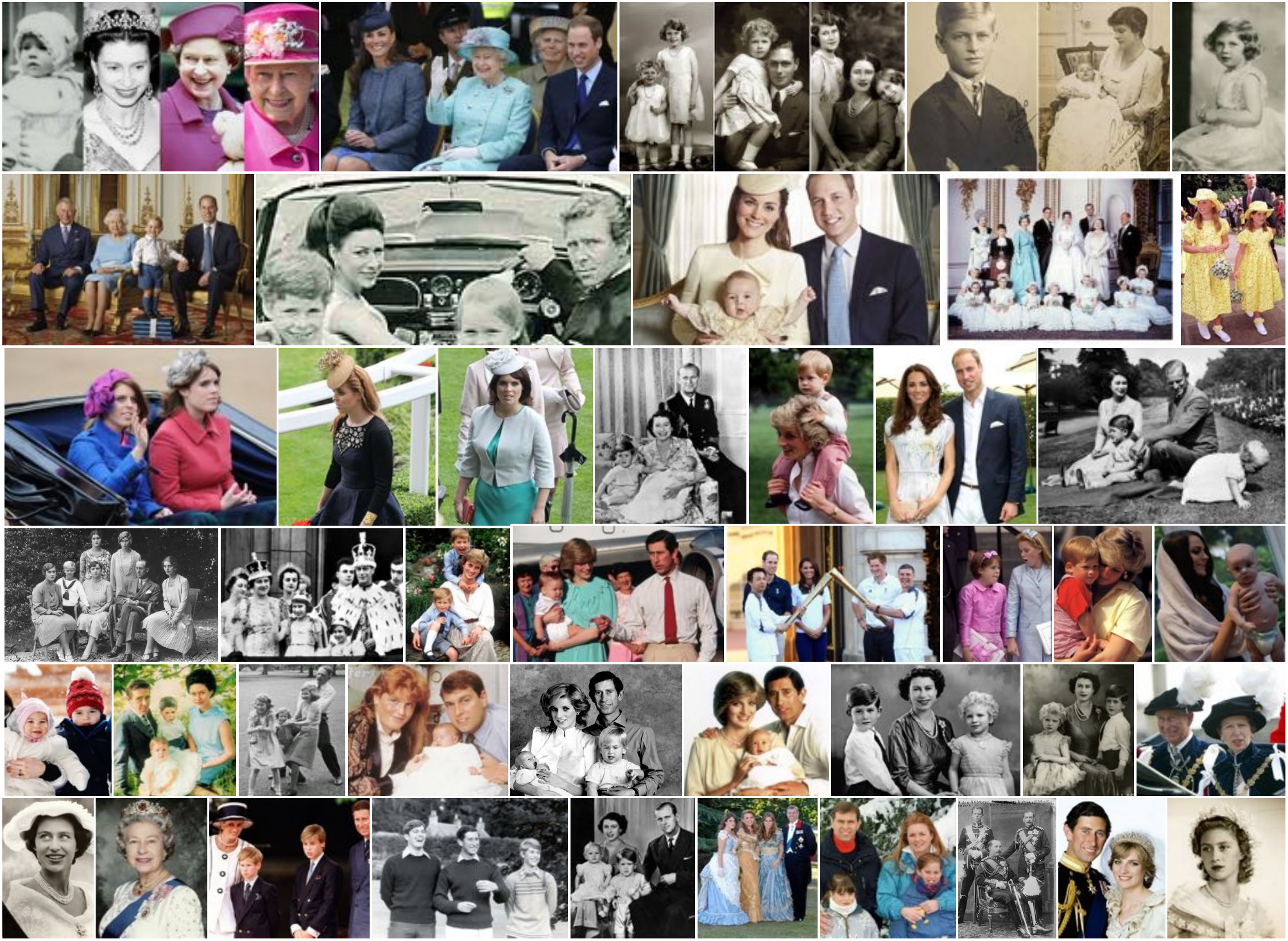}
    
    \end{subfigure}%
     \caption{\textbf{Visualization depicting family structure and photos of the Royal Family.} There are several members in the tree (top) and many photos in total (bottom).} \label{chapter:fiw:fig:pairs} 
\end{figure}

\begin{figure}[ht!] 
\centering
	\includegraphics[width=.9\linewidth]{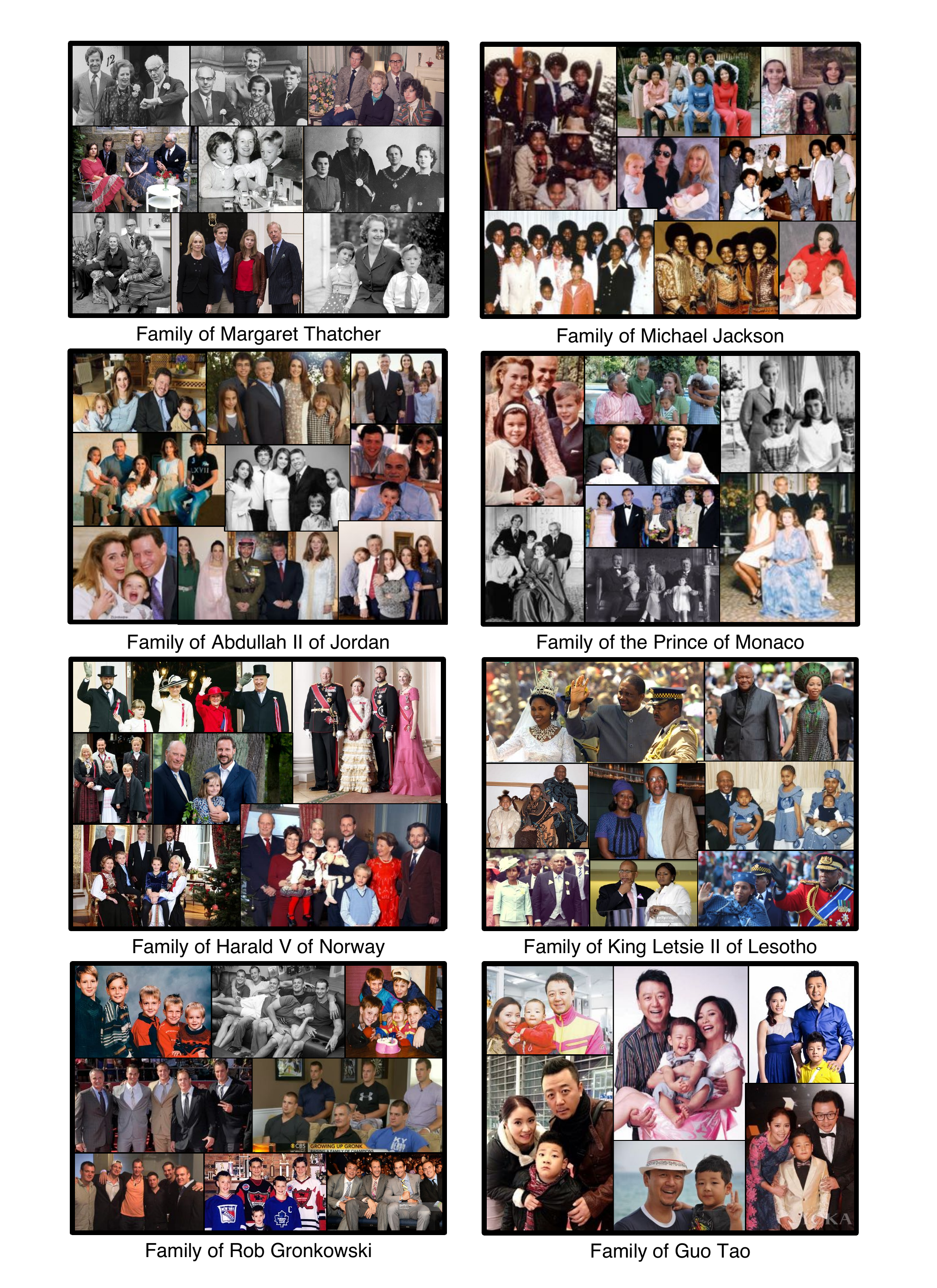}
	\caption{\textbf{Family photo montage.} Samples photos for 8 of 1,000 families in \gls{fiw}.} 
	\label{chapter:fiw:fig:bigmontage}
\end{figure} 
\pagebreak

Preparing for \textit{Step 2}, we set two requirements for the data: (1) rich text metadata describing subjects in each photo and (2) a profile photo per new member. Profile photos enabled label expansion for each new members (\ie single face and single name align with higher confidence). 

\subsubsection{Step 2: Data Preparation}\label{subsubsec:preparing}
The goal here was to maximize the amount of side information available for clustering in \textit{Step 3}. Thus, we took advantage of both labeled (\ie faces \& names) and unlabeled data (\ie detected faces \& text metadata) to automatically infer labels for many unlabeled faces (see \textit{Data Preparation} in Fig. \ref{chapter:fiw:fig:labeling_process}). We next describe each component involved in this step. \\\vspace{-3.6mm}

\begin{figure}[ht!] 
\centering
	\includegraphics[width=.9\linewidth]{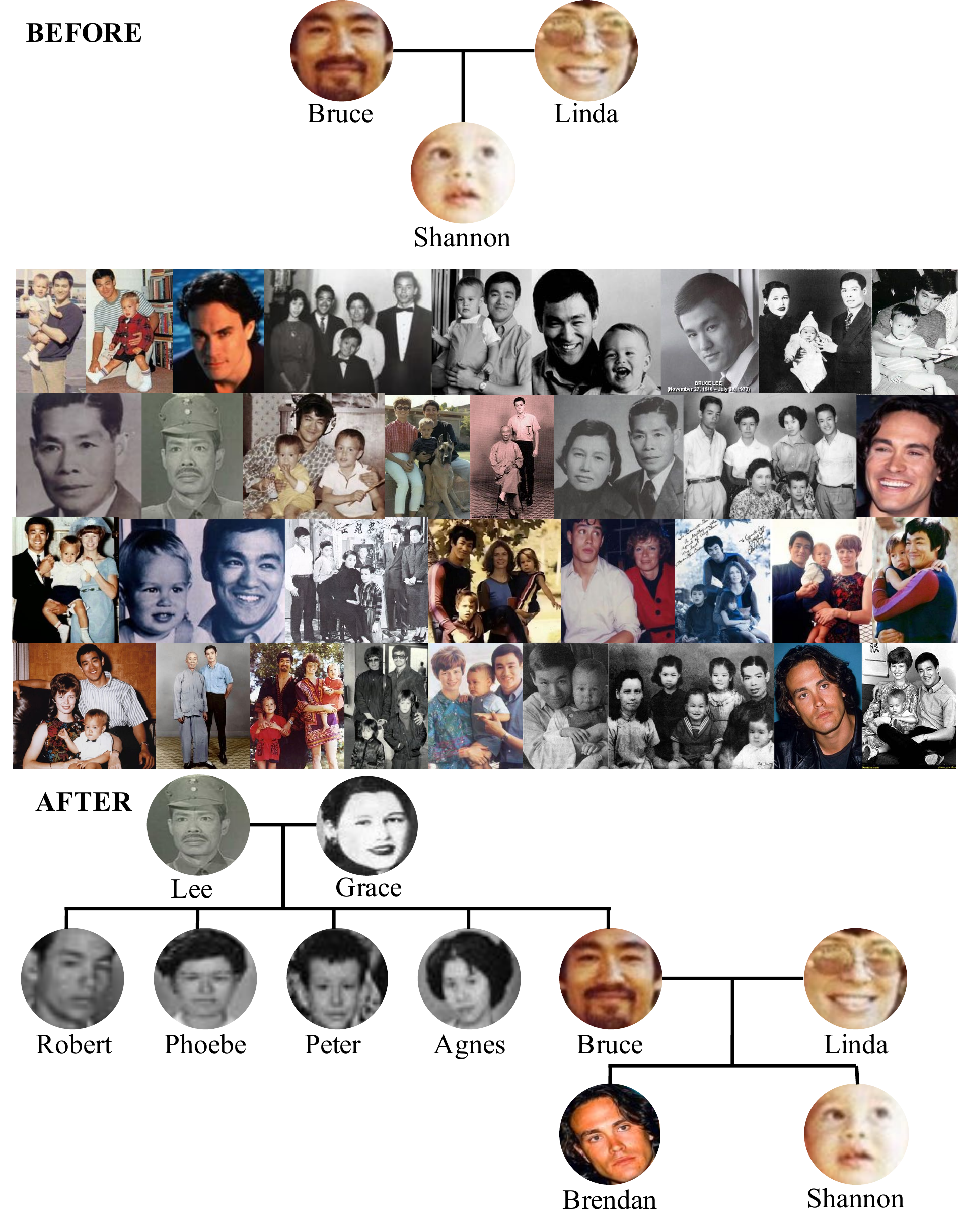}
	\caption{\textbf{Bruce Lee family tree before (top) and after (bottom) extension.} The photos in the middle were added to existing photos using our proposed semi-automatic labeling scheme. This increased both the samples per members and the total number (\ie from just 3 to 10 members).}
	\label{chapter:fiw:fig:lee_family}
\end{figure} 

\noindent\textbf{Text metadata} (\ie image captions) were collected for all photos in \textit{Step 1}. With this, a list of names for each family was generated via a Name Entity Recognition (NER) classifier~\cite{finkel2005incorporating}. Then, a \gls{lut} of candidate names for each member was generated-- \ie keys as member IDs (MIDs) and values as possible references to that member (\eg \textit{Bruce} aka \textit{Bruce Lee} aka \textit{Brandon's father}). One challenge stemmed from name variations (\eg a person legally named \textit{Joseph} might be called \textit{Joe}); additionally, there were name titles (\eg \textit{Queen Elizabeth II} might be called \textit{Elizabeth II}, \textit{Elizabeth}, or, in older photos, \textit{Princess Elizabeth}). Additionally, nicknames posed additional challenges (\eg \textit{Robert Gronkowski}, commonly referred to by his nickname \textit{Gronk}). To address this, a \gls{lut} with detected references for each subject was first compiled, and then further augmented with additional tags (\eg adding titles and surnames). \glspl{lut} were later referenced to find evidence in the text metadata of a member's presence in a photo.\\

\noindent\textbf{New MIDs} found in profile photos (\eg $PID_1$ in \figref{chapter:fiw:fig:labeling_process})-- when processing a family, each image that has a single face detected and just one name in its metadata was considered a profile photo. Profile photos were processed first. The name detected in the metadata was compared to all names for members stored in the LUTs. If there were no matches, the subject was then added as a new member in that family. A LUT of names was then generated for each new member, and the name of highest frequency (\ie number of detections in all metadata) recorded as the name corresponding to their assigned MID (\eg $MID_6$ for the sixth member).\\

\noindent\textbf{Unlabeled and labeled faces} were encoded as 4,096D features from the $fc_7$-layer of the pre-trained VGG-Face CNN model~\cite{Parkhi15}. \textit{One-vs-rest}  \gls{svm} models were trained for each member using labeled samples from all other members of that family as the negatives. Next, profile photos were processed (\ie 1 name and 1 face). Names that match an existing label were added to corresponding MID data pools, while mismatched names were added as a new MID with a \gls{lut} generated. This shows the benefit of including profile pictures for each new member, which makes it so all family members were known. It is important to note that \glspl{svm} were updated each time a new labeled face was added.\\

\noindent\textbf{Discovering labels} continues in a similar fashion, except now the \glspl{svm} play a more critical role. Now moving on to images with 2 faces and 2 names, the 2 \glspl{svm} of the respective members were used to classify the 2 faces. Provided high scores and no conflicts, labels were inferred. Cases with low confidence or conflicts were skipped, leaving those faces to be labeled via clustering. Next, photos with 3 faces and 3 names were processed, then 4 faces and 4 names, and so on and so fourth. After all one-to-one cases were processed, photos with a different number of names and faces were processed. For each photo, only \glspl{svm} that correspond to a \gls{lut} with matching names were used. Thus, justifying a requirement of \textit{Step 1}-- collect rich metadata in terms of specifying members present in photos.

\begin{table}[t!]
\centering
\caption{\textbf{Database counts and attributes.} Comparison of \gls{fiw} with related datasets.}
  \scriptsize
  \centering
  \begin{tabular}{l*{2}{p{.85cm}<{\centering}}*{2}{p{.6cm}<{\centering}}{p{1.2cm}<{\centering}}} 
\toprule
  Dataset & No. Family & No. People &  No. Faces &  Age Varies & Family Trees \\ \midrule
  CornellKin\cite{fang2010towards} & 150  & 300 & 300 & \xmark & \xmark \\
  UBKinFace\cite{Ming_CVPR11_Genealogical,Xia201144}& 200 & 400 & 600 & \ 
\Large\checkmark 
 & \xmark \\
  KFW-I\cite{lu2014neighborhood}& \xmark & 533 & 1,066 & \xmark & \xmark\\
  KFW-II\cite{lu2014neighborhood}& \xmark & 1,000 & 2,000 & \xmark & \xmark \\
  TSKinFace\cite{qin2015tri} & 787 & 2,589 & \xmark & \Large\checkmark & \Large\checkmark \\
  Family101\cite{fang2013kinship} & 101 & 607 & 14,816 & \Large\checkmark & \Large\checkmark \\
  \gls{fiw}~\cite{FIW} & \textbf{1,000} & \textbf{10,676} & \textbf{30,725} & \Large\checkmark & \Large\checkmark\\
\bottomrule
\end{tabular} 
 \label{chapter:fiw:tab:compared}
\end{table}

Notice that some families benefited far more than others in this process. Nonetheless, roughly 25\% of the 2,973 added faces were correctly labeled by this simple multi-modal process.

\begin{table}[b!] 
\centering
\caption{\textbf{Ethnicity distribution for the 1,000 of \gls{fiw}.} \textit{Mix} families contain $>$2 ethnicity (\eg Bruce (\textit{Asian}) and Linda (\textit{Caucasian}) Lee with 2 children.}
	\scriptsize
	\centering
	\begin{tabular}{cccccc}\toprule
	 Caucasian & Spanish/Latino& Asian&African/AA&Arabic&Mix\\ \midrule
	64\%&10.7\%&9.1\% &8.2\% &2.0\%&6.0\% \\\bottomrule
	\end{tabular} 
 \label{chapter:fiw:tab:ethnicity_distribution}       
\end{table}

\begin{table}[t!]
\centering 
\caption{\textbf{Speedup analysis.} Previous (white) versus new (shaded) labeling processes  compared in terms of inputs (keyboard and mouse clicks) and time (hours:minutes:seconds).}\label{chapter:fiw:tab:methods_compared}
         \resizebox{\textwidth}{!}{%
\begin{tabular}{l  c  c  c  c  c c  c  c }
\toprule
&Bruce Lee &	Michael Jordan	&John Malone&	Craig Mccaw&	Marco Reus	&British Royal&	Michael Jackson & Total\\\hline
Inputs (count)&	551	&97	&153	&178	&35&	1,838&	1,272 & 4,124\\
\rowcolor{mygray2}\textbf{Inputs (count)}&	\textbf{12}&\textbf{6}	&\textbf{10}	&\textbf{15}&	\textbf{7}&	\textbf{21}	&\textbf{24}&\textbf{95}\\\hline
Time (h:m:s)&	0:15:08&	0:5:31	&0:5:18	&0:6:16	&0:4:24&	1:25:23	&0:44:52 & 2:46:52\\
\rowcolor{mygray2}\textbf{Time (h:m:s)}&	\textbf{0:1:11}	&\textbf{0:0:31}	&\textbf{0:1:05}	&\textbf{0:0:56}	&\textbf{0:0:31}	&\textbf{0:6:44}	&\textbf{0:7:13}&\textbf{0:18:11}\\\bottomrule
\end{tabular}}
 \label{chapter:fiw:tab:label_clock}       
\end{table}

\subsubsection{Step 3: Label Generation}\label{subsubsec:labeling}
Label proposals were generated for unlabeled faces using the proposed semi-supervised clustering method. To get the most out of our model we automatically labeled additional data in \textit{Step 2}, while identifying all new members being added to each family. Hence, the number of members (\ie $k$) was known for each family.

More details, including the objective function and solution, given in Section \ref{chapter:fiw:sec:autolabels}.

\subsubsection{Step 4: Label Validation}\label{subsubsec:validating}
Finally, clusters (\ie labels) were validated by a human. This was a three-part process: assign an MID to each cluster; validate each cluster, which was displayed in a grid of faces in the order of confidence score; specify gender and relationships of newly added members. As shown in \figref{chapter:fiw:fig:labeling_process}, a JAVA interface was designed to generate ground-truth for new data with just a few clicks of the mouse and minimal time per family. The inputs were cluster assignments for a family, with faces listed in order of confidence (\ie cosine distance from centroid). MIDs were assigned in \textit{Step 2} (\ie inferred from text, \gls{svm} scores, or both), which must also be validated. The outputs were labels for each PID and an updated relationship matrix (\figref{chapter:fiw:fig:label_types}).

\subsubsection{Discussion}
Seven families of various sizes were used to compare the old~\cite{FIW} and proposed labeling schemes-- old scheme took 4,124 inputs in  about 2.75 hours, and just 95 inputs in about 18.1 minutes via the new (see Table ~\ref{chapter:fiw:tab:methods_compared}). Collecting and labeling the data for the extended \gls{fiw} was done by a single person in days; it initially took a small team several months with the old scheme. Thus, demonstrating a significant savings in manual labor and time (note that greater amounts of data was originally collected, however, relative savings in time and manual labor clearly yields from process used in this extended version). A possible future direction is to use this scheme to extend families of \gls{fiw} with video data. Another possibility is to use this method to extend the number of families, which, if on the order of thousands or more, then automating \textit{Step 1} could further reduce savings (\ie web scrape for family information (\eg \textit{Wiki}) and photos (\eg \textit{Google}, \textit{Bing}, \etc)).

\section{Semi-Supervised Face Clustering}\label{chapter:fiw:sec:autolabels}
Labeling is a human-necessary and expensive task to benchmark data sets. Here we aim to accelerate the process by using some labeled data in advance. In this part, we demonstrate a novel semi-supervised clustering for labeling. Let $X=\{x_i\}$ be the data matrix with $n$ instances and $m$ features and $S$ be a $n' \times K'$ side information matrix, which denotes $n'$ labeled data instances into $K'$ classes. Our goal was to make use of $S$ to guide the remaining data into $K$ classes, with $K' \le K$.

\subsection{Objective Function}
Inspired by our previous work~\cite{Liu15ICDM, liu2017partition}, a partition level constraint is used to make the learnt partition agree with partial human labels as much as possible. To demonstrate the effectiveness of our labeling mode, K-means with cosine similarity is employed as the core clustering method to handle high-dimensional data due to its high efficiency and robustness. Our objective function is 
\begin{equation}\label{eq:obj}
\min \sum_{k=1}^K\sum_{x_i\in \mathcal{C}_k}f_{cos}(x_i,m_k)+\lambda U_c(S, H \otimes S),
\end{equation}
where $f_{cos}$ is the cosine similarity, $H$ is the final partition, $H_S=H \otimes S$ is part of $H$ which the instances are also in the side information $S$, $m_k$ is the centroid of $\mathcal{C}_k$, $U_c$ is the well-known Categorical Utility Function~\cite{Mirkin01ML} and $\lambda$ is the trade-off parameter.

To better understand the last term in Eq.~\ref{eq:obj}, we give the detailed calculation of $U_c$. Given two partitions $S$ and $H_S$ containing $K'$ and $K$ clusters, respectively. Let $n^{(S)}_{kj}$ denote the number of data objects belonging to both cluster $C_{j}^{(S)}$ in $S$ and cluster $C_k$ in $H_S$, $n_{k+}=\sum_{j=1}^{K'}n^{(S)}_{kj}$, and $n^{(S)}_{+j}=\sum_{k=1}^{K}n^{(S)}_{kj}$, $1\leq j\leq K'$, $1\leq k\leq K$. Let $p^{(S)}_{kj}=n^{(S)}_{kj}/n'$, $p_{k+}=n_{k+}/n'$, and $p^{(S)}_{+j}=n^{(S)}_{+j}/n'$. We then have a normalized contingency matrix (NCM), based on which a wide range of utility functions can be accordingly defined. For instance, the widely used category utility function can be computed as follows:
\begin{equation}\label{equ:Uc}
  U_c(H_S,S)=\sum_{k=1}^K p_{k+}\sum_{j=1}^{K'} (\frac{p^{(S)}_{kj}}{p_{k+}})^2-\sum_{j=1}^{K'}(p^{(S)}_{+j})^2.
\end{equation}
It is worthy to note that $U_c$ measures the similarity of two partitions, rather than two instances. The larger value of $U_c$ indicates the higher similarity.

\subsection{Solution}
We notice that the first term in Eq.~\ref{eq:obj} is the standard K-means with cosine similarity. Could we still apply K-means optimization to solve the problem in Eq.~\ref{eq:obj}? The answer is yes! Due to our previous work~\cite{Wu15TKDE}, we provide a new insight of $U_c$ by the following lemma. 

\begin{lemma}
Given a fixed partition $S$, we have
\begin{equation}
   U_c(H_S, S) = -||S-H_SG||^2_{\textup{F}}+\textup{constant},
\end{equation}
where $G$ is the centroid matrix of $S$ according to $H_S$.
\end{lemma}

By the above lemma, the second term in Eq.~\ref{eq:obj} can also be transformed into a K-means problem with squared Euclidean distance. Then a K-means-like algorithm can be used on the augmented matrix with modified distance function and centroid update rule for the final partition. 

First an augmented matrix $D$ is introduced as follows.
\begin{equation}\label{eq:D}
  D = \begin{bmatrix} X_S & S\\ X_T & 0  \end{bmatrix}\ \ \textup{with}\ \ X = \begin{bmatrix} X_S \\ X_T  \end{bmatrix},
\end{equation}
where $d_i$ is the $i^{th}$ row of $D$, which has of two parts, $d_i^{(1)}$ and $d_i^{(2)}$ (\ie $d_i^{(1)}=(d_{i,1}, \cdots, d_{i,d_m})$ presents the feature space and $d_i^{(2)} =(d_{i,d_m+1}, \cdots, d_{i,d_m+K'})$ denotes the label space). Zeros in $D$ are the artificial elements, rather than the true values so that all zeros contribute to the computation of the distance and centroids, which inevitably interfere with the cluster structure. To make the zeros in $D$ not involved in the calculation, we give the new update rule for the centroids of $D$. Let $m_k=(m_k^{(1)}, m_k^{(2)})$ be the $k^{th}$ centroid $\mathcal{C}_k$ of $D$, we modify the computation of centroids as follows.
\begin{equation}\label{eq:centroid}
  m_k^{(1)} = \frac{\sum_{d_i\in \mathcal{C}_k}d_i^{(1)}}{|\mathcal{C}_k |},\ \ m_k^{(2)} = \frac{\sum_{d_i\in \mathcal{C}_k}d_i^{(2)}}{|\mathcal{C}_k\cap X_S|}.
\end{equation} 
and the distance function is also adjusted as 
\begin{equation}
  f(d_i,m_k) = f_{cos}(d_i^{(1)}, m_k^{(1)}) + \boldsymbol{1}(d_i \in S)f_{sqE}(d_i^{(2)}, m_k^{(2)}),
\end{equation}
where $\bf{1}$ returns 1 if the condition is satisfied, otherwise 0.

The correctness and convergence of the modified K-means is similar to one in ~\cite{Liu15ICDM}.

\section{Experiments}\label{chapter:fiw:sec:experimental}
We conduct the following experiments: benchmark kinship verification and family classification; evaluate the proposed semi-supervised clustering method at the core of the new labeling scheme; fine-tune CNNs using \gls{fiw} and evaluate on KinWild I \& II (\ie transfer-learning); measure human performance on kinship verification and compare to top scoring algorithms.

The subsequent subsections are organized as follows. First, we review the visual features, metric learning methods, and deep learning common in all experiments. Then, we dive into the experiments mentioned above. We introduce each independently, but with the same structure: experimental settings, experiment-specific training philosophy, and then the results and analysis. 

\begin{table}[t!]\centering
	\caption {\textbf{Averaged verification accuracy scores (\%) for 5-fold experiment on \gls{fiw}.} Note that there was no family overlap between folds.}
	\ra{1.3}
\resizebox{\textwidth}{!}{%
	\begin{tabular}{@{}rcccccccccc@{\hskip 0.08in}c@{\hskip 0.08in}c@{\hskip 0.08in}ccr@{}}\toprule
		& \multicolumn{3}{c}{{siblings}} & \phantom{a}& \multicolumn{4}{c}{{parent-child}} &
		\phantom{b} & \multicolumn{4}{c}{{grandparent-grandchild}}&
		\phantom{c} & \multicolumn{1}{c}{}\\
		\cmidrule{2-4} \cmidrule{6-9} \cmidrule{11-14} 
		Method&B-B & S-S & SIBS&& F-D & F-S & M-D & M-S && GF-GD & GF-GS & GM-GD & GM-GS &&Acc. $\pm$ Std. \\\midrule    
		LBP~\cite{ahonen2006face}&55.52 &57.49 & 55.39 &&55.05 &53.77 &55.69 &54.65 && 55.79 &55.92 &54.00 &55.36 &&55.33 $\pm$ 1.01 \\
		SIFT~\cite{1467360}&57.86&59.34&56.91&&56.37&56.24 &55.05 &56.45 && 57.25 &55.35& 57.29 &56.74 &&56.80 $\pm$ 1.17\\		
		ResNet-22~\cite{wen2016discriminative}&65.57 &69.65&60.12 &&59.45 & 60.27& 61.45& 59.37&&55.37 &58.15 &59.74 & 59.70 &&61.34 $\pm$ 3.81\\
		VGG-Face~\cite{Parkhi15}&69.67& 75.35& 66.52&& 64.25&63.85&66.43&62.80 &&62.06& 63.79 & 57.40& 61.64 &&64.89 $\pm$ 4.68\\
		+ITML~\cite{davis2007information}  &57.15 &61.61 &56.98 &&58.07 &54.73 &57.26 &59.09 && 62.52 &59.60 & 62.08 & 59.92&& 59.00 $\pm$ 2.44\\
		+LPP~\cite{niyogi2004locality} & 67.61&66.22 &71.01 && 62.54 & 61.39&65.04 &63.54 && 63.50 &59.96 &60.00 &63.53 &&64.03 $\pm$ 3.32\\
		+LMNN~\cite{weinberger2009distance} & 67.11 & 68.33 & 66.88 && 65.66 & 67.08 & 68.07 & 66.16 && 61.90 & 60.44 & 63.68 &60.15 &&65.04 $\pm$ 3.00\\
		+GmDAE~\cite{peng2013marginalized} & 68.05 & 68.55 & 67.33 && 66.53 & 68.30 & 68.15 & 66.71 && 62.10 & 63.93 &63.84 & 63.10 &&66.05 $\pm$ 2.36\\
		+DLML~\cite{ding2015discriminative} & 68.03 & 68.87& 67.97 && 65.96 & 68.00 & 68.51 & 67.21 && 62.90 & 63.96 & 63.11 & 63.55&&66.19 $\pm$ 2.36\\
		 +mDML~\cite{kinFG2017}&69.10 &70.15 & 68.11 && 67.90 & 66.24 & 70.39 &  67.40 &&  65.20 & \textbf{66.78} & 63.11& 63.45&&67.07 $\pm$ 2.44\\
		 ResNet+CF~\cite{robinson2017recognizing}&69.88 &69. 54&69.54&& 68.15 &67.73 &71.09 &68.63 &&\textbf{66.37}& 66.45& \textbf{64.81}&64.39 &&67.87 $\pm$ 2.15\\
		 SphereFace\cite{liu2017sphereface}&\textbf{71.94} &\textbf{77.30} &\textbf{70.23} && \textbf{69.25} & \textbf{68.50}& \textbf{71.81}&\textbf{69.49} && 66.07 & 66.36 &64.58 &\textbf{65.40} &&\textbf{69.18} $\pm$ 3.68\\
		\bottomrule
\end{tabular}}
\label{chapter:fiw:tab:Veri}
\end{table}

\subsection{Experimental Setting}
For the sake of organization, all low-level features and metric learning approaches used throughout are listed and described in this section. Most are in two or more experiments, however, even those used for verification, for example, are still treated as common information, and thus is described alongside other items of preliminary information. Following traditional ``shallow'' methods, we review specifications of the pre-trained CNNs used as off-the-shelf face encoder.

\subsubsection{Feature Representations}
Detected and aligned faces were normalized and encoded using low-level and CNN-based features. We next describe the descriptors used in this work-- SIFT, LBP, pre-trained VGG-Face and ResNet CNNs-- each having been widely used in visual kinship and facial recognition problems.\\\vspace{-3mm}

\noindent\textbf{SIFT~\cite{lowe2004distinctive}} is amongst the most widely used feature type in object and face recognition. Here we follow the settings of~\cite{lu2014neighborhood}: resize images to 64$\times$64, then extract features from 16$\times$16 blocks with a stride of 8 (\ie 49 blocks that yields 128$\times$49 = 6,272D face feature).\\\vspace{-3mm}

\noindent\textbf{LBP~\cite{ahonen2006face}} is renown for its effectiveness in tasks such as texture analysis and face recognition. We again follow the settings of \cite{lu2014neighborhood}: resize images to 64$\times$64, divide into 16$\times$16 non-overlapping blocks, and use a radius of 2 and sampling number of 8. Each block was represented as a 256D histograms (\ie 256$\times$16 = 4,096D face encoding).\\\vspace{-3mm}

\noindent\textbf{VGG-Face~\cite{Parkhi15}}, a pre-trained CNN with the topology of VGG-16: made-up of small convolutional kernels (\ie 3$\times$3) with a convolutional stride of 1 pixel. VGG-Face was trained on  $\approx$2.6M face images of 2,622 different celebrities. VGG has worked well on various face databases-- 97.3\% in accuracy on \textit{YouTube Faces}~\cite{wolf2011face}; 98.95\% accuracy on \textit{Labeled Faces in the Wild}~\cite{huang2007labeled}. By removing the top two layers-- softmax and last fully-connected layer (aka fc8-layer or $fc_8$)-- the CNN can be used as an \textit{off-the-shelf} face encoder~\cite{robinson2016pre}. Thus, models get trained on an auxiliary resource and employed on target data. Here, we fed faces through to the fc7-layer (aka $fc_7$), yielding a 4,096D face encoding. \\\vspace{-3mm}
\begin{figure}[!t]
\centering
                \includegraphics[width=.7\linewidth, trim={0cm 2mm 0 0},clip]{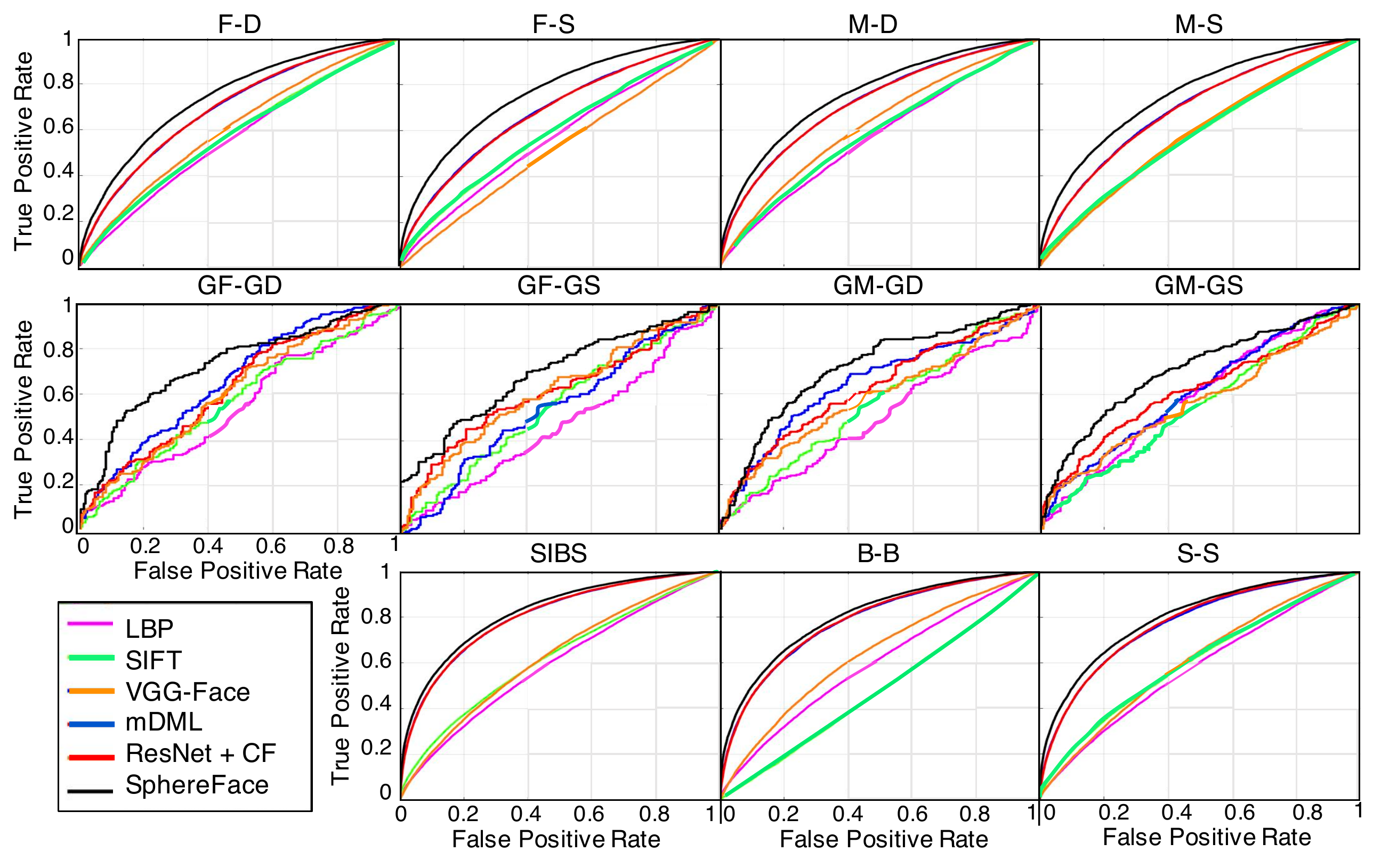}      
        \caption{\textbf{Relationship type specific ROC curves.} Notice the fine-tuned Sphereface dominates, while the sample counts for the \emph{grandparent}-\emph{grandchild} were less as indicative of the jagged curves.}
        \label{chapter:fiw:fig:roc}
\end{figure}

\noindent\textbf{ResNet-22~\cite{wen2016discriminative}} is a 22-layer residual CNN trained on CASIA-Webface~\cite{yi2014learning}. ResNet-22 has a different network topology than VGG (\ie more layers made possible via skipping connections in residual blocks to ensure that the signal stays intact by superimposing an identity tensor). Faces were fed through to layer $fc_5$ (512D encoding).

\subsubsection{Metric Learning}
Metric learning has been commonly used and, frequently, designed for kinship problems. Four metric learning and graph embedding methods used previously for face-based problems include: Information theoretic metric learning (ITML)~\cite{davis2007information}, Discriminative Low-rank Metric Learning (DLML)~\cite{ding2015discriminative}, Locality Preserving Projections (LPP)~\cite{niyogi2004locality}, and Large Margin Nearest Neighbor (LMNN)~\cite{weinberger2009distance}. 

\subsubsection{Deep Learning}
\noindent\textbf{Fine-Tuned CNNs.} Centerface (CF) \cite{wen2016discriminative} loss enhances the discriminative power of deeply learned features by adding a supervision signal to reduce the intra-class variations. SphereFace uses an angular softmax loss, and has most recently claimed state-of-the-art in facial recognition~\cite{liu2017sphereface}. We fine-tune both these CNNs on \gls{fiw}.

Additionally, we include two state-of-the-art methods based on \gls{ae}, graph regularized marginalized Stacked \gls{ae} (GmDAE)~\cite{peng2013marginalized}, and marginalized \gls{dae} based metric learning (mDML)~\cite{kinFG2017}.

\subsection{Kinship Verification}
Kinship verification is a binary classification problem (\ie \textit{true} or \textit{false}, aka \textit{kin} or \textit{non-kin}, respectfully). It is the \textit{one-to-one} view of kinship recognition, which we explain next. 

\subsubsection{Experimental Setting}
The protocol followed is conventional in face-based tasks: 5-fold cross validation with no family-overlap between folds. There are 11 relationship types evaluated (summarized in \tabref{chapter:fiw:tab:pair_count1}). 

For each pair type, we added negative (\ie \textit{non-kin}) pairs to the 5-folds-- we randomly mismatched pairs in each fold until the number of negative and positive pairs are the same in each fold (\ie negative pairs are added at random until it makes up 50\% of the respective fold). Thus, the total number of positive and negative labels are equivalent.

For this task we included each feature, metric learning approach, and deep learning model listed above. We then fine-tuned the pre-trained CNN models on the \gls{fiw} dataset, which is described in detail in the next subsection. To compare features, we computed cosine similarity between each pair, which was then compared to a threshold to classify each pair as either \textit{kin} or \textit{non-kin}. 

Verification accuracy (\ie average of 5-folds) and \gls{roc} curves were used to evaluate. 

\subsubsection{Training Philosophy}
For ResNet-22 + CF, we fine-tuned the Centerface model on our \gls{fiw} data. Training was done using four Titan X GPUs with a batch size of 256. The learning rate was initially set to 0.01, then drops to 0.001 and 0.0001 at the 800 and 1,200 iterations, respectively. Training was complete after 1,600 iterations. The weight decay was set to 0.0005. For SphereFace~\cite{liu2017sphereface}, the settings are similar to ResNet-22+CF (\ie same batch size, learning rate, weight decay, and number of iterations), and with the angular margin set to 4.

\subsubsection{Results}
As listed in Table \ref{chapter:fiw:tab:Veri}, \textit{sibling} pair types tended to score the highest, followed by \textit{parent}-\textit{child} types, and then \textit{grandparent}-\textit{grandchild}. Thus, the wider the generational gap, the wider between appearances of faces.  

SphereFace, which was fine-tuned on \gls{fiw}, outperformed other benchmarks with an average accuracy of 69.18\%, which is 1.31\% and 2.11\% better than ResNet-22+CF and mDML, respectively, which were the top scoring methods prior to the recent release of SphereFace. Also, out of the pre-trained CNNs, VGG-Face scored 3.55\% higher than ResNet-22, and both outperformed the low-level features (\ie LBP \& SIFT). From such, encodings from VGG-Face were used as features for the metric learning and AE methods. Besides LMNN and DLML, which improved score by 0.15\% and 1.30\%, the other metric learning methods actually worsened the performance of the descriptors extracted from the pre-trained VGG-Face CNN. This infers that faces encoded via VGG-Face are more discriminative when used \textit{off}-\textit{the}-\textit{shelf} than when metrics are learned on top.

We show a significant boost in performance when fine-tuning CNNs on \gls{fiw} data-- all features from CNNs outperform the conventional shallow methods. The results show that the deep learning models better encode the complex representation needed to discriminate between \textit{kin}/\textit{non-kin} (see \figref{chapter:fiw:fig:roc}). An improvement to these benchmarks, perhaps via a deep network designed specifically for this task, is certainly a direction for future work.

\subsection{Family Classification}
Family classification is a \textit{one-to-many} problem. The goal is to determine which family an unseen subject came from. In other words, a set of families with a missing member to the model is provided. Then, the missing (\ie unseen) members get classified as being from one of the families (\ie closed form, as we currently assume that all members at test time belong to one of the families modeled during training). We next review some details for this task.

\label{chapter:fiw:sec:famrec}
\subsubsection{Experimental Setting}
Data from 564 families leaving a different single member out in each fold for testing, while data from all the other members was used for training (\ie leave-one-out w.r.t. family members). Families with at least 5 members were used. Thus, the data was split into 5-folds with no family overlap between folds (\ie a minimum of 4 family members for training and 1 for testing). Each fold contained roughly 2,700 images-- about that many faces used to test each split, while the rest, about 12,800 faces, were used for training (\ie a total of 13,420 images).

\subsubsection{Training Philosophy}
VGG-Face and ResNet-22 CNNs were fine-tuned on \gls{fiw} by replacing the loss layers of the pre-trained CNNs with a softmax loss to predict the 564 family classes. There were a few differences: VGG-16 used a fixed learning rate of 0.0001, a batch size of 128, and trained for 800 iterations on one Titan X GPU; ResNet-22 used the same batch size and number of iterations, but with a larger learning rate 0.001, which was fixed too. For ResNet-22 + CF and SphereFace, we followed the same training process used for verification.

\begin{table}[t!]
\centering
\caption{\textbf{Family classification results.} Accuracy scores (\%) using 564 families.}
  \footnotesize
  \centering
  \begin{tabular}{rlc} \toprule
  Run ID & Network(s) &  Acc. \\\midrule
  Run-1 & VGG-Face, $fc_7$ (4,096D)+\textit{one-vs-rest} SVMs& 3.04\\
  Run-2 & VGG-Face, replaced softmax (564D)+fine-tuned & 10.42\\
 Run-3 & ResNet-22  + softmax (564D) & 14.17\\
  Run-4 &  SphereFace (564D) &  13.86\\
  Run-5 &   ResNet-22 + CF (512D) + softmax (564D) &  \textbf{16.18} \\
\bottomrule
\end{tabular} 
 \label{chapter:fiw:tab:famrec}
\end{table}

\subsubsection{Results}
We report the accuracy scores for five runs (see Table \ref{chapter:fiw:tab:famrec}). As shown, the top-1 accuracy for modeling \textit{one-vs-rest} linear SVMs on top of deep VGG-Face features was just 3.04\%. Then, by replacing the softmax layer to target the number of families (\ie 564), and fine-tuning on \gls{fiw}, the top-1 accuracy was improved (\ie +7.38\% to 10.42\%). ResNet-22, also fine-tuned by replacing softmax layer, showed the second to highest accuracy with 14.17\%, which outscored the top performing CNN on verification (\ie SphereFace). The top performance was obtained with the fine-tuned ResNet-22 using Centerface (CF) loss with 16.18\%.

\subsection{Proposed Semi-Supervised Clustering}\label{subsec:semi}
To demonstrate the effectiveness of our semi-supervised model, we cluster \gls{fiw} data using various amounts of \textit{family-level} labels as side information. We simulate two settings for evaluation-- all data and just unlabeled data-- shown as bold and dotted lines, respectively (see Fig. \ref{chapter:fiw:fig:nmiplot}).
	
\subsubsection{Experimental Setting}	
We used $23,979$ faces from $996$ family classes. Faces were encoded using a pre-trained VGG-Face (\ie $fc_7$). We varied the ratio of unlabeled data to side information across the horizontal axis up to 50\% percent of labeled clusters, while the y-axis denotes the clustering performance on the rest of the unlabeled data by NMI. We compared to a pair-wise constrained clustering method, LCVQE~\cite{Pelleg07ECML}, which is also a K-means-based constrained clustering method and transforms the partition level side information into 'must-link' and 'cannot-link' constraints. We used K-means as a baseline (\ie no side-information).
\subsubsection{Results}	
\begin{figure}[t!]
\centering
\includegraphics[width=.6\linewidth]{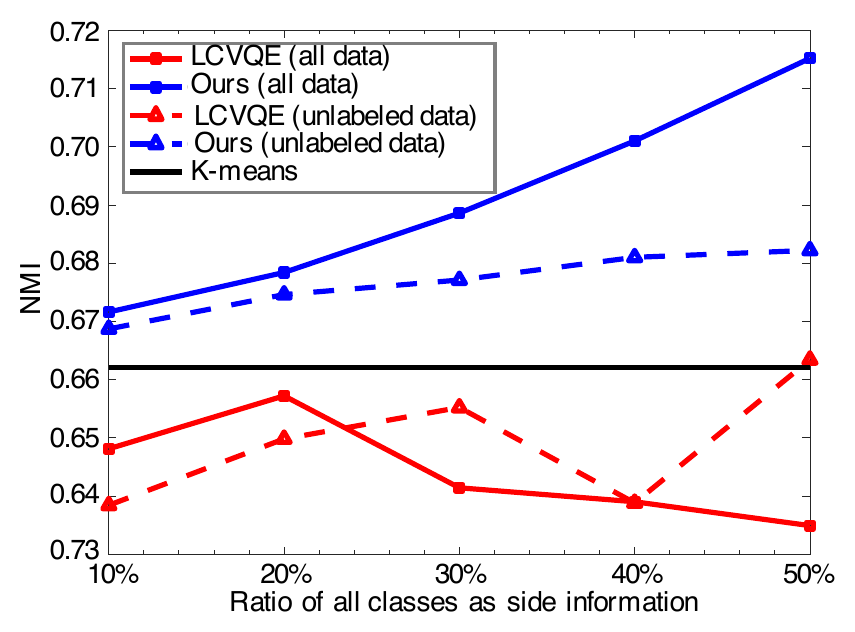}
   \caption{\textbf{Results for clustering families using different amounts of side information.} As clearly depicted, our method obtains the top performance. Moreover, a distinct increase in NMI for our method is shown with an increase in the amounts of side information.}
\label{chapter:fiw:fig:nmiplot}
\end{figure}

Fig. \ref{chapter:fiw:fig:nmiplot} shows a clear boost in performance for our method with more side information. Even on the unlabeled data, our method outperforms K-means, further validating the effectiveness of our method for semi-automatic labeling tasks. For LCVQE, the pair-wise constraints make the cluster structure unpredictable, vulnerable to deviate from the true one, and, thus, perform worse than the baseline. This shows that imposing hard constraints on side information, like 'must-link' and 'cannot-link', may even damper results. On the contrary, our model leverages the side information to only improve when more is added.

\subsection{Transfer-Learning Experiment}\label{subsec:transfer}
To demonstrate that \gls{fiw} generalizes well, we fine-tune the ResNet CNN model on the entire dataset and assess the model on a smaller, non-overlapping image collection. Specifically, we achieve state-of-the-art performance using a fine-tuned CNN to encode faces of the renown KinWild datasets (see \tabref{chapter:fiw:tab:kinwild_eval}). For KinWild I, we get a 4\% increase in performance (\ie from 78.4\% to 82.4\%). For KinWild II, there was a 5.6\% improvement to 86.6\%. 

\begin{table}[t!]

\centering
	\caption{\textbf{Transfer learning experiment}. Accuracy (\%) for KinWild I \& II. CNN fine-tuned on \gls{fiw} top scorer. Note that these results were up-to-date when journal (\ie \cite{robinson2018visual}) was released, but is no longer. See \tabref{chap:fiwmm:tab:kinwild} for most up-to-date scores.}
	\begin{tabular}{@{}r|@{\hskip 0.05in}c@{\hskip 0.05in}c@{\hskip 0.05in}c@{\hskip 0.05in}c@{\hskip 0.05in}r|c@{\hskip 0.05in}c@{\hskip 0.05in}c@{\hskip 0.05in}c@{\hskip 0.05in}r@{}}\toprule
	 	\multicolumn{1}{c}{} & \multicolumn{5}{c}{\textit{KinWild-I}} & \multicolumn{5}{c}{\textit{KinWild-II}} \\
		\cmidrule{2-6}  \cmidrule{7-11}
		Method & FD & FS & MD & MS & Avg. & FD & FS & MD & MS & Avg. \\\midrule
		LBP~\cite{ahonen2006face}&72.8&79.5&71.7&68.1&73.0 & 70.8&78.4&69.0&73.2&72.9\\
		SIFT ~\cite{1467360}&73.9&81.4&76.4&71.1&75.7 & 72.2&78.8&82.2&79.6&78.2\\
		NRML (LBP)~\cite{lu2014neighborhood} &81.4&69.8&67.2&72.9&72.8&	79.2	&71.6&	72.2	&68.4	&72.9\\
		NRML (HOG) &83.7&74.6&71.6&80.0&77.5&	80.8	&72.8	&74.8	&70.4	&74.7\\
		BIU (LBP)~\cite{lu2015fg} &85.5&76.5&69.9&74.4&76.6&	84.2&	79.5&	76.0	&77.0	&79.2\\
		BIU (HOG) &	\textbf{86.9}&76.5&70.6&79.8&78.4&87.5&80.8&	79.8	&75.6&81.0\\
		VGG-Face~\cite{Parkhi15}& 72.0&77.6&	78.3&80.6&77.1 & 68.8&74.4&	76.6&74.6&	73.6\\
		ResNet + CF~\cite{wen2016discriminative} &78.0&\textbf{83.7}&\textbf{87.0}&\textbf{80.8}&\textbf{82.4}&\textbf{87.7}&\textbf{86.0}&\textbf{86.7}&\textbf{87.4}&\textbf{86.6}\\
		\bottomrule
\end{tabular}
\label{chapter:fiw:tab:kinwild_eval}
\end{table}

Notice the significant boost in accuracy for KinWild I for type F-D, and especially compared with M-D. Clearly, the small sample size of these types is not properly represented in KinWild, while \gls{fiw} yields far less variance between scores of parent-child types. Regardless of the high score of \textit{Bar Ilan University} (BIU) for type F-D , our fine-tuned network performs better on all other types, in average accuracy, and while providing less variation in type-specific scores. Again, this variance is caused by the small sample size, as there is less variation in score for the parent-child types of \gls{fiw}.

\subsection{Human assessment using FIW}\label{chap:fiw:sec:humanassess}
We evaluated humans in kinship verification with a subset of \gls{fiw} pairs. Although others conducted similar experiments~\cite{lu2014neighborhood, Ming_CVPR11_Genealogical, dal2006kin}, this was done with a larger sample set made up of more relationship types (\textit{Case 1}). Additionally, an evaluation was done for the Boolean case only (\textit{Case 2}).  We now discuss experimental settings, results, and analyses of both human experiments. 

\subsubsection{Experimental Setting}
First a list of pairs from \gls{fiw} with a fair data distribution was sampled (\ie different and diverse families with faces of various ages). Faces for both positive and negative pairs were from different photos. Also, we used no more than one positive and negative sample per member. We rigorously examined and, in some ways, handcrafted the list to best control the experiment (\ie replaced face images of poorer quality and famous people). Thus, efforts were spent to better ensure a fair, unbiased assessment. We also only used faces to avoid evidence besides facial appearance influencing human responses~\cite{best2014unconstrained}. The same list of images was used for both cases: evaluating pairs per specific relationship types and for the Boolean case only.

A Google Form was used to collect responses, and the university and social media networks to recruit volunteers. Answers were anonymous, although demographic information was collected (\ie ethnicity, country of origin, and gender). Some volunteers completed both experiments. However, scores and answers were not revealed. Also, there was nearly a year between when the two experiments were conducted, with the Boolean case being a follow-up experiment to analyze how specific relationship types influence responses. 
Users chose from predefined responses: \textit{Related}, \textit{Unrelated}, or \textit{Skip}. Participants were asked to \textit{Skip} if they had prior knowledge of one or both subjects, regardless of knowledge about the relationships (\ie skip any pair containing an identifiable face). Face pair-types were processed in no special order: a type-by-type basis for \textit{Case 1}, then shuffled at random for \textit{Case 2}. There was a total of 406 face pairs sampled from the 11 categories (see \tabref{chapter:fiw:tab:humancounts}). 

We had 75 and 110 volunteers for \textit{Case 1} and \textit{2}, respectively. No training of any sort was provided. In both cases, the distribution of demographics was approximately 45\% Caucasian, 35\% Asian, 10\% Hispanic/ Latino, 4\% African American, and 1\% Arab; 65\% born in the United States, 30\% from China, and 1-2\% from South America, Middle East, and the Philippines; 55\% males and 45\% Females. No specific demographics were targeted (\ie a matter who volunteered on social media, per request of the authors, \etc). Future work could involve a greater emphasis on demographics, both in overall distribution of volunteers and intended analysis. Here, we hope to lay the framework for such a study, along with other interesting directions that assessing human ability to recognize kinship can take.

To compare human performance to benchmarks, we fine-tune SphereFace CNN on the 764 families that were not included in face pairs used for the human evaluation.

\begin{table}[t!]
\centering
\caption{\textbf{Face pair counts for human evaluation on kinship verification.} SIBS represents all siblings of the same generation, PC are parent-child, and GPGC are grandparent-grandchild.}
  \footnotesize
  \centering
  \begin{tabular}{lcccc} \toprule
  &SIBS& PC&GPGC &Total\\\midrule
  No. Face Pairs (per type)& 50 &36&28&–-\\
No. Face Pairs (total) & 150 & 144 & 112 & 370 \\
\bottomrule
\end{tabular} 
 \label{chapter:fiw:tab:humancounts}
\end{table}

\subsection{Discussion}
The label structure of \gls{fiw} is dynamic-- labels can be parsed to use the data in various ways. For instance, siblings can be split between those who share one and both parents. Even a slight change in paradigm can drastically change the study-- use both parents for verification (\ie tri-subject verification~\cite{qin2015tri}); use child photos only to test with for family classification. Besides, we still need to improve our visual recognition capability for kinship in current benchmarks. Then, it only seems natural to aim for fine-grained categorization of entire family trees (\ie the ultimate achievement). On a different note, generative modeling is another interesting research track to pursue (\eg given a couple and predict the offspring, or samples of their baby and predict the baby's appearance as an adult). Even other pair types (\eg great- and great-great-grandparents, cousins, aunts, uncles, \etc). Also, the labeling framework introduced in this work could be used to add video data to the families of \gls{fiw}, which can be served as a resource for template- based search and retrieval, or even consider emotional responses and facial expressions of family members.

We expect that as researchers advance this problem, \gls{fiw} and its uses too will advance, and especially when considering the potential for interdisciplinary collaborations-- Whether nature-based studies, generative or predictive modeling, or security-based. We hope \gls{fiw} inspires new types of problems, and anticipate the list of uses to only grow when \gls{fiw} is in the hands of researchers worldwide. In the end, the aim here is to attract more experts to the problem of kinship recognition.

\textit{Families In the Wild} (FIW) is the first large-scale dataset available for visual kinship recognition. We annotated complex hierarchical relationships with only a small team in a fast and efficient manner-- providing the largest labeled collection of family photos to-date.  \gls{fiw} was structured to support multiple tasks with its dynamic label structure. We provided several benchmarks for kinship verification and family classification. Pre-trained CNNs were used as \textit{off-the-shelf} face encoders, which outperformed conventional methods. Results for both tasks were further improved by fine-tuning the CNN models on \gls{fiw}. We measured human observers and compared their performance to the machine vision algorithms, showing that CNNs surpass humans in recognizing kinship.


\section{Data challenges and incentives}
Challenges date back to 2011, where multi-modal data for twins was collected annually and in a highly controlled setting (\ie \emph{Twins Day}~\cite{vijayan2011twins}). Also, starting in 2014 were data challenges on unconstrained face data~\cite{lu2014kinship}. Then, Lu~\etal attracted many with a \gls{fg} challenge with \gls{kfw}~\cite{lu2015fg}. Robinson~\etal expanded the data challenges as part of a 2017 ACM MM Workshop using the first large-scale visual kinship recognition dataset~\cite{robinson2017recognizing}, which was followed by three consecutive \gls{fg} challenges - an annual effort that still occurs nowadays~\cite{robinson2020recognizing} (\ie 2018-2020, with 2020 still accessible via Codalab\footnote{\href{https://competitions.codalab.org/competitions/21843}{https://competitions.codalab.org/competitions/21843}}). Besides, over five hundred teams partook in \gls{rfiw} on Kaggle.\footnote{\href{https://www.kaggle.com/c/Recognizing-Faces-in-the-Wild}{https://www.kaggle.com/c/Recognizing-Faces-in-the-Wild}} Recently, there have been several tutorials at top-tier conferences (\ie ACM MM18~\cite{robinson2018recognize}, CVPR 2019\footnote{\href{https://web.northeastern.edu/smilelab/fiw/cvpr19_tutorial/}{https://web.northeastern.edu/smilelab/fiw/cvpr19\_tutorial/}}, and \gls{fg} 2019\footnote{\href{http://fg2019.org/participate/workshops-and-tutorials/visual-recognition-of-families-in-the-wild/}{http://fg2019.org/visual-recognition-of-families-in-the-wild}}). The human evaluations were done using volunteers in a non-competitive forum.


\section{Experimental}\label{sec:sota:visualkin}

\begin{table}[t!]
    \centering
    \caption{\textbf{Kinship verification (T1) counts.} Number of unique pairs (\textbf{P}), families (\textbf{F}), and face samples (\textbf{S}), with an increase in counts and types since~\cite{robinson2017recognizing}.}
         \resizebox{\textwidth}{!}{%
    \begin{tabular}{p{.1in}m{.1in}|ccc|cccc|cccc|c}
        & & BB& SS& SIBS& FD & FS & MD & MS & GFGD & GFGS & GMGD & GMGS& Total\\\hline
     \parbox[t]{2mm}{\multirow{3}{*}{\rotatebox[origin=c]{90}{\emph{Train}}}}&\textbf{P} & 991  & 1,029 &1,588 & 712 & 721& 736& 716 & 136 & 124 & 116 & 114 &6,983\\
    \multirow{3}{*}{} &\textbf{F}  &303 & 304 & 286 & 401 & 404 & 399 & 402 & 81 & 73&71 & 66 &2790\\
    \multirow{3}{*}{} &\textbf{S} &39,608& 27,844 & 35,337& 30,746  &46,583 & 29,778&  46,969& 2,003 &  2,097  &1,741 & 1,834  &264,540\\\hline
    
    \parbox[t]{2mm}{\multirow{3}{*}{\rotatebox[origin=c]{90}{\emph{Val}}}} &\textbf{P}  & 433 & 433 & 206& 220 & 261 & 200 & 234 & 53 & 48 & 56 & 42 & 2,186 \\
    
    \multirow{3}{*}{} &\textbf{F}  &74  & 57& 90 & 134& 135& 124& 130& 32& 29& 36&27 &868\\
    \multirow{3}{*}{} &\textbf{S}  & 8,340 & 5,982 & 21,204& 7,575 &9,399&8,441 &7,587 & 762 &879 & 714 & 701 & 71,584\\\hline

    \parbox[t]{2mm}{\multirow{3}{*}{\rotatebox[origin=c]{90}{\emph{Test}}}} &\textbf{P}  &  469& 469 & 217 & 202& 257 & 230 & 237 & 40 & 31 & 36 & 33&2,221 \\
    \multirow{3}{*}{} &\textbf{F}  & 149  & 150  & 89 & 126 & 133 & 136 & 132 & 22 & 21 & 20 & 22 & 1,190\\
    \multirow{3}{*}{} &\textbf{S}  & 3,459 &2,956 &967 &3,019&3,273&3,184& 2,660 &121&96&71&84&39,743\\
    
    \end{tabular}}\label{chap:fiwmm:tab:track1:counts} 
\end{table}

The organization of this section is as follows. First, we examine studies involving a human's ability to recognize kinship as imagery. Thus, deeming the soft-attribute of kinship as being detectable by the eye. Next, we review kin-based task protocols - each complete with a problem statement, data splits, metrics, and baseline solutions. We then highlight commonalities in problem formulation and proposed solutions for the various tasks. Following this, we describe traditional and deep solutions. We then put this in perspective with the \gls{rfiw} data challenge series - four editions (\ie 2017~\cite{robinson2017recognizing}-2020~\cite{robinson2020recognizing} and Kaggle Competition\footnote{\href{https://www.kaggle.com/c/recognizing-faces-in-the-wild}{https://www.kaggle.com/c/recognizing-faces-in-the-wild}} held just prior to the 2020 \gls{rfiw}). Finally, we discuss recent attempts to predict the appearance of family members' faces.

\subsection{Task Evaluations, Protocols, Benchmarks}\label{sec:task:eval}
We next describe each kin-based task separately: the problem statement and motivation, data splits and protocols, and benchmark experiments (\ie baselines). A brief section on the common experimental settings precedes the detailed descriptions of settings unique to the task and follow in separate subsections.

\subsubsection{Common settings}
The \gls{fiw} dataset provides the most extensive set of face pairs for kin-based face recognition. \gls{fiw} provides the data needed to train modern-day data-driven deep models~\cite{duan2017advnet, li2017kinnet, wu2018kinship}. With over 12,000 family photos for 1,000 disjoint family trees the data contains various counts for faces, samples, members, and relationships per family. Hence, the faces of the image collection are cropped out and organized by family-- faces of each family member ranges from one-to-many. \gls{fiw} is split into three parts:  \emph{train}, \emph{val}, and \emph{test}. Specifically, 60\% of the families were assigned to the  \emph{train} set; the remaining 40\% was split evenly between \emph{val} and \emph{test}. The three sets are disjoint in family and identity. The test set remains ``blind'', with automatic scoring of submissions added to the leadership board of the codalab competition. Note that the splits are consistent across tasks, so the same families makeup the ``blind'' set.

\begin{table}[ht!]
\centering
\caption {\textbf{Kinship verification (T1) results.} Averaged verification accuracy scores of \gls{rfiw}.}
\label{tab:benchmark:track1}
         \resizebox{\textwidth}{!}{%
\begin{tabular}{r|ccc|cccc|cccc|c}
  Methods& BB & SS & SIBS & FD & FS & MD & MS & GFGD & GFGS & GMGD & GMGS  & Avg. \\
  \midrule
  ArcFace~\cite{wang2018additive} (baseline)& 0.57 & 0.64 & 0.50 & 0.61 & 0.66 & 0.69 & 0.62 & 0.66 &0.71& 0.73 & \textbf{0.68}  & 0.64\\
    stefhoer~\cite{id2} & 0.66 & 0.65 & 0.76& \textbf{0.77} & 0.80 & 0.77 & \textbf{0.78} & 0.70 & \textbf{0.73} & 0.64 & 0.60  & 0.74\\
     ustc-nelslip~\cite{id6}& 0.75 & 0.74 & 0.72& 0.76 & 0.82 & 0.75 & 0.75 & \textbf{0.79} & 0.69 & \textbf{0.76} & 0.67  & 0.76\\
     DeepBlueAI~\cite{id3}& 0.77 & 0.77 & 0.75 & 0.74 & 0.81 & 0.75 & 0.74 & 0.72 & \textbf{0.73} & 0.67 & \textbf{0.68}  & 0.76\\
  vuvko~\cite{id4}& \textbf{0.80} & \textbf{0.80} & \textbf{0.77}& 0.75 & \textbf{0.81} & \textbf{0.78} & 0.74 & 0.78 & 0.69 & \textbf{0.76} & 0.60  & \textbf{0.78}\\
\end{tabular}}
\end{table}

As part of pre-processing, faces for all three sets were encoded via ArcFace \gls{cnn}~\cite{wang2018additive}  (\ie 512 D). All pre-processing and the model weights were from the original work.\footnote{\href{https://github.com/ZhaoJ9014/face.evoLVe.PyTorch}{https://github.com/ZhaoJ9014/face.evoLVe.PyTorch}} Also common is the use of cosine similarity to determine closeness of a pair of facial features $p_1$ and $p_2$~\cite{nguyen2010cosine}. This is defined as
$$
\text{CS}(\pmb p_1, \pmb p_2) = \frac {\pmb p_1 \cdot \pmb p_2}{||\pmb p_1|| \cdot ||\pmb p_2||}.
$$

Scores were then either compared to threshold $\gamma$ (\ie $\text{CS}(p_1, p_2) > \gamma$ infers \emph{KIN}; else, \emph{NON-KIN}) or sorted (\ie ranked list). This concludes the common experimental settings.

Teams were allowed up to six final submissions per task. Submissions were accompanied by a brief (text) description of the system used to generate results. 

\subsubsection{Kinship verification (T1)}\label{sec:kinver}
Kinship verification aims to determine whether a face pair are blood relatives. This classical Boolean problem has two possible outcomes, \emph{KIN} or \emph{NON-KIN}. Hence, this is the \textit{one-to-one} view of kin-based problems. The classical problem can be further extended by considering the type of kin relation between a pair of faces, rather than treating all kin relations equally~\cite{robinson2018recognize}.

Prior research mainly considered parent-child kinship types, \ie \gls{fd}, \gls{fs}, \gls{md}, \gls{ms}. Less attention has been given to sibling pairs, \ie \gls{ss}, \gls{bb}, and \gls{sibs}. Research findings in psychology and computer vision found that different relationship types share different familial features~\cite{shao2011genealogical}. Hence, each relationship type can be modeled and evaluated independently. Thus, additional kinship types would further both our understanding and capabilities of automatic kinship recognition. With \gls{fiw}, the number of facial pairs accessible for kinship verification has dramatically increased. Additionally, benchmarks now include grandparent-grandchildren types, \ie \gls{gfgd}, \gls{gfgs}, \gls{gmgd}, \gls{gmgs}.

\vspace{1mm}\noindent\textbf{Data splits.}

The two datasets used (\ie \gls{kfw} and \gls{fiw}) follow the same settings. Both provide a list of pairs labeled as KIN and NON-KIN. The differences are in the number of pair types and overall size of splits. Data specifications are in Table~\ref{chap:fiwmm:tab:track1:counts}.

\gls{kfw} provides two sets (\ie \gls{kfw} I \& II) and the four parent-child pair types. \gls{fiw} spans eleven different relationship types - the types used in 2020 \gls{rfiw} (\tabref{chapter:fiw:fig:allpairs}). The {\emph test} set is made up of an equal number of positive and negative pairs and with no family (and, hence, no identity) overlap between sets.

\vspace{1mm}\noindent\textbf{Settings and metrics.}\label{subsec:track1:settings}

\begin{figure}[t!]
    \centering
    \includegraphics[width =.65\linewidth]{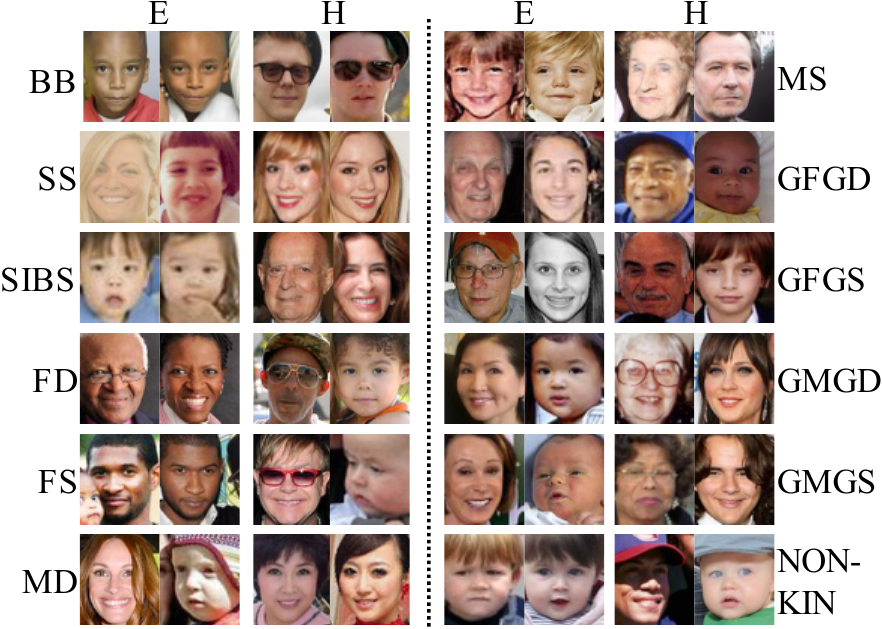}
    \caption{\textbf{Kinship verification (T1) sample pairs.} Sample pairs with similarity scores near the threshold (\ie hard (H) samples), along with highly confident predictions (\ie easy (E) samples) in verification task.}
    \label{fig:track1:samples}
    \vspace{1mm}
\end{figure}

Verification accuracy is used to evaluate. Specifically,

$$
\text{Accuracy}_j = \frac{\text{\# correct predictions for j-th type}}{\text{Total \# of pairs for j-th type}},
$$
where $j\in\{$4 relationship types and $\O$ for \gls{kfw} and 11 relationship types and $\O$ for \gls{fiw}$\}$ (listed in Fig.~\ref{fig:track1:samples}). Then, the overall accuracy is calculated as a weighted sum (\ie weight by the pair count to determine the average accuracy).

\vspace{1mm}\noindent\textbf{Baseline and results.}
The threshold was determined by the value that maximizes the accuracy on the \emph{held-out} data in all cases. The results on \gls{kfw} I and II are show in \tabref{chapter:fiw:tab:kinwild_eval}. The results for \gls{fiw} are in Table~\ref{tab:benchmark:track1}, with sample pairs that either 100\% or 20\% of all teams got correct are shown in Fig.~\ref{fig:track1:samples:submitted}. 

\subsubsection{Tri-Subject verification (T2)}\label{sec:trisubject}
Tri-Subject Verification (T2) focuses on a different view of kinship verification-- the goal is to decide if a child is related to a pair of parents. First introduced in~\cite{qin2015tri},  it makes a more realistic assumption, as having knowledge of one parent often means the other potential parent(s) can be easily inferred.

Triplet pairs consist of Father ({F}) / Mother ({M}) - Child ({C}) ({FMC}) pairs, where the child {C} could be either a Son ({S}) or a Daughter ({D}) (\ie triplet pairs are {FMS} and {FMD}).

\vspace{1mm}\noindent\textbf{Data splits.} Following the procedure in \cite{qin2015tri}, we create positive (have kin relation) triplets by matching each husband-wife spouse pair with their biological children, and negative (no kin relation) triplets by shuffling the positive triplets until every spouse pair is matched with a child which is not theirs (Table~\ref{tbl:track2:counts}).

The number of potential negative samples far exceeds the number of potential positive examples-- We post-process the positive triplets before generating negatives to ensure balance among individuals, families, and spouse pairs, since a naive data selection procedure which weights every face sample similarly would result in some individuals and families being severely over-represented due to an abundance of face samples for some identities and families. 
The post-processing is done by limiting the number of samples of any triplet $(F, M, C)$, where $F$, $M$, and $C$ are identities of a father, mother, and child to 5, then limiting the appearance of each $(F, M)$ spouse-pair to 15, and then finally limiting the number of triplet samples from each family to 30. The \emph{test} set has an equal number of positive and negative pairs. Lastly, note that there is no family or subject identity overlapping between any of the sets.

\vspace{1mm}\noindent\textbf{Settings and metrics.} Per convention in face verification, we offer 3 modes (\ie the same as in task 1 listed in Section~\ref{subsec:track1:settings}). Again, the metric used is verification accuracy, which is first calculated per triplet-pair type (\ie FMD and FMS). Then, the weighted sum (\ie average accuracy) determines the leader-board.

\begin{figure}[t!]
    \centering
    \includegraphics[width =.65\linewidth]{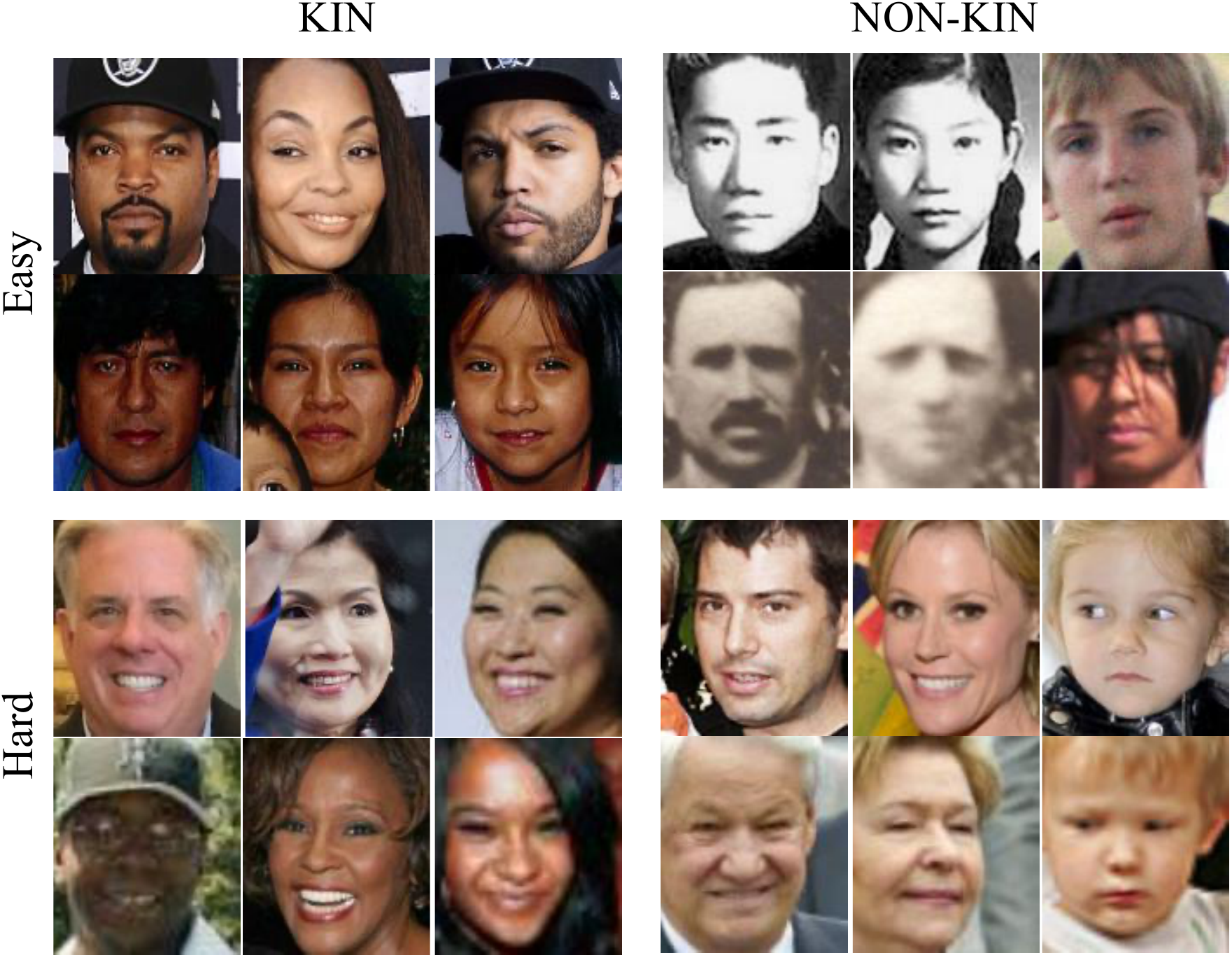}
    \caption{\textbf{Triplets with extreme scores (\ie correct and incorrect).} Each show FMS (top rows) and FMD (bottom) for tri-subject (T2).}
    \label{fig:track2:montage}
\end{figure}

\vspace{1mm}\noindent\textbf{Baseline and results.}
Baseline results are shown in Table~\ref{tab:benchmark:track2}, with samples of easier and more challenging samples for both \emph{KIN} and \emph{NON-KIN} triplets in Fig.~\ref{fig:track2:montage} and \ref{fig:track2:samples:submitted}. A score was assigned to triplet $(F_i, M_i, C_i)$ in the validation and \emph{test} sets using the formula 
$$\text{Score}_{i} =  avg(\cos{(F_i, C_i)}, \cos{(M_i, C_i)}),$$
where $F_i$, $M_i$ and $C_i$ are the feature vectors of the father, mother, and child images respectively from the i-th triplet. 
Scores were compared to a threshold $\gamma$ to infer a label (\ie predict KIN if the score was above the threshold; else, NON-KIN). 
The threshold was determined experimentally on the \emph{val} set and used for \emph{test}.

\subsubsection{Search and retrieval (T3)}\label{sec:search}
A newer task, search and retrieval (T3), is posed as a \textit{many-to-many}, \ie one-to-many samples per subject. Thus, we imitate template-based evaluations on the probe side, but with the gallery now labeled by family. Furthermore, the goal is to find relatives of search subjects (\ie \textit{probes}) in a search pool (\ie \textit{gallery}).

\begin{figure}[t!]
    \centering
    \includegraphics[width = .76\linewidth]{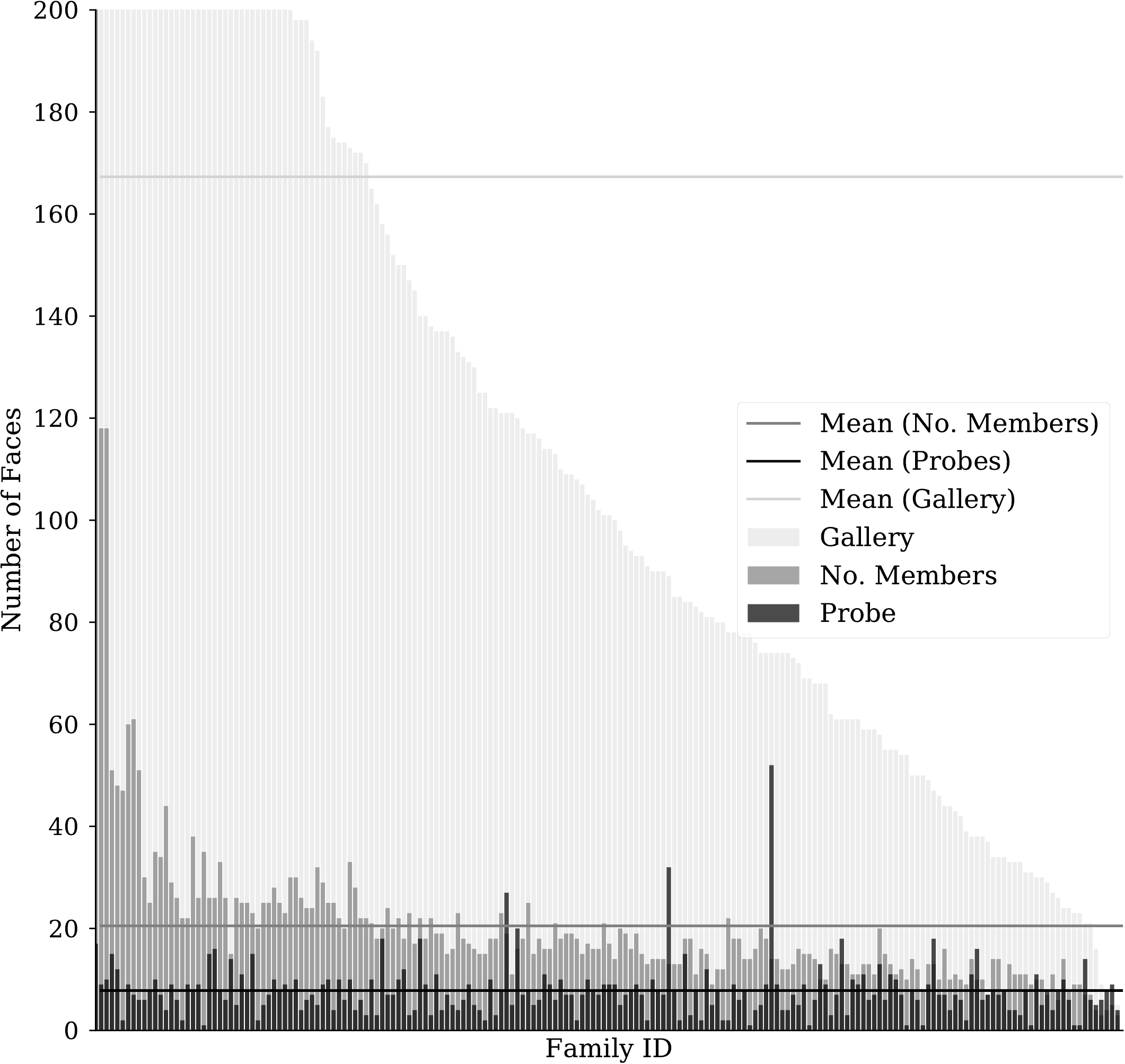}
    \caption{\textbf{Plot of face counts per family in {\emph test} set of T3}. The probes have about 8 faces on average, while the number of family members in the gallery nears 20 on average, with a total average of 170 faces.}
    \label{fig:track3:counts}
\end{figure}

The protocol of T3 could be used to find parents and other relatives of unknown, missing children. The gallery contains 31,787 facial images from 190 families (Fig.~\ref{fig:track3:counts}): inputs are subject labels (\ie probes), and outputs are ranked lists of all faces in the gallery. The number of relatives varies for each subject, ranging anywhere from 0 to 20+. Furthermore, probes have one-to-many samples-- the means of fusing samples of probes is an open research question. This \textit{many-to-many} task is currently set up in closed form (\ie every probe has a relative(s) in the gallery).

\vspace{1mm}\noindent\textbf{Data splits.}
This task will be composed of search subjects (\ie \textit{probes}) from different families. \textit{Probes} are supported by several samples of query subject, text description of family (\eg ethnicity, some relationships between selected members, \etc), and list of relatives present in the \textit{gallery}. The \textit{test} set will only consist of sets of images for the probes. Diversity in terms of ethnicity is ensured for both sets. Again, three disjoint sets were split (Table~\ref{tbl:track3:counts}).

\vspace{1mm}\noindent\textbf{Settings and metrics.} 
Each subject (\ie probe) is searched independently-- 190 subjects with one-to-many faces. Hence, 190 families make up the \textit{test} set. Following template conventions of other \textit{many-to-many} face evaluations, face images of unique subjects are separated by identity, with a gallery containing various number of relatives with a variable number of faces each~\cite{whitelam2017iarpa}.

Mean average precision (MAP) was the underlying metric used for comparisons. Mathematically speaking, scores for each of the $N$ missing children are calculated as follows:
$$\text{AP}(f)=\frac{1}{P_F}\sum^{P_F}_{tp=1}Prec(tp)=\frac{1}{P_F}\sum^{P_F}_{tp=1}\frac{tp}{rank(tp)},$$
where average precision (AP) is a function of family $f$ with a total of ${P_F}$ \gls{tpr}. MAP is then
$$\text{MAP} = \frac{1}{N}\sum^{N}_{f=1}\text{AP}(f).$$

\begin{table}[!b]
    \centering
    
    \caption{\textbf{Tri-subject verification (T2) counts}. No. pairs (\textbf{P}), families (\textbf{F}), face samples (\textbf{S}).}
    \begin{tabular}{p{.1in}m{.1in}ccc}
    & &FM-S &FM-D &Total\\\hline
     \parbox[t]{2mm}{
     \multirow{3}{*}{\rotatebox[origin=c]{90}{train}}}&\textbf{P} & 662  & 639 &1,331 \\
    \multirow{3}{*}{} &\textbf{F}  &375 & 364 & 739\\
    \multirow{3}{*}{} &\textbf{S} &8,575& 8,588 &  17,163\\\hline
    
    \parbox[t]{2mm}{
    \multirow{3}{*}{\rotatebox[origin=c]{90}{val}}} &\textbf{P}  & 202 & 177 & 379 \\
    \multirow{3}{*}{} &\textbf{F}  &116  & 117& 233\\
    \multirow{3}{*}{} &\textbf{S}  & 2,859 & 2,493 & 5,352 \\\hline
    \parbox[t]{2mm}{
    \multirow{3}{*}{\rotatebox[origin=c]{90}{test}}} &\textbf{P}  &  205& 178 & 383  \\
    \multirow{3}{*}{} &\textbf{F}  & 116  & 114  & 230 \\
    \multirow{3}{*}{} &\textbf{S}  & 2,805 &2,400 &5,205\\\hline
    
    \end{tabular}\label{tbl:track2:counts} 
\end{table}

\vspace{1mm}\noindent\textbf{Baseline and results.}
Submissions consisted of a matrix with a row per \emph{probe} listing the indices of all subjects in the \textit{test} gallery as a ranked list. Results are listed in Table \ref{tbl:t3:benchmarks} with sample inputs and predictions shown in Fig.~\ref{fig:track3:montage}.

\begin{figure}[t!]
    \centering
    \includegraphics[width =.65\linewidth]{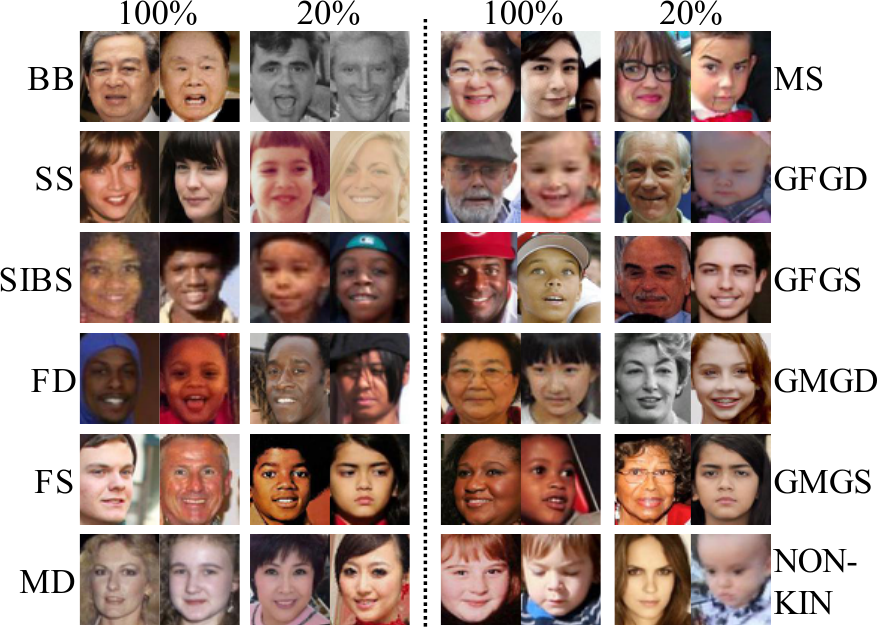}
    \caption{\textbf{Qualitative analysis of T1.} Samples of each relationship type that all of the teams either got correct (100\%) or mostly not (20\%) for the eleven pair types of \gls{fiw} and NON-KIN.}
    \label{fig:track1:samples:submitted}
\end{figure}

\section{Methodologies}\label{sec:methods}
Many formulated kinship recognition problems in the visual domain as multi-view, multi-task, and multi-modal, which is typically to increase the amount of information obtainable, even when the final target is among other targets during training (\ie auxiliary tasks that complement the knowledge obtained from recognition, alone). For instance, the \gls{dkmr} was proposed as a jointly-trained model on top of a graph optimization algorithm~\cite{liang2017using}. Clearly, deep learning has overcome the traditional metric-learning approaches from about 2017~~\cite{li2017kinnet, qin2018heterogeneous, wei2019adversarial, zhu2019visual, mukherjee2019kinship} and still today~\cite{zhang2020advkin, laiadi2020tensor, wang2020kinship, multiperson2020, AsianConferenceonPatternRecognitionAucklandN}. We will first review the traditional methods, and then deep learning, for discriminating problems; an overview of the kin-based generative modeling is given at the end of the section.

\subsection{Traditional approaches}
The main focus of the survey is on large data resources, along with the modern-day complex, data-driven modeling (\ie deep learning). However, such respective work makes up the latter half of the decade. Feature and metric-learning dominated the first half of this past decade in research of visual kinship recognition-- before the release of \gls{fiw}. For completeness, we will introduce several methods that predate the deep methods on \gls{fiw}.

\vspace{1mm}\noindent\textbf{Handcrafted features}. Fang~\etal proposed using features such as geometric differences between face parts, color features, and handcrafted features that were the basis for the metrics to be learned in the years to come~\cite{fang2010towards}. Furthermore, and as mentioned, many of the smaller datasets are limited in diversity (\ie all similar demographics) and with pairs from the same photos, from which some proposed color-based features~\cite{crispim2020verifying}. Still, papers that hone-in on the smaller data employ more classical approaches, such as representation learning via binary trees~\cite{RAVIKUMAR2020e03751}.

\vspace{1mm}\noindent\textbf{Metric learning}. Metric learning methods are popular solutions in kin-based vision problems. The general idea is to optimize a metric between classes. In kinship verification, the classes are KIN and NON-KIN (\ie true match and imposter, respectively). Lu~\etal proposed \gls{nrml} for kinship verification which aims for a
contractive deep belief net (fcDBN) made by stacking fcRBMs to learn weights in a greedy, layer-by-layer fashion using both local and global features~\cite{kohli2016hierarchical}.

Wu~\etal combined color and texture features for kinship verification with extreme learning machines (ELM) for robustness on small data~\cite{wuX2018kinship}. Mahpod~\etal proposed a hybrid asymmetric distance learning (MHDL) scheme, combining symmetric and asymmetric multiview distances~\cite{mahpod2018kinship}. Most recently, Hu~\etal proposed treated-different features as multiple views via a multi-view geometric mean metric learning (MvGMML)~\cite{hu2019multi}.

For more details on the traditional methods see~\cite{qin2019literature}. 

\begin{table}[!b]
\centering
\caption {\textbf{Verification scores.} Results for tri-subject (\ie T2).}
\label{tab:benchmark:track2}
\begin{tabular}{r|cc|c}
  &FMS & FMD & Avg. \\
  \midrule
  
  Sphereface~\cite{wang2018additive} (baseline) & 0.68 & 0.68 & 0.68 \\ 
    stefhoer~\cite{id2} & 0.74 & 0.72 & 0.73 \\
  DeepBlueAI~\cite{id3}  & 0.77 & 0.76 & 0.77 \\
 ustc-nelslip~\cite{id6}  & \textbf{0.80} & \textbf{0.78} & \textbf{0.79} \\
\end{tabular}
\end{table}

\subsection{Deep learning approaches}
The 2012 AlexNet~\cite{krizhevsky2012imagenet} sparked the deep learning era. As done in many problems, deep learning grew more popular with the big data provided with \gls{fiw}. Still on \gls{kfw}, we first review the deep metric done using small amounts of training data, and then discuss the data-driven work done using \gls{fiw}.

There are many commonalities between the different solutions proposed as part of the \gls{rfiw} challenge. Typically, a ResNet-based~\cite{he2016deep} backbone; if not, then together with FaceNet~\cite{schroff2015facenet}. Nonetheless, the story as seen in the timeline is split in half (\ie with the latter half dominated by modern-day deep learning approaches) and quite significantly, metrics learned on top of hand-crafted features dominated the charts as \gls{sota} for many years~\cite{8803754}. As recent as 2017, metric-learning was a go-to approach for kin-based problems, whether a single metric or multiple (\eg \gls{lm3l}~\cite{hu2017local}). Even so, geometric and distant features in pixel space (\eg key point coordinates on neutral face~\cite{KALJAHI2019100008})-- directly related to insufficient data for modern-day data-driven machinery (\ie deep learning). 

\begin{figure}[t!]
    \centering
    \includegraphics[width =.65\linewidth]{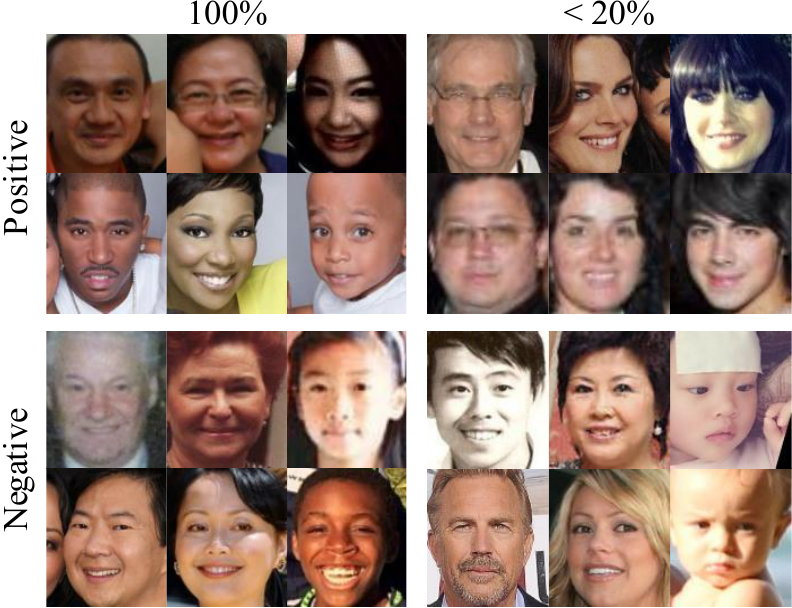}
    \caption{\textbf{Sample of T2.} Samples that all teams got correct (left) and mostly incorrect (right) for FMS (top rows) and FMD (bottom).}
    \label{fig:track2:samples:submitted}
\end{figure}

Provided a deep \gls{cnn} trained to classify face identity, the encodings produced encapsulated much information of the subject. However, instead of looking for absolute closeness in embedding space as the ideal case for a set of samples of a single class (\ie identity), in kin-based tasks we hope to detect when similarities between a pair (or group) of faces (\ie encoded) reflect that of the various relationship-types. For this, many tend to fine-tune models initially trained on a larger \gls{fr}-based database, such as
VGG-Face~\cite{schroff2015facenet}, VGG2~\cite{cao2018vggface2}, and MSCeleb~\cite{guo2016ms}. We next speak on various flavors of deep learning.
		
\begin{table}[!t]
    \centering
    \caption{\textbf{Tri-subject (T2) counts.} Individuals \textbf{I}, families \textbf{F}, face samples \textbf{S}.}
    \begin{tabular}{p{.1in}m{.1in}ccc}
    & &Probe &Gallery &Total\\\hline
     \parbox[t]{2mm}{
     \multirow{3}{*}{\rotatebox[origin=c]{90}{train}}}&\textbf{I} & --  & 3,021 & 3,021 \\
    \multirow{3}{*}{} &\textbf{F}  &-- & 571 & 571\\
    \multirow{3}{*}{} &\textbf{S} & --& 15,845 & 15,845 \\\hline
    
    \parbox[t]{2mm}{
    \multirow{3}{*}{\rotatebox[origin=c]{90}{val}}} &\textbf{I}  & 192 & 802  & 994  \\
    \multirow{3}{*}{} &\textbf{F} & 192 & 192 & 192  \\
    \multirow{3}{*}{} &\textbf{S}  &1,086  &4,030 &5,116 \\\hline
    
    \parbox[t]{2mm}{
    \multirow{3}{*}{\rotatebox[origin=c]{90}{test}}} &\textbf{I}& 190 & 783  & 9d73 \\
    \multirow{3}{*}{} &\textbf{F} &190  & 190  & 190   \\
    \multirow{3}{*}{} &\textbf{S}  &1,487  & 31,787 & 33,274\\\hline
    
    \end{tabular}\label{tbl:track3:counts} 
\end{table}
\vspace{1mm}
\noindent\textbf{Pre-trained CNNs.} Besides that most solutions involve the renowned Siamese training model, many of which still incorporate a cosine loss as in the seminal work done at Bell Lab's mid-90s~\cite{bromley1994signature}, \ie multiple inputs to networks with shared weights for which metric is learned on top (Fig~\ref{fig:siamese}). In the simplest form, Siamese-based \gls{cnn} models map two or more samples by a single \gls{cnn} to a real-number vector space $\mathrm{R}^d$ (\ie a function $f(\cdot)$ to encode an image (\ie facial [encoding, embedding, feature] of size $d$, especially in the context of facial representation, all refer to the $f(x_i)=z_i\in\mathrm{R}^d$. Generally, and in most methods proposed in \gls{rfiw}, the shared model is pre-trained data for another, yet similar task (\ie facial recognition). With that, the \gls{cnn} that now serves as an encoder, maps $k$ samples to its $d$-dimensional space learned to discriminate between faces. With the Siamese frozen-- whether entire network, with a couple of layers on top set with a small learning rate, or popped off by adding a path that splits off prior to later rejoin or just remove entirely-- the goal then is to learn a metric optimal for recognizing family members by face cues. Clearly, there are several design choices-- with simple solutions in those with an \emph{off-the-shelf} \gls{cnn} with no additional training (\ie trained for \gls{fr}, so naively assuming that the best way to detect kinship is to detect faces that look like the source). However simple, and with many cases a fair assumption, the naive approach outperformed previous \gls{sota} methods prior to \gls{fiw} providing the number of data samples needed to suffice the capacity of most deep learning approaches. In light of this, the \gls{cnn} then serves as the method for feature extraction-- claiming to provide the best face representations for the task. As previously described of the wave of metric-based and subspace-modeling methods, we can then further refine the output of the feature extractor by extending the composition function by adding and training mappings in the embedding space and while again, often with the weights of the pre-trained \gls{cnn} $f$ held static. From this, kin-based tasks can be targeted by learning filters, mappings, and even metrics from the embedding space on up (\ie build up from the embedding space from where face embeddings are compared in some fashion). 

\vspace{1mm}
\noindent\textbf{Deep metric learning.} 
Lu~\etal proposed to learn a distance metric for $K$ feature types via $K$ MLPs - learn to project each feature using the optimal thresholds determined independently~\cite{lu2017discriminative}. This method, which was called discriminative deep metric learning (DMML), proved effective on the \gls{kfw} settings of minimal training data (Table~\ref{chapter:fiw:tab:kinwild_eval}).

\vspace{1mm}
\noindent\textbf{Fine-tuning.} 
There is an abundant of public \gls{fr} data (\eg LFW, VGG, MSCeleb~\cite{guo2016ms}) with some labeled by \emph{soft attributes} (\eg age~\cite{zhifei2017cvpr}, gender~\cite{cheng2019exploiting}, attribute, and diverse demographics~\cite{robinson2020face, DBLP:journals/corr/abs-1812-00194}). With this, and provided the known concept of deep learning tending to learn transferable features~\cite{zhang2018survey}, the use of fine-tuning pre-trained has been done by many. For instance, a SphereFace loss, which is a multi-class loss, is first used to train a large \gls{cnn} to do facial recognition on identities of an auxiliary dataset, and then having the layers near the top fine-tuned to recognize the families of the \gls{fiw} training set via

\begin{equation}
    \mathcal{L}_{\text{family}}(\theta) = -\frac{1}{B} \sum_{i=1}^{B}\log\frac{\exp^{W_{y_i}^Tx_i+b_{y_i}}}{\sum_{j=1}^{N}\exp^{W_{y_i}^Tx_i+b_{j}}},
    \label{eq:arcfamily}
\end{equation}
where $B$ is the batch size, $N$ is the number of families, $x_i$ is the face encoding from family $y_i$, $W$ is the weight matrix (\ie $W_j$ denotes the $j^{th}$ column) and $b$ is the bias term. In the end, verifying kinship between a face pair can be done using the model to encode the faces and cosine distance to measure their closeness. If family labels are unavailable, which is another setting of the verification task, approaches tend to use Siamese concepts on top of the pre-trained \gls{cnn} (Fig.~\ref{fig:siamese}). Specifically, sharing weights for two or more samples, and penalizing based on the closeness between a set of samples upon being encoded by the network, has shown to be an effective means of staging a network for the verification task. In return, Siamese; furthermore, the relationship between the pairs with respect to labels at training differences is in preprocessing, method of fusion (\eg \emph{early} versus \emph{late}).
\begin{figure}[t!]
    \centering
    \includegraphics[width = .65\linewidth]{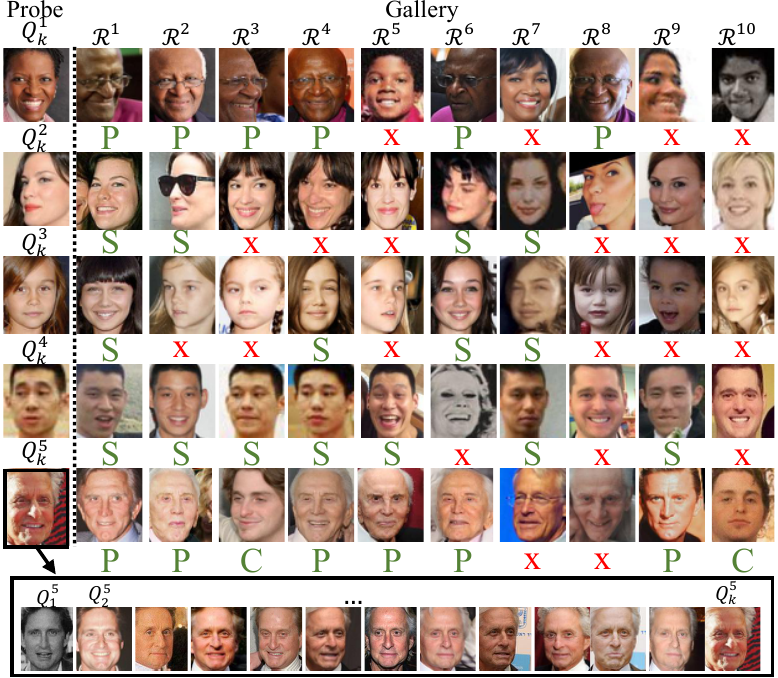}
    \caption{\textbf{T3 sample results (Rank 10).} Each query (row) has one or more faces, for the probe returns and ranks all samples in the gallery - here we show top 10. \gls{fp} are labeled by \textcolor{red}{x}, while true matches list the relationship type in green: \textcolor{ao(english)}{P} for parent; \textcolor{ao(english)}{C} for child; \textcolor{ao(english)}{S} for sibling.}
    \label{fig:track3:montage}
\end{figure}

\begin{table}[t!]
	\centering

	\caption{\textbf{T3 results.} Performance ratings for \gls{sota} methods.}

	\begin{tabular}{r|cc} 
	      \textbf{Methods}  &\textbf{mAP} & \textbf{Rank@5} 	\tabularnewline \hline
		  Baseline (Sphereface)~\cite{wang2018additive} & 0.02 & 0.10	\tabularnewline
		  DeepBlueAI~\cite{id3} & 0.06 & 0.32	\tabularnewline
		  HCMUS notweeb~\cite{id9} & 0.07 & 0.28		\tabularnewline
		  ustc-nelslip~\cite{id8} & 0.08 & 0.38		\tabularnewline
		  vuvko~\cite{id4} & \textbf{0.18} & \textbf{0.60}		\tabularnewline
	\end{tabular}
	\label{tbl:t3:benchmarks}
\end{table}

In~\cite{id8}, Track I and III completed in succession, such that a wider sweep of \gls{cnn} backbones, loss functions, and fusion methods were assessed in Track 1, to both gain deeper understanding to make decisions pertaining to Track III. Mainly, ResNet50 and SENet50 were evaluated separately, each with additional fully-connected layers with two losses on top, \gls{bce} and Focal loss. \gls{bce}, a widely used loss that does as its name implies: uses the measure of entropy of a distribution, say $q(y)$ for $c\in{1, \dots, C}$ classes as $\mathrm{H}(q) = \sum_{c=1}^Cq(y_c)*\log(q(y_c))$. Since we have no knowledge of the true distribution, we aim to match samples of the \emph{true} distribution $p(y)$. Hence, cross-entropy is entropy between $p(y)$ and $q(y)$.

Yu~\etal found that \gls{bce} loss outperformed Focal Loss for all fusion schemes and settings in Track I~\cite{id8}. Intuitively, this makes sense as Track I, a Boolean task, has an equal number of positive and negative pairs-- imbalanced data motivated Focal Loss, which is not an issue for verification. Then, transferring over the model, loss, and fusion settings that worked best for Track I to Track III and used as is. The difference is in the ranking scheme (\ie provided multiple faces per query, the average of all faces and each gallery sample determined the score at the subject-level.

\vspace{1mm}
\noindent\textbf{Deep representation.} 
Training a set of \glspl{cnn}, each targeting specific regions (or parts) of the face, was proposed as a solution for \gls{kfw}~\cite{zhang122015kinship}.
Then, 
\gls{hsl} tackled various tasks of kinship recognition via multi-view learning, with the different views set as different relationship types dubbed~\gls{msml}.~\cite{qin2018heterogeneous}. Similarity, \gls{svdd} was proposed as a \gls{sml} loss function, allowing detailed information to be extracted as geometric and appearance-based features for kinship verification~\cite{qin2019social}. 
Duan~\etal proposed a coarse-to-fine scheme for which \glspl{cnn} at different levels (\ie layers) were transferred from being trained using a \gls{fr} dataset and then fine-tuned for kinship using a loss function based on \gls{nrml}~\cite{duan2017face}. In fact, many recent works leveraged existing \gls{fr} methodologies (\eg \gls{cnn} trained to classify faces) as a prior, then fine-tune using the kin-based image data as the source in a transfer-learning regime~\cite{zhang2019deep}.

\begin{figure}[!t]
    \centering
    \includegraphics[width=.75\linewidth]{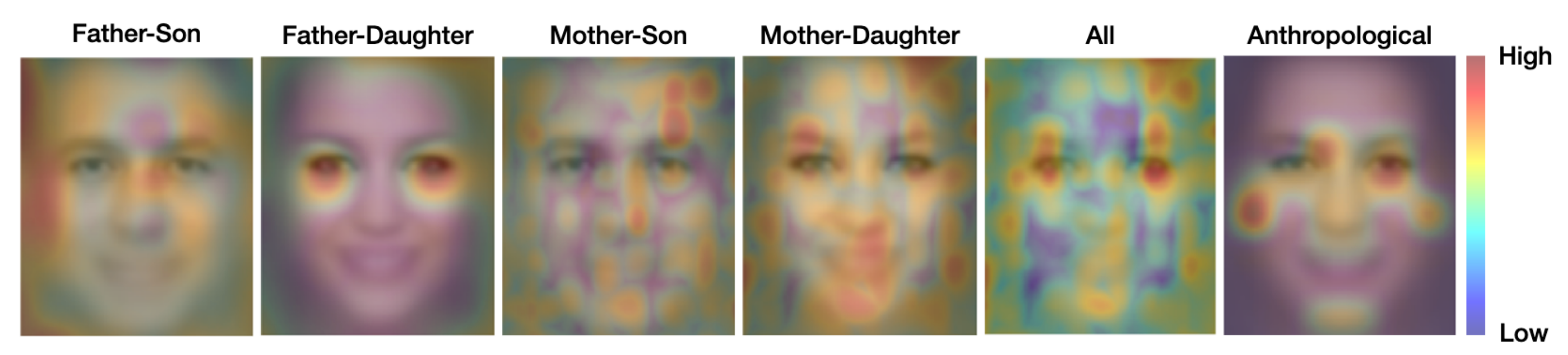}
    \caption{\textbf{Activations from mapping image-to-latent space (from~\cite{dehghan2014look}).} The salience mapped from the activation response and superimposed on the average face. Family101 dataset was used for this experiment~\cite{fang2013kinship}. The end result depicted here were dubbed the \emph{genetic features} from latent space of a trained Gated \gls{ae}.}
    \label{fig:genetic:features}
\end{figure}

Several lines of research specifically focused on the {one-to-one} kinship verification problem by learning a face encoder robust in detecting kinship relationship via \gls{ae} (\eg \gls{dae}~\cite{wang2018cross, chergui2020kinship, chukinship}), \ie deep representation learning methods~\cite{kohli2018deep}.  Dehghan~\etal was amongst the first, proposing to train a Gated \gls{ae} to encode faces as \emph{genetic features}, and weighting according to the salience for the respective relationship type~\cite{dehghan2014look}. Fig.~\ref{fig:genetic:features} depicts the salience, with high being most similar regions and low dissimilar.
Besides still-faces, deep learning approaches were also proposed for recognizing kinship pairs using facial cues in video data~\cite{sun2018video}. A sequence recurrent \gls{nn} was trained for kinship verification in videos using a novel attention mechanism~\cite{lv2019attentive}. With videos, there comes more bits of information; however, the range of bits (\ie the underlying variation of the data) should be optimized to maximize the information gain. In other words, video data introduces another space for fusion in the choosing of the best frame(s) to describe and represent~\cite{gong2019video}. Graphical neural network (GNN) with a metric learned on top proved to be one of the most effective deep learning models employed for kin-based vision problems~\cite{liang2018weighted}. Not just on the large-scale \gls{fiw} data, but a graph-based kinship reasoning (GKR) network proved effective on \gls{kfw}~\cite{li2020graph} (Table~\ref{chapter:fiw:tab:kinwild_eval}).

\noindent\textbf{Approaches to data challenges.}
\gls{rfiw} serves as a platform for experts to publish and junior scholars to get started. The first edition of \gls{rfiw} was in 2017~\cite{robinson2017recognizing} - a data challenge workshop in conjunction with the \emph{ACM Conference on Multimedia}. Ever since, \gls{rfiw} has been held annually (\ie 2018-2020 held in conjunction with \gls{fg} as a data challenge), with each year building on the prior. Let us review series highlights over the years, and then focus on the top teams of the 2020 edition.

\begin{figure}[t!]
    \centering
    \includegraphics[width=.55\linewidth]{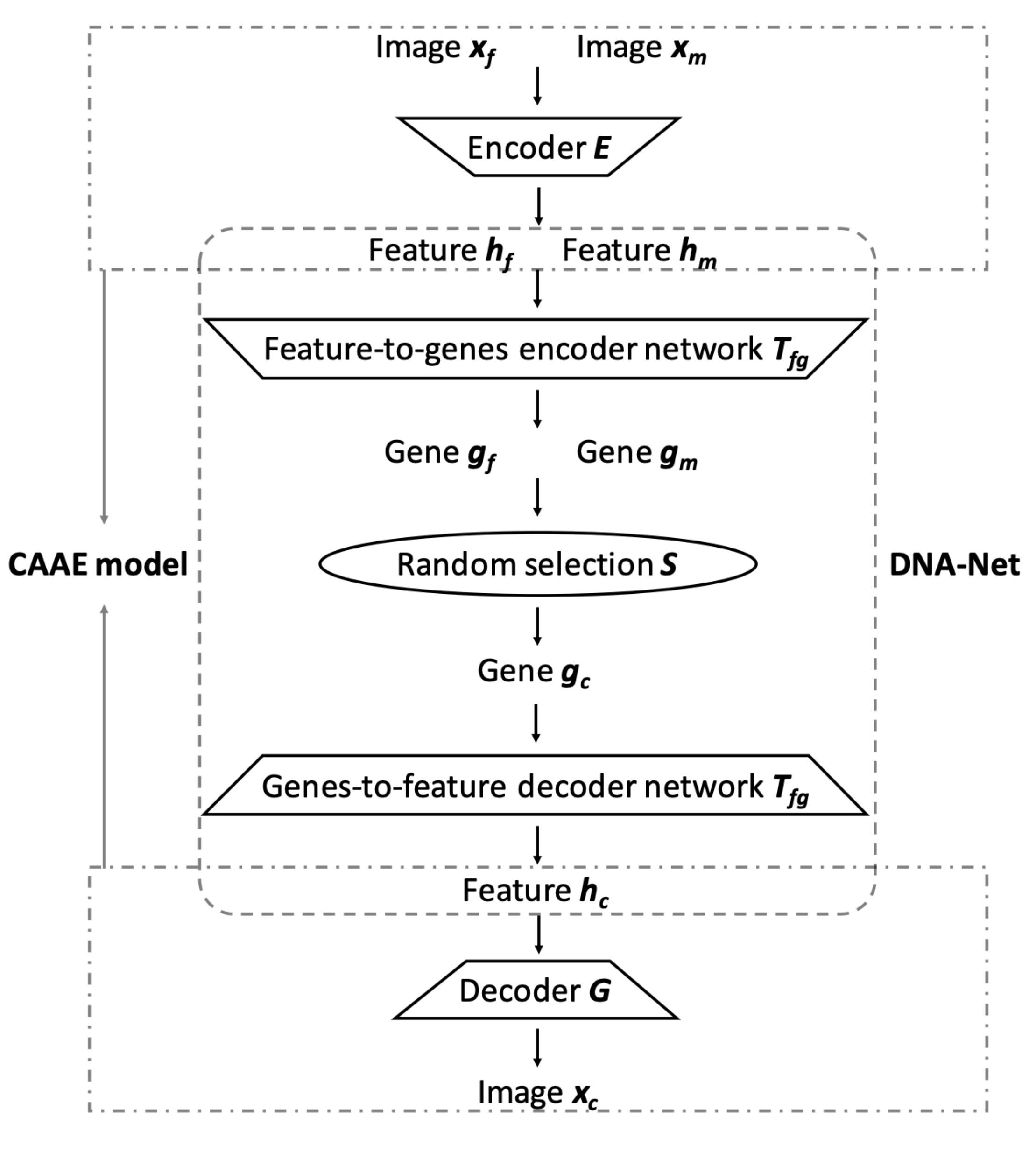}
    \caption{\textbf{Model to synthesize children faces from a parent-pair (visualizations from~\cite{gao2019will}).} Notice that the output of encoder $E$ is the concatenation of features from prospective parents, the father $h_f$ and mother $h_m$ joined by $\oplus$ such that the two embeddings encoded by the Siamese network are fused (\ie $2*\mathbb{R}^d\rightarrow\mathbb{R}^{2d}$) before passed as input to the \gls{caae} model.}\label{fig:dna:net}
\end{figure}
\begin{figure}[h!t]
    \centering
    \begin{subfigure}[t]{0.65\textwidth}
        \centering
        \includegraphics[width=.85\textwidth]{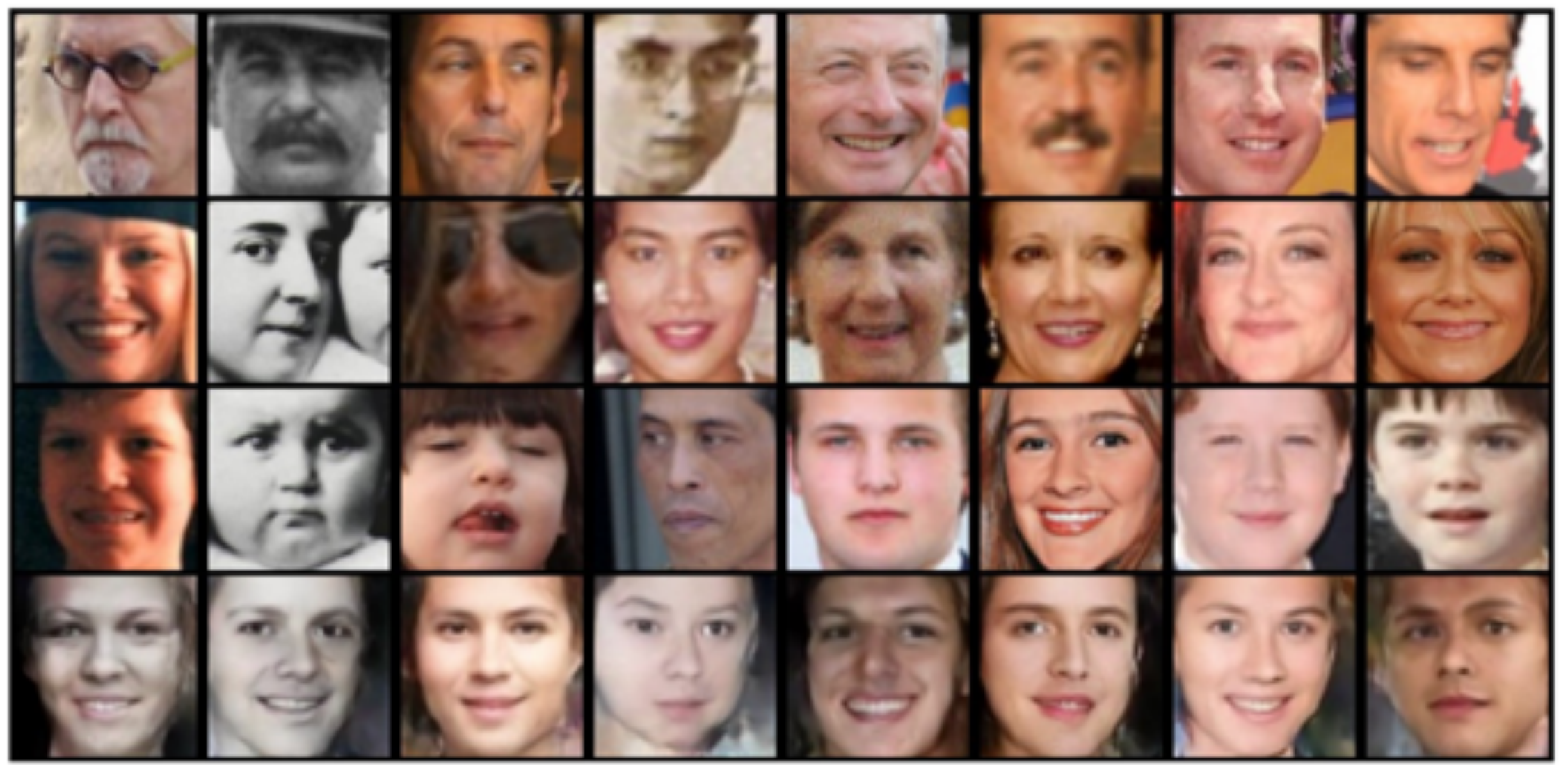}
        \caption{Random synthesis.}
        \label{fig:montage:random}
    \end{subfigure}%
    \\
    \begin{subfigure}[t]{0.65\textwidth}
        \centering
        \includegraphics[width=.85\textwidth]{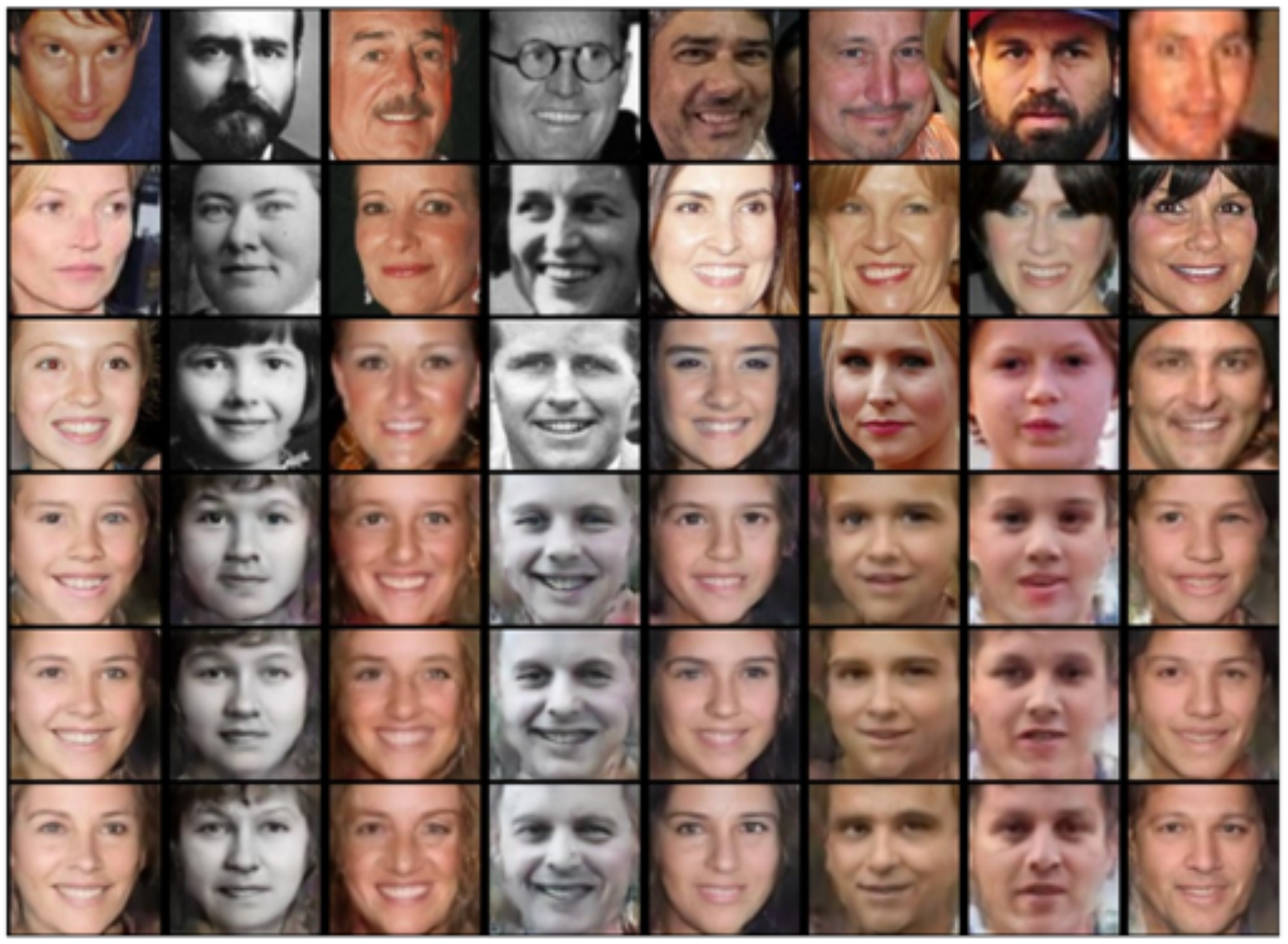}
        \caption{Age (\ie 10, 20, 30 years old from row 4-6, respectfully).}
        \label{fig:montage:age}
    \end{subfigure}%
    \\
    \begin{subfigure}[t]{0.65\textwidth}
        \centering
        \includegraphics[width=.85\textwidth]{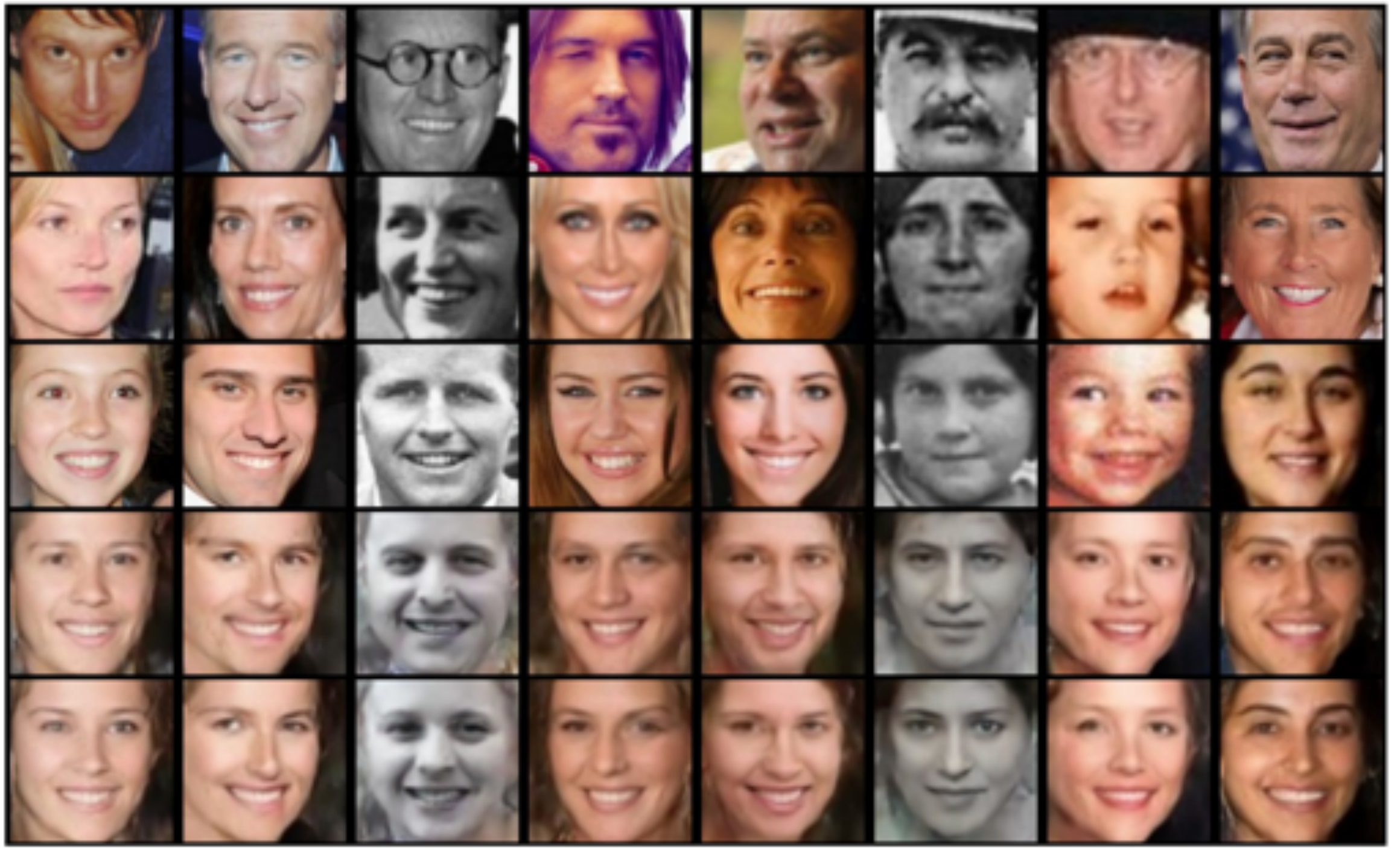}
        \caption{Gender (\ie male-to-female from row 4-5, respectfully).}
         \label{fig:montage:gender}
    \end{subfigure}
    \caption{\textbf{Synthesized results from~\cite{gao2019will}.} Columns correspond to families, with fathers on first row, mothers on second, and real and generated children on third row and bottom, respectively (a). See subcaption for specifics on age (b) and gender (c).}
    \label{fig:dna:net:sample:results}
\end{figure}

From the start, solutions for \gls{rfiw} typically involved \glspl{cnn} pre-trained for \gls{fr}. For the top performing submission of the 2017 \gls{rfiw}, Yong~\etal used an ensemble of deep \glspl{cnn} with data augmentation and mining techniques called KinNet~\cite{li2017kinnet}. Specifically, the authors proposed to train four ResNet models (\ie 80, 101, 152, and 269 layers) for \gls{fr} to then fine-tune for kinship verification via a triplet loss targeting intra-family relationships. KinNet used two tricks during training: (1) augmentation using imaging processing techniques (\eg gamma correction, down/up sampling of pixels, blurring) and hard-negative mining for selecting triplets. In the end, KinNet scored an impressive average of 74.9\%. It is important to note that the data has changed since this first edition of RFIW (\eg \emph{grandparent}-\emph{grandchild} types were not included). Thus, a comparison with the proceeding years would be unfair. 

In 2018, Dahan~\etal got the top performance 68.2\%~\cite{dahan2018selfkin}. Specifically, the authors trained a VGG-Face model with the novel \emph{local features conv-layer} that fused the Siamese inputs by summing the features. In other words, conventional conv-layers share weights in image space, whereas these authors proposed learning local weights to produce pair and location specific features.

In 2019, Laiadi~\etal extended XQDA-to-TXQDA to operate on multilinear data in a low dimensional and discriminative tensor subspace. TXQDA uses multilinear projections of tensors to a space with greater separation between data classes, is enhanced in a way and helps lightening the small sample size problem (\ie results for both \gls{kfw} and \gls{fiw})~\cite{laiadi2019kinship}. Nandy~\etal followed a Siamese learning approach~\cite{nandy2019kinship}, which we will next learn was the most common in 2020.

\begin{figure}[t!]
\centering
\begin{subfigure}[t]{0.25\linewidth}
\centering
\includegraphics[width=\linewidth]{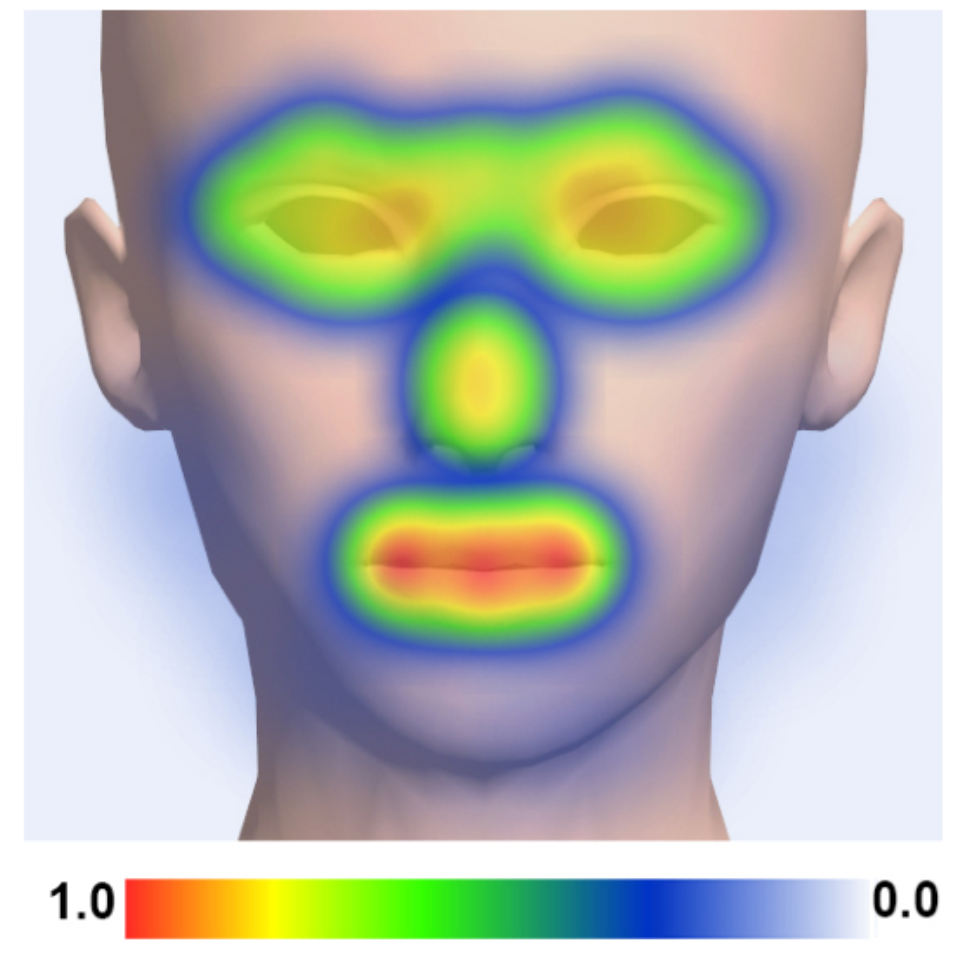}
\caption{Generated child}
\end{subfigure}
\quad
\begin{subfigure}[t]{0.25\linewidth}
\centering
\includegraphics[width=\linewidth]{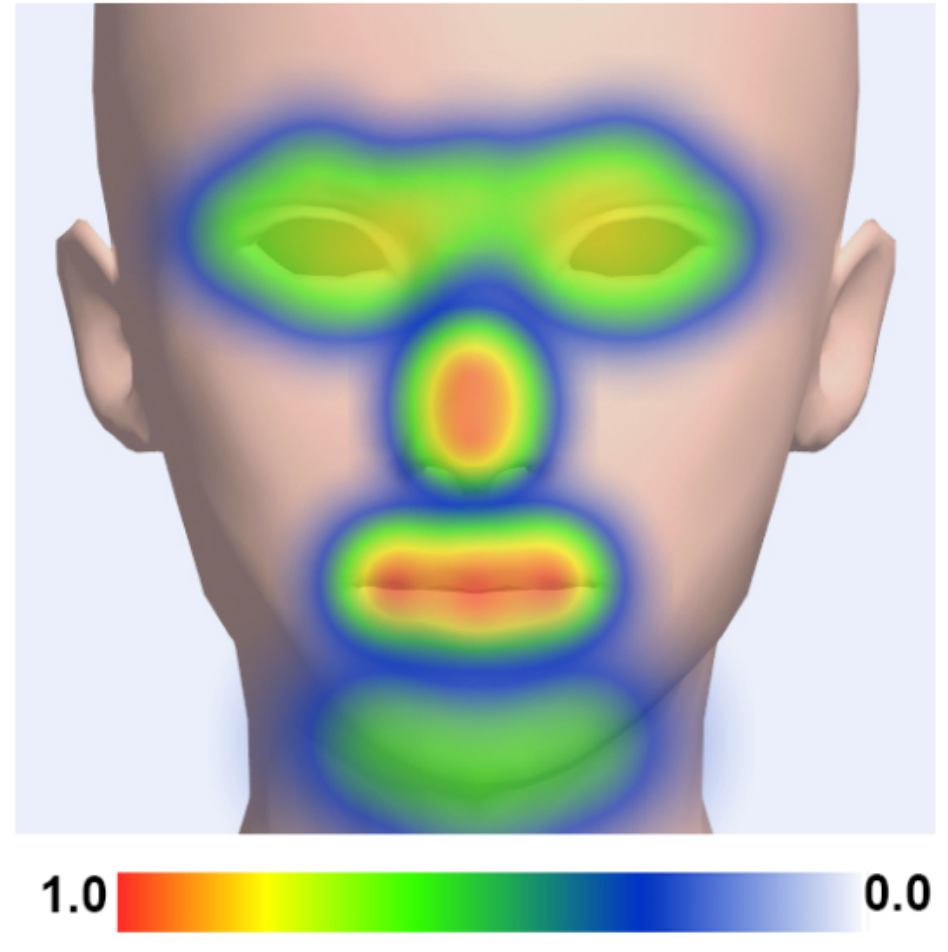}
\caption{Real child}
\end{subfigure}
\caption{\textbf{Salience map per key-points~\cite{gao2019will}.} Best viewed in color.}
\label{chap:fiwmm:fig:heritablity}
\end{figure}

The 2020 edition saw a great increase in interest and participation. Several methods were used as solutions for two or more tasks. 
Shadrikov~\etal treated the different pair types as multiple tasks, training a local expert for each on top of a ResNet50, simultaneously. The authors used T1 data as validation, but then deployed on T2 and T3 as well. Another multi-task approach applied different fusion techniques in deep feature space~\cite{id6, id8}. 
Zhipeng~\etal used two pre-trained \glspl{cnn}, fused the two face encodings by different types of arithmetic, and generated solutions for all three tasks~\cite{id6, id8}. In \cite{id3}, the distance between faces was then determined euclidean distance (instead of the typical cosine similarity); also, SENet~\cite{iandola2016squeezenet} was used as the backbone showing a modest boost over ResNet50 on the validation, but dropping on the test. Like in~\cite{robinson2018visual}, except the authors now used Arcface instead of Sphereface, \"{H}\"{o}orman~\etal fine-tuned a \gls{cnn} using families as the classes\cite{id3} - ultimately placing second in verification (\ie T1) and tri-subject (\ie T2). Yu~\etal put emphasis on the dependence of family identification accuracy for cross-gender versus same-gender pairs types~\cite{id2}. These researchers constructed a Kinship \emph{comparator} module that consisted of eleven ``local expert networks'' connected in series-- eleven networks corresponding to the eleven relationship types of T1. In the end, Stefhoer registered the highest score in the subcategories of father-daughter and mother-son in T1. Yu~\etal also used a Siamese network, \ie encoded features from face images via a \gls{cnn} with shared weights~\cite{id6}. ResNet50 or SENet50 was used as the backbone, both pre-trained on VGGFace2~\cite{cao2018vggface2}. Team ustc-nelslip also employed two loss functions, binary cross-entropy and focal loss, and fused the features using two algebraic formulae leading to \( 2 \times 2 \times 2 = 8 \) independent ``models.'' A unique feature was the construction of a ``jury system'' to combine outputs of different models to improve accuracy. With \cite{id4} the top-scorer in T2. Nguyen~\etal competed in T1 and T3\cite{id9}-- the authors use a Siamese CNN with FaceNet (Inception-ResNet-v1 trained with triplet loss) and VGG-Face (Resnet-50) pre-trained. The authors also implement ArcFace~\cite{wang2018additive} - a family of loss functions based on the geodesic distance between feature vectors which aim to discriminate the latent representation of deep \glspl{nn}. Samples that were unanimously classified correct or most incorrect are in Fig.~\ref{fig:track1:samples:submitted},~\ref{fig:track2:samples:submitted}, and~\ref{fig:track3:montage}, along with average performance ratings in Table~\ref{tab:benchmark:track1},~\ref{tab:benchmark:track2}, and~\ref{tbl:t3:benchmarks} for T1, T2, and T3, respectfully.

\subsubsection{In summary}
Of all the proposed methods, there is a common factor: the larger the age gap the higher percentage of \gls{fp} during evaluation. As mentioned, this was addressed early on with UB Face~\cite{shao2011genealogical} and, although fundamental to the analysis of results over the years, proposed models tended to acknowledge this as a challenge, but with no added mechanism to make robust to age-variations between \emph{parent}-\emph{children}. That is, until Wang~\etal proposed using \gls{gan} technology to synthesize younger versions of an input face. Specifically, and a clearly effective data augmentation approach, the authors trained generators for both genders to account for this while training a deep \gls{cnn} with a maximum margin loss to do boolean classification (\ie \emph{KIN} / \emph{NON-KIN}). As formalized in their work, domain $A$, for \emph{aged}, was the source and domain $Y$m, for \emph{young}, was the target. Provided paired data, the \emph{parent} aimed to transform $x_i\in A\rightarrow x_j\in Y$ with data distribution $x\sim p_A\rightarrow x\sim p_Y$. Having noticed that \gls{fiw}, which most closely matches real-world data, does not necessarily have parents at older ages (\ie \emph{aged}), thus, the inputs could very well be parents as juveniles, or even during infancy. To mitigate the problem, with no age-labels provided in \gls{fiw}, focus was directed to constrain the output such to influence younger aged faces less so, than if faced with an elderly parent.

\subsection{Generative modeling approaches}\label{sec:synthesis}
The dynamics of the offspring synthesis problem has a great distinction from tradition one-to-one mapping - two parents with directional relationships are input as prior knowledge to predict the appearance of their child. Such a two-to-one problem raises the question on how to best fuse knowledge from a pair of faces. Let us now even consider information for various family members - the fusion then should consider directed relationships inherent to family trees. Current face synthesizers conditioned on kinship assumes knowledge of one~\cite{ozkan2018kinshipgan} or both~\cite{gao2019will, ertugrul2017will} parents. 

Ozkan~\etal proposed KIN-\gls{gan} to synthesize a child's face from a sample of a single parent~\cite{ozkan2018kinshipgan}. The problem is inherently difficult, for the variation embedded in many complex factors nearly changes from one sample-to-the-next. Nonetheless, trying to solve the problem with just one parent is insensible-- it takes two to tango in nature and, thus, such a formulation is out of scope before the problem is even started. Noticing this, \cite{ertugrul2017will} proposed means of modeling as a two-to-one mapping. Similarly, Gao~\etal aimed to mimic the nature of reproduction with a model dubbed DNA-Net~\cite{gao2019will}. DNA-Net fuses latent representation of a parent pair at the feature-level, which is used as input to \gls{caae} model trained on top (Fig~\ref{fig:dna:net}). The parents' signals are fused at the output of encoder E by concatenation of their features, and are then fed to the \gls{caae} model to produce a single feature representing the face encoding of the child. Finally, the child's encoding is decoded by G to the predicted facial image.

Note that DNA-Net was dubbed by the authors in the effective work proposed; however fair when speaking in general terms (\ie infrequent situation in research), we suddenly see naming schemes such as this, \emph{genetic features}, among few others is too strong. Nonetheless, there is a clear analogy, so for the sake of story-telling and system depiction, Gao~\etal dubbed this as a single face is synthesized from face pair. The choice in \gls{caae} made it so the generator could synthesize children as a function of age and sex (Fig.~\ref{fig:dna:net:sample:results}). Note that treating sex as a continuous spectrum, opposed to discrete labels, is both appropriate and more precise (\ie provided an extreme pair, one female and the other male, there exists many cases in between, which is, in fact, where most of society falls~\cite{merler2019diversity}). As a part of the work to support DNA-Net, the authors compared salience in detecting kinship of type \emph{parent}-\emph{child} at specific facial features (\ie eyes, nose, mouth, and chin). Hu invariant moments were used as the shapes of the four facial parts~\cite{hu1962visual}, from which the accumulative cosine distances yielded \emph{heritability maps} (Fig.~\ref{chap:fiwmm:fig:heritablity}). 

\chapter{FIW-MM}\label{chap:fiwmm}

\section{Overview}\label{chapMM:sec:introduction}
So far, we have covered vast works in automated kinship recognition that assumes a genetic relatedness between individuals detectable by facial cues - a state in technology unimaginable just over a decade ago. Much of the progress in the difficult tasks of kin-based recognition was by the availability of labeled family data with sufficient counts and concurrent advances in face recognition~\cite{liu2017sphereface, masi2018deep}-- proposed systems inherently gain if it based on a \gls{fr} model that experienced a gain itself via progress in conventional \gls{fr}. In other words, \gls{fr} (\ie determine if face pair are of the same identity) and visual kinship recognition (\ie determine whether a pair of faces are of the same bloodline) both target facial cues to determine whether or not a face is a match to a gallery (\ie test sample(s)). Conventional \gls{fr} is the more general, simpler by definition and protocols, and with a higher relevance to vast use-cases of the two. \gls{fr} also has a data need that is more readily accessible. So, it is absolutely to no surprise that \gls{fr} tends to be ahead of kinship recognition technology, which results in there being lots of research findings in \gls{fr} that can be transferred or referenced when devising hypotheses in kinship problems.

The seminal work in visual kinship recognition introduced the first image dataset~\cite{fang2010towards}. Thereafter, larger and more difficult datasets were released, such as \gls{fiw}~\cite{robinson2018recognize} and \gls{tsk}~\cite{qin2015tri}. In response, vision researchers developed methods and models to match the rising level of difficulty in kinship datasets~\cite{robinson2020recognizing, robinsonKinsurvey2020}.

Along with conventional \gls{fr} and the different sub-problems that model visual knowledge from facial cues, speaker-based problems have recently grown popular based on audiovisual data (\eg speaker separation~\cite{ephrat2018looking}, speaker identification~\cite{nagrani2017voxceleb, chung2018voxceleb2}, cross-modal audio-to-visual or vice-versa~\cite{Nagrani18a}, emotion recognition in \gls{mm}~\cite{Albanie18a, hao2020visual}, and several others ~\cite{Wiles18, Wiles18a}). The sudden surge of attention to audio-visual data has brought together experts who specialize in biometric signals to share thoughts, combine knowledge, and propose solutions that best fuse multi-domain knowledge for optimal decision making~\cite{song2019review, petridis2017end}. This work shows that the addition of \gls{mm} for kin-based recognition can improve the current \gls{sota}. 

Studies in speech recognition found that identical twins were hard-negatives and confused in classification tasks consistently~\cite{zuo2012formant}. Although results were strongly in favor of the hypothesis that twins sound alike, the experiments were done on a small sample set of 8 pairs. Interestingly,~\cite{zuo2012formant} showed similar results for 2 of the 8 pairs the authors dubbed `separated twins' (\ie twins that were brought up in different households per full-time custody bargains that divided the twins, but had claimed to have spoken regularly on the phone and often spent weekends together throughout their entire childhood). Given the above, a fair hypothesis would be that subjects living together develop common speaking habits (\eg phrases, frequently used words and jargon, and even accents). Thus, \emph{nurture} plays its role, with hints that \emph{nature} does as well. A recent study used computer vision and speech recognition technologies to verify kinship~\cite{wu2019audio}. The authors showed significant gain in kinship verification performance when \gls{mm} (\ie video-audio) is utilized using modern-day deep learning techniques that leverage both modalities, over just the still face images. In the end, Wu~\etal used 400 video-pairs of \emph{parent}-\emph{child} pairs to show promise in the use of multi-modal systems for kinship recognition - which, to the best of our knowledge, has been the only attempt of using visual-audio data for recognizing kinship.\footnote{The dataset collected for~\cite{wu2019audio} is not available for public use.}

\begin{figure}
\centering
    \includegraphics[width=.9\linewidth]{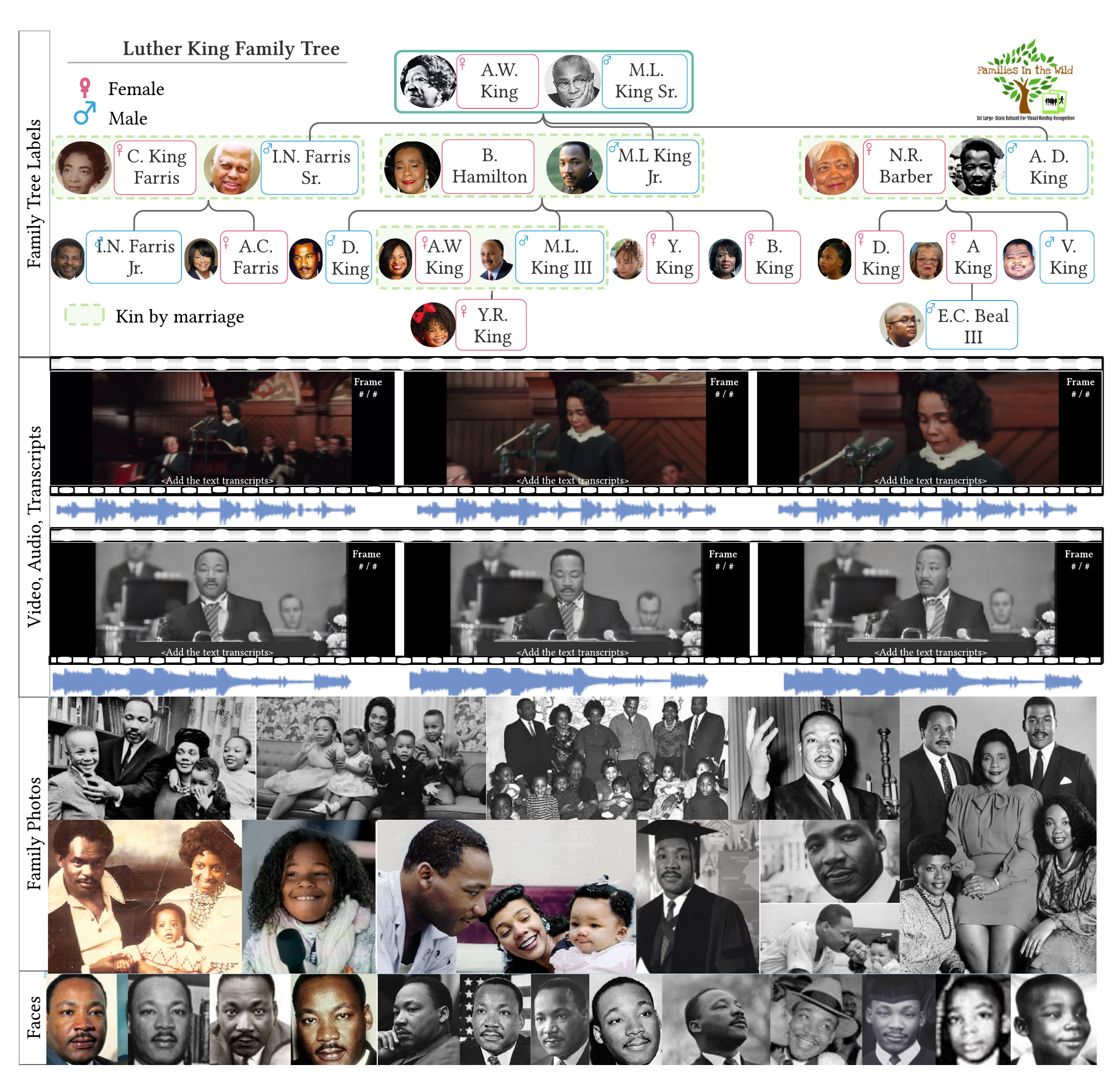}
  \caption{\textbf{Sample family of \gls{fiwmm}.} Top-to-bottom: \emph{family-tree labels} show faces of members in the immediate family, with subjects of the same generations in the same row; \emph{videos, audio, and contextual} exemplify sample video pairs of Dr. King Jr. and his daughter Andrea with tracklets of faces in the visual domain and audio data aligned frame-by-frame; \emph{family photos} that contain Dr. Luther King Jr. randomly selected (note, cropped to fit); \emph{faces} of Dr. King Jr. from adolescence-to-adulthood. Multiple faces are available for most subjects. Best viewed electronically.}
  \label{chapMM:fig:teaser}
\end{figure}

Our contributions to the \gls{fr}, biometric, anthropology, and \gls{mm} communities were 3-fold.
\begin{itemize}
\item {\verb|Built multimedia database|}: a large-scale dataset for kinship recognition was built using existing paired image data and an automatic labeling scheme. Media of different modalities is now available: video-tracks, audio segments, visual-audio clips, and text transcripts. Specifically, we extended the \gls{fiw} imageset. Additionally, we restructured the \gls{mm} family database to better encapsulate the added metadata and paired data respectively at the subject and instance levels.

\item {\verb|Recorded protocols and benchmarks|}: a new paradigm for kinship recognition suited for \gls{mm} data as a step towards experimenting with real-world settings. Specifically, the problem has evolved from instance to template-based. Thus, we are the first to measure kinship recognition capabilities using a large-scale, multimodal template-based collection. 

\item {\verb|Showed the advantage of all modalities|}: Following the improved protocols and, thus, experimental settings, we demonstrate an increase in system performance from still-images, to still-images and videos, and then with audio speech signals added as well-- a clear benefit of each added modality is shown. Our analysis highlights the shortcomings of the different media types for future work to address.

\end{itemize}

We believe this will attract a wider range of scholars to kin-based and multimedia problems. \gls{fiwmm} will be accessible online in various formats.\footnote{\gls{fiwmm} - the data, code, trained models, and other resources - will be available upon publication of this work. }

\section{Related Work}\label{sec:related:work}
Early on, problems of recognizing kinship started with domesticated animals (\eg dogs~\cite{hepper1994long} and sheep~\cite{POINDRON200799,poindron2007maternal}), as many species have a natural ability to recognize their kin through various signals (\eg touch, smell, visual, and acoustics). From this, we hypothesized that different types of media, besides image-level or conventional speech recognition, can be leveraged to better detect kinship in humans. Knowledge extracted from still-images and stationary speech signals are lacking an abundance of evidence. A more complex signal which helps improve decision making, such as dynamic features across video frames, can attribute inheritable characteristics (e.g., expressions, mannerisms, and accents from different emotions). Nonetheless, such technology will take effort to acquire. We demonstrate the ability of the added modalities with face tracks from videos and standard audio features from speech signals.

We next review existing work in visual kinship recognition on still-images, and then more recent advances in the acquiring and modeling of visual-audio data for \gls{fr}.

\subsection{Kinship recognition}
Computer vision researchers began using facial cues to recognize kinship about a decade ago. Specifically, Feng~\etal proposed to model the geometry, color, and low-level visual descriptors extracted from faces to discriminate between KIN and NON-KIN~\cite{fang2010towards}. Others then formulated the problem as various paradigms (\eg transfer subspace learning~\cite{Xia201144, xia2012understanding}, 3D face modeling~\cite{vijayan2011twins}, low-level feature descriptions~\cite{zhou2011kinship}, sparse encoding~\cite{fang2013kinship}, metric learning \cite{lu2014neighborhood}, tri-subject verification~\cite{qin2015tri}, adversarial learning~\cite{zhang2020advkin}, ensemble learning \cite{wang2020kinship}, video understanding~\cite{sun2018video, zhang2014talking, georgopoulos2020investigating}, and, most recently, video-audio understanding~\cite{wu2019audio}). A common factor of the aforementioned was the attempt to improve discriminatory power for classifying a pair of faces as either KIN or NON-KIN; another commonality was the limited sample size and, thus, unrealistic experimental settings.  

Robinson~\etal introduced a large-scale image dataset to recognize families in still-imagery called \gls{fiw}~\cite{FIW,robinson2018visual}. \gls{fiw} contains 1,000 families with an average of 13 family photos, 5 family members, and 26 faces. It came with benchmarks for 11 pairwise types, with the top performance of the baselines being a fine-tuned CNNs (\ie SphereFace~\cite{liu2017sphereface} and Center-loss~\cite{wen2016discriminative}). This was the beginning of big data in kin-based vision tasks-- deep learning could then be used to overcome observed failure cases~\cite{wang2018cross, wu2018kinship}. Furthermore, new applications such as child appearance prediction~\cite{gao2019will, ghatas2020gankin} and familial privacy protection~\cite{mingaaai2020} were done recently.

Nowadays, \gls{fiw} continues to challenge researchers with various views of image-based tasks. A myriad of methods demonstrated the ability of machinery to use still-images to determine kinship in a pair or group of subjects. Nonetheless, only so much information can be extracted from still-images. The dynamics of faces in video data (\eg mannerisms expressed across frames) contain additional information, and audio as well as text transcripts (\ie contextual data describing the speech and other sounds) can widen the range of cues we model to discriminate between relatives and non-relatives. We propose the first large-scale multimedia dataset for kinship recognition. Specifically, we leveraged the familial data of the \gls{fiw} image database to build upon the existing resource~\cite{FIW,robinson2018visual}, using the still-images of \gls{fiw} and adding video, audio, audiovisual, and text data of subjects. Note that video, audio, and visual-audio differ in that the latter has the face speaking and the speech spoken are aligned, while the others are independent, unaligned clips. After its predecessor, we dubbed the database \gls{fiwmm} (\figref{chapMM:fig:teaser}). En route to bridging research-and-reality, we follow the protocols of \gls{fiw}~\cite{robinson2020recognizing}, but now with the capacity to be template-based (\ie per \gls{nist} in~\cite{maze2018iarpa}).

Besides the different use-cases, and independent research work it made possible, \gls{fiw} was used as part of an annual data challenge motivated to attract more attention from and provide more incentive for the research community; namely, the \gls{rfiw} series, which has been held each year since 2017~\cite{robinson2017recognizing} to 2020~\cite{robinson2020face}. There have been many great attempts on the still-images as a result~\cite{KinNet, AdvNet}. Recent surveys~\cite{qin2019literature}, tutorials~\cite{robinson2018recognize}, and challenge papers~\cite{robinson2020recognizing, lu2015fg,wu2016kinship, lu2014kinship} elaborate on \gls{rfiw} and the various submissions in detail.

\begin{figure}[t!]
    \centering
    \includegraphics[width=\linewidth]{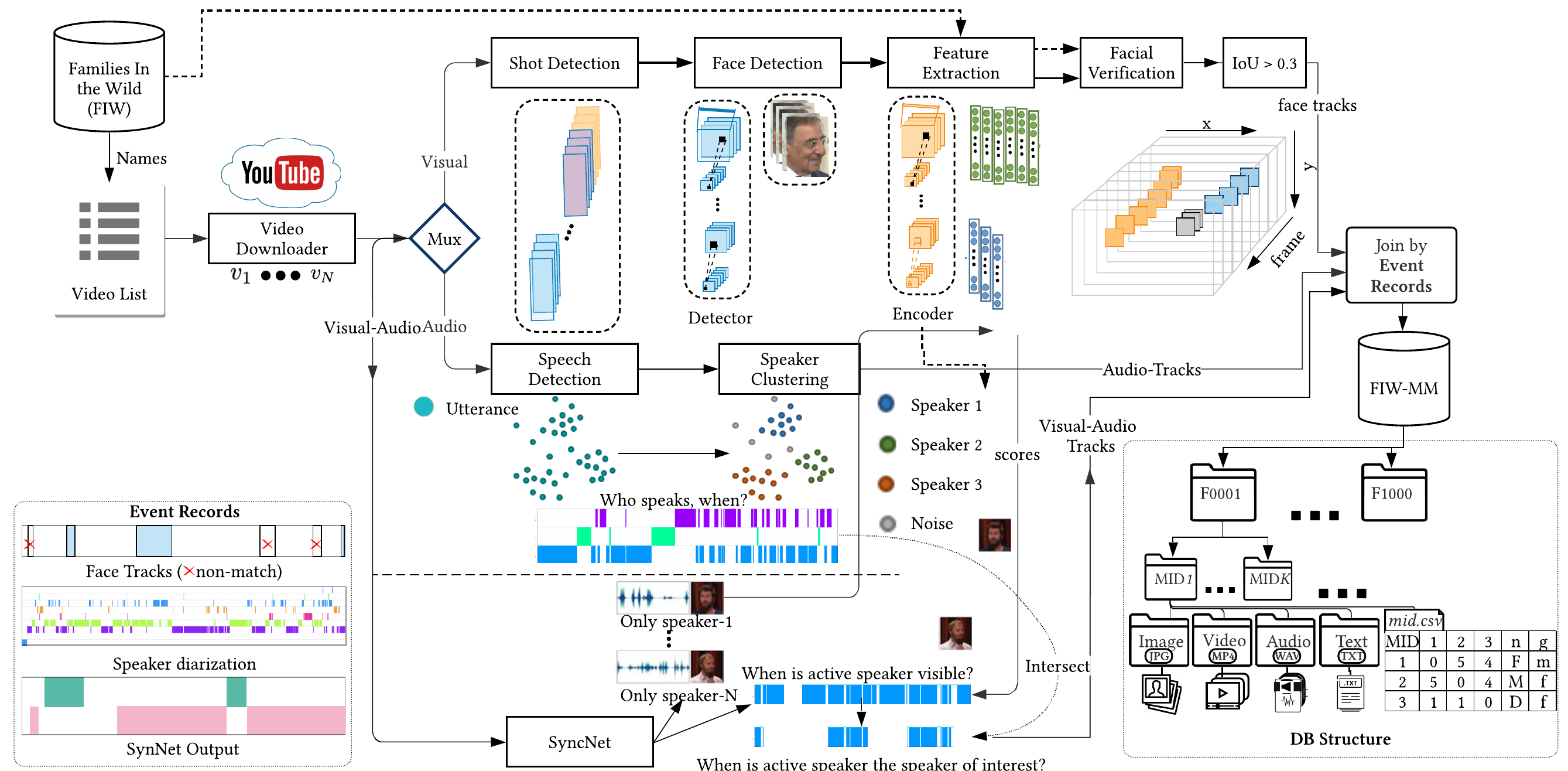}
    \caption{\textbf{Automated labeling framework.} For each of the 1,000 families, there are a set of $K$ members. For this, the template of a member consists of all media available. Tag numbers 1-6 correspond to sections in Section~\ref{subsec:datapipeline}}\label{chapMM:fig:workflow}
    \vspace{5mm}
\end{figure}

\subsection{Audio-visual data}
The archetypal big data resources for audio-visual identification problems are Voxceleb~\cite{nagrani2017voxceleb} and Voxceleb2~\cite{chung2018voxceleb2}. Similar to \gls{fiwmm}, the datasets were acquired by extending still-image collections (\ie Voxceleb and Voxceleb2 extended of the VGGFace~\cite{Parkhi15} and VGGFace2~\cite{Cao18}, respectively). Currently, the primary usage of Voxceleb is in speaker-based tasks, such as using the audio-visual data to detect and classify the speaker by the \emph{who} and the \emph{when}~\cite{ephrat2018looking}. Additional speaker centric problems have been proposed using the Voxceleb collections, like to enhance speech signals~\cite{afouras2018conversation}, to detect \emph{when} and \emph{where} the speaking face is visible~\cite{Chung17a}, and when the audio and mouth motions infer the lips and sound are in sync~\cite{Afouras18b}. Nonetheless, the lip-reading task predates the larger Voxceleb with older lip-reading datasets~\cite{Chung16, Chung17}. 

It is worth highlighting that these audio-visual databases were instrumental in applied research as well (\eg generating talking faces~\cite{Chung17b}, where the input is a still-image face and a stream of audio, and the output are frames mocking the audio with the faces as if the input face was regurgitating the audio clip). In~\cite{wiles2018x2face}, face frames were generated from a still-image and audio clip, with pose information added as a control signal for the synthesized output. Furthermore, Voxceleb predicted emotion labels via its own signals to automatically infer ground-truth~\cite{Albanie18}.

Previous attempts to tackle kinship recognition have also been made with audio-visual data. Most relevant was in~\cite{wu2019audio}, where the authors built a collection made-up of 400 pairs. Wu~\etal certainly demonstrated the core hypothesis of this work-- multimedia can enhance our ability to automatically detect kinship in humans-- as was clearly demonstrated in their work~\cite{wu2019audio}. However, the sample size was limited in the number of pairs and in the types of labels, as there is no family tree structure, nor multiple samples per member (\ie age-varying), as is the case in our much larger and comprehensive \gls{fiwmm}. 

\section{The FIW-MM Database}
The automated labeling pipeline in \figref{chapMM:fig:workflow} leveraged existing name labels available in \gls{fiw}, and with each of these subjects represented by one-to-many face samples. Hence, the visual evidence of \gls{fiw} was modeled and used to annotate the \gls{mm} data by recording \emph{when} and \emph{where} events of interest occurred. In other words, we curated \gls{fiw} by parsing videos to cropped face tracks, entire speaker instances, aligned visual-audio data, and the spoken words transcribed to text. All of these instances were organized as time-stamps representative of the start and end video frames, along with the bounding box information for the face tracks in all frames including and between the initial and final boundary frames. In this way overlap in samples was clearly identifiable. Furthermore, these instances were organized by family (\ie \gls{fid}) and specific member (\ie \gls{mid}) to allow possible outcomes to be limited to boolean (\ie \emph{genuine} or \emph{imposter} match). Family folders contained a folder for each of its \gls{mid}, which held separate folders for face images, non-speaking face tracks, speech samples, transcribed conversations, and face tracks actively speaking (\ie visual-audio).
Of course, relationship matrix and the genders for each \gls{mid} remained in the family folder as was done back for the original \gls{fiw}~\cite{FIW}. The structure of \gls{fiwmm} is shown bottom-right of \figref{chapMM:fig:workflow}. Also, and along with being defined when introduced, various acronyms and symbols used in the following sections are listed under Acronyms at beginning of manuscript.

Next, the specifications, the pipeline used to automatically annotate the data, along with a few strategies used to further reduce the manual input requirements are described.

\subsection{Specifications}
The goal was to extend \gls{fiw} in the number of samples, the media types, and in the possible experimental settings. Recall that \gls{fiw} provides name metadata and face images for an average of $>$13 members of 1,000 families~\cite{robinson2018recognize}. Thus, we aimed to accumulate paired \gls{mm} data for existing persons of \gls{fiw}. Specifically, we acquired paired \gls{mm} data for members of 150 families, with at least two members for each. Otherwise, if we only had samples for a single member for a family, the audio information could not be compared amongst any pair in the respective family. Hence, the requirement was set at least two members per family.

With complete access to \gls{fiw} for research purposes\footnote{\href{https://web.northeastern.edu/smilelab/fiw/download.html}{https://web.northeastern.edu/smilelab/fiw/download.html}}, we leveraged this data as the knowledge needed to build \gls{fiwmm} with minimal manual labor and zero financial cost. For this, we employed \gls{sota} models and algorithms in speech and vision throughout the data pipeline. Our next steps are the modules and feedback loops making up the pipeline om \figref{chapMM:fig:workflow}.

\begin{table}[!t]
    \centering
    \caption{\textbf{Database statistics.} Types are split based on the span in generation of the relationship.}\label{chapMM:tab:data-counts}
     \resizebox{\textwidth}{!}{%
    \begin{tabular}{r c c c c c c c c c c c c c c c c}
        \toprule
        & \multicolumn{3}{c}{\emph{1$^{st}$-generation}} & \multicolumn{4}{c}{\emph{2$^{nd}$-generation}}  & \multicolumn{4}{c}{\emph{3$^{rd}$-generation}}& \multicolumn{4}{c}{\emph{4$^{th}$-generation}}\\
        
        \cmidrule(l){2-4} \cmidrule(l){5-8} \cmidrule(l){9-12}\cmidrule(l){13-16} 
         & BB & SS & SIBS & FD & FS & MD & MS & GFGD& GFGS& GMGD & GGMGS & GGFGGD& GGFGGS& GGMGGD & GGMGGS & Total\\ 
        \midrule

       \# Subjects& 883 & 824 & 1,542 & 1,914 & 1,954 & 1,892 & 2,041 & 426&463 & 483 & 526 & 39&30 &45  &  37&13,099\tabularnewline
       \# Families&  345&  334&  472&  666& 676 & 665 &  670&154 & 174& 178 & 191 & 9&10 &11  &10  & 953 \tabularnewline

        \# still-images&40,386 & 31,315 &46,188  & 83,157 & 89,157 & 57,494 &63,116 &8,007 & 6,775 & 6,373 &6,686 &408 & 410 &798& 797 &441,067  \tabularnewline
        
       \# Clips& 123 & 79 & 81 & 155 & 134 & 147 & 138 &16 & 18& 25 & 15 &2 &4 & 0 & 0 & 937\tabularnewline
       \# Pairs& 641 & 621 & 1,138  & 1,151 & 1,253 & 1,177  & 1,207  &263 &280 & 292 & 324  & 28 &18 & 36  &28  & 8,457\tabularnewline
        \bottomrule
    \end{tabular}}
\end{table}

\subsection{Data pipeline}\label{subsec:datapipeline}
Inspired by previous work, such as \gls{fiw}~\cite{robinson2018visual} (\ie labeling families) and VoxCeleb~\cite{nagrani2017voxceleb} (\ie labeling audio-visual data), aspects of both were merged as the basis of our pipeline design and, in essence, one of three branches that make up our data collection pipeline. Specifically, the merging of the aforementioned pipelines make up the \emph{audio-visual} branch, which processes end-to-end and in parallel with the \emph{visual} and \emph{audio} branches. The notion of branches is used for clarity in the following description, as each respective branch is concerned with the modality for which it is referred.

The following subsections cover the details of the pipeline built to acquire \gls{fiwmm} as the sequence of modules it grew to - the steps are covered in order of process (\ie from left-to-right in \figref{chapMM:fig:workflow}). Philosophically, all data was assumed to be of type \emph{non-match} (\ie zero amount of \gls{mm} data to start). Then, there are various checkpoints throughout the branches that add data which was found to be a \emph{match} with high confidence. Under the pretext that \gls{fiwmm} will be a resource used by experts from different data domains, all data points that \emph{match} are saved (\ie visual tracks, audio, and audio-visual). Nonetheless, overlapping segments are clearly annotated such to remove repeated samples (\ie visual-audio will also be present in sets containing just visual and just audio). At the same time, if one or both modalities are of interest, then the maximum amount of data points is readily available. Note that no data points are repeated in sets created for included benchmarks. Also, the following subsections are numbered according to the yellow circle call-outs in \figref{chapMM:fig:workflow}.

\vspace{2mm}
\noindent\textbf{1) Raw data resource.}
\gls{fiw} has still-image data for 1,000 families with over 13,000 family members (\ie subjects) in total. From the families, we chose a subset of 150 for which 2-5 members appeared in 1-3 YouTube videos, with a total of 500 subjects in 605 videos. The importance of this step was assuring that there were at least 2 members per family with \gls{mm}; otherwise, the added modalities would have no basis to match about. Also, ethnicity for these 500 subjects were manually collected at this time. Video URLs were queried under unique \glspl{vid} (\ie $v_1, \dots, v_N$ for $N$ videos). Generally speaking, the videos were either interview-style (\eg with a news anchor or alone in a plain room answering scripted questions) or face-time clips (\ie self-recordings of subject speaking directly to the camera, as is the normal case when face-timing).

Our scripts used Pypi's youtube-dl\footnote{\href{https://github.com/ytdl-org/youtube-dl/}{https://github.com/ytdl-org/youtube-dl/}} to download YouTube vidoes by URL, which were then archived under corresponding \gls{vid}. Allowed with the \gls{mm} (\ie in MKV file format), time-stamped captions were also scraped when available-- later parsed as transcribed words spoken by the subject. Alongside the text, the MKV files were processed to three files: a copy of the original MKV for the \emph{audio-visual} branch, and then an audio only (WAV) and visual only (MP4) extracted with \emph{ffmpeg}. From the start, all video data was assumed a constant 25 FPS.

\begin{table}[t!h]

    \centering
    \caption{\textbf{Task-specific counts.} For individuals (\textbf{I}), families (\textbf{F}), still-face images (\textbf{S}), video-clips (\textbf{V}), audio snippets (\textbf{A}), audio snippets (\textbf{VA}) in the set of probes (\textbf{P}), gallery (\textbf{G}), and in total (\textbf{T}).}\label{chapMM:tab:counts} 
         \resizebox{\textwidth}{!}{%
\begin{tabular}{m{.5mm}m{.5mm}ccccccccccccccc}
\toprule
        & & \multicolumn{5}{c}{\emph{Train}} & \multicolumn{5}{c}{\emph{Val}}  & \multicolumn{5}{c}{\emph{Test}}\tabularnewline
        
        \cmidrule(l){8-12} \cmidrule(l){13-17}
         & & \textbf{I} & \textbf{F} & \textbf{S} & \textbf{V} & \textbf{A} & \textbf{I} & \textbf{F} & \textbf{S} & \textbf{V} & \textbf{A} & \textbf{I} & \textbf{F} & \textbf{S} & \textbf{V} & \textbf{A}\tabularnewline \midrule
\parbox[t]{1mm}{\multirow{1}{*}{\rotatebox[origin=c]{90}{\emph{T1}}}} &     &    &         &         &         &         &         &         &         &          &          &   &          &          &          &          \tabularnewline[-1em]

  \multirow{1}{*}{}      &      \textbf{T} & 2,976 & 571 & 16,464 & 290 & 7,217 & 955 & 190 & 5,458 & 72 & 3,308 & 972 & 192 & 5,231 & 91 & 1,775
  \tabularnewline

        \multirow{1}{*}{}    &    &    &         &         &         &         &         &         &         &          &          &   &          &          &          &           \tabularnewline[-1em]
        \midrule
         
   \parbox[t]{2mm}{\multirow{3}{*}{\rotatebox[origin=c]{90}{\emph{T3}}}} &      \textbf{P} & 571 & 571 & 3,039 & 47 & 1,843 & 190 & 190 & 1,334 & 16 & 789 & 192 & 192 & 993 & 23 & 876       \tabularnewline
   \multirow{3}{*}{}       &    \textbf{G} & 2,475 & 571 & 13,571 & 244 & 5,581 & 791 & 190 & 4,538 & 56 & 2,519 & 800 & 192 & 4,705 & 69 & 899  \tabularnewline
    \multirow{3}{*}{}      &      \textbf{T} & 3,046 & 571 & 16,610 & 291 & 7,424 & 981 & 190 & 5,872 & 72 & 3,308 & 992 & 192 & 5,698 & 92 & 1,775 \tabularnewline
\bottomrule
\end{tabular}}
\end{table}

\vspace{2mm}
\noindent\textbf{2) Event records.} 
Before passing data down any branch, blank (sequential) tabular records were created for the duration of the video with tuples as index (\ie time and frame number)-- one record per branch (\ie audio, visual, and audio-visual event records). These are essential for refinement processes that are later activated via a feedback mechanism. In essence, the mutual information across records at a given instance (\ie frame or time-stamp) are used to imply matches, contradiction, and non-matches across modalities (\ie a means to propagate labels across modalities). This usage of set theory helps both to validate true matches and filter out non-matches: others have too leveraged logic and sets to parse videos~\cite{haq2019movie}; however, opposed to high-level semantics such as types of objects present, we reference output of simpler tasks (\eg face or no face, speech or non-speech, same or different face or voice)-- this increases random chance and thus reduces low confident decisions.

\vspace{2mm}
\noindent\textbf{3) Visual branch.} 
We first split a video into scenes using two global measures under the assumption that, statistically, neighboring frames will match as close as 90\%: HSV (\ie color) and local binary patterns~\cite{ahonen2006face} (\ie texture) features were extracted and used to parameterize two probabilistic representations per frame, which were compared using KL-Divergence and compared to a threshold of 0.1~\cite{sanchez2017multimedia}. This produces a set of shots for each of the $V$ videos of size $C$, \ie $v_c\in\{1, \dots, C\}$ represents all shots detected in the $i$-th video. From there the first, last and the frame in between closest to the centroid (in color and texture) of the entire track (\ie the beginning, end, and the assumed best representation for the respective clip).  The three frames per clip are then passed through a MTCNN face detector~\cite{zhang2016joint}, and clips with no faces detected in at least one of these frames. In addition, the set of clips is filtered further by comparing detected faces to the ground-truth faces of \gls{fiw}. Again, clips with no matches are discarded. Note that this was a means to quickly drop unwanted data. To compare faces, faces were encoded with ArcFace via the architecture, settings, training details, and \emph{matcher} in~\cite{deng2019arcface}. Specifically,

\begin{equation}\label{eg:matcher}
    d_{bool}({x}_i, {x}_j) = d({x}_i, {x}_j) \leq \theta,
\end{equation}
where the \textit{matcher} $d_{boolean}$ compared the $i$-th face detected to $j$-th \gls{fiw} face encoding ${x}$~\cite{LFWTech}. In other words, $d_{boolean}$ is the decision boundary in similarity score (or metric distance) space-- if threshold $\theta$ is satisfied, assume match; else, non-match. Note that it is currently assumed that $i$ and $j$ are from different sets (\ie with $J$ labeled samples of a subject from \gls{fiw} and $I$ face detections in the new video data). The \emph{matcher} in Eq~\ref{eg:matcher} was set as cosine similarity the closeness of the L2 normalized~\cite{wang2017normface} encodings by
$
d_{bool}({x}_i, {x}_j) = 1 - d({x}_i, {x}_j) = \frac{z_i\cdot z_j}{||z_i||_2||z_j||_2} > \theta
$, where $z$ represents an encoded piece of media. At this stage, $\theta=0.2$ was manually set for a high recall. The matching process - including the usage of ArcFace to encode faces - is the \textit{matcher} used throughout. 

Next, the MTCNN outputs were generated for all frames in clips, while saving the bounding box coordinates, fiducials (\ie 5 points), and confidence scores. Next, only continuous face tracks in clips were kept. For this, the ROI was set on the previous location of the face, and then IoU was calculated frame-by-frame, each value must surpass a threshold of 0.3. Finally, up to 25 faces were sampled uniformly from track (\ie opposed to choosing the top $K$ based on pose information, as this yielded redundancy in similar frontal posed faces). Each was then passed to $d_{bool}$ with each of the $I$ labeled faces (\ie producing $K\times I$ score matrix). The mean across $I$ samples was calculated to produce a single score per the $K$ faces, at which point the value at the 25-percentile was compared to $\theta=0.25$. The fusion of scores was done in such a way to both consider all the existing labeled faces equally, while avoiding a few (of the $K$) low-quality detections having any weight. Upon this process, and with the aid of \gls{sota} techniques mentioned throughout, this step alone yielded many face tracks matching with a high confidence.

\vspace{2mm}
\noindent\textbf{4) Audio branch.}
Audio data, in its raw form, is extracted from the videos and saved as high-quality wave files. We first set out to do speaker diarization on each video: we aimed to have a record indicating the presence of speech, from which change in speaker is marked, and, ultimately, the number of speakers in the video along with \emph{who} speaks \emph{when}. Note, we assume no audio labels. Thus, the speakers are arbitrarily tagged per video. 

Put differently, the first purpose of this branch is to find the number of speakers per video, with predictions based on the detected speakers on who spoke when: a speech detector determined the \emph{when}, and then clusters all the different speech segments to determine the number of speakers and, thus, which speech segment to assign to which of the speakers (\ie the \emph{who}). The former was implemented using PyPi's SpeechRecognizer\footnote{\href{https://github.com/Uberi/speech_recognition}{https://github.com/Uberi/speech\_recognition}}, with the latter based on models from~\cite{chung2020in}. See supplemental for further detail. Finally, parsing through segments and marking as $\text{speaker}_a$, $\text{speaker}_b$, ..., $\text{speaker}_j$, where $j$ is the number of speakers in a given clip. The time-stamps are used to detect speakers of interest.

\vspace{2mm}
\noindent\textbf{5) Visual-audio branch.} 
Focused is on detecting when the speaker is in the field of view. Thus, its purpose was to detect the boundaries (\ie start and end frames) for which the face and speech are aligned. An intuitive way to do this is to relate the faces detected and the lip movement with the audio-- which is at the core of many speaker identification methods in \gls{mm}~\cite{zhu2020deep}. To acquire this, videos were processed using SyncNet~\cite{Chung16a}) with the settings and trained weights from~\cite{li2017targeting}. The output were tracks: first trimming the video about the boundaries detected, and then cropping out the faces using the detected bounding box coordinates extended 130\% in all four directions. From this, each track is static spatially, and with each face detection captured within. This modification made it so individual tracks were of constant size and location in pixel space; opposed to producing tracks with moving coordinates to preserve the face in the field of view (\ie the added 30\% covered this). We then had three sets of coordinates saved (\ie the original detection, the extended version, and the set accounting for relative offsets for the crop). Similar to the \emph{visual branch}, labeled faces from \gls{fiw} were then used to determine the tracks belonging to the subject of interest. Once compared and, thus, filtered, all cropped tracks were manually inspected. Upon this the paired data was assumed.

\begin{table}[!t]
      \centering
        	\caption{\textbf{\gls{tar} at specific \gls{far}.} Scores are for template-based settings: still-images only (left column), +videos (middle), and +video+audio (right). Higher is better.}\label{subtab:task1:results}
\centering
     \resizebox{\textwidth}{!}{%
\begin{tabular}{r | ccc | ccc| ccc |ccc| ccc |ccc |ccc| ccc }
\toprule
    \gls{far}/\gls{tar} (\%)& 
    \multicolumn{3}{c}{\gls{bb}}& \multicolumn{3}{c}{\gls{ss}}& \multicolumn{3}{c}{\gls{sibs}}& 
    \multicolumn{3}{c}{\gls{fd}}& \multicolumn{3}{c}{\gls{fs}}& \multicolumn{3}{c}{\gls{md}}& \multicolumn{3}{c}{\gls{ms}} & 
    \multicolumn{3}{c}{Average}\tabularnewline\midrule
        0.5 (EER)  &97.8&97.8&\textbf{97.8} &91.5&92.3&\textbf{92.3} &91.7&90.8&\textbf{90.8} &\textbf{79.8}&77.8&77.8 &85.3&85.3&\textbf{85.3} &\textbf{90.6}&88.8&88.8 &81.3&82.6&\textbf{82.6} &88.3&87.9&\textbf{87.9}\tabularnewline

        0.3  &94.1&94.1&94.1 &88.0&87.2&87.2 &82.9&83.9&83.9 &63.5&66.5&66.5 &77.1&79.1&79.1 &82.4&82.0&82.0 &68.9&70.1&70.1 &79.6&80.4&\textbf{80.4}\tabularnewline
        0.1  &88.1&87.4&87.4 &76.1&76.1&76.1 &68.7&68.2&68.2 &34.5&36.9&36.9 &54.3&54.3&54.3 &62.2&63.1&63.1 &46.1&46.5&46.5 &61.4&61.8&\textbf{61.8} \tabularnewline
        0.01  &70.4&70.4&70.4 &54.7&55.6&55.6 &44.2&46.1&46.1 &5.9&7.9&7.9 &23.6&24.0&24.0 &28.3&31.3&31.3 &11.6&13.3&13.3 &34.1&35.5&\textbf{35.5} \tabularnewline
        0.001  &54.8&\textbf{57.0}&\textbf{57.0} &47.9&\textbf{48.7}&\textbf{48.7} &\textbf{29.5}&29.0&29.0 &2.0&\textbf{2.5}&\textbf{2.5} &9.3&\textbf{10.9}&\textbf{10.9} &14.2&\textbf{14.6}&\textbf{14.6} &3.3&\textbf{4.6}&\textbf{4.6} &23.0&\textbf{23.9}&\textbf{23.9} \tabularnewline
\end{tabular}
}
\end{table}

\section{Problem Definitions and Protocols}\label{sec:experimental}
The \gls{fiwmm} database extends the large-scale imageset \gls{fiw}~\cite{FIW, robinson2018visual}. Specifically, the images and names of \gls{fiw}, as explained in the previous section, allowed labeled multimedia data to be acquired via an automated process. Following the protocols of the recent \gls{rfiw} data challenge~\cite{robinson2020recognizing}, we benchmark two kin-based tasks:  verification and search \& retrieval. Differences from \gls{fiw} and, thus, from the experimental settings of \gls{rfiw}, are protocols based on still-imagery-- uni-modal and the experiments are organized as \emph{one-shot} problems. In contrast, \gls{fiwmm} offers multiple modalities, resulting in many more samples and sample types (\tabref{chapMM:tab:data-counts}). Furthermore, in an attempt to further bridge the gap between research-and-reality, the protocols we explain next is the first attempt in kinship recognition to follow template-based protocols~\cite{maze2018iarpa}.

As for experimental tasks, kinship verification has been the primary focus. More recently, the emergence of the more challenging but more practically awarding task of \emph{searching for missing children} task~\cite{robinson2020recognizing}, \ie search and retrieval. We benchmark \gls{fiwmm} for both these tasks. However, opposed to the single-shot setting followed up until now, we use templates~\cite{maze2018iarpa}; hence, a means to move experiments closer to settings for operational use-cases.

Template-based experiments are organized as follows: Known subjects (\ie prior knowledge of identity and family) are first enrolled in a \emph{gallery} $G$. At inference, the aim in the search and retrieval task is to compare an unseen \emph{probe} $P$ to subjects enrolled in $G$; the verification task compares a list of \emph{probes} to individual \emph{gallery} subjects (\ie one-to-one) with the solution space of either \emph{KIN} or \emph{NON-KIN}; kinship identification compares the \emph{probe} to the entire \emph{gallery} (\ie one-to-many), with the end result being a ranked list of family members. In all cases, at least one family member exists in $G$, making for a closed-set recognition problem.

Specifically, template $X$ holds all of the media for a subject (\ie face images, videos, audio-clips, and text transcripts). Hence, $X$ consists of samples $x$, where each $x$ is an independent piece of media represented as an encoding $z$. For instance, still-image $x$ encoded as $z$ by $\mathcal{F}(x)=z$, where $\mathcal{F}$ maps faces to a learned feature space (\ie $\mathcal{F}(x)\in\mathbb{R}^\mathrm{d}$, where the dimension $\mathrm{d}$ represents the size of the respective encoding). The same is done for face tracks in videos, which were fused to a single encoding by average pooling. Put formally, a face track is represented as $\bar{z}=\frac{1}{m}\sum_x\mathcal{F}(x)$, where $m$ is the frame count. Similarly, an audio segment (\ie a clip where subject speaks without interruptions or major pauses) is treated as a single piece of media $x$ via average pooling all encodings to form a single representation per clip. Note that a video may consist of several independent visual, audio, and visual-audio (\ie aligned) tracks. Thus, there are many independent media samples for both the visual and audio modalities. Again, subjects are represented by these templates $X$ made up of these various media samples $x$, such that the $j^{th}$ subject can be represented by $k$ media samples as follows: $X_j={\mathcal{F}_t(x_1), \mathcal{F}_t(x_2), \dots, \mathcal{F}_t(x_k)}$, where $t$ corresponds to the media type and, hence, the corresponding encoder. From this, $|X_j|$ is the total number of encodings for subject $j$. The \emph{gallery} $G$ consists of a set of subjects by $G=\{(X_1, y_1)^l, (X_2, y_2)^l, \dots, (X_n, y_n)^l\}$, where $y$ are identity labels for each of the $N$ subjects, and $l \in \{1,2, \dots L \}$ are ground-truth for $L$ families. To establish a precise definition for problems of kinship, each tuple also contains a tag representing the set of $L$ families (\ie $(X_j, y_j)^l$), where $l\in\{1, 2, \dots, L\}$. Further partitioning of the data is done per requirements of a task. For instance, for the verification, the $m^{th}$ pair of tuples from the same family $\mathbb{P}_m=((X_i, y_i)\bigcap(X_j, y_j))$, where $i\neq j$, inherit labels KIN (\ie \emph{match}) and relationship type.

Following the 2020 \gls{rfiw}, each task consists of a train, validation, and test set. These sets are disjoint in family and subject IDs, and are roughly split 60\%, 20\%, and 20\% for the train, validation, and test set, respectively. Thus, the splitting is done using the family labels, and the resulting partitioning of sets is static for all tasks.

\subsection{Kinship verification}
\label{sec:track1}
Kinship verification is a challenging task within a complex topic. It inherits all the challenges of traditional \gls{fr}, with aspects amplified in difficulty due to kinship being a soft attribute with high variation, bias in nature, and directional in the variety of relationship types. The most fundamental question asked in kinship verification, and re-asked in all other kinship discrimination tasks is whether a face pair is related. Therefore, kinship verification is a boolean classification of pairs (\ie ${y}\in\{\emph{KIN}\bigcup\emph{NON-KIN}\}$). Knowledge of the relationship type is assumed to be known. Thus, provided the output of the model for a given pair is \emph{KIN}, then the specific type is implied. Future efforts could incorporate relationship-type signals to advance capabilities of kinship detection systems; however, and as stated upfront, verification provides the simplest of all the benchmarks and, up until now, is the most popular~\cite{robinson2020recognizing}.

\subsubsection{Data splits and settings}
The data is organized as pairs, with pairs a part of a set of common relationship-type. Specifically, pairs are of type \gls{bb}, \gls{ss}, or \gls{sibs} of mixed-sex (\ie same generation), or \gls{fd}, \gls{fs}, \gls{md}, or \gls{ms} (\ie difference on 1-generation). Counts for all types of relationship pairs are listed in \tabref{chapMM:tab:data-counts}, with the aforementioned types (\ie same and 1-generation) used in experiments provided sample sizes are such to allow for fair representations.  Data splits (\ie train, validation, and test) and their sample counts are listed in \tabref{chapMM:tab:counts}. The task here has no concept of query and gallery.

\subsubsection{Metrics}
The one-to-one paradigm (\ie kinship verification) is the main view vision researchers aim to solve. The task is to determine whether a face-pair are blood relatives (\ie \emph{true kin}). Conventionally, a query consists of a single face image $x_1$, which is then paired with a second face $x_2$ to predict against (\ie a one-shot, boolean classification problem with labels $y\in\{\text{KIN}, \text{NON-KIN}\}$). Put formally, given a set of face-pairs $(x_1, x_2)_s^m$, where the number of sample pairs $s\in\{1, 2, \dots, S\}$ of relationship-type $m\in M\rightarrow\{BB, SS, \dots, GMGD, GMGS\}$ (\ie $|M|=11$). A set of pair-lists $\mathbb{P}=\{[(x_1, x_2)_1^m]_\mathbbm{1}, [(x_1, x_2)_2^m]_\mathbbm{1}, \dots ,[(x_1, x_2)_S^m]_\mathbbm{1}\}$ for the $M$ types, and with the label determined by the indicator function $\mathbbm{1}$ for a single pair $\mathbb{P}_s\rightarrow\{ 0,1\}$, \ie

\begin{equation}\label{eq:indicator}
\mathbbm{1}(\mathbb{P}_s) = 
\begin{cases} 
            0 & \hspace{5mm} \emph{NON-KIN} \\
            1 & \hspace{5mm} \emph{KIN}\\
\end{cases}.
\end{equation}

Note, a $\mathbb{P}_s$ consists of a pair of templates and, thus, the task is to determine whether the media of the templates provide evidence of the two subjects being blood relatives; notice Eq.~(\ref{eq:indicator}) is the template \emph{matcher} defined in Eq.~(\ref{eg:matcher}).

As described, \gls{fiwmm} is organized as templates with many samples from various modalities (i.e., still-face, face-tracks, audio, and transcripts (contextual). Specifically, true IDs $y$ are paired with a template of all media available for the respective subject. In contrast with conventional kinship recognition, where one image is compared to another, the one-to-one paradigm is based on templates (\ie one template is compared to another). For consistency, given $\mathbb{P}_s^m=((X_i, y_i), (X_j, y_j))$ as a pair of templates for different subjects (\ie $X_i$ and $X_j$, where $i\neq j$).

\Gls{det} curves, along with average verification accuracy, were used for kinship verification as also were \gls{tar} across intervals of \gls{far} (\tabref{subtab:task1:results}).

\newcolumntype{g}{>{\columncolor{Gray}}c}
\begin{table}[!t]
      \centering
        \scriptsize
        \caption{\textbf{Identification results, with \glspl{ta} highlighted.} Accuracy scores for different ranks are listed (\ie higher is better). Also, MAP scores are provided for each.}\label{subtab:task3:results}


	\begin{tabular}{m{14mm}ccccccc} 
	         &&	  \multicolumn{5}{c}{\emph{Rank}} &\tabularnewline \cline{3-7}\tabularnewline[-1em]
		&	& \textbf{@1} &	 \textbf{@5} & \textbf{@10}	&  \textbf{@20} &  \textbf{@50} &  \textbf{mAP}\tabularnewline[-.25em]
	\midrule

	\emph{img} &mean & 0.29 & 0.43 & 0.54 & 0.64 & 0.78 & 0.13\tabularnewline
     & median & 0.28 & 0.44 & 0.52 & 0.64 & 0.77 & 0.13\tabularnewline
     & max & 0.11 & 0.19 & 0.28 & 0.34 & 0.52 & 0.06\tabularnewline
     	 & \cellcolor{LightCyan} TA & \cellcolor{LightCyan} 0.31 &\cellcolor{LightCyan}  0.43 & \cellcolor{LightCyan} 0.52 &\cellcolor{LightCyan}  0.63 & \cellcolor{LightCyan} 0.74 &\cellcolor{LightCyan}  0.14\tabularnewline
    \midrule

    \emph{img+video} & mean & 0.30 & 0.44 & 0.52 & 0.64 & 0.77 & 0.14\tabularnewline
     & median & 0.28 & 0.44 & 0.50 & 0.63 & 0.76 & 0.14\tabularnewline
     & max & 0.13 & 0.21 & 0.26 & 0.30 & 0.44 & 0.06\tabularnewline
          & \cellcolor{LightCyan} TA & \cellcolor{LightCyan} 0.34 & \cellcolor{LightCyan} 0.46 & \cellcolor{LightCyan} 0.55 & \cellcolor{LightCyan} 0.68 & \cellcolor{LightCyan} 0.75 & \cellcolor{LightCyan} 0.16\tabularnewline
    \midrule
     
    \emph{img+video+audio} & mean & 0.30 & 0.44 & 0.52 & 0.64 & 0.77 & 0.14\tabularnewline
     & median & 0.28 & 0.44 & 0.50 & 0.63 & 0.76 & 0.14\tabularnewline
     & max & 0.13 & 0.21 & 0.26 & 0.30 & 0.44 & 0.06\tabularnewline
     & \cellcolor{LightCyan} TA & \cellcolor{LightCyan} \textbf{0.56} &\cellcolor{LightCyan}  \textbf{0.59} &\cellcolor{LightCyan}  \cellcolor{LightCyan} \textbf{0.63} &\cellcolor{LightCyan}  \textbf{0.74} &\cellcolor{LightCyan}  \textbf{0.78} & \cellcolor{LightCyan} \textbf{0.24}\tabularnewline
     \bottomrule
\end{tabular} 
\end{table}

\begin{figure}[t!]
\centering
  \begin{subfigure}[b]{0.48\linewidth}
  \centering
  \includegraphics[width=\linewidth]{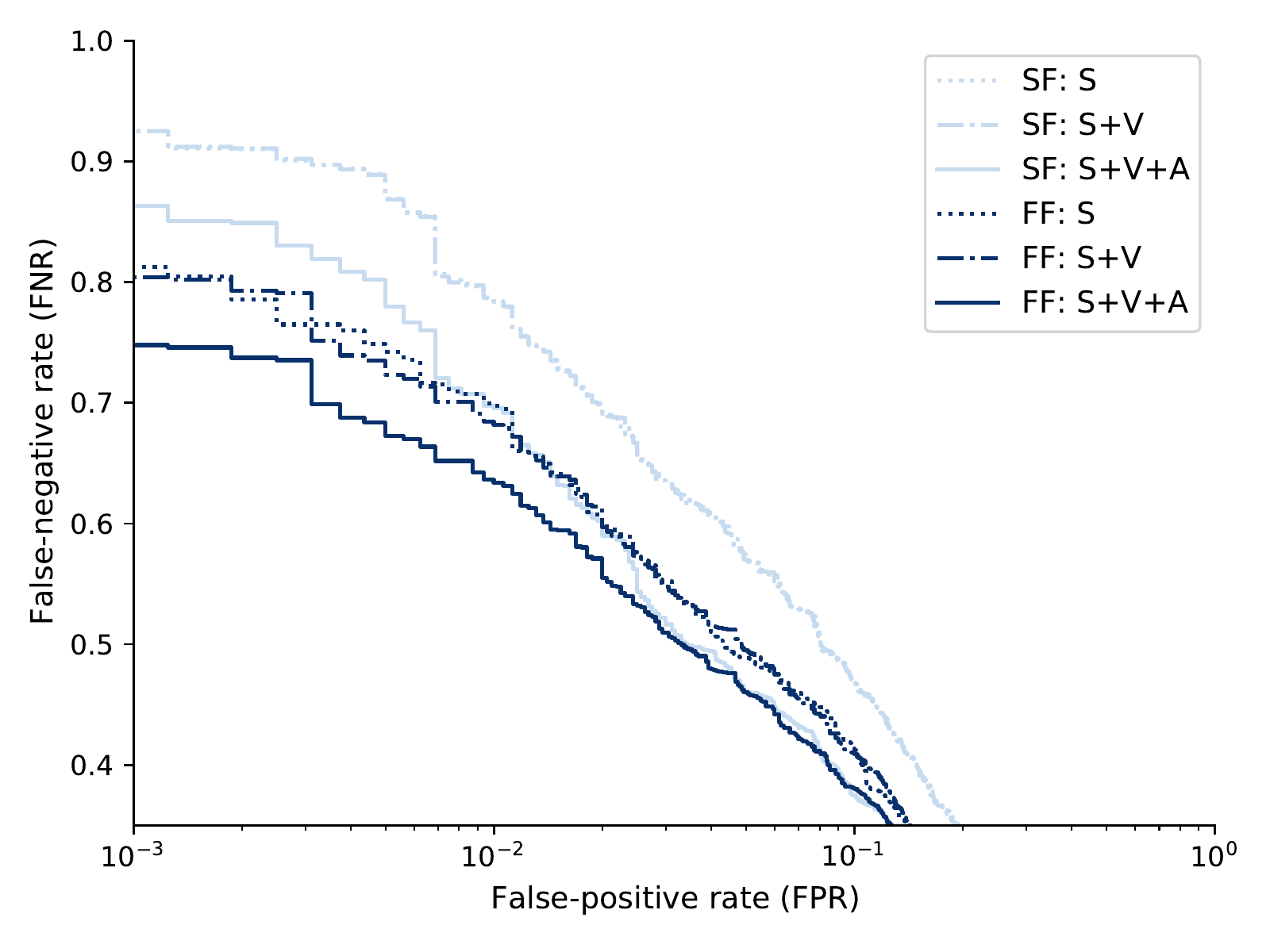}
  \caption{{\gls{det} curve}}\label{chapMM:fig:track1:det}
  \end{subfigure}
\begin{subfigure}[b]{0.48\linewidth}
  \centering
  \includegraphics[width=\linewidth]{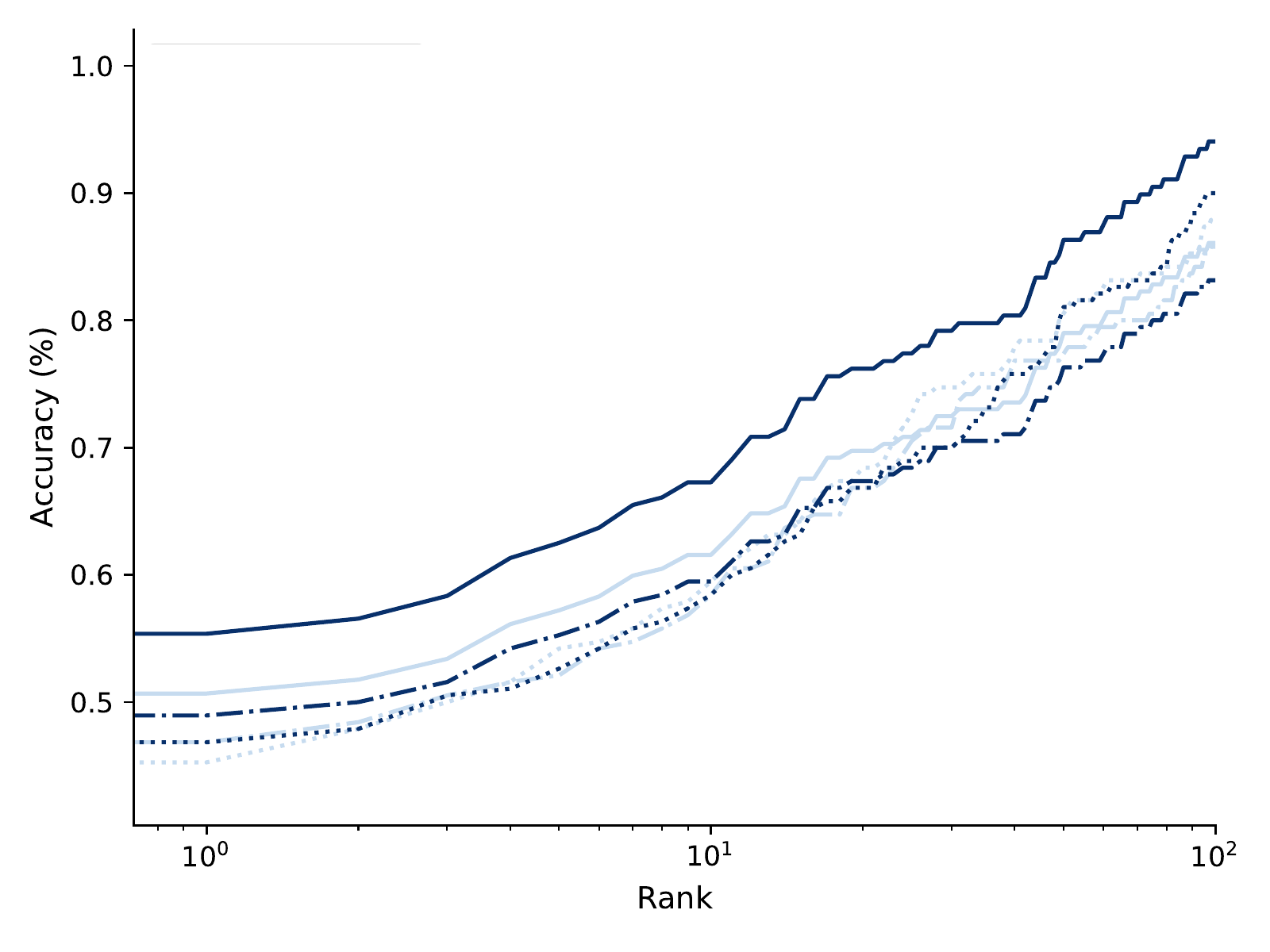}
  \caption{{CMC curve}}\label{chapMM:fig:track3:cmc}
  \end{subfigure}
    \caption{\textbf{Plotted results.} Shown are score fusion (SF) and feature fusion (FF) as late and early fusion methods, respectively. Included are still-images $S$, video clips $V$, and audio segments $A$, with still-images and video were fused $S+V$, and also still-images, audio, and video were fused $S+V+A$. Clearly, both tasks benefit from early fusion: the DET curve (left) summarizes the verification task by plotting FNR as a function of FPR (\ie lower is better); search and retrieval is summarized as a CMC (right) by showing the accuracy as a function of rank (\ie higher is better).}\label{chapMM:fig:plots}
\end{figure}

\subsection{Search \& retrieval (missing child)}
\subsubsection{Overview}
Kinship identification is organized as a \textit{many-to-many} search and retrieval task, with each subject having one-to-many media samples. Thus, we imitate template-based evaluation protocols~\cite{maze2018iarpa}. Furthermore, the goal is to find relatives of search subjects (\ie \textit{probes}) in a search pool (\ie \textit{gallery}).

\subsubsection{Data splits and settings}
A gallery $G=\{g_i\}, ( i=1,...,N)$ is queried by a set of probes $P=\{p_j\}, ( j=1,...,M)$ for search and retrieval, where $g_i$ is the $i$-th template in $G$ and $p_j$ is the template of the $j$-th query subject. As mentioned, a template consists of samples of various modalities. Given a template of \gls{mm}, various schemes were applied to integrate the ID information from all media components of $P$.

\subsubsection{Metrics} 
Scores of $N$ missing children are calculated as
$$
AP(l)=\frac{1}{P_L}\sum^{P_L}_{tp=1}Prec(tp)=\frac{1}{P_L}\sum^{P_L}_{tp=1}\frac{tp}{rank(tp)},
$$
where \gls{ap} is a function of family $l\in L$ (\ie $|L| = {P_L}$) for a given \gls{tpr}. All \gls{ap} scores are averaged to find the mean \gls{ap} (\ie MAP):
$$MAP = \frac{1}{N}\sum^{N}_{l=1}AP(l).$$

Also, \gls{cmc} curves as a function of rank enable for analysis between different attempts~\cite{decann2013relating}, along with the accuracy at rank 1, 5, and 10.

\section{Benchmarks}
\subsection{Methodology}

The problems of \gls{fiwmm} have various views– multi-source and multi-modal.  The former varies in samples and in treating the different media-types independently until the matching function outputs scores (\ie late-fusion).  The latter demands a method for early fusion (\eg feature-level) which should enhance performance by leveraging informative samples while ignoring noisy and less discriminative samples. We next describe the modality-specific features (\ie encoding different media types), and early fusion.   

\subsubsection{Visual features}
\gls{fr} performance traditionally focuses on verification-- popularized by the Labeled Faces in the Wild dataset~\cite{LFWTech} (images) and the YouTubeFaces dataset~\cite{wolf2011face} (videos). In contrast, the newer IJB-[A,B,C] \gls{fr} datasets~\cite{maze2018iarpa} unifies evaluation of one-to-many face identification with one-to-one face verification over templates (\ie sets of imagery and videos for a subject). Then, visual kinship recognition research followed a similar path, addressing the simpler verification task. \gls{fiwmm}
provides the data needed to run template-based kin recognition experiments.

We demonstrate results from a variety of naive fusion techniques (\eg average pooling of features or voting of scores). To no surprise, the score-based fusion outperforms the naive feature-level fusion schemes. Specifically, the mean of all scores, both within a template and comparing templates (\tabref{subtab:task1:results}). The gain from each added modality is clear from just the naive score-fusion.

As mentioned, naive fusion methods at the feature level are an ineffective way of combining knowledge. Provided a collection of media - media that varies in  modality, quality and discriminative power - a simple, unweighted average across the items of a template does not exploit all available information. To better fuse the template, we adapt a model to the template to best represent the subject for verification or identification of family members. Details are provided right after the description of audio features. 

\subsubsection{Audio features} All speech segments were encoded a \gls{sota} deep learning architecture~\cite{chung2020in}. Specifically, we trained  SqueezeNet~\cite{iandola2016squeezenet} as a 34-layer ResNet~\cite{he2016deep} with an \emph{angular prototypical loss} and optimized with Adam~\cite{kingma2014adam} to transform WAV-encoded audio files to a single encoding, \ie $f(x)=z\in\mathcal{R}^d$, where $d=512$. \emph{Angular prototypical loss }~\cite{snell2017prototypical} learns a metric alongside softmax to minimize within-class scatter (\ie penalty formed as the sum of euclidean distances from all samples of a subject in a mini-batch from the mean centroid of the respective mini-batch). Specifically, a support set $S$ and a query $Q$ are set in each mini-batch on a subject-by-subject basis, with $Q$ made-up of a single utterance to compare with the centroid of $S$ that consists of all other samples in the mini-batch for that class. \emph{Angular prototypical} takes advantage of the perks of using centroid prototypes, while enhancing by following \gls{ge2e}~\cite{wan2018generalized} usage of a cosine-based similarity metric. This is scale invariant, is more robust to feature variance, and facilitates stability in convergence during training~\cite{wang2017deep}.

\subsubsection{Feature fusion}
\Gls{ta}~\cite{whitelam2017iarpa} is a form of transfer learning that fuses the deep encodings of many labeled faces from a source domain with a template specific \glspl{svm} trained on the target domain. For kinship verification, we employ \emph{probe adaptation}; for identification (\ie search \& retrieval), we do \emph{gallery adaptation}. Thus, \gls{ta} enabled early fusion of different media types in both tasks.

Specifically, a similarity function $s(P, Q)$, for probe $P$ and reference template $Q$, is learned for a given probe (\ie template). An \gls{svm} is trained on top of the face encodings with media in $x^+$ as the positive samples and the set of negatives $x^-$ being single sample from subjects in the train set (\ie $x^+ \ll x^-$). For verification, this process repeats for another \gls{svm} $Q$ (\ie the template of the subject in question). Negatives were set in same way. Then, let $P(q)$ represent the evaluation of media encodings of template $Q$ upon being trained on $P$. We do this in both direction via
\begin{equation}
s(P,Q) = \frac{1}{2}P(q) + \frac{1}{2}Q(p).
\end{equation}
The score produced is the result of the templates fused together from media to an \gls{svm} and then to a score.

The benefit of \glspl{svm} is in the kernel. Specifically, the linear, max-margin modeling scheme of a vanilla \gls{svm} has proven effective at separating non-linear feature space of two classes; (\ie $i$ and $j$, where $y_{ij}=\pm1$ for instances of the same ($+$) and different ($-$) classes. Thus, the implicit embedding function (\ie kernel) $K(x_i, x_j, y_{ij})=\varphi(x_i, y_i)\varphi(x_j, y_j)$ projects the encoding pair to a non-linear space such that the \gls{svm} learns the best hyperplane $\mathbf{w}^T K(x_i, x_j, y_{ij}) + b = 0$ to separate the two classes. This is done by (1) maximizing the margin and (2) minimizing the loss on the training set-- weights $\mathbf{w}$ is learned, while bias term $b$ we set to 1 (\ie concatenated on $\mathbf{w}$ as an added dimension). Also, $K(x_i, x_j, y_{ij})=\exp{\frac{||x_i- x_j||^2}{2\sigma^2}}$ for ${y_{ij}\in\{-, +\}}$ as the respective class (\ie Gaussian RBF kernel~\cite{scholkopf2002learning} projects all encodings to a higher dimensional space); then the predicted class is inferred as $\hat{y}=\mathbf{w}^T\varphi(x_i)\varphi(x_j) + b$. We used \emph{dlib's}~\cite{king2009dlib}-- L2 regularized cosine-loss with class-weighted hinge-loss, \ie
    
  \begin{align}
    \nonumber \min_\mathbf{w}{\frac{1}{2}\mathbf{w}^T\mathbf{w}} + \lambda_{+}\sum_{i=1}^{N_{+}}\max\large{[}0, 1 - y_i\mathbf{w}^Tf(x_i)\large{]}^2 \\
    + \lambda_{-}\sum_{j=1}^{N_{-}}\max\large{[}0, 1 - y_j\mathbf{w}^Tf(x_j)\large{]}^2.
    \label{eq:hinge}
  \end{align}

Adapting this to the notion of a gallery, the protocols are set for \emph{gallery adaptation}: train a similarity function $s(P, G)$ from a probe $P$ to gallery $G$. A gallery of templates $G=\{X_1, X_2, \dots, X_m\}$ are used to train the \gls{svm} (\ie the scoring function $s(P, X_i)$). The difference between \emph{probe adaptation} and \emph{gallery adaptation} is in the negative sets. Along with the sample per subject trained against for \emph{probe adaptation}, \emph{global adaptation} samples all other templates in $G$ as additional negatives. Again, $N_{+} << N_{-}$. The class imbalance is handled via class-weighted hinge-loss in Eq~\ref{eq:hinge}, with $\lambda_{+}=\lambda\frac{N_{+} + N_{-}}{2N_{+}}$, $\lambda_{-}=\lambda\frac{N_{+} + N_{-}}{2N_{-}}$, which are regularization constants  inversely proportionate to class frequency. The constant $\lambda$ trades-off between the regularization and loss, which we set to 10 as in previous work~\cite{whitelam2017iarpa}.

\subsection{Results}
The ability of a system to discriminate is improved with each added modality (Fig~\ref{chapMM:fig:plots}, \tabref{subtab:task1:results} and~\ref{subtab:task3:results}). Considering the benchmarks use conventional speech and \gls{fr} technology, and our hypothesis that video and audio boosts discrimination, much promise reflects-- these notable improvements would likely continue to climb provided a more sophisticated or specific solution. It would be interesting to fuse earlier on than done here, and train machinery jointly with audio-visual data. This way, more complex dynamics of facial appearance, along with the corresponding sound of voice, could further improve and give additional insights. 

\begin{figure}[!t]
\centering
    \includegraphics[width=\linewidth]{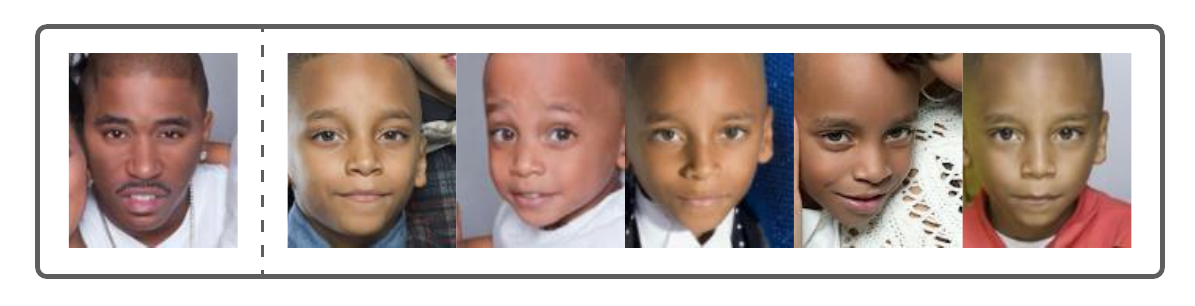}
  \caption{\textbf{Random hard sample.} Template of a true FS pair that was incorrectly classified using score fusion, but correct for TA (\ie feature fusion). Here only a single face is available for the father (left), while all instances of the son are at a young age (right).}
  \label{chapMM:fig:fs:visual:qual}
\end{figure}

\begin{figure}[!t]
\centering
    \includegraphics[width=\linewidth]{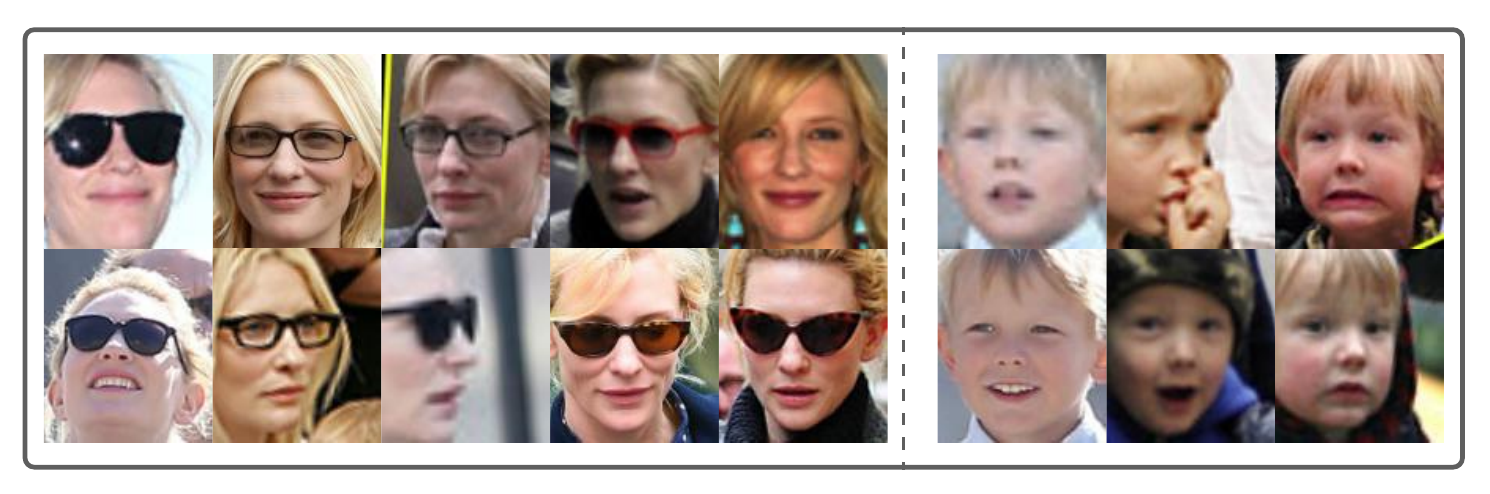}
  \caption{\textbf{Hard sample pair.} A true MS pair incorrectly classified using late fusion, while correctly identified as KIN when using the early fusion via TA. Challenges here are in the young age of the son and the majority of the faces of the mother occluded by sunglasses (\ie score fusion puts equal weight on all samples, where TA learns to better discriminate).}
  \label{chapMM:fig:ta:visual:qual}
\end{figure}

There is a trend in the type of samples that were corrected when comparing the score fused to the feature fused (\ie \gls{ta}) results. As shown in \figref{chapMM:fig:fs:visual:qual}, the challenge of recognizing kinship from samples of one or more member at a young age is mitigated. \gls{ta} learns to better discriminate in such conventional failure cases. Additionally, some templates made up of multiple instances, often are better than others when comparing. Hence, \gls{ta} does not simply average all instances with equal weighting, as done in late (\ie score) fusion-- seen in cases containing a minority of samples that are more discriminate than its majority (Figure~~\ref{chapMM:fig:ta:visual:qual}).

\subsection{Discussion}
The template-based protocol adds practical value by mimicking the more likely structure posed in operational settings, per \gls{nist}~\cite{maze2018iarpa}. Besides, several other factors make it a more interesting formulation and therefore, a higher potential for researchers to show-off their creativity. For instance, opposed to using a single sample per subject (\ie one-shot learning), each now is represented in a set of media (\ie a template). The questions now arise - how to best fuse knowledge and incorporate evidence from different modalities, and how to best learn from all available MM data? Another consequence of using templates is that the random chance is increased from (1) the knowledge added to pool (or fuse) from the added modalities, and (2) the gallery size reduces from tens of thousands by nearly ten-fold. The latter is not an implication of lesser difficulty, but the byproduct of reducing bias in data~\cite{robinson2020face}. That is, opposed to having one-to-many samples per subject, there is just one template. Mitigating certain sources of data imbalance (\ie whether there are thirty samples or just one) a system’s ability to recognize a particular pairing or group affects the metric evenly for all. In other words, a system may easily recognize a specific parent-child pair - regardless of the number of face samples and, consequently, the number of face pairs. Hence, the impact on the metric is proportional to the number of unique pairs, not sample pairs.

\section{Future Work}
\gls{fiwmm} pushes the bar for possibilities in automatic kinship recognition and understanding. One immediate next step for research involves the benefit of gathering experts of different domains, such as those in sequence-to-sequence modeling, whether visual (\ie video), audio (\ie speech), contextual (\eg conversations, parts-of-speech, \etc), or early-fusing pairs or groups. Let us next discuss a variety of ways the data is foreseen to benefit and bring together different research communities, and beyond (\ie its inherent commercial potential).

We expect \gls{fiwmm} will bring experts of anthropology and genealogy-based together with those researching \gls{mm}, machine learning, and vision topics toward helping to identify the hidden patterns that relate families in the \gls{mm} data. Particularly, let us consider audio. As we have shown, pre-trained models from the speech recognition domain provide a means to acquire audio features with discriminative power that boosts kinship recognition systems which use only visual evidence. Furthermore, high-level semantics (\ie attributes) like accents, commonly used phrases, and speaker demeanor, could boost the performance of a system and also provide insights by interpretation. Similarly, studies focused on familial language components and inherited changes; or even the same generation (\ie commonalities and differences in the spoken tendencies of siblings) can be quite revealing as well. Hence, the potential only grows with audio-visual data, both in model complexity (\ie capacity) and practical uses: one could model mannerisms from the dynamics captured across video frames, and provide the answers to questions like ``do I have my mother's smile?''.

The data mining potential of \gls{fiwmm} is noteworthy. The family trees, abundance of data points, rich metadata for individuals and relationships among \gls{mm} data-- \gls{fiwmm} could serve as a basis for group-based (\ie social) data mining. Additional data can enhance or target specific nature-based studies, traditional ML-based audio, visual, and audio-visual tasks, or even further extend this dataset. Fusing audio-visual data is an ongoing, unanswered problem ~\cite{song2019review}. Note the following: (1) the model training, for instance, with one or multiple incomplete modalities, (2) the data processing and balancing, and (3) the underlying roots of the problem to the high-level semantics, similar to contemporary biometrics systems with audio-visual data-- \gls{fiwmm} and, thus, this work in its entirety, poses more problems than it solves.  We actually introduce a much larger problem space than that of solutions.

Note that FIW-MM and thus this work in its entirety pose more problems than it solves: from the model training, to improvements made when dealing with incomplete modalities, and even the data processing and data imbalance; from the underlying roots of the problem to the high-level semantics, similar to contemporary biometrics systems with audio-visual data; we introduce a much larger problem space than that of solutions. 

Another direction is fusion. For experiments, we included early and late fusion by joining the different media as features and scores, respectively. Scores were fused naively, ignoring the signal type, and assuming all samples and media types should be weighted uniformly. Fusion can incorporate more sophisticated techniques: cross-modality, selectively choosing the highest quality samples, or a decision tree based on modalities. This concept alone is vast in empty solution space-- whether data fusion, where the input is then clips of aligned audio-visual data; early-fusion, which was exemplified with \gls{ta} fusing the features; or late-fusion, also demonstrated by averaging scores, but could have just as easily been guided by a more clever decision tree mechanism. Besides, meta-knowledge, like relationship types (\eg directional relationships that inherently exist), genders, age, and other attributes, could indicate final decisions. Hence, there are an abundance of fusion paradigms-- none are trivial, yet most hold promise. 

Research topics to spawn off the proposed is vast, to say the least; the specifics suggested here are limited by our perception. We expect scholars and experts of different domains to seek out paradigms not thought of by us in the moment. Hence, whether it be an improved variant of adapting templates and feature fusion (\eg like in~\cite{xiong2017good}), deciding when to fuse, a new method of integration, along with the integration details, are all open research questions. 

In the end, the data resource outweighs the benchmarks. This is by design, as this resource will be readily available for research purposes - even a complete characterization of the contents as is (\ie ablation studies like on the effects of template sizes, media type versus relationship types, or even high-level interpretations (\eg smiling faces versus neutral). 

\section{Conclusion}
We introduced new paradigms (\ie template-based) for kinship recognition via the proposed \gls{fiwmm} database. \gls{fiwmm} contains audio, video, audio-visual, and text captions for 2+ members from 150 / 1,000 families of \gls{fiw}. Our labeling pipeline uses multi-modal evidence and a simple feedback schema to leverage the labeled data of \gls{fiw} to propagate ground truth for the added modalities. Benchmarks show improved performance with each added media type, and then further by early fusion. \gls{fiwmm} marks a major milestone for kin-based problems by welcoming experts of other data domains. In addition, \gls{fiwmm} supports a number of \gls{mm} recognition tasks due to its rich metadata, template-based structure and multiple modalities.

One motivation of this survey is to establish cohesive views of the major milestones via protocols that are clearly defined, data splits that are ready for download, and trained models that make baselines reproducible.\footnote{\href{https://github.com/visionjo/pykinship}{https://github.com/visionjo/pykinship}} Hence, components to reproduce experiments that we report make up part of the supplemental material. We cover the edge cases that challenge \gls{sota}, including an examination of the different settings and training tricks that further our abilities in kin-based detection from faces.

\part{Post Processing}\label{part:post}
\chapter{Kinship Recognition - State of Technology}\label{chap:krstateoftechnology}

\section{Overview}
To review the current state of technology accessible for automatic kinship recognition in multimedia there are two separate aspects of the problem in need of elaboration: our limitations and the challenges for which they are set and details of the real-world uses-cases that have been mentioned briefly throughout this dissertation.

Hence, having reviewed the means and the results, let us now examine how it fits in practice. Specifically, let us now summarize the technical challenges still relevant in problems of automatic kinship recognition. We cover the challenges as they exist: general challenges as seen collectively, the way in which the challenges put limits on capabilities for current \gls{sota} machinery, and in which ways challenges of the problem are set by nature, the environment, and inherited by the source and structure of the data in itself. We then transition to a discussion on applications, \ie actual applications and potential ideas for use-cases. Finally, we conclude the chapter with a reflection of the aforementioned topics in the form of a discussion.

\section{Technical Challenges}\label{sec:challenges}
Like conventional \gls{fr}, unconstrained faces \emph{in the wild}~\cite{LFWTech} yield more difficult - imagery collected from sources outside a controlled laboratory environment is subject to more variations in pose, illumination, and scale. For faces, there are even more variables to further complicate the problem, such as expression and age. Furthermore, preparing to run such benchmarks to mimic real-world use-cases (\ie designing experiments and preparing the data) is, in itself, a challenge. Inheriting these challenges, but adding even more variations inherent in nature and in true data distributions of kinship, it is unsurprising that visual kinship recognition is a difficult problem. Nonetheless, great efforts over the last decade have been spent not just on solving the problems in kinship recognition, but also critiquing kinship research and its direction. We now elaborate on the challenges to keep this technology from making the transition of research-to-reality.

\subsection{Current limitations of SOTA}\label{sec:limitations}
Still, we are close to achieving a performance-rating necessary for some applications (Section~\ref{sec:applications}). From this, we perceive that bridging the gap between research-and-reality (\ie transitioning from research-to-practice) is happening. Upon a clear assessment of the state of progress in research, we highlight barriers still in need of overcoming, along with sharing edge cases as means of highlighting common errors. Hence, we aim to inspire by explicitly depicting weaknesses in current \gls{sota} systems.

A clear limitation, however, is that most solutions for visual kinship recognition assume the relationship type \emph{apriori}. Sometimes this could be practical, like if given a known source to decide whether or not the face, when paired with a target, is \emph{KIN} or \emph{NON-KIN}. Nonetheless, when considering the broader HCI incentive, along with data mining with social context, it is desirable to predict the exact type of relationship (\ie not just \emph{KIN} or \emph{NON-KIN}). Nonetheless, a high confidence in knowing whether a relationship does exist could serve as powerful prior knowledge when classifying the specific type.

Let us now consider the renowned \gls{kfw} dataset. Although the dataset has had a great impact in research, for having attracted many to the problem and, thus, has motivated many outstanding works, there are a few clear flaws in relating the results to real-world data. More than half of all true pairs making up \gls{kfw} are faces from the same photo. Researchers have then questioned the validity of the patterns being learned, showing that naive approaches such as color features~\cite{lopez2016comments} or detecting whether or not faces are from the same photo~\cite{dawson2018same} outperform \gls{sota} on most datasets, including \gls{kfw}. Thus, another clear limitation of some data resources is in the data distribution itself -- a technical challenge we soon cover in-depth (Section~\ref{subsec:datachallenges}).

\subsection{The nature}

\begin{figure}[!t]
    \centering
    \includegraphics[width=.65\linewidth]{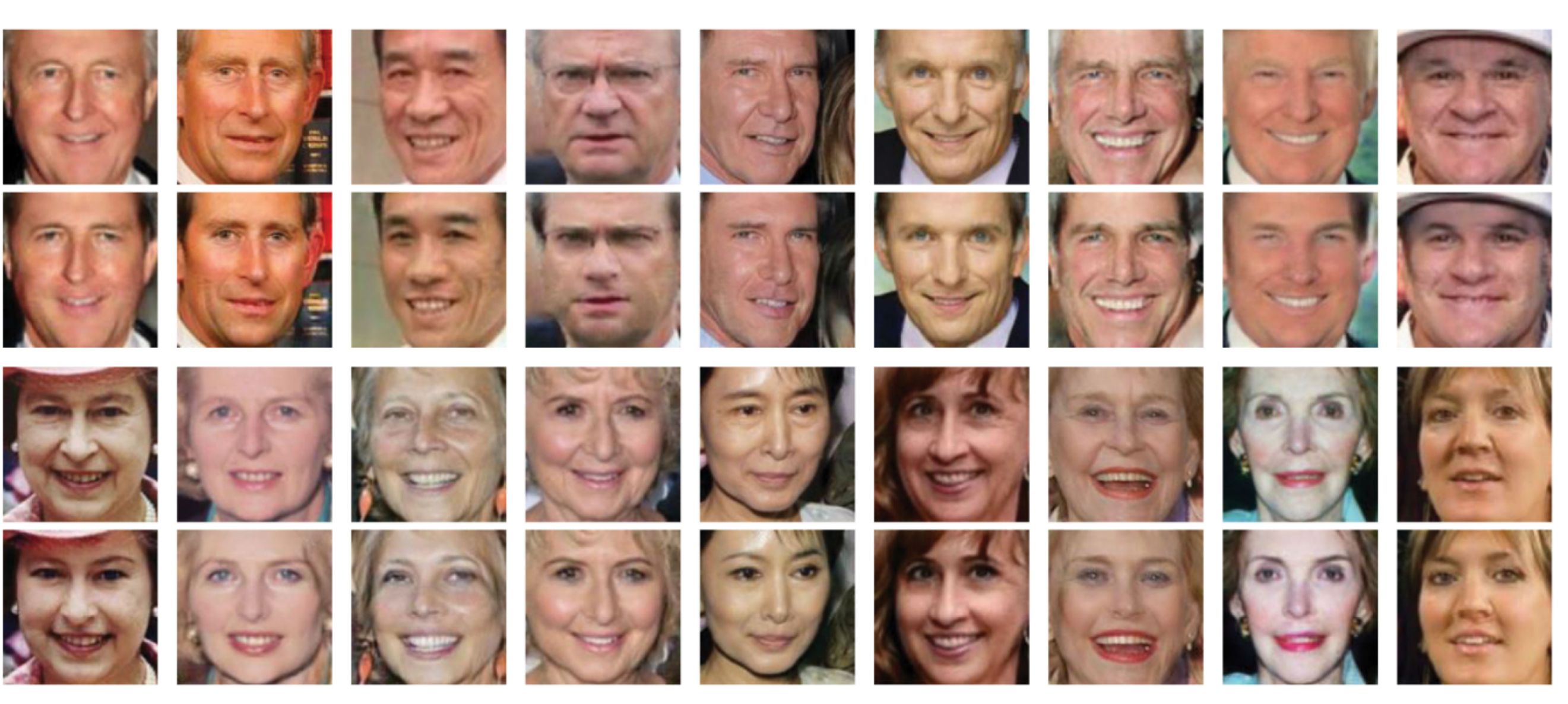}
    \caption{\textbf{Sample faces synthesized to improve predictive power for faces of elderly adults (visualization from~\cite{wang2018cross}).} Two models (\ie one per gender) were trained to synthesize input faces as younger-- male fathers (\ie rows 1-2) and female mothers (\ie rows 3-4), the top sample is the original and the generated is below.}
    \label{chap:fiwmm:fig:cross:generation}
\end{figure}

\noindent\textbf{Demographics and inherent bias.}
A challenge are issues of bias in \gls{fr} machinery. However, no study of bias in demographics (\eg ethnicity and gender) for kin-based data: a study that should be conducted, like Robinson~\etal found variations in score sensitivities across subgroups in \gls{fr}~\cite{robinson2020face}.

\noindent\textbf{Effects of age variations.}
Family members with a large age-gap makes for more of a challenge. Wang~\etal demonstrated a benefit in having a face image synthesized at younger ages~\cite{wang2018cross}. Their ablation study revealed cumulative improvements as $x\sim p_Y$ was bounded to $>$20 years of age, then to $>$30, and up to $>$50. Improved results came with increasing the size of the domain (\ie the respective age considered young, which is orthogonal to those considered old). \figref{chap:fiwmm:fig:cross:generation} depicts samples of parents synthesized for kinship verification. Other augmentation techniques also proved useful, like transforming faces to their basis to then invert, rotate, and change ocular geometry~\cite{dal2015allocentric}.

\subsection{The environment}
The challenges from age variations in \gls{fr} not only intensify in kin-based problems, but also change in novel ways. For instance, let us assume a comparison in the faces of a grandmother and a prospective grandson. The age of each and age gap between the two are subject to variation. In other words, the problem inherits the same challenges of \gls{fr} such that considerations for directed relationships of concern-- the grandmother might be in her early years when the picture was captured, just as the grandson might even be a grandfather himself at the time the picture was taken.

Nurture adds additional challenges to the problem: For instance, a pair of brothers inherited the nose from their mother; one boy experienced a broken nose perhaps more than once; suddenly, that boy no longer has a nose that resembles the mother. Where such challenges exist in conventional \gls{fr}, the relative cost is greater with losing an inherited distinguishable feature from a prospective parent(s) in kin-based problems.  
\begin{figure}
    \centering
    \includegraphics[width=.75\linewidth]{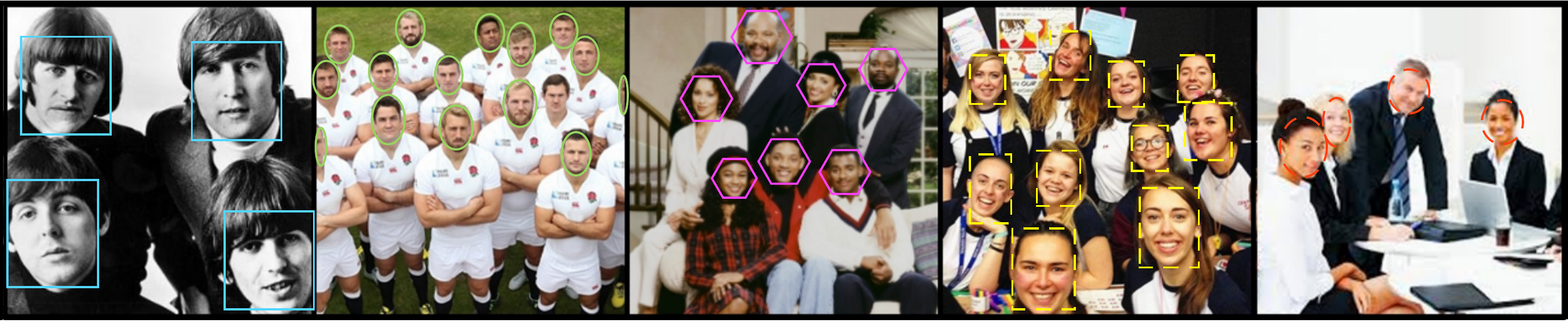}
    \caption{\textbf{\gls{fsp} data (modified from~\cite{dawson2018same}}). Data tagged via constraints, \emph{must} and \emph{cannot}-link: $\approx$1M data points scraped from the web via 125 non-kin queries (\eg \emph{school student}, \emph{sports team}).}
    \label{chap:fiwmm:fig:fspdata}
\end{figure}

Biology-based research has focused on the problem of kinship recognition from a vast array of viewpoints. For instance, work that precedes the work done in machine vision, focused on a human's ability to recognize kinship-- specifically, the ability of younger siblings to better distinguish between \emph{KIN} and \emph{NON-KIN} in strangers~\cite{kaminski2010firstborns} (\ie having seen the first-born their entire lives trains them). An interesting hypothesis indeed, which is supported in the reported experiments (minimal sample set, but typical of human evaluations done in face-based research). Intuitively, the contrary could also be true (\ie the role of the older sibling, watching after their younger sibling would better train for this ability). In any case, the authors propose a theory conditioned on age; difference in age could play a significant role in such a study, as we agree this could be the case for a much older sibling (\ie already developing an ability to discriminate between faces), the same argument of realizing the key differences as a means of recognizing kin in a sibling at a young age could be argued both ways. Furthermore, the authors discarded samples of subjects with no siblings and more than two siblings-- on the one hand the intent to control the experiment with less variation is understandable - on the other, subjects without siblings would serve as a meaningful baseline, while those with a number of siblings only strengthens the case for the oldest being the most keen on recognizing kin (\ie having grown watching over their younger siblings).

\subsection{The data and its distribution}\label{subsec:datachallenges}
Within-family variations are vast. As such, one cannot infer that the inherited traits from one father-son pair would mimic inherited traits of another father-son pair. Furthermore, the factors introducing added complexity vary across different ethnic groups. 

To capture the true data distributions of visual kinship as seen around the world is a great challenge, where many efforts have exhibited exploitable flaws. For instance, using color features claimed \gls{sota} on the \gls{kfw} dataset, as faces of true-relatives often were cropped from the same photos~\cite{lopez2016comments}\cite{wu2016usefulness}. The same motivation ushered in a different paradigm as means to measure unintended data leakage in the unnatural domain inherited by samples being of the same image or different. To say the least - this was a crafty piece of work that acquired an abundance of cheap data by image-level constraints that impose faces in the same photo as \emph{matches}, which means it is a binary problem with classes for the \emph{same} and \emph{different} photo. In other words, by the paired data acquired by finding images with one-to-many faces from the web (\figref{chap:fiwmm:fig:fspdata}), Dawson~\etal proposed training a detector to determine whether a face pair was from the \emph{same} or \emph{different} photo. Then, the boolean class model was directly evaluated on kin-based image sets, with the only difference in the target classes (\ie \emph{same} and \emph{different} assumed to be \emph{KIN} and \emph{NON-KIN}). Thus, showing \gls{sota} ratings on a majority of existing kinship data-- again, hypothesis that public benchmarks were subject to unintended data leakage, and one that is intrinsic to the distribution of classes (\ie \emph{KIN} and \emph{NON-KIN}). In the end, \gls{fsp} proved competitive on KFW-I, KFW-II, Cornell KF, and \gls{tsk}; however, \gls{fsp} lacks sufficient training to perform well on the multi-image \gls{fiw} data (\ie 58.6\%, which was the first, smallest version of the \gls{fiw} dataset). In fact, at the core of \gls{fiw} specifications, as defined in its earliest paper~\cite{FIW}, the concept of same and different photo was one considered in the creation of \gls{fiw}-- mentioned as part of motivation for the data in other recent literature reviews on kin-based image datasets~\cite{dawson2018same}.

\section{Applications}\label{sec:applications}
We next review the use-cases for kinship recognition technology.

\vspace{1mm}
\noindent\textbf{Entertainment and personal knowledge.}
 AncestryDNA claimed $>$15 billion people in its DNA network: their $>$3M paying subscribers (and $>$16M people DNA tested), resulted in the establishment of 100M family trees that form 13B connections across 80 countries.\footnote{\href{https://www.ancestry.com/corporate/about-ancestry/company-facts}{www.ancestry.com/corporate/about-ancestry/company-facts}} As of 2019, Ancestry launched AncestryHealth as a means to infer inheritable health conditions via DNA. Clearly, there is high interest in learning about one's family roots-- which started from curiosity (\ie knowing where one fits, recalling the aforementioned words of Furstenberg~\cite{furstenberg2020kinship}), but now includes learning about one's health from their DNA. Acquiring sufficient data to support DNA and imagery would be difficult. However, provided more reliable kinship recognition capabilities, such technology would certainly enhance popular services such as those provided by billion dollar companies (\eg \href{https://www.ancestry.com/}{ancestry.com}).

\vspace{1mm}
\noindent\textbf{Connect families.}
Identify unknown children being exploited online; reconnect families separated by the modern-day refugee crisis~\cite{mcnatt2018impact}; find unknown relatives, whether directly or indirectly. Statistics show that people want to learn of missing family ties. Furthermore, unfortunate scenarios leave family members desperate to reconnect with lost member(s). Alternatively, law-enforcement could use kinship to solve other high-profile crimes-- the decades long mystery of who the \emph{Golden State Killer} was got solved by using DNA to build his family tree~\cite{goldenstate}.

\vspace{1mm}
\noindent\textbf{Soft attribute as prior knowledge for traditional FR.} Whether it be to enhance \gls{fr} capabilities~\cite{taherkhani2018deep}, to learn to discriminate between hard negatives (\eg brothers), or to narrow the search (\eg \gls{fr} failed to identify bombers of the 2013 Boston Marathon) - but had we known they were brothers, the search space could have been drastically reduced. Hence, kinship provides a powerful cue to help boost existing \gls{fr} systems.

\vspace{1mm}
\noindent\textbf{Nature-based studies.}
With the new millennium came the ability of 3D scans of facial appearances of ten pairs of twins to be compared via landmark features (\ie anteroposterior and vertical facial parameters)~\cite{naini2004three}. About ten years later, this inspired Dehghan~\etal to ask: \emph{Who do I look like?} And then attempt to solve the question using computer vision (\ie gated \gls{ae}~\cite{dehghan2014look}).

\vspace{1mm}
\noindent\textbf{Kin-based face synthesis.} 
An early attempt to predict the appearance of a child from prospective parents was in~\cite{frowd2008predict}. Specifically, Frowd~\etal proposed EvoFit, which used classic shape-based modeling and \emph{eigenfaces} to project a pair of faces via statistical appearance-based modeling. In all fairness, the generative task was heavily influenced by~\cite{cootes2001active}, as many face synthesis tasks were throughout the years, and especially in 2006 the EvoFit came out. In short, EvoFit learned its weights from face samples collected in a tightly controlled setting-- per the requirement that 223 landmarks were precisely marked for all faces. As seminar as EvoFit was in its own right, this early attempt to predict the appearance of children was seemingly ahead of its time, in available machinery (lacking the data-driven, highly complex modeling techniques of today), in resources available to reproducible (\ie no public data released with paper), and in the problem statement itself. In other words, considering EvoFit was proposed before our 2010 timeline means it predated the first benchmark in kinship verification. With that, we believe the small impact of this work was due to its timing and, in return, the lack of complete support for the problem, so if others did want to partake they too would have to collect data. Meaning, it was impossible to reproduce results directly. Regardless how minimal the impact was in citations and usage of other researchers, the work certainly showed promise considering the results were from a minimally-sized data pool. Thus, had a widely used benchmark been practiced, or provided the data constraints were handled (\ie inability to generalize + inability for others to reproduce), then EvoFit could have attracted much more attention. Perhaps, our 2010 time-line would have had to start a few years prior. Nonetheless, this is only speculation and, therefore, we can only hypothesize the \emph{what ifs} after the fact.

\section{Discussion}
After we surveyed in~\cite{robinsonKinsurvey2020}, it was clear that a decade of research in visual kinship recognition resulted in an increased interest with an increase in data resources that were available. Clearly, the problems alone are challenging, even when compared to other machine-vision tasks (\eg conventional \gls{fr}). Furthermore, the task of designing, collecting, and annotating labels is exceptionally difficult for kin-based problems. Thus, as contributions in data are proposed, interest seems to spike in response. With the release of the large-scale \gls{fiw} dataset, for the first time, a data resource attempts to closely mimic data distributions of families around the globe. Moreover, \gls{fiw} provides the data needed for the modern day, deep learning models. \gls{fiw}, having had many existing datasets to learn from, remains the largest and most dynamic. However, the release of \gls{fiw} was only the beginning, as efforts were then spent on annual challenges (\ie four consecutive years, 2017-2020, and also a Kaggle competition). With the resource and incentive provided by challenges, motivation for researchers to engage is ever so high and thus, we present this survey-  not only as a means to realize the aspects that have been effective and vice versa-  but also for a solid foundation for the next decade to build upon well-defined protocols and problem statements, each supported with source code, enabling even a wider audience to get started and contributing to the problem.

The deep learning revolution has only begun for visual kinship recognition - how to embed, how to fuse, how to interrupt - how do experts across disciplines engage by leveraging for a deeper understanding in inheritance from a strictly scientific point-of-view (\ie anthropology)! Hence, if we can devise the right tools for the right scholars synergy is bound to reveal insights in the nature of faces within families. Considering the many benchmarks that have a lot of room for improvement, along with the many social and relational data mining that is made possible with soft-attribute labels such as those in \gls{fiw}, it is an exciting time for junior, senior, and practical researchers to reap benefits alongside its place with pure business, product, and patent design.


\chapter{Bias in Face Recognition}\label{chap:bias}
\glsresetall
\section{Overview}
\label{chap:bias:sec:introduction}
The more our society becomes integrated with \gls{ml}, the higher the interest in topics such as bias, fairness, and even the implications for the underlying formalization of existing or prospective \gls{ml} standards~\cite{10.1007/978-3-030-13469-3_68, anne2018women, wang2018racial}. Thus, an effect of vast companies growing more dependent on \gls{ml} is an ever-increasing concern about the biased and unfair algorithms, \eg untrustworthy and prejudiced \gls{fr}~\cite{nagpal2019deep,snow2018}.

A common trend in both research and mainstream has grown clear: the more we depend on technology that accelerates or automates everyday tasks, the more attention concepts such as biased and unfair algorithms should receive~\cite{lazo2020towards}. Furthermore, systems deployed for sensitive tasks, like biometrics~\cite{drozdowski2020demographic} (\eg \gls{fr}), need to be fully considered and understood. The perspective here recognizes the tendency of the researcher, the reporter, and the consumer to maintain transparency. Nowadays, in \gls{fr} \glspl{cnn} are trained on a large number of faces identified by a detection system (\chapref{chap:facedetection}). Recall that the goal is to encode faces in an N-dimensional space that pulls together samples of the same identity and pushes those that are different further apart. So, \glspl{cnn} are trained to encode faces from the face (\ie image-space) to encoded representation (\ie feature-space). Face images are mapped to feature vectors and evaluated via a similarity scoring function, and the pairs with a similarity score above the decision threshold are assumed \emph{genuine} and all others are classified as \emph{imposters}.

\begin{figure}
    \centering
    \includegraphics[width=.55\linewidth]{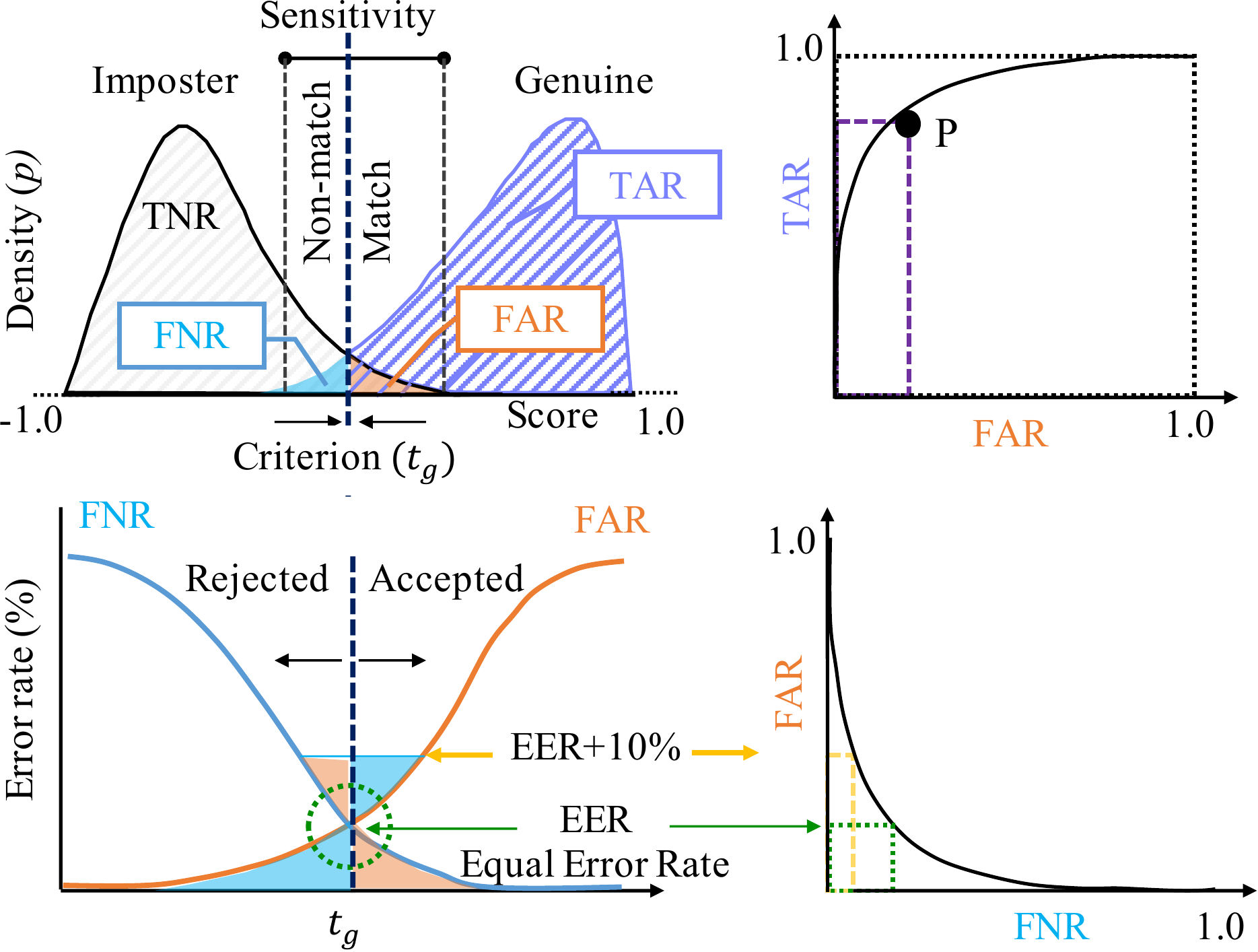}
    \caption{\textbf{Depiction of the biometrics.} The SDM shows the sensitivity related to a single threshold $t_g$ (\emph{top-left}). The area to the right of the threshold considers all accepted pairs, both correctly and incorrectly predicted. \gls{tar} as a function of \gls{far} is a common way to report ratings for given false-rates (\emph{top-right}). Equally common in FR is the trade-off between \gls{fnr} and \gls{far} (\emph{bottom-left} and \emph{bottom-right}).}
    \label{chap:bias:fig:metrics}
\end{figure}



Typically, a fixed threshold sets the decision boundary by which to compare scores (\figref{chap:bias:fig:metrics}). As such, features of the same identity must satisfy a criterion based on a single value~\cite{wang2018cosface, liu2017sphereface, deng2019arcface, wang2018additive}. However, we found that an individual (\ie global) threshold-- a crude measure that leads to skewed results across demographics and other attributes-- are determined using a held-out set. In other words, the threshold for the \gls{fr} \emph{Matching} module is set according to the desired target: in research the threshold that yields that highest accuracy on the validation set; in practice the threshold is determined by the value that yields the desired rate, which depends on the use-case. For example, in the use-case where \gls{fr} is used to enable entry via access control may have a smaller threshold that will realize fewer \emph{genuine} samples than \gls{fr} used for tagging photos per software recommendation-- the falsely accepted instances of the latter use-case will have minimal, if any, negative effects. An important note to consider for the concept of having a held-out set is that it typically shares the same distribution with the test set, meaning it favors the same demographics as the held-out set had. That skew (\ie the difference in the performance of an algorithm of particular demographics) is our definition of bias. A key question is: \emph{is \gls{fr} too biased, or not?}

\begin{table}[!t]
    \centering
    \caption{\textbf{Database stats and nomenclature.} \textit{Header:} Subgroup definitions. \textit{Top:} Statistics of \gls{bfw}. \textit{Bottom:} Number of pairs for each partition. Columns grouped by ethnicity and then further split by gender.}\label{chap:bias:tab:ethnic-splits}
     \resizebox{\textwidth}{!}{%
    \begin{tabular}{r c c c c c c c c l}
        \toprule
        & \multicolumn{2}{c}{Asian (A)} & \multicolumn{2}{c}{Black (B)}  & \multicolumn{2}{c}{Indian (I)}& \multicolumn{2}{c}{White (W)}\\
        \cmidrule(l){2-3} \cmidrule(l){4-5} \cmidrule(l){6-7}\cmidrule(l){8-9} 
         & Female (AF) & Male (AM) & BF & BM& IF & IM & WF & WM&Aggregated\\ 
        \midrule

       \# Faces  &  2,500&  2,500& 2,500 & 2,500& 2,500 & 2,500 & 2,500 & 2,500 &20,000 \\ 
        \# Subjects & 100& 100& 100  & 100  & 100  & 100& 100 &100&800  \\ 
        \# Faces / Subject  & 25 & 25    & 25 & 25 & 25  & 25  &  25 & 25 & 25\\ 
\specialrule{.01em}{.05em}{.05em}
            \# Positive Pairs &  30,000&  30,000& 30,000 &30,000 & 30,000 &30,000&30,000 & 30,000 &240,000 \\ 
        \# Negative Pairs & 85,135&  85,232& 85,016  & 85,141  & 85,287  & 85,152& 85,223 &85,193&681,379  \\ 

        \# Pairs (Total) & 115,135 & 115,232    &115016 &115,141 & 115287  & 115,152  &  115,223& 115193 & 921,379\\ 
        \bottomrule
    \end{tabular}
    }

\end{table}


Now, provided two or more faces features encoded by a \gls{cnn}, a distance (or similarity score) $s$ must be learned such to act as a decision boundary to separate the genuine pairs score from the imposters score. Ideally, genuine and imposter scores would be completely separable. However, this is not the case in practice. It is this score-threshold (\ie $\theta$) that determines whether or not the pair should be accepted. The implications are for faces features: to be assumed as the same, genuine class the score (or distance) must satisfy a criterion in the form of a single value~\cite{wang2018cosface, deng2019arcface, liu2017sphereface, wang2018additive}.  Mathematically, the decision $D$ in similarity space is defined as

\begin{equation*}
D=\begin{cases}
          \text{accept} \quad &\text{if} \, s \ge \theta \\
          \text{reject} \quad &\text{if} \, s < \theta \\
     \end{cases}.
\end{equation*}

The importance of $\theta$ should not be overlooked - a hyper-parameter that is a decision boundary in metric space. The optimal value depends on the specific use-case (\ie larger thresholds yield a lower probability that a sample is predicted as a true match). Regardless, the choice in threshold has a clear trade-off between \gls{fp} and \gls{fn} rates. For instance, a system that is claimed to perform at an error rate of 1 and 10,000, \ie one in every ten-thousand instances are incorrectly matched. We would then set our system by determining the threshold based on held-out data samples that allow the desired target error rate to be matched (\figref{chap:bias:fig:metrics}). The problem, per convention, is which assumes an average result across a held-out that then only holds true on the distribution of the source used. Then, specific cohorts (\eg ethnicity, gender, and other demographics) are unequally weighted due to an unequal representation. So, that single, global threshold, which is a sort of crude measure to begin with, is skewed to different cohorts that are not fairly represented by the source. In the end, these systems favor certain demographics, and it is the bias for which a change in cohort causes a change in the average performance of an algorithm.

Making matters more challenging, precise definition of race and ethnicity vary from source-to-source. For example, the US Census Bureau allows an individual to self-identify race.\footnote{\scriptsize\href{https://www.census.gov/mso/www/training/pdf/race-ethnicity-onepager.pdf}{www.census.gov/mso/www/training/pdf/race-ethnicity-onepager}} Even gender, our attempt to encapsulate the complexities of the sex of a human as one of two labels. Others have addressed the oversimplified class labels by representing gender as a continuous value between 0 and 1 - rarely is a person entirely \emph{M} or \emph{F}, but most are somewhere in between~\cite{merler2019diversity}. For this work, we define subgroups as specific sub-populations with face characteristics similar to others in a region. Specifically, we focus on 8 subgroups (\figref{chap:bias:fig:avg-faces}).

The adverse effects of a global threshold are two-fold: \textbf{(1)} mappings produced by \glspl{cnn} are nonuniform. Therefore, distances between pairs of faces in different demographics vary in distribution of similarity scores (Fig~\ref{chap:bias:fig:detection-model}); \textbf{(2)} evaluation set is imbalanced. Subgroups that make up a majority of the population will carry most weight on the reported performance ratings. Reported results favor the common traits over the underrepresented. Demographics like gender, ethnicity, race, and age are underrepresented in most public datasets~\cite{merler2019diversity, wang2018racial}.

For \textbf{(1)}, we propose subgroup-specific (\ie optimal) thresholds while addressing \textbf{(2)} with a new benchmark dataset to measure bias in \gls{fr}, \gls{bfw} (\tabref{chap:bias:tab:bfw:counts} and~\ref{chap:bias:tab:bfw:attributes}). \gls{bfw} serves as a proxy for fair evaluations for \gls{fr} while enabling per subgroup ratings to be reported. We use \gls{bfw} to gain an understanding of the extent to which bias is present in \gls{soa} \gls{cnn}s used \gls{fr}. Then, we suggest a mechanism to mitigate problems of bias with more balanced performance ratings for different demographics. Specifically, we propose using an adaptive threshold that varies depending on the characteristics of detected facial attributes (\ie gender and ethnicity). We show an increase in accuracy with a balanced performance for different subgroups. Similarly, we show a positive effect of adjusting the similarity threshold based on the facial features of matched faces. Thus, selective use of similarity thresholds in current \gls{soa} \gls{fr} systems provides more intuition in \gls{fr} research with a method easy to adopt in practice.

The contributions of this work are three-fold. (1) We built a balanced dataset as a proxy to measure verification performance per subgroup for studies of bias in \gls{fr}. (2) We revealed an unwanted bias in scores of face pairs - a bias that causes ratings to skew across demographics. For this, we showed that an adaptive threshold per subgroup balances performance (\ie the typical use of a global threshold unfavorable, which we address via optimal thresholds). (3) We surveyed humans to demonstrate bias in human perception (NIH-certified, \textit{Protect Humans in Research}).

The adverse effects of a global threshold are three-fold: \textbf{(1)} the evaluation set is typically imbalanced. The demographics of the majority are weighted more in the reported performance ratings. Therefore, reported results skew to rarer traits that are more common in the underrepresented subgroups-- a phenomena that should be considered for different subgroups (\ie gender, ethnicity, race, age). \textbf{(2)} the mappings produced by a \gls{cnn} have various levels of sensitivity in the metric (\figref{chap:bias:fig:detection-model}). Therefore, the range of distances between true pairs varies across demographics. (\textbf{3}) a global threshold is referenced when comparing face encodings. Since the optimal score shifts for different demographics, there ought to be variable thresholds (\ie sliding threshold set according to demographic information). Furthermore, the validation set used to determine global threshold and the test set used to report results should be understood - this issue will be unnoticed in performance ratings if the validation and test are from the same distribution and, thus, the resulting performance ratings are based in favor of the majority. This leads to performance ratings that are incomplete and even misleading.

To address the lack of a balanced data (\ie \textbf{(1)}), we propose to evaluate on our dataset, which was built specifically for measuring biases in demographics for \gls{fv} systems in a systematic, reproducible way. With it, we introduce a new benchmark for \gls{fr}, called \gls{bfw} (\figref{chap:bias:fig:face-montage}). \gls{bfw} serves as a platform to fairly evaluate \gls{fr} systems and enable demographic-specific ratings to be reported (\tabref{chap:bias:tab:bfw:attributes}). We use \gls{bfw} to gain a deeper understanding of the extent of bias present in facial embeddings extracted from a \gls{soa} \gls{cnn} model. We then suggest a mechanism to counter the biased feature space to mitigate problems of bias with more balanced performance ratings across demographics, and all the while improving the overall accuracy. Specifically, we unlearn demographic knowledge in face encodings, while preserving identity information. Thus, we learn to map the encodings to a lower dimensional space containing less knowledge of subgroups. The byproducts are then embeddings that preserve the privacy of its subject's ethnicity and gender (\ie subgroups). It is this feature adaptation scheme proposed to address items (\textbf{2}-\textbf{3}).

Our contributions in topics of bias (\ie this~\cite{robinson2020face}) are the following:
\begin{itemize}
    \item Demonstrate a bias in an existing \gls{soa} \gls{cnn} with our \gls{bfw} dataset (\tabref{chap:bias:tab:bfw:counts}). We propose a feature learning scheme that employs domain adaptation to debias face encodings and, most importantly, balances performances across subgroups such to boost the overall performance.
    \item Hide attribute information in encodings-- a byproduct of the proposed debiasing scheme is the reducing knowledge of attributes. Beyond privacy, it removes other potential biases, whether unintended (\eg models trained on top) or intended bias (\eg human consciously using).
    \item Provide insights with analysis of hard samples overcome by the proposed debiasing scheme. Evidence in the form of salience mapping and face pairs are shown and discussed.
    \item Develop code-base as public Git Hub (\ie \href{https://github.com/visionjo/facerec-bias-bfw}{https://github.com/visionjo/facerec-bias-bfw}); provided form (\ie link \href{https://forms.gle/3HDBikmz36i9DnFf7}{https://forms.gle/3HDBikmz36i9DnFf7}) for dataset download requests, where paired data and related resources used in \emph{facerec-bias-bfw} repo are available.
\end{itemize}

\subsection{Organization}
The rest of the chapter is organized as follows. In~\secref{chap:bias:sec:relatedworks}, we review work related to bias in \gls{fr}, along with works related to the problem and solution spaces. We then cover our \gls{bfw} dataset-- the motivation, specifications, and described (\secref{cbap:bias:sec:bfw}). Then, in~\secref{chap:bias:sec:proposed}, we introduce the proposed methodology.~\secref{chap:bias:sec:experimental} follows this with settings and results of the experiments, along with the details of our \gls{bfw} database. Finally, we conclude and discuss next steps in~\secref{chap:bias:sec:conclusions}.

\section{Related Work}\label{chap:bias:sec:relatedworks}
We next review the research related to bias and privacy in \gls{fr}, both for humans and machines, and along with some background information required to understand the motivation and overall solution of the proposed model. Specifically, we first briefly discuss problems of bias in general \gls{ml}, then that which is specific to \gls{fr}. Following this, we support our hypothesis that human too possess a similar bias (\ie more familiar to those of subgroup most frequently seen in the past). Then, we describe several works in domain adaptation-- the domain for which our proposed solution is best characterized. Finally, we cover problems of privacy in \gls{fr}.

\subsection{Bias in machine learning}
    The progress and commercial value of \gls{ml} are exciting. However, due to inherent biases in ML, society is not readily able to trust completely in its widespread use. The exact definitions and implications of bias vary between sources, as do its causes and types. A common theme is that bias hinders performance ratings in ways that skew to a particular sub-population. In essence, the source varies, whether from humans~\cite{windmann1998subconscious}, data or label types~\cite{tommasi2017deeper}, \gls{ml} models~\cite{amini2019uncovering, kim2019learning}, or evaluation protocols~\cite{stock2018convnets}. For instance, a vehicle-detection model might miss cars if training data were mostly trucks. In practice, many \gls{ml} systems learn biased data, which could be detrimental to society.

\subsection{Bias in facial recognition}
Biases in \gls{fr} focus on characterizing performance across various \emph{soft attributes}, such as gender, ethnicity, or age~\cite{drozdowski2020demographic}. Researchers have spent great efforts proposing problem statements and solutions to problems of bias in \gls{fr} technology. We focus on the demographics (or subgroups) of gender and ethnicity. The inherent problems here are two-fold. First, gender is handled as a boolean label, which is a gross approximation of individual uniqueness in question of sexuality - a spectrum of real numbers would be more appropriate~\cite{merler2019diversity}. Secondly, the definitions of race and ethnicity are loosely defined. The US Census Bureau allows an individual to self-identify race.\footnote{\href{https://www.census.gov/mso/www/training/pdf/race-ethnicity-onepager.pdf}{https://www.census.gov}} We define it as a group of people having facial characteristics similar to those found in a region. The result is various types of biases in FR systems in favor of or against particular demographics remain a question. 
    
Balakrishnan~\etal trained a generator to manipulate latent space features in that the controlled attributes were skin-tone, length of hair, and hair color\cite{balakrishnan2020towards}. Another synthesis solution proposed was to generate faces across various ages as a means to augment training data~\cite{georgopoulos2020enhancing}. Terhorst~\etal recognized the same phenomena our work is found on-- the variation in sensitivities of scores for different demographics~\cite{terhorst2020post}. Specifically, the authors propose a score normalization scheme to handle the problem of inaccurate performance ratings when demographic-specific performances are compared to the average-- a problem highlighted in paper~\cite{robinson2020face}.

Some aim to characterize the amount of bias in a system, whether it be for gender~\cite{serna2020insidebias, albiero2020analysis, das2018mitigating}, ethnicity, age~\cite{srinivas2019face}, or two or more of the aforementioned~\cite{nagpal2019deep,  acien2018measuring, gong2019debface, savchenko2019efficient, Nagpal_2020_CVPR_Workshops}. A recent \emph{European Conference on Computer Vision} (ECCV) challenge provided incentive and for researchers to propose solutions for problems of bias with respect to ethnicity, gender, age, pose, and with and without sunglasses~\cite{sixta2020fairface}. Other recent works in \gls{fr} technology introduce additional modalities, such as profile information, to the problem of bias~\cite{pena2020bias, pena2020faircvtest}. Another research question concerns the measuring of biases in \gls{fr} systems~\cite{acien2018measuring, serna2020insidebias}. Some focus on templates~\cite{8987331}. Some debias at the score level~\cite{terhorst2020post}. Some focus to debias pre-trained models~\cite{Sadeghi_2020_CVPR_Workshops}. Wang~\etal introduced a reinforcement learning-based race balance network (RL-RBN) to find optimal margins for non-Caucasians as a Markov decision process before passing to the deep model that learns policies for an agent to select margins as an
approximation of the Q-value function; \ie the skewness of feature scatter between races can be reduced~\cite{wang2020mitigating}. Even more HCI-based views have been introduced as semi-supervised bias detection systems that act as tools with humans in-the-loop~\cite{law2020designing}. 

Yin~\etal proposed to augment the feature space of underrepresented classes using different classes with a diverse collection of samples~\cite{yin2019feature}. This was to encourage distributions of underrepresented classes to resemble the others more closely. Similarly, others formulated the imbalanced class problem as one-shot learning, where a \gls{gan} was trained to generate face features to augment classes with fewer samples~\cite{ding2018one}. \gls{gapf} was proposed to create fair representations of the data in a quantifiable way, allowing for the finding of a de-correlation scheme from the data without access to its statistics~\cite{huang2018generative}. Wang~\etal defined subgroups at a finer level (\ie Chinese, Japanese, Korean), and determined the familiarity of faces inter-subgroup~\cite{wang2018they}. Genders have also been used to make subgroups (\eg for analysis of gender-based face encodings~\cite{muthukumar2019}). Most recently,~\cite{wang2018racial} proposed to adapt domains to bridge the bias gap by knowledge transfer, which was supported by a novel data collection, \gls{rfw}. The release of \gls{rfw} occurred after \gls{bfw} was built - although similar in terms of demographics, \gls{rfw} uses faces from MSCeleb~\cite{guo2016ms} for testing, and assumes CASIA-Face~\cite{yi2014learning} and VGG2~\cite{Cao18} were used to train. In contrast, our \gls{bfw} assumes VGG2 as the test set. Furthermore, \gls{bfw} balance subgroups: \gls{rfw} splits subgroups by gender and race, while \gls{bfw} has gender, race, or both). 

\begin{figure}[t!]
    \centering
    \includegraphics[width=.5\linewidth]{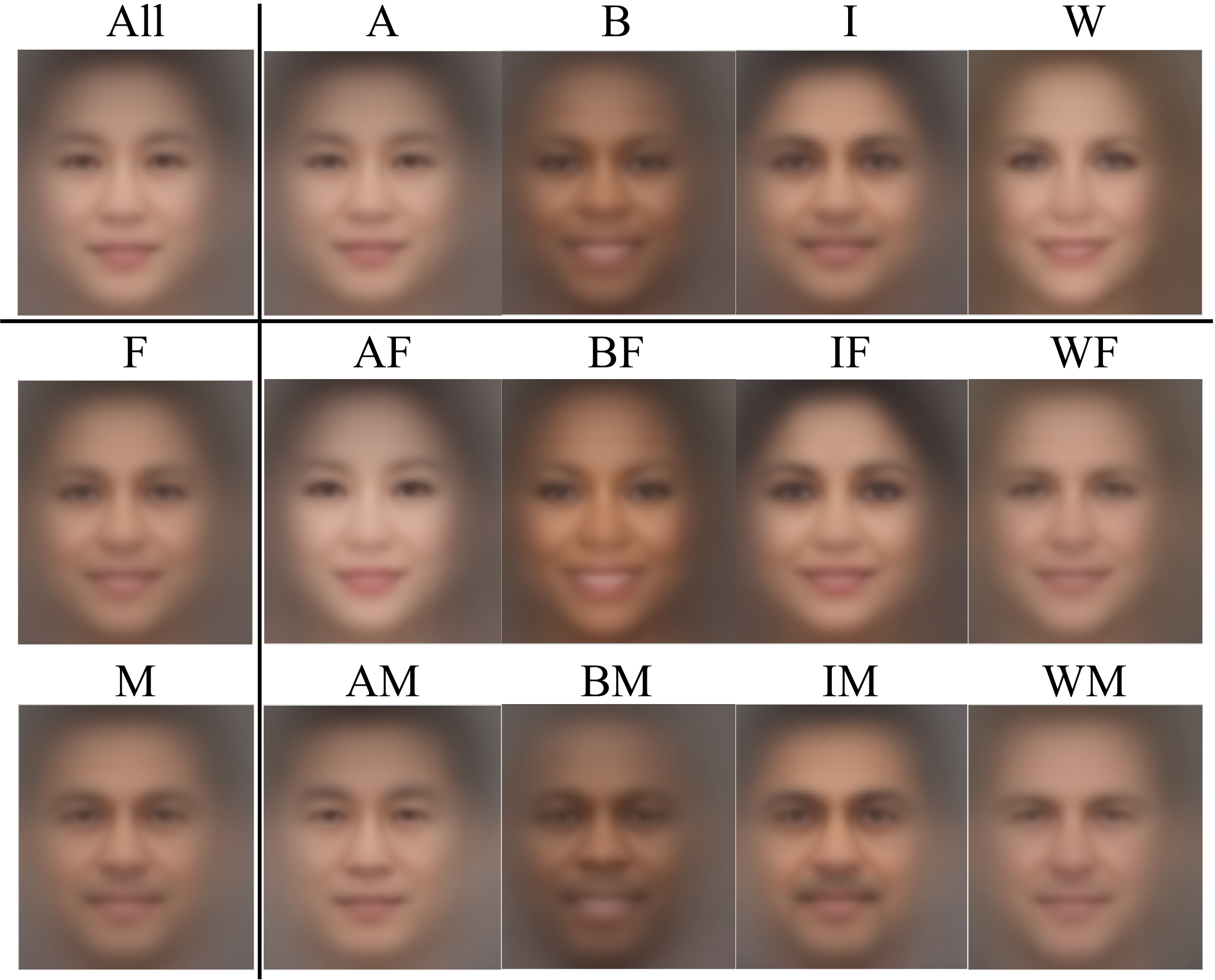}
    \caption{\textbf{BFW dataset.} Average face per subgroup: \emph{top-left}: the entire \gls{bfw}; \emph{top-row} per ethnicity;  \emph{left-column}: per gender. The others represent the ethnicity and gender, respectively. \tabref{chap:bias:tab:ethnic-splits} defines the acronyms of subgroups.}
    \label{chap:bias:fig:avg-faces}
\end{figure}

Most similar to us is~\cite{srinivas2019face, das2018, demogPairs, lopez2019dataset} - each was motivated by insufficient paired data for studying bias in \gls{fr}. Then, problems were addressed using labeled data from existing image collections. Uniquely, Hupont~\etal curated a set of faces based on racial demographics (\ie \gls{a}, \gls{b}, and \gls{w}) called \gls{dp}~\cite{demogPairs}. In contrast,~\cite{srinivas2019face} honed in on adults versus children called \gls{itwcc}. Like the proposed \gls{bfw}, both were built by sampling existing databases, but with the addition of tags for the respective subgroups of interest. Aside from the additional data of \gls{bfw} (\ie added subgroup \gls{i}, along with other subjects with more faces for all subgroups), we also further split subgroups by gender. Furthermore, we focus on the problem of facial verification and the different levels of sensitivity in cosine similarity scores per subgroup.

\subsection{Human bias in machine learning}
Bias is not unique to \gls{ml} - humans are also susceptible to a perceived bias. Biases exist across race, gender, and even age~\cite{10.1007/978-3-030-13469-3_68, bar2006, meissner2001, nicholls2018}. Wang~\etal showed machines surpass human performance in discriminating between Japanese, Chinese, or Korean faces by nearly 150\%~\cite{wang2018they}, as humans just pass random (\ie 38.89\% accuracy).

We expect the human bias to skew to their genders and races. For this, we measure human perception with face pairs of different subgroups (Section~\ref{subsec:human-assessment}). The results concur with~\cite{wang2018they}, as we also recorded overall averages below random ($<$50\%).

\glsunset{if}\glsunset{im}\glsunset{af}\glsunset{am}\glsunset{bf}\glsunset{bm}\glsunset{wf}\glsunset{wm}

\begin{figure}[t!] 
\glsunset{fpr}
\centering
\includegraphics[width=.9\linewidth]{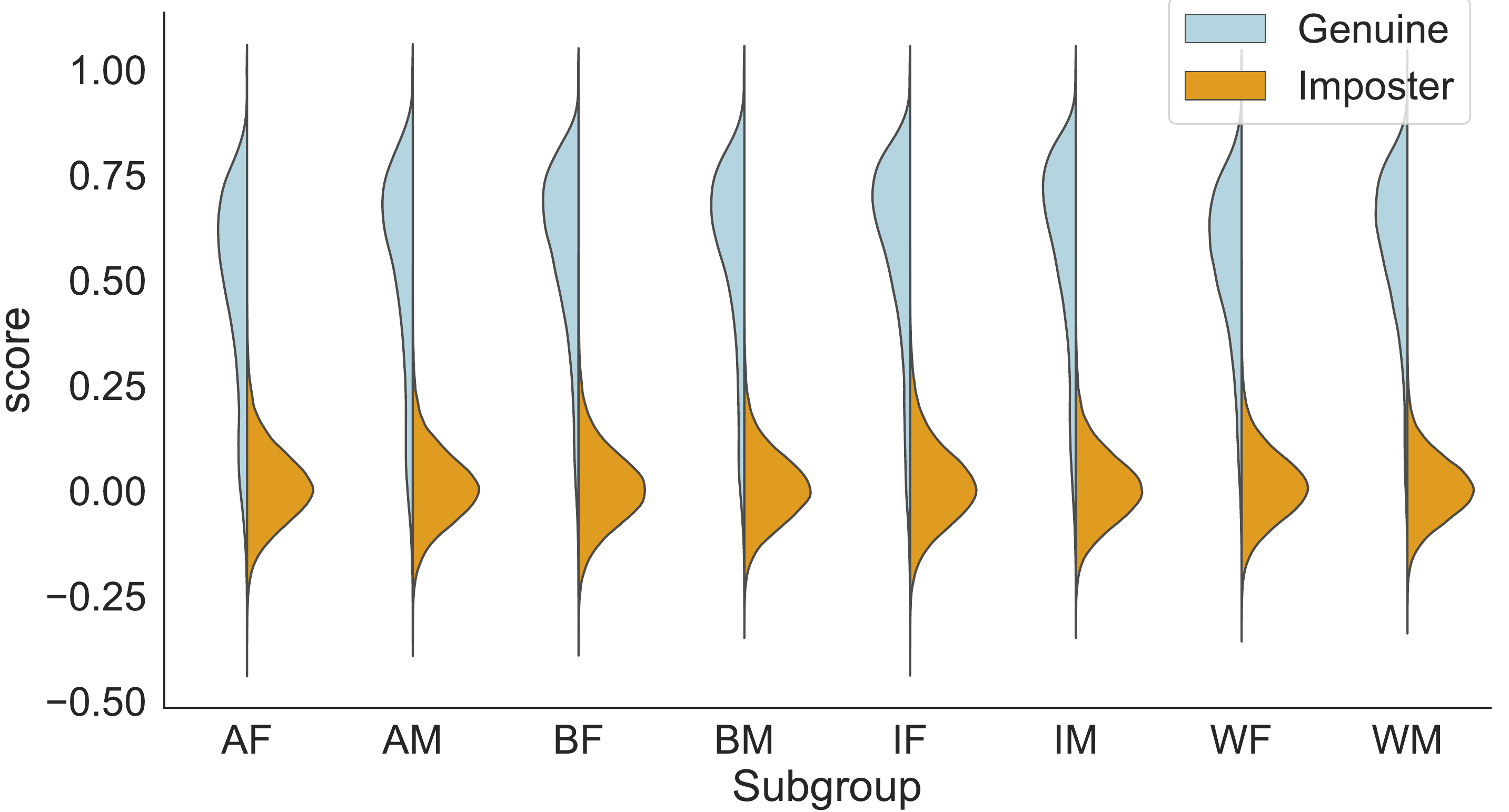}
	\caption{\textbf{\Gls{sdm} across subgroups.} Scores of \emph{imposters} have medians $\approx$0.3 but with variations in upper percentiles; \emph{genuine} pairs vary in mean and spread (\eg AF has more area of overlap). A threshold varying across different subgroups yields a constant \gls{far}.} \label{chap:bias:fig:detection-model} 
\end{figure}

\subsection{Imbalanced data and data problems in FR}

The impact from the quality of fairness depends on the context for which it is used. Furthermore, various paradigms have been proposed as means to a solution: some alter the data distribution to yield classifiers of equal performances for all classes (\ie re-sampling, like by under-sampling and over-sampling~\cite{drummond2003c4}); others alter the data itself (\ie algorithms that adjust classification costs). For instance, Oquab~\etal re-sampled at the image patch-level~\cite{oquab2014learning}. Specifically, the aim was to balance foregrounds and backgrounds for object recognition. On the other hand, Rudd~\etal proposed the mixed objective optimization network (MOON) architecture~\cite{rudd2016moon} that learns to classify attributes of faces by treating the problem as a multi-task (\ie a task per attribute) attribute to more balanced performances when training on data that has an imbalanced distribution across attributes. The Cluster-based Large Margin Local Embedding (CLMLE)~\cite{huang2019deep} sampled clusters of samples in the feature-space that were used to regularize the models at the decision boundaries of underrepresented classes. See literature reviews for details on approaches that alter at the data or algorithmic-level~\cite{he2009learning,he2013imbalanced, krawczyk2016learning}. 

More specific to faces, Drozdowski~\etal summarizes that the cohorts of concern in biometrics are demographic (\eg sex, age, and race), person-specific (\eg pose or expression~\cite{xu2020investigating}, and accessories like
eye-wear or makeup), and environmental (\eg camera-model, sensor size,  illumination,
occlusion)~\cite{drozdowski2020demographic}. Albiero~\etal found empirical support that having training data that is well balanced in gender does not mean that results of a gender-balanced test set will be balanced~\cite{albiero2020does}. Our studies focus on the effect demographic has in \gls{fv} by assessing demographic-specific classification ratings. Our \gls{bfw} data resource allows us to analyze existing \gls{soa} deep \glspl{cnn} on different demographics (or subgroups). We provide practical insight: FR benchmarks often report with misleading ratings-- ratings are dependent on the demographics of the population.

To match the capacity of modern-day deep models several large \gls{fr} datasets were released~\cite{guo2016ms, schroff2015facenet, Cao18, maze2018iarpa}. More recently, several have reported on the imbalance in demographics are the data, and proposed balanced resources for \gls{fr}-based tasks~\cite{hupont2019demogpairs, wang2018racial, karkkainen2019fairface, merler2019diversity}. Diversity in Faces (D\emph{i}F) came first~\cite{merler2019diversity}, which did not include identity labels. Moreover, D\emph{i}F is no longer available for download. Others released data with demographics balanced, but for the task of predicting the demographic and, thus, do not include identity labels~\cite{wang2018racial, karkkainen2019fairface}. Hupont~\etal proposed DemogPairs is balanced across 6 subgroups of 600 identities from CASIA-WebFace (CASIA-W)~\cite{yi2014learning}, VGG~\cite{schroff2015facenet}, and VGG2~\cite{Cao18}. Similar to DemogPairs, except with 8 subgroups, 800 identities, and with number of faces per identity the same for all, our \gls{bfw} data was the latest release for measuring bias in \gls{fr} technology. Furthermore, recognizing that existing \gls{soa} models are already trained on a public resource, we built \gls{bfw} from just VGG2 to minimize conflicts in overlap between train and test. \tabref{chap:bias:tab:bfw:attributes} compares our data with the others.

\subsection{Domain adaptation and feature alignment}
Domain adaptation (DA)~\cite{ding2018graph,peng2017visda,saito2019semi} employs labeled data from the source domain to make it generalize well to the typically label-scarce target domain; hence, a common solution to relieve annotation costs. DA can be roughly classified as the semi-supervised DA~\cite{saito2019semi} or the unsupervised one~\cite{shu2018dirt} according to the access to target labels. The crucial challenge toward DA is the distribution shift of features across domains (\ie domain gap), which violates the distribution-sharing assumption of conventional machine learning problems. In our case, the different domains are the different subgroups (\tabref{chap:bias:tab:bfw:counts}).

To bridge such gap, some of feature alignment (FA) methods attempt to project the raw data into a shared subspace where certain feature divergence or distance is minimized to confuse them. Various methods, such as Correlation Alignment (CORAL)~\cite{sun2015subspace}, Maximum Mean Discrepancy (MMD)~\cite{long2013transfer}, and Geodesic Flow Kernel (GFK)~\cite{gong2012geodesic,gopalan2011domain}, have been developed in this line. Currently, adversarial domain alignment methods (\ie DANN~\cite{ganin2016domain}, ADDA~\cite{tzeng2017adversarial}) have attracted increasing attention by designing a zero-sum game between a domain classifier (\ie discriminator) and a feature generator. The features of different domains will be mixed if the discriminator cannot differentiate the source and target features. More recently, learning well-clustered target features proved to be helpful in conditional distributions alignment. Both DIRT-T~\cite{shu2018dirt} and MME~\cite{saito2019semi} methods applied entropy loss on target features to implicitly group them as multiple clusters in the feature space to keep the discriminative structures through adaptation. Inspired by FA, we aim to align the score distributions of different subgroups by adjusting score sensitivities (\figref{chap:bias:fig:detection-model}).

\subsection{Protecting demographic information in \gls{fr}}
\gls{ml} is growing more accessible. As such, it grows more in our day-to-day lives. The levels of sensitivities of the use-case ought to be put under careful consideration - not only the model, but the data too~\cite{guler2019privacy}. That means, intermediate results, also known as features are included. \gls{fr} is an \gls{ml} problem that has made great progress in recent years, being used \emph{off-the-shelf} in many applications. It is time to carefully consider data privacy concerns - with priority on topics of biometrics. Provided careless or too little action is spent to protect the user behind the face image, the more the chance that data may be used by an adversary maliciously~\cite{bowyer2004face}.

Several have recently attempted to solve problems of bias in demographic-based classifiers (\eg ethnicity and gender classifiers). Furthermore, attempts to disguise demographic information in facial encodings while preserving \gls{fv} abilities have been proposed for privacy and protection purposes~\cite{dhar2020adversarial, 8698605}. In other words, prior works recognized the importance of preserving the identity information in facial features, while ridding it of evidence of demographics. Our model inherently does this as part of the objective aims for an inability to recognize subgroups.

The aforementioned assume the facial encoding are accessible - this makes sense in terms of reduced computations (\ie no need to encode each time) and storage (\ie encodings are smaller representations of the image). However, several works aimed to hide attribute information in image space; for instance, Othman~\etal learned to morph faces to suppress gender information in the image-space while preserving the identification~\cite{10.1007/978-3-319-16181-5_52}. Guo~\etal proposed a mapping from image-to-noise, both encrypting the image such that the encoder still decodes the identity but without the ability to determine gender by machine or human~\cite{GUO2019320}. Ma~\etal viewed the communication between servers as the point of concern for privacy-  a lightweight privacy preserving adaptive boosting (AdaBoost) \gls{fr} framework (POR) based on additive fusion for secret sharing to encrypt model parameters, while using a cascade of classifiers to address different protocols~\cite{ma2019lightweight}.

\section{Balanced Faces In the Wild (BFW)}\label{cbap:bias:sec:bfw}
\begin{figure}[t!]
    \centering\glsunsetall
    \begin{subfigure}[t]{.2\linewidth}
        \centering
        \includegraphics[width=\linewidth]{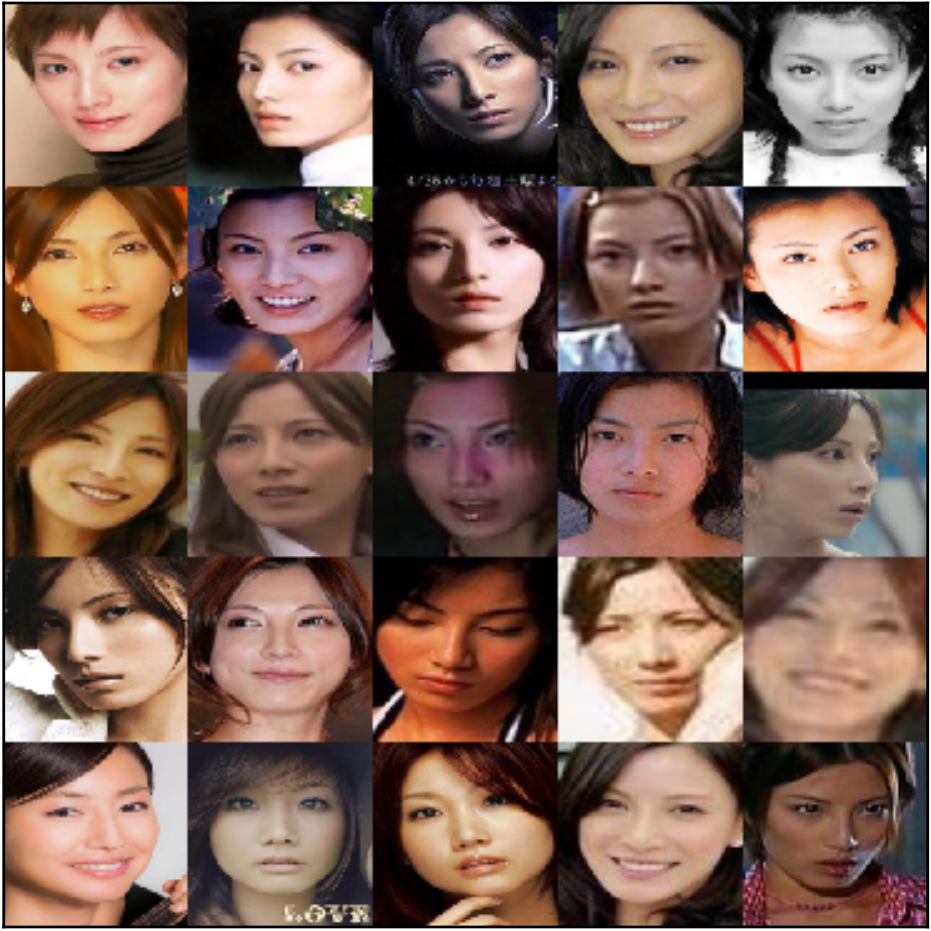}
        \caption{\gls{af}.}
        \label{subfig:af}
    \end{subfigure}%
    \begin{subfigure}[t]{.2\linewidth}
        \centering
        \includegraphics[width=\linewidth]{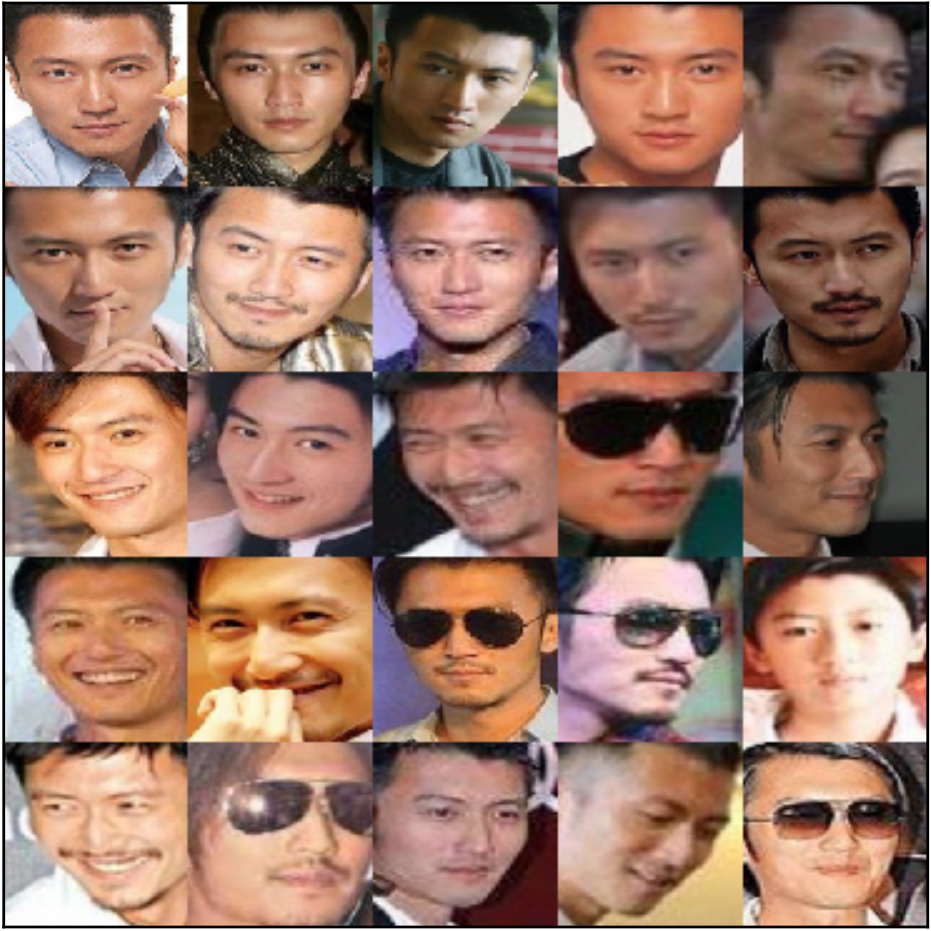}
        \caption{\gls{am}.}
         \label{subfig:am}
    \end{subfigure}
    \begin{subfigure}[t]{.2\linewidth}
        \centering
        \includegraphics[width=\linewidth]{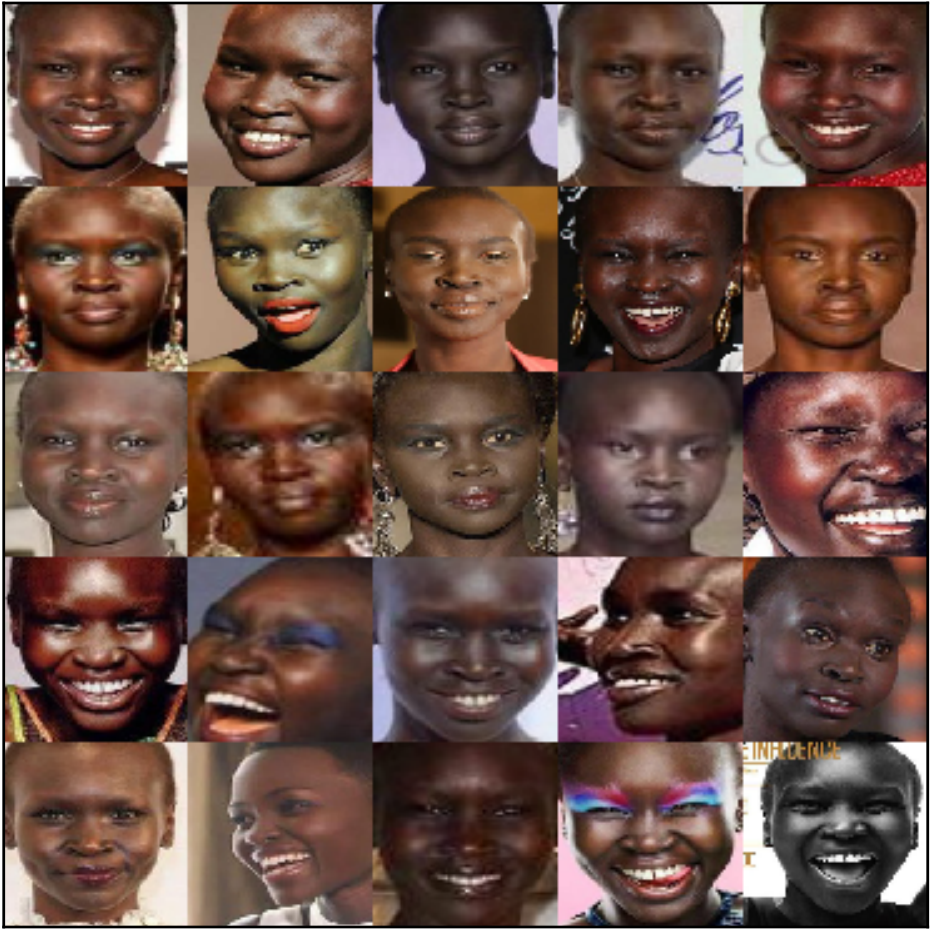}
        \caption{\gls{bf}.}
        \label{subfig:bf}
    \end{subfigure}%
    \begin{subfigure}[t]{.2\linewidth}
        \centering
        \includegraphics[width=\linewidth]{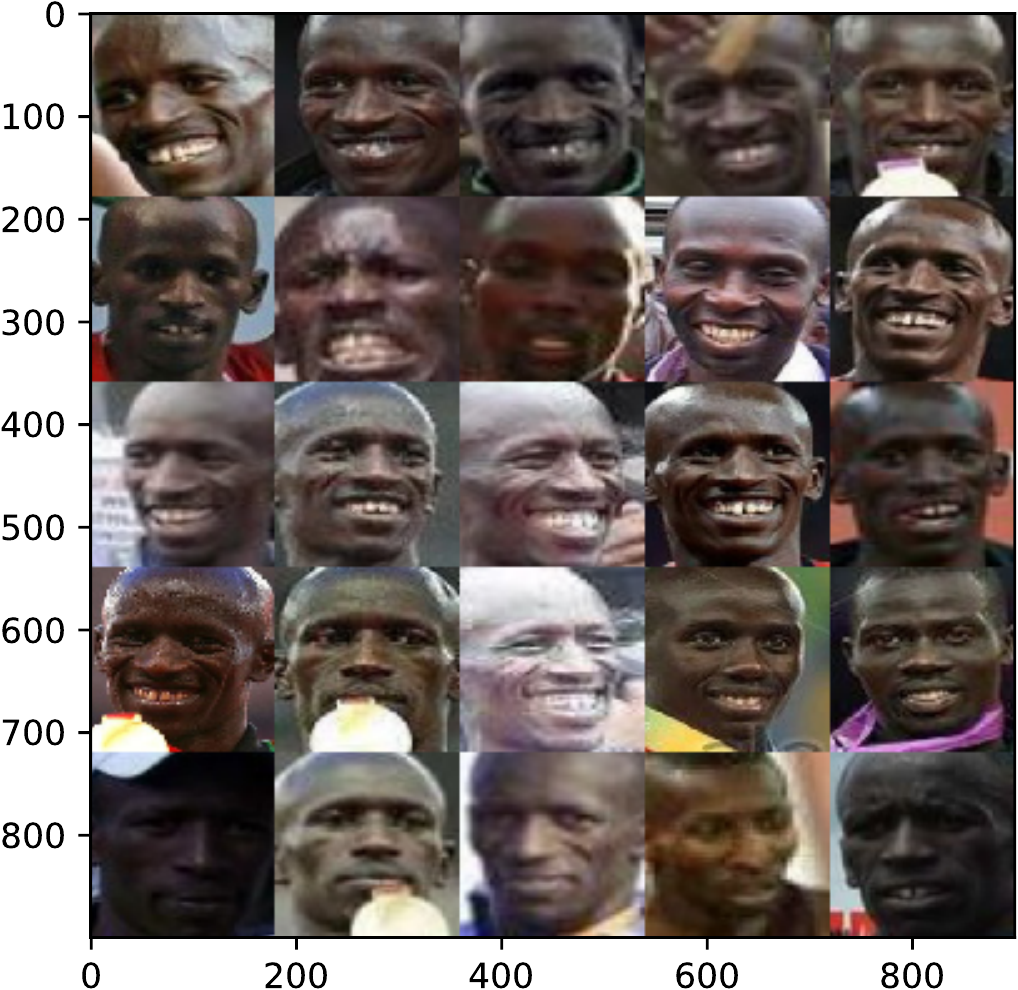}
        \caption{\gls{bm}.}
         \label{subfig:bm}
    \end{subfigure}\\
    \begin{subfigure}[t]{.2\linewidth}
        \centering
        \includegraphics[width=\linewidth]{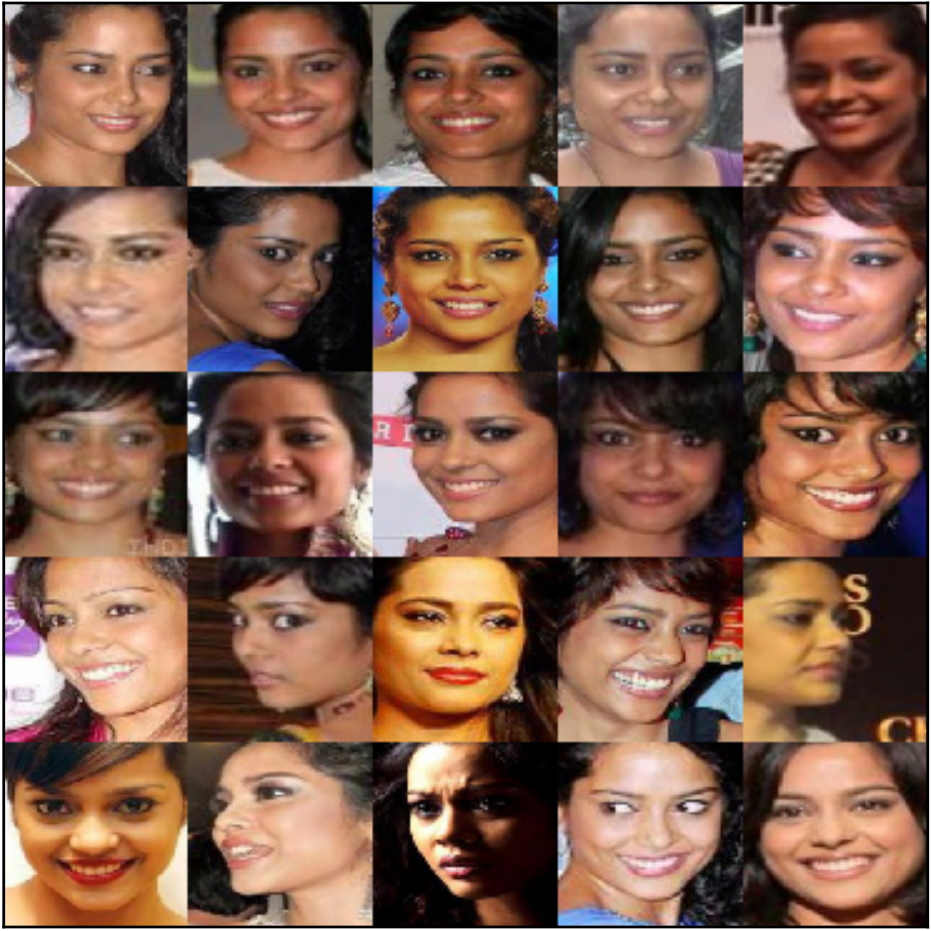}
        \caption{\gls{if}.}
        \label{subfig:if}
    \end{subfigure}%
    \begin{subfigure}[t]{.2\linewidth}
        \centering
        \includegraphics[width=\linewidth]{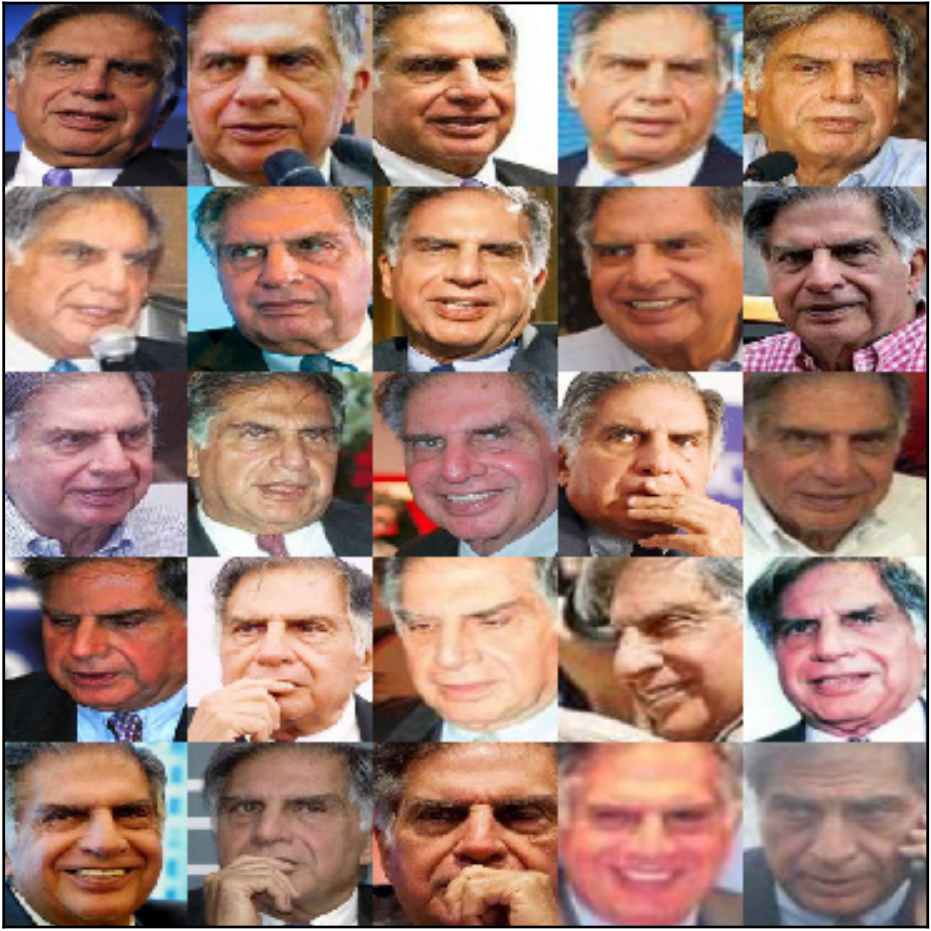}
        \caption{\gls{im}.}
         \label{subfig:ism}
    \end{subfigure}
        \begin{subfigure}[t]{.2\linewidth}
        \centering
        \includegraphics[width=\linewidth]{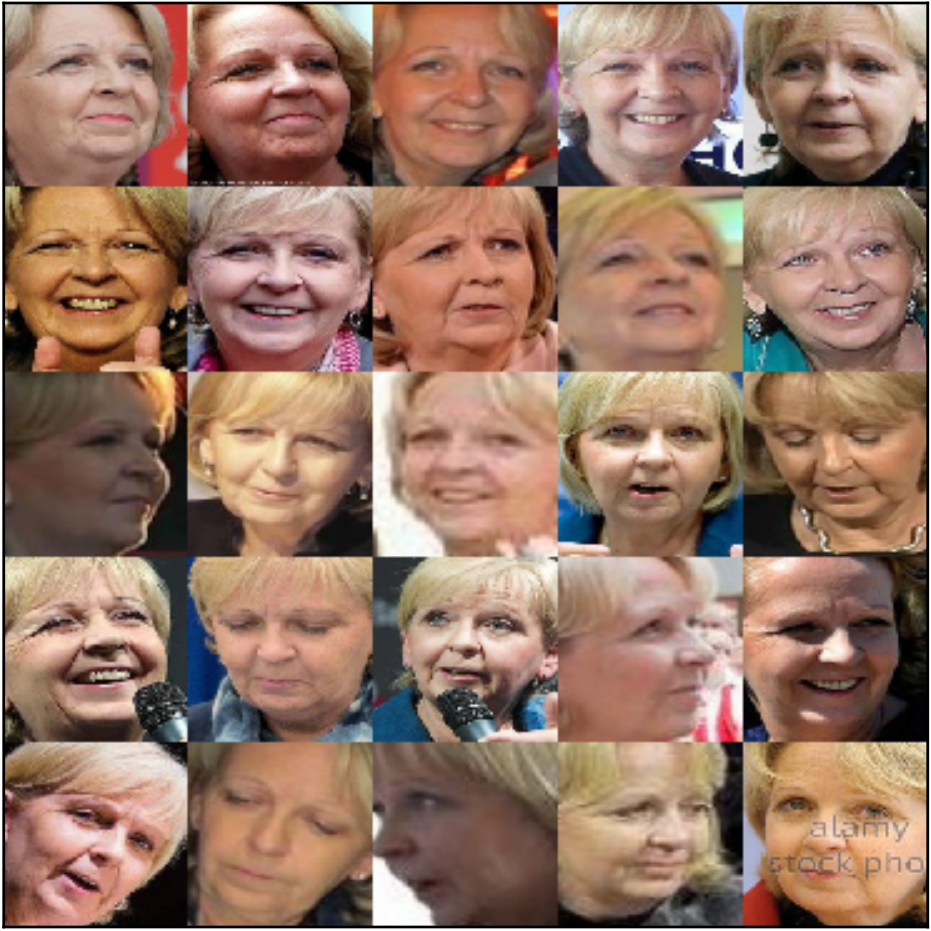}
        \caption{\gls{wf}.}
        \label{subfig:wf}
    \end{subfigure}%
    \begin{subfigure}[t]{.2\linewidth}
        \centering
        \includegraphics[width=\linewidth]{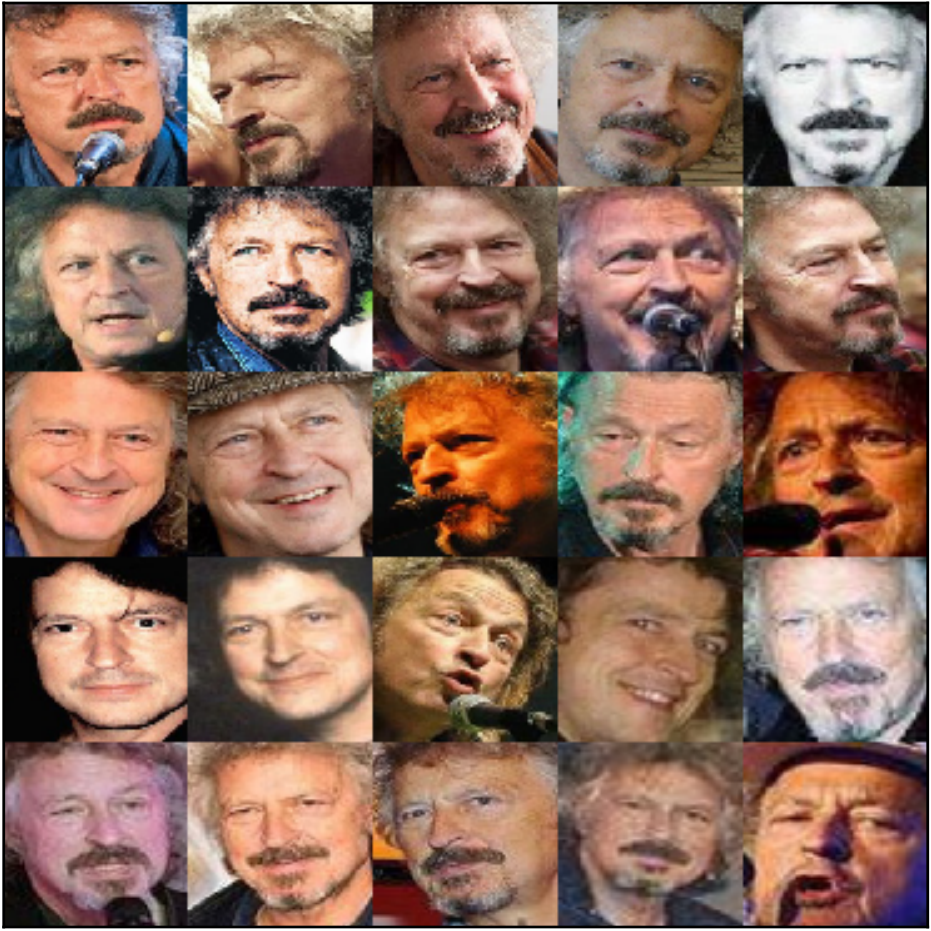}
        \caption{\gls{wm}.}
         \label{subfig:wm}
    \end{subfigure}
    \caption{\textbf{Samples of \gls{bfw}.} Per subgroup: the 25 samples for a random subject are shown.}\label{chap:bias:fig:face-montage}
\end{figure}

\gls{bfw} provides balanced data across ethnicity (\ie Asian (A), Black (B), Indian (I), and White (W)) and gender (\ie Female (F) and Male (M))-- a total of eight demographics referred to as subgroups (\figref{chap:bias:fig:face-montage}). As listed in \tabref{chap:bias:tab:bfw:attributes}, \gls{bfw} has an equal number of subjects per subgroup (\ie 100 subjects per subgroup) and faces per subject (\ie 25 faces per subject). Note that the key difference between \gls{bfw} and \gls{dp} is the additional attributes and the increase in labeled data; the differences from RFW and FairFace are in the identity labels and distributions (\tabref{chap:bias:tab:bfw:counts}).

\begin{figure}[t!]
    \centering
    \includegraphics[width=.4\linewidth]{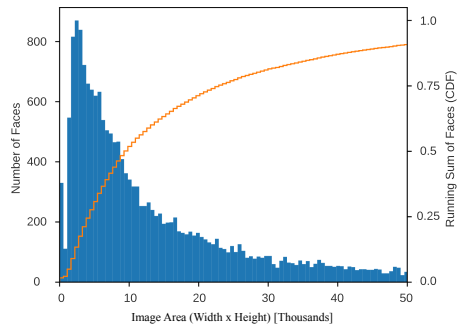}
    \caption{\textbf{\Gls{bfw} statistics (\ie pixel counts).} Histogram of image areas in pixels (blue plot). The orange curve shows the cumulative count of images up to a given area.}
    \label{chap:bias:fig:statistics}
\end{figure}

\gls{bfw} was built with VGG2~\cite{Cao18} by using a set of classifiers on the list of names, and then the corresponding face data. Specifically, we ran a name-ethnicity classifier~\cite{ambekar2009name} to generate the initial list of subject proposals. Then, the list was further refined by processing the corresponding faces with ethnicity~\cite{fu2014learning} and gender~\cite{levi2015age} classifiers. Next, we manually validated, keeping only those that were true members of the respective subgroup. Faces for each subject were then limited to a total of 25 faces that were selected at random, with the distribution of the resolution of the detected faces (\ie area of the bounding boxes) shown in \figref{chap:bias:fig:statistics}. Thus, \gls{bfw} was obtained with minimal human input, having had the proposal lists generated by automatic machinery.


The subgroups of \gls{bfw} were determined based on physical features most common amongst the respective subgroup~\cite{robinson2020face}, which can be regarded as multiple domains due to the feature distributions mismatch across these subgroups. However, the assumption that a discrete label has the capacity to describe an individual is, at best, imprecise. Nonetheless, the assumption allows for a finer-grain analysis of subgroup and is a step in the right direction. Thus, we refute any claim that our efforts here are the final solution; however, we insist that the data and proposed machinery are merely an attempt to establish a foundation for future work to extend. In any case, the two gender for the four ethnic groups make up the eight subgroups of the \gls{bfw} dataset (\figref{chap:bias:fig:face-montage}). Formally put, the tasks addressed have labels for gender $l^g\in\{F, M\}$ and ethnicity $l^e\in\{A, B, I, W\}$, where the $K$ subgroups (\ie demographics) are then $K=|l_g|*|l_e|=8$.

We next review the details behind the process of building \gls{bfw} without any financial burden, while maintaining limited amounts of human labor requirements.


\subsection{The data}
Problems of bias in \gls{fr} motivated us to build \gls{bfw}. Inspired by \gls{dp}~\cite{demogPairs}, the data is made up of evenly split subgroups, but with an increase in subgroups (\ie \gls{if} and \gls{im}), subjects per subgroup, and face pairs (\tabref{chap:bias:tab:ethnic-splits} and \figref{chap:bias:fig:avg-faces}).

\vspace{.5mm}
\noindent\textbf{Compiling subject list.} 
Subjects were sampled from VGG2~\cite{Cao18} - unlike others built from multiple sources, \gls{bfw} has fewer potential conflicts in train and test overlap with existing models. To find candidates for the different subgroups, we first parsed the list of names using a pre-trained ethnicity model~\cite{ambekar2009name}. This was then further refined by processing faces using ethnicity~\cite{fu2014learning} and gender~\cite{levi2015age} classifiers. This resulted in hundreds of candidates per subgroup, which allowed us to manually filter 100 subjects per the 8 subgroups.

\vspace{.5mm}
\noindent\textbf{Detecting faces.} Faces were detected using MTCNN~\cite{zhang2016joint}.\footnote{\href{https://github.com/polarisZhao/mtcnn-pytorch}{https://github.com/polarisZhao/mtcnn-pytorch}} Then, faces were assigned to one of two sets. Faces within detected bounding box (BB) regions extended out 130\% in each direction, with zero-padding as the boundary condition made-up one set. The second set were faces aligned and cropped for Sphereface~\cite{liu2017sphereface} (see the next step). Also, coordinates of the BB and the five landmarks from \gls{mtcnn} were stored as part of the static, raw data. For samples with multiple face detections, we used the BB area times the confidence score of the \gls{mtcnn} to determine the face most likely to be the subject of interest, with the others set aside and labeled \textit{miss-detection}.

\vspace{.5mm}
\noindent\textbf{Validating labels.} 
Faces of \gls{bfw} were encoded using the original implementation of the \gls{soa} Sphereface~\cite{liu2017sphereface}. For this, each face was aligned to predefined eye locations via an affine transformation. Then, faces were fed through the \gls{cnn} twice (\ie the original and horizontally flipped), with two features fused by average pooling (\ie 512 D). A matrix of cosine similarity scores was then generated for each subject and removed samples (\ie rows) with median scores below threshold $\theta=0.2$ (set manually). Mathematically, the $n^{th}$ sample for the $j^{th}$ subject with $N_j$ faces was removed if the ordinal rank of its score $n = \frac{P\times N}{100}\geq\theta$, where $P=50$. In other words, the median (\ie 50 percentile) of all scores for a face with respect to all of faces for the respective subject must pass a threshold of $\theta=0.2$; else, the face is dropped. This allowed us to quickly prune \gls{fp} face detections. Following~\cite{FIW, robinson2018visual}, we built a JAVA tool to visually validate the remaining faces. For this, the faces were ordered by decreasing confidence, with confidence set as the average score, and then displayed as image icons on top toggling buttons arranged as a grid in a sliding pane window. Labeling then consisted of going subject-by-subject and flagging faces of \emph{imposters}.

\begin{figure}[!t] 
	\centering    
 \glsunset{fpr}
  \glsunset{fnr}
	\includegraphics[width=.5\linewidth]{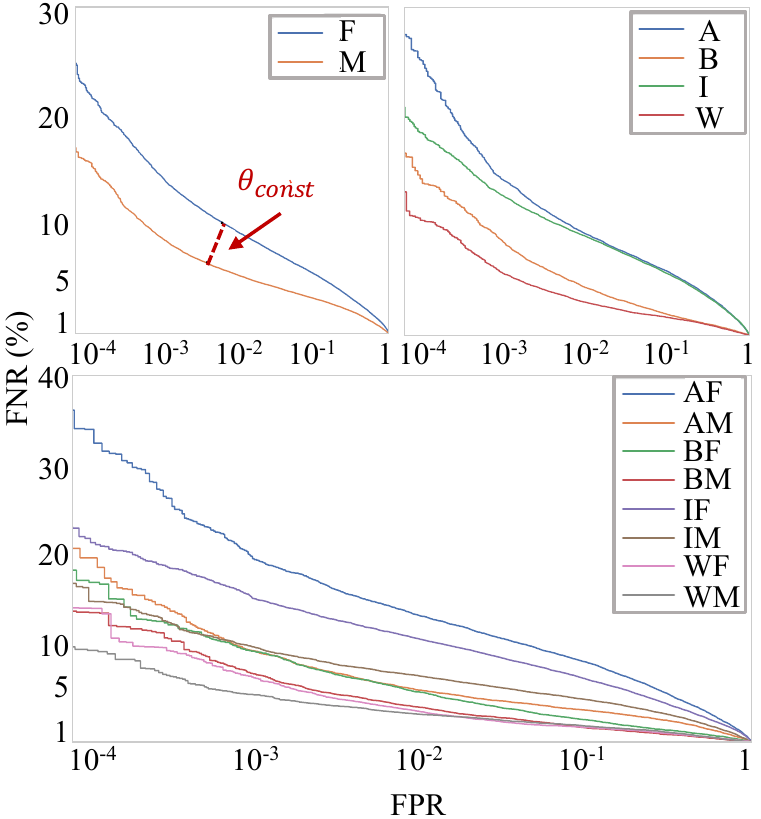}
		\caption{\textbf{\gls{det} curves.} \emph{Top-left}: per gender. \emph{Top-right}: per ethnicity. \emph{Bottom}: per subgroup (\ie combined). Dashed line shows about 2$\times$ difference in \gls{fpr} for the same threshold $\theta_{const}$. \gls{fnr} is the match error count (closer to the bottom is better).}
		\glsreset{det}\glsreset{fpr}
\label{chap:bias:fig:detcurves} 

\end{figure} 

\noindent\textbf{Sampling faces and creating folds.} We created lists of pairs in five-folds with subjects split evenly per person and without overlap across folds. Furthermore, a balance in the number of faces per subgroup was obtained by sampling twenty-five faces at random from each. Next, we generated a list of all the face pairs per subject, resulting in $\sum_{l=1}^{L}\sum_{k=1}^{K_d} {N_k \choose 2}$ positive pairs, where the number of faces of all $K_l$ subjects $N_k=25$  for each of the $L$ subgroups (\tabref{chap:bias:tab:ethnic-splits}). Next, we assigned subjects to a fold. To preserve balance across folds, we sorted subjects by the number of pairs and then started assigning to alternating folds from the one with the most samples. Note, this left no overlap in identity between folds. Later, a negative set from samples within the same subgroup randomly matched until the count met that of the positive. Finally, we doubled the number with negative pairs from across subgroups but in the same fold.

\subsection{Problem formulation}\label{subsec:pf} 
\Gls{fv} is the special case of the two-class (\ie boolean) classification. Hence, pairs are labeled as the ``same'' or ``different'' \textit{genuine} pairs (\ie \textit{match}) or \textit{imposter} (\ie \textit{mismatch}), respectively. This formulation (\ie \gls{fv}) is highly practical for applications like access control, re-identification, and surveillance. Typically, training a separate model for each unique subject is unfeasible. Firstly, the computational costs compound as the number of subjects increase.  Secondly, such a scheme would require model retraining each time a new person is added. Instead, we train models to encode facial images in a feature space that captures the uniqueness of a face, to then determine the outcome based on the output of a scoring (or distance) function. Formally put:
\begin{equation}\label{chap:bias:eg:matcher2}
    f_{boolean}(\vec{x}_i, \vec{x}_j) = d(\vec{x}_i, \vec{x}_j) \leq \theta,
\end{equation}

\noindent where $f_{boolean}$ is the \textit{matcher} of the feature vector $\vec{x}$ for the $i^{th}$ and $j^{th}$ sample~\cite{LFWTech}.

Cosine similarity is used as the \emph{matcher} in Eq~\ref{chap:bias:eg:matcher2} the closeness of $i^{th}$ and $j^{th}$ features, \ie
$
s_l= \frac{f_i\cdot f_j}{||f_i||_2||f_j||_2}
$ is the closeness of the $l^{th}$ pair. 

\begin{figure}[t!]
	\centering    
	\includegraphics[width=.5\linewidth]{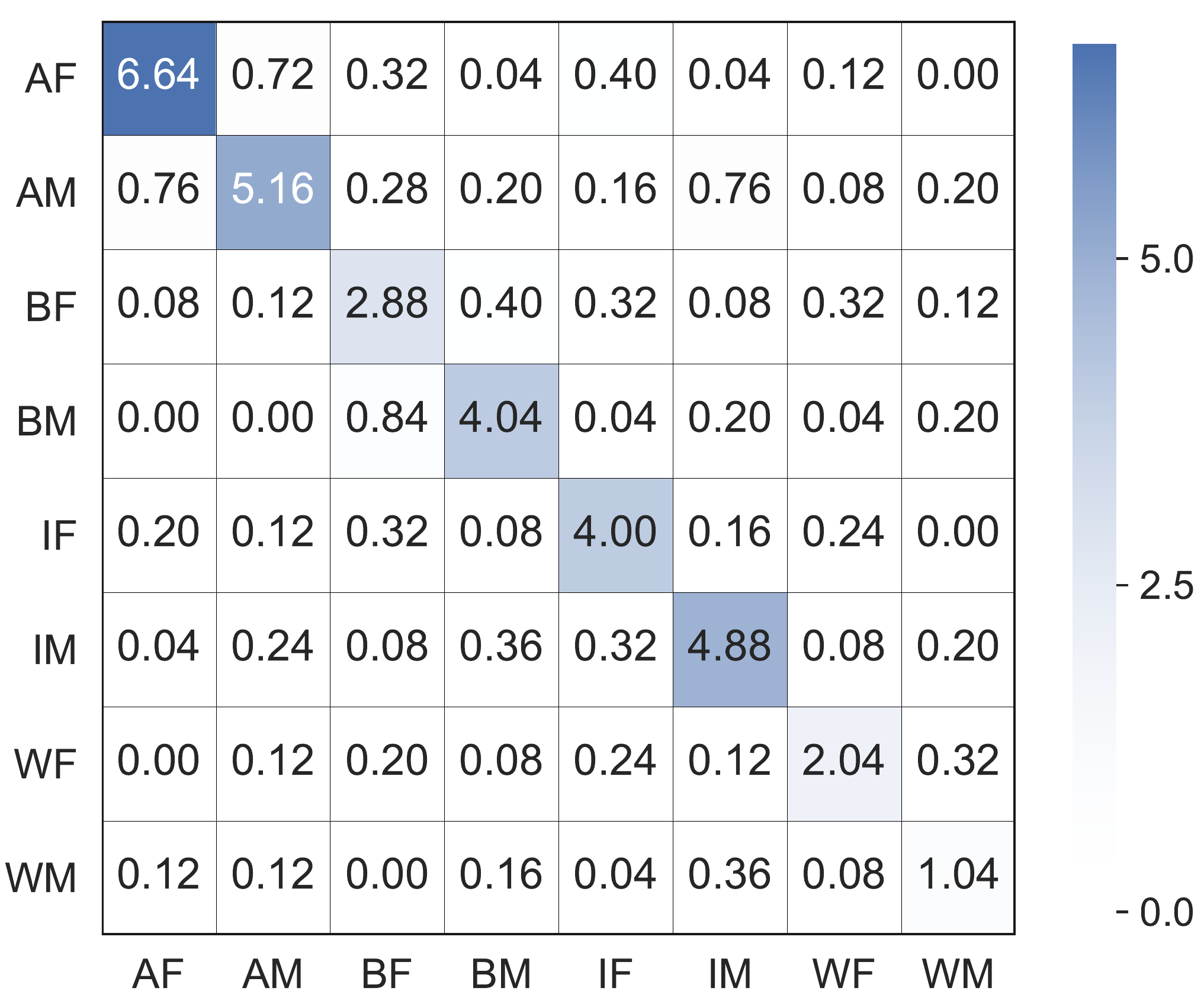}
		\caption{\textbf{Confusion matrix.} Error (Rank 1, \%) for all \gls{bfw} faces versus all others. Errors concentrate intra-subgroup - consistent with the \gls{sdm} (\figref{chap:bias:fig:detection-model}). Although subgroups are challenging to define, this shows the ones chosen are meaningful for \gls{fr}.}
		\label{chap:bias:fig:confusion:rank1} 
\end{figure} 

\subsection{Human assessment}\label{subsec:human-assessment}
To focus on the human evaluation experiment, we honed-in on pairs from two groups, White Americans (W) and Chinese from China (C). This minimized variability compared to the broader groups of Whites and Asians, which was thought to be best, provided only a small subset of the data was used on fewer humans than subjects in \gls{bfw}.


Samples were collected from individuals recruited from multiple sources (\eg social media, email lists, family, and friends) - a total of 120 participants were sampled at random from all submissions that were completed and done by a W or C participant. Specifically, there were 60 W and 60 C, both with \gls{m} and \gls{f} split evenly. A total of 50 face pairs of non-famous ``look-alikes'' were collected from the internet, with 20 ({\emph WA}) and 20 ({\emph C}) pairs with, again, \gls{m} and \gls{f} split evenly. The other 10 were of a different demographic (\eg Hispanic/ Latino, Japanese, African). The survey was created, distributed, and recorded via \emph{\href{https://paperform.co}{PaperForm}}. It is important to note that participants were only aware of the task (\ie to verify whether or not a face-pair was a \emph{match} or \emph{non-match}, but with no knowledge of it being a study on the bias).

\begin{table}
    \centering

     \caption{\textbf{Data statistics, notation, and scores for subgroups of our \gls{bfw} data.} \textit{Top:} Specifications of \gls{bfw} and subgroup definitions. \textit{Middle:} Number of pairs. \textit{Bottom:} Accuracy fo a global threshold $t_g$, the value of the optimal threshold $t_o$, and accuracy using $t_o$ per subgroup. Columns grouped by race and then further split by gender. Notice the inconsistent ratings across subgroups.}
\resizebox{\textwidth}{!}{%
    \begin{tabular}{r c c c c c c c c l}
    
        \toprule 
        & \multicolumn{2}{c}{Asian (A)} & \multicolumn{2}{c}{Black (B)}  & \multicolumn{2}{c}{Indian (I)}& \multicolumn{2}{c}{White (W)}\\
        \cmidrule(l){2-3} \cmidrule(l){4-5} \cmidrule(l){6-7}\cmidrule(l){8-9} 
         & Female (AF) & Male (AM) & BF & BM& IF & IM & WF & WM&Aggregated\\ 
        \midrule

       No. Faces  &  2,500&  2,500& 2,500 &2,500 & 2,500 & 2,500 & 2,500 &2,500 &20,000 \\ 
        No. Subjects & 100& 100& 100  & 100  & 100  & 100& 100 &100&800  \\ 
        No. Faces / subject  & 25& 25    & 25 & 25 & 25  & 25  &  25 & 25 & 25\\ 
        \midrule
            No. Positive &  30,000& 	30,000& 30,000 &30,000 & 30,000 &30,000&30,000 & 30,000 &240,000 \\ 
        No. Negative &85,135&  85,232& 85,016  & 85,141 & 85,287  & 85,152& 85,223 &	85,193&681,379  \\ 
        Total & 115,135 & 115,232    &115,016 &115,141 & 115,287  & 115,152  &  115,223& 115,193& 921,379\\ 
        \midrule

        Acc$@t_g$ & 0.876 & 0.944  &0.934 &0.942 &0.922&0.949 &0.916  &0.918&0.925$\pm$0.022 \\ 
        $t_o$ & 0.235 &  0.274 & 0.267&0.254  &0.299 & 0.295& 0.242 &0.222&0.261$\pm$0.025\\ 
        Acc$@t_o$ & 0.916 &0.964 &0.955 &0.971 &0.933 &0.958 & 0.969 &0.973 & 0.955 $\pm$ 0.018\\ 
        \bottomrule
    \end{tabular}
    }\label{chap:bias:tab:bfw:counts} 
\end{table}
    
\glsreset{fv}

\section{Methodology}\label{chap:bias:sec:proposed}
To discuss the bias and privacy concerns addressed by the proposed, we first introduce \gls{fv}. Specifically, we review the problem statement, the supporting facial image dataset, along with the objectives of the proposed framework set to achieve the solutions sought in this work. That is to say, to preserve identity information while balancing the sensitivities of encodings for the different demographics (\ie subgroups), and in a way, to remove knowledge of the subgroups from the facial encoding - the typical representation available for operational cases.

\subsection{Problem statement}
 FV systems make decisions based on the likeliness a pair of faces are of the same identity. In fact, the core procedure of verification systems are often similar to the \gls{fr} employed for various applications. Specifically, a \gls{cnn} is trained on a closed set of identities, and then later used to encode faces (\ie map face images to features). The encodings are then compared in closeness to produce a single score-- often via cosine similarity in \gls{fr}~\cite{nguyen2010cosine}. It is imperative to learn the optimal score threshold separating true from false pairs. The threshold is the decision boundary in score space, \ie the \emph{matching function}.

\subsubsection{The matching function}\label{subsec:metrics}
A real-valued similarity score $\mathrm{R}$ assumes a discrete label of $Y=1$ for \emph{genuine} pairs (\ie a true-match) or $Y=0$ for \emph{imposter} (\ie untrue-match). The real-value is mapped to a discrete label by $\hat{Y}=\mathbbm{I}\{\mathrm{R}>\theta\}$ for some predefined threshold $\theta$. The aforementioned can be expressed as \emph{matcher} $\mathrel{d}$ operating as

\begin{equation}\label{chap:bias:eg:matcher}
    f_{boolean}(\vec{x}_i, \vec{x}_j) = \mathrel{d}(\vec{x}_i, \vec{x}_j) \leq \theta,
\end{equation}

\noindent where the face encodings in $\vec{x}$ being the $i^{th}$ and $j^{th}$ sample-- a conventional scheme in the \gls{fr} research communities~\cite{LFWTech}. We use cosine similarity as the \emph{matcher} in Eq.~\ref{chap:bias:eg:matcher}, which produces a score in closeness for the $i^{th}$ and $j^{th}$ faces (\ie $l$-{th} face pair) by
$d(\vec{x}_i, \vec{x}_j)=s_l= \frac{f_i\cdot f_j}{||f_i||_2||f_j||_2}$. The decision boundary formed by threshold $\theta$ controls the level of \emph{acceptance} and \emph{rejection}. Thus, $\theta$ inherits a trade-off between the sensitivity and specificity. The operating point chosen is done so, most always, depending on the purpose of the respective system (\ie perhaps higher sensitivity (or lower threshold) for threat-related tasks, in that flagging a few extra is worth not overlooking the one true positive). Specifically, the trade-off involves \gls{fnr}, a Type 1 Error where \emph{genuine} attempts to pass but is falsely rejected. Mathematically, it relates by
$$
\text{FNR}=\frac{\text{FN}}{\text{P}}=\frac{\text{FN}}{\text{FN}+\text{TP}}=1-\text{TAR} = 1 - \frac{\text{TP}}{\text{FN}+\text{TP}},
$$
\noindent with positive counts P, \gls{tp}, and \gls{fn}.

\begin{figure}[t!]
    \centering
    \includegraphics[width=.85\linewidth]{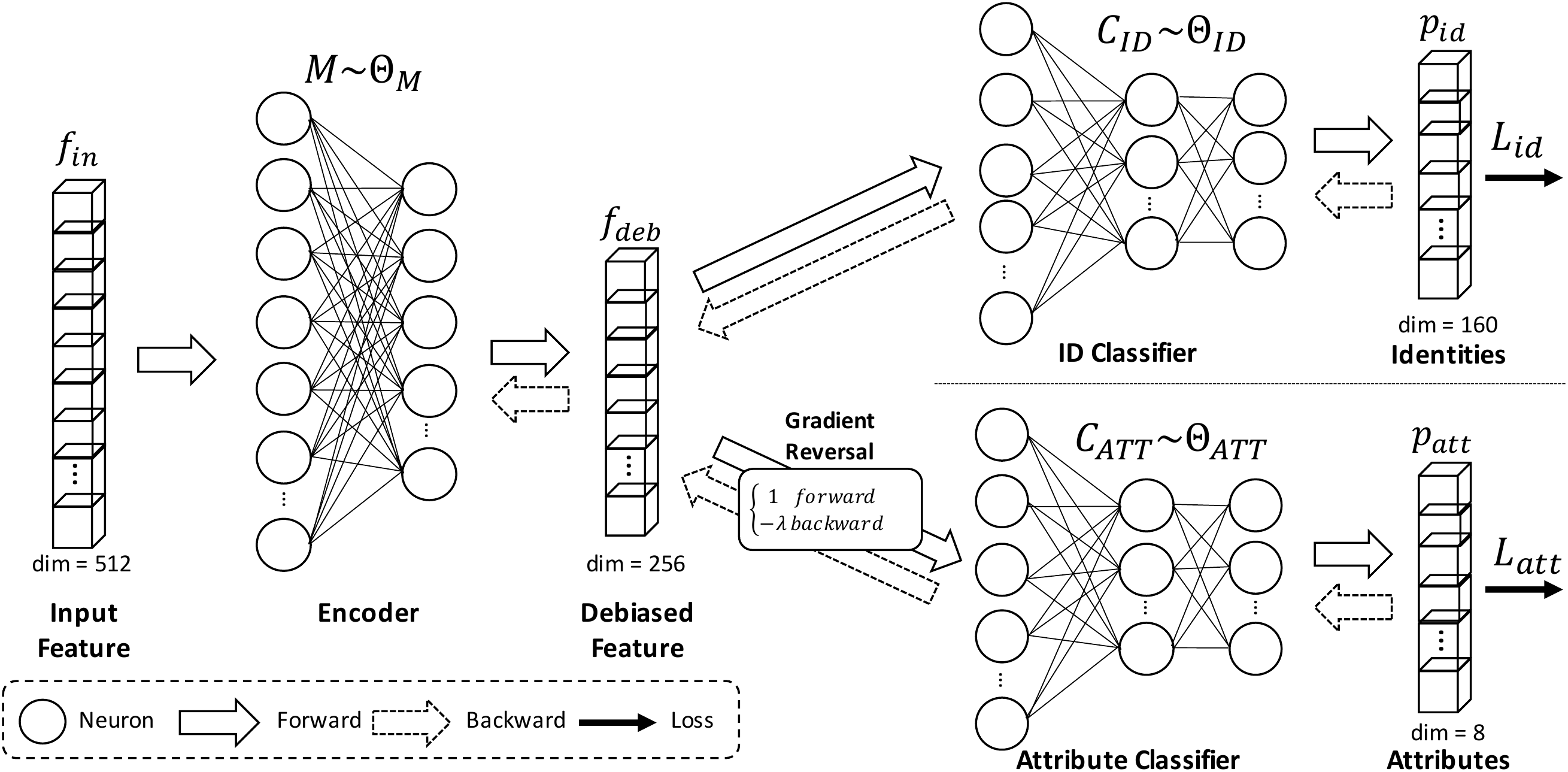}
    \caption{\textbf{Debiasing framework.} The framework used to learn a projection that casts facial encodings to a space that (1) preserves identity information (\ie $C_{ID}$) and (2) removes knowledge of subgroup (\ie $C_{ATT}$). The benefits of this are two-fold: ability to verify pairs of faces fairly across attributes and an inability to classify attribute for privacy and safety purposes. Note, that the \emph{gradient reversal}~\cite{ganin2015unsupervised} flips the sign of the error back-propagated from $C_{ATT}$ to $M$ by scalar $\lambda$ during training.}\label{chap:bias:fig:framework}
\end{figure}

The other error type contributes to the \gls{fpr}, is referred to as the Type II Error as an \emph{imposter}, and is falsely accepted: $$
\text{FAR}=\frac{\text{FP}}{\text{N}}=\frac{\text{FP}}{\text{FP}+\text{TN}}=1-\text{TNR}=1-\frac{\text{TN}}{\text{FP}+\text{TN}},
$$
\noindent where the counts negatives is N, and the metrics are \gls{tn}, \gls{fp}, \gls{tnr}, and \gls{far}. A \emph{matching} function is the module typically characterized using the listed metrics (\figref{chap:bias:fig:metrics}). The geometric relationships of the metrics related to the score distributions and the choice in threshold show the trade-offs in error rates (\figref{chap:bias:fig:metrics}).
    
 The hyperparameter $\theta$ is often determined for a desired error rate on a held-out validation, and is use-case specific. In research, it is typically set to acquire the best performance possible. Some analyze $\theta$ as a range of values, for a more complete characterization is often obtainable with evaluation curves that inherently show performance trade-offs. However, the held-out validation and test sets typically share data distributions as a single source partitioned into subsets (\ie train, validation, test). Regardless, the decision boundary in score space that maximizes the performance is transferred to the pin-point (\ie 1D) decision boundary - in our case, with cosine similarity, the value is a floating point value spanning [0, 1].

\subsubsection{Feature alignment}
Domain $\mathcal{D}$ can be represented by the tuple $\mathcal{D}=\{(\mathbf{x}_i,y_i) \in \mathcal{X} \times \mathcal{Y} \}_{i=1}^{N}$, with $\mathcal{X}$ and $\mathcal{Y}$ representing the input feature space and output label space, respectively. The objective of \gls{fr} algorithms is to learn a mapping function (\ie an hypothesis): ${\eta} : \mathcal{X} \rightarrow \mathcal{Y}$, which assigns each sample vector with a semantic identity label.

In domain adaptation, a model is trained on source data and deployed on target data, where an abundance of paired data is available to train a model for a task similar to that of the target. Mathematically, the labeled source domain $\mathcal{D}_S$ and the unlabeled target domain $\mathcal{D}_T$ can be denoted as  $\mathcal{D}_S = \{(\mathbf{x}_i^s, y_i^s) \in \mathcal{X}_S \times \mathcal{Y}_S\}^{N_S}_{i=1}$ and  $\mathcal{D}_T = \{\mathbf{x}_i^t \in \mathcal{X}_T \}^{N_T}_{i=1}$ with the sample count $N_S=|\mathcal{D}_S|$ and $N_T=|\mathcal{D}_T|$ corresponding to the $i$-th sample (\ie $\mathbf{x}_i \in \mathbb{R}^{d}$) and label (\ie $y_i \in \{1, ..., K\}$). $\mathcal{D}_S$ and $\mathcal{D}_T$ are further defined by tasks $\mathcal{T}_S$ and $\mathcal{T}_T$, which is indicative of the exact label type(s) and the specific $K$ classes of interest. The goal is to learn an objective ${\eta}_S: \mathcal{X}_S \rightarrow \mathcal{Y}_S$, and then transfer to target $\mathcal{D}_T$ for $\mathcal{T}_T$. By this, knowledge is leveraged from both $\mathcal{D}_S$ for $\mathcal{D}_T$, with the goal of obtaining $\eta_T$. Since such two domains share different marginal distributions, \textit{i.e.}, $p({\mathbf{x}}^s)\not=p({\mathbf{x}}^t)$, as well as distinct conditional distributions, \textit{i.e.}, $p(y^t|{\mathbf{x}}^t)\not=p(y^t|{\mathbf{x}}^t)$, the model trained only by the labeled source domain samples usually performs poorly on the unlabeled target domain. A typical solution towards such domain gap is to learn a model $f$ that aligns the features in a shared subspace: $p({f(\mathbf{x}}^s))\approx p({f(\mathbf{x}}^t))$.

\subsection{Proposed framework}
Given samples with identity and subgroup labels-- $\mathcal{D} = \{\mathbf{x}_i, y_i^{id}, y_i^{att} \}^{N}_{i=1}$ , where $\mathbf{x} \in \mathbb{R}^{d}$, $y^{id} \in \{1,...,I\}$ and $y^{att} \in \{1,...,K\}$-- are used for the two objectives of the proposed framework (\figref{chap:bias:fig:framework}). Hence, we aim to learn a mapping $\mathbf{f}_{deb} = M(\mathbf{x}, \Theta_M)$ to a lower dimensional space $\mathbf{f}_{deb} \in \mathbb{R}^{d/2}$ that preserves identity information of the target. This is dubbed the identity loss $\mathcal{L}_{ID}$. We also learn to do so without subgroup information, which we call the attribute (or subgroup) loss $\mathcal{L}_{ATT}$. The total loss (\ie the final objective $\mathcal{L} = \mathcal{L}_{ID} + \mathcal{L}_{ATT}$) is formed by summing the two aforementioned loss functions:

\begin{align}
\mathcal{L}_{ID} = - &\frac{1}{N} \sum_{i=1}^{N}  \sum_{k=1}^{I}  \mathbf{1}_{[k=y_i^{id}]}{\mathrm{\log}{({p}({y}=y_i^{id}| \mathbf{x}_i)})} ,\label{e2} \\
\mathcal{L}_{ATT}= - &\frac{1}{N} \sum_{i=1}^{N}  \sum_{k=1}^{K}  \mathbf{1}_{[k=y_i^{att}]}{\mathrm{\log}{( {p}({y}=y_i^{att}| \mathbf{x}_i})} , \label{e3}
\end{align}

\noindent where ${p}({y}=y_i^{id}| \mathbf{x}_i)$ and ${p}({y}=y_i^{att}| \mathbf{x}_i)$ represent the softmax conditional probability of its identity and attribute.

We added $\mathcal{L}_{ATT}$ to debias the features to remove variation in scores that were previously handled with a variable threshold. Further, a byproduct are these features that preserve identity information without containing knowledge of subgroup-- a critical concern in the privacy and protection of biometric data.

There are three groups of parameters (\ie $\Theta_M$, $\Theta_{ID}$ and $\Theta_{ATT}$) required to be optimized by the objective (\figref{chap:bias:fig:framework}). Both the identity classifier $C_{ID}$ and attribute classifier $C_{ATT}$ are used to find a feature space that remains accurate to identity and not for subgroup by minimizing the empirical risk of $\mathcal{L}_{ID}$ and $\mathcal{L}_{ATT}$:

\begin{align}
    {\Theta}_{ID}^{*} =&\mathop {\arg \min} \limits_{{\Theta}_{ID}}  \mathcal{L}_{ID},\label{e4}\\
    {\Theta}_{ATT}^{*} =&\mathop {\arg \min} \limits_{{\Theta}_{ATT}}  \mathcal{L}_{ATT}.\label{e5}
\end{align}

It is important to note that the task of $\mathcal{L}_{ATT}$ is to be incorrect (\ie learn a mapping that contains no knowledge of subgroup). Thus, a gradient reversal layer~\cite{ganin2015unsupervised} that acts as the identity during the forward pass, while inverting the sign of the gradient back-propagated with a scalar $\lambda$ as the adversarial loss during training:

\begin{equation}
{\Theta}_{M}^{*} =\mathop {\arg \min} \limits_{{\Theta}_{M}} - \lambda \mathcal{L}_{ATT} +  \mathcal{L}_{ID}. \label{e6}
\end{equation}

Although the proposed learning scheme is simple, it proved effective for both objectives we seek to solve. The effectiveness is clearly demonstrated in the results and analysis.

\section{Results and Analysis}
A single \gls{cnn} was used as a means to control the experiments. For this, Sphereface~\cite{liu2017sphereface} trained on CASIA-Web~\cite{yi2014learning}, and evaluated on \gls{lfw}~\cite{LFWTech} (\%99.22 accuracy), encoded all of the faces.\footnote{\href{$https://github.com/clcarwin/sphereface\_pytorch$}{https://github.com/clcarwin/sphereface\_pytorch}} As reported in~\cite{wang2018racial}, \gls{lfw} has about 13\%, 14\%, 3\%, and 70\% ratio in Asian, Black, Indian, and White, respectively. Furthermore, CASIA-Web is even more unbalanced (again, as reported in~\cite{wang2018racial}), with about  3\%, 11\%, 2\%, and 85\% for the same subgroups.
\glsunset{det}

\noindent\paragraph{\gls{det} analysis.}
\gls{det} curves (5-fold, averaged) show per-subgroup trade-offs (\figref{chap:bias:fig:detcurves}). Note that \gls{m} performs better than \gls{f}, precisely as one would expect from the tails of score-distributions for \emph{genuine} pairs (\figref{chap:bias:fig:detection-model}). \Gls{af} and \gls{if} perform the worst.
\glsunset{m}
\glsunset{f}


\glsunset{am}\glsunset{af}\glsunset{bm}\glsunset{bf}\glsunset{im}\glsunset{if}\glsunset{wm}\glsunset{wf}
\begin{figure}[t!]
    \centering
    \begin{subfigure}[t]{.3\linewidth}
    \includegraphics[width=\linewidth]{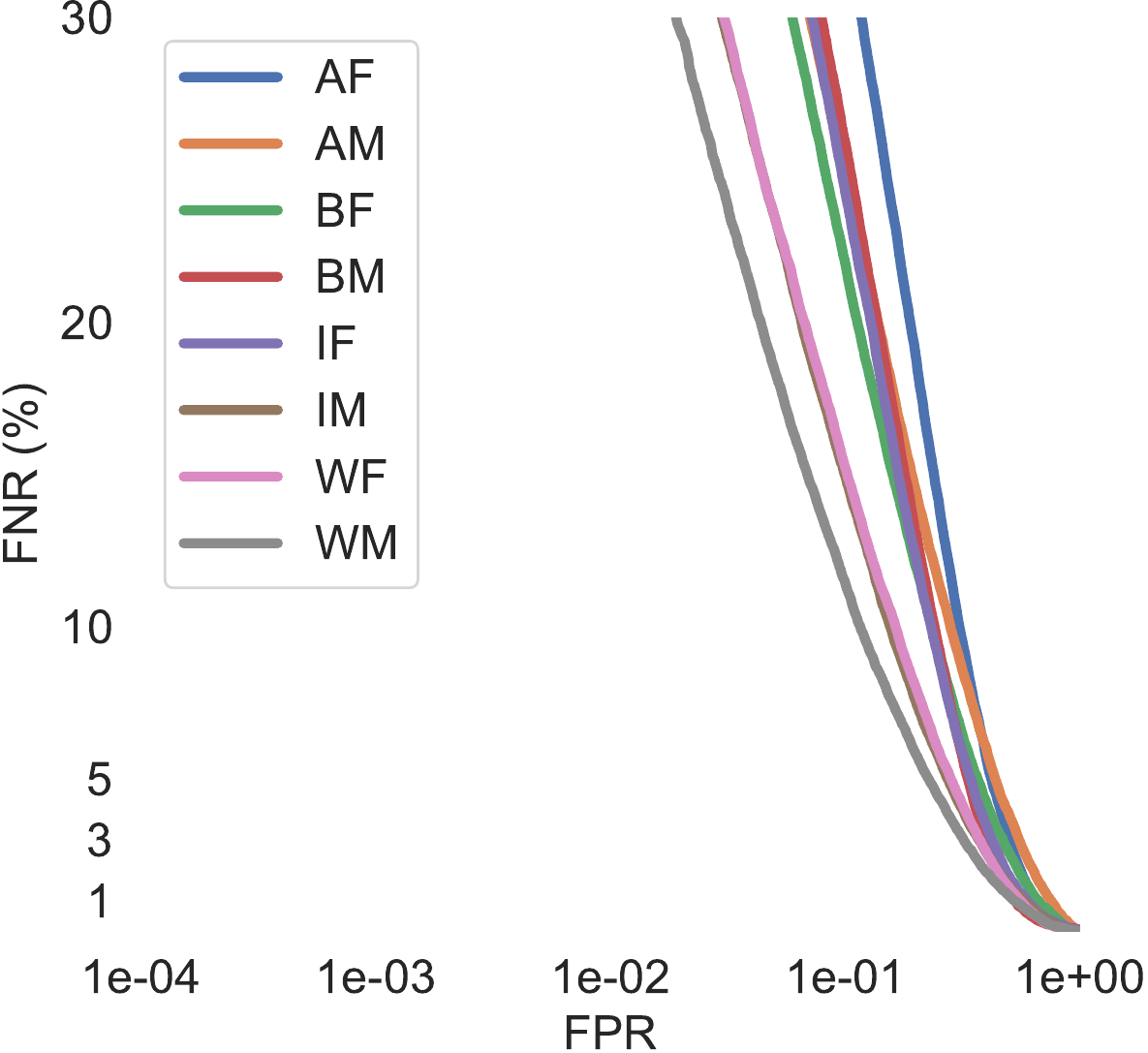}
    \caption{VGG16~\cite{simonyan2014very}}
 \end{subfigure}
    \begin{subfigure}[t]{.28\linewidth}
    \includegraphics[width=\linewidth]{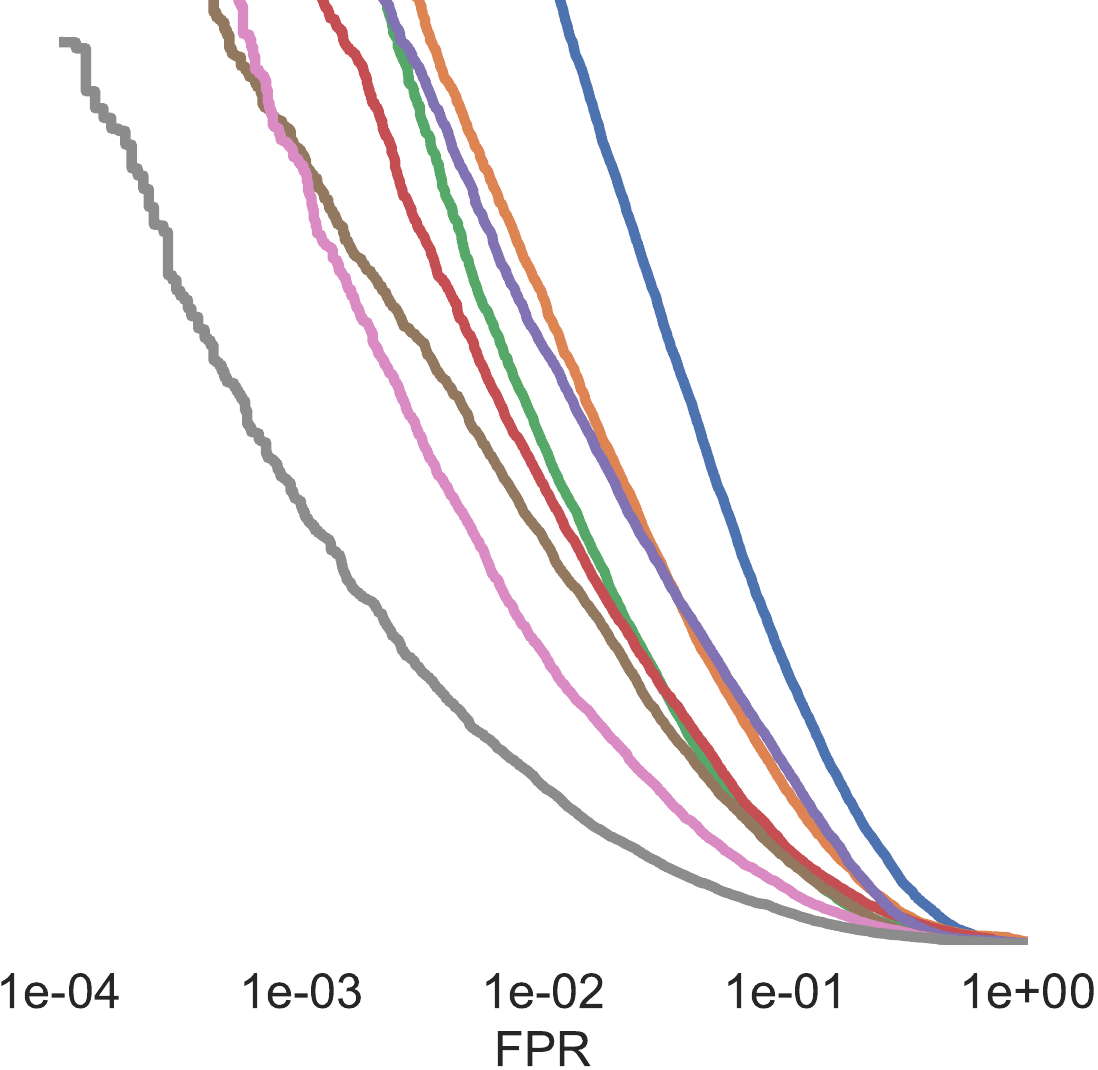}
    \caption{ResNet50~\cite{he2016deep}}
  \end{subfigure}
    \begin{subfigure}[t]{.28\linewidth}
    \includegraphics[width=\linewidth]{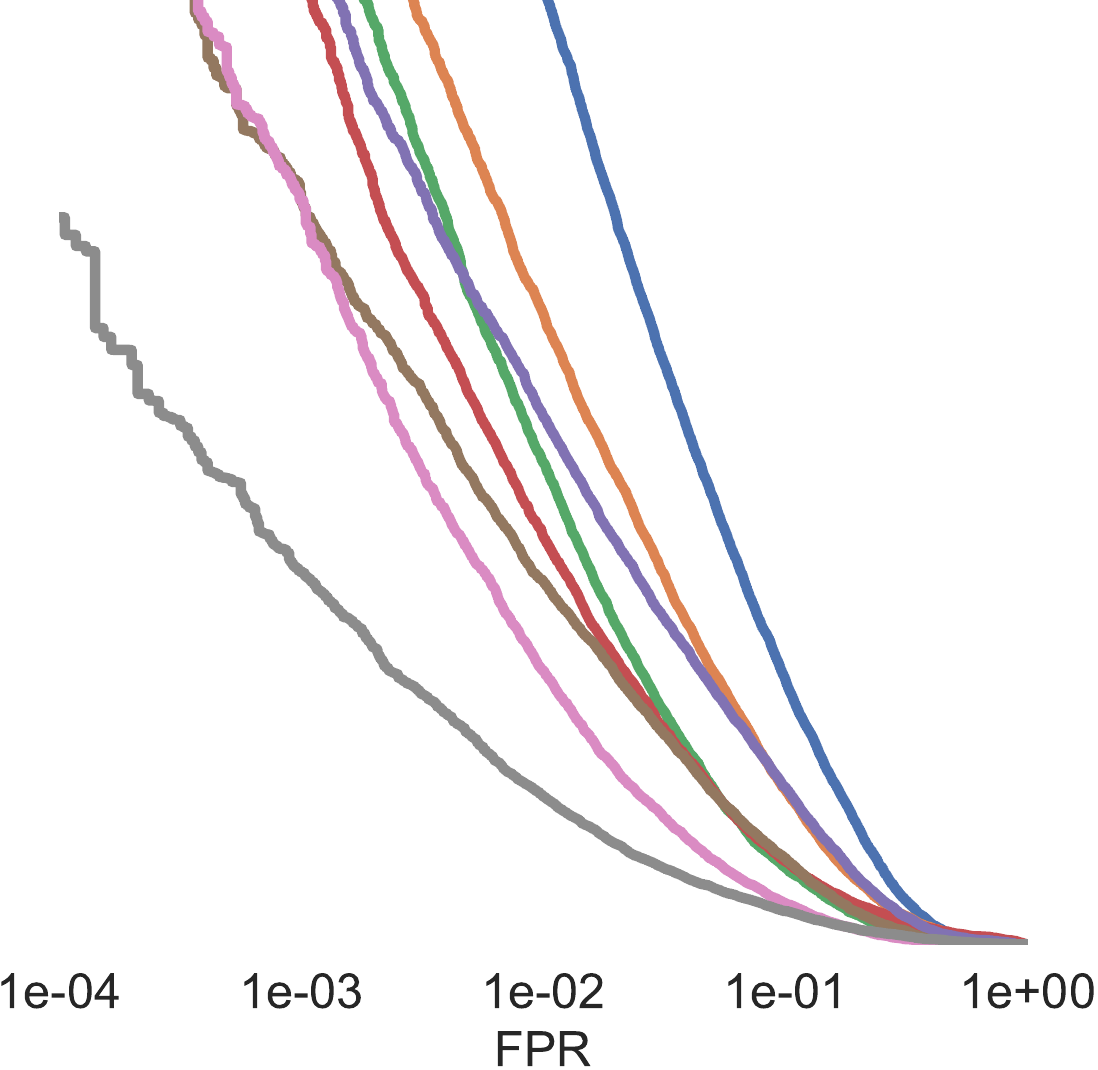}
    \caption{SENet~\cite{hu2018squeeze}}
    \end{subfigure}
    \caption{\textbf{\gls{det} curves for different CNNs.} \gls{fnr} (\%) (vertical) vs \gls{fpr}  (horizontal, log-scale) for VGG2~\cite{Cao18} models with different backbones (VGG16~\cite{simonyan2014very}, Resnet50~\cite{he2016deep}, SEnet50~\cite{hu2018squeeze}, in that order). Lower is better. For each plot, \gls{wm} is the top-performer, while \gls{af} is the worst. The ordering of the curves is roughly the same for each backbone.}\label{chap:bias:fig:sdm-appendix-a}
\end{figure}

\noindent\paragraph{Score analysis.}
\figref{chap:bias:fig:detection-model} shows score distributions for faces of the same (\ie \emph{genuine}) and different (\ie \emph{imposter}) identity, with a subgroup per \gls{sdm} graph. Notice that score distributions for imposters tend to peak about zero for all subgroups, and with minimal deviation comparing modes of the different plots. On the other hand, the score distribution of the \emph{genuine} pairs varies across subgroups in location (\ie score value) and spread (\ie overall shape).
\figref{chap:bias:fig:confusion:rank1} shows the confusion matrix of the subgroups. A vast majority of errors occur in the intra-subgroup. It is interesting to note that while the definition of  each group  based on ethnicity and race may not be crisply defined, the confusion matrix indicates that in practice, the \gls{cnn} finds that the groups are effectively separate. The categories are, therefore, meaningful for \gls{fr}.

\begin{figure}[t!] 
	\centering    
	\includegraphics[width=.55\linewidth] {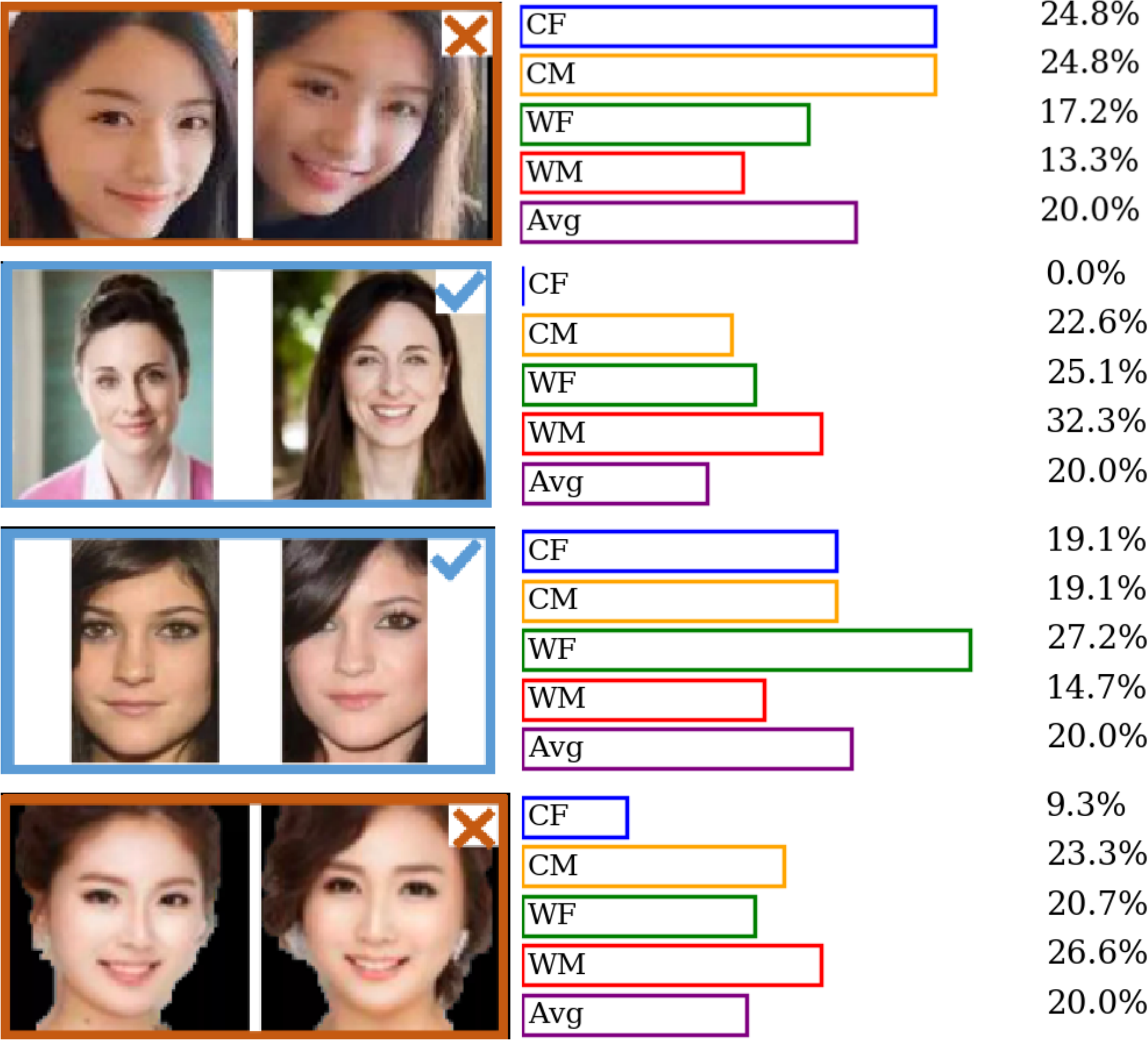}
		\caption{\textbf{Human assessment (qualitative).} $\checkmark$ for \emph{match}; $\times$ for \emph{non-match}. Accuracy scores shown as bar plots. Humans are most successful at recognizing their own subgroup, with a few exceptions (\eg bottom).}
		\label{chap:bias:fig:human-eval} 

\end{figure} 

The gender-based \gls{det} curves show performances with a fixed threshold (dashed line). Other curves relate similarity (lines omitted to declutter). For many \gls{fr} applications, systems operate at the highest \gls{fpr} allowed. The constant threshold shows that a single threshold produces different operating points (\ie \gls{fpr}) across subgroups, which is undesirable. 
If this is the case in an industrial system, one would expect a difference in about double the FPs reported based on subgroup alone. The potential ramifications of such a bias should be considered, which it has not as of lately-- even noticed in main-stream media ~\cite{england2019,snow2018}.

\begin{figure}[t!h]
    \centering
    \includegraphics[width=.5\linewidth,trim=0 0in 0 0,clip]{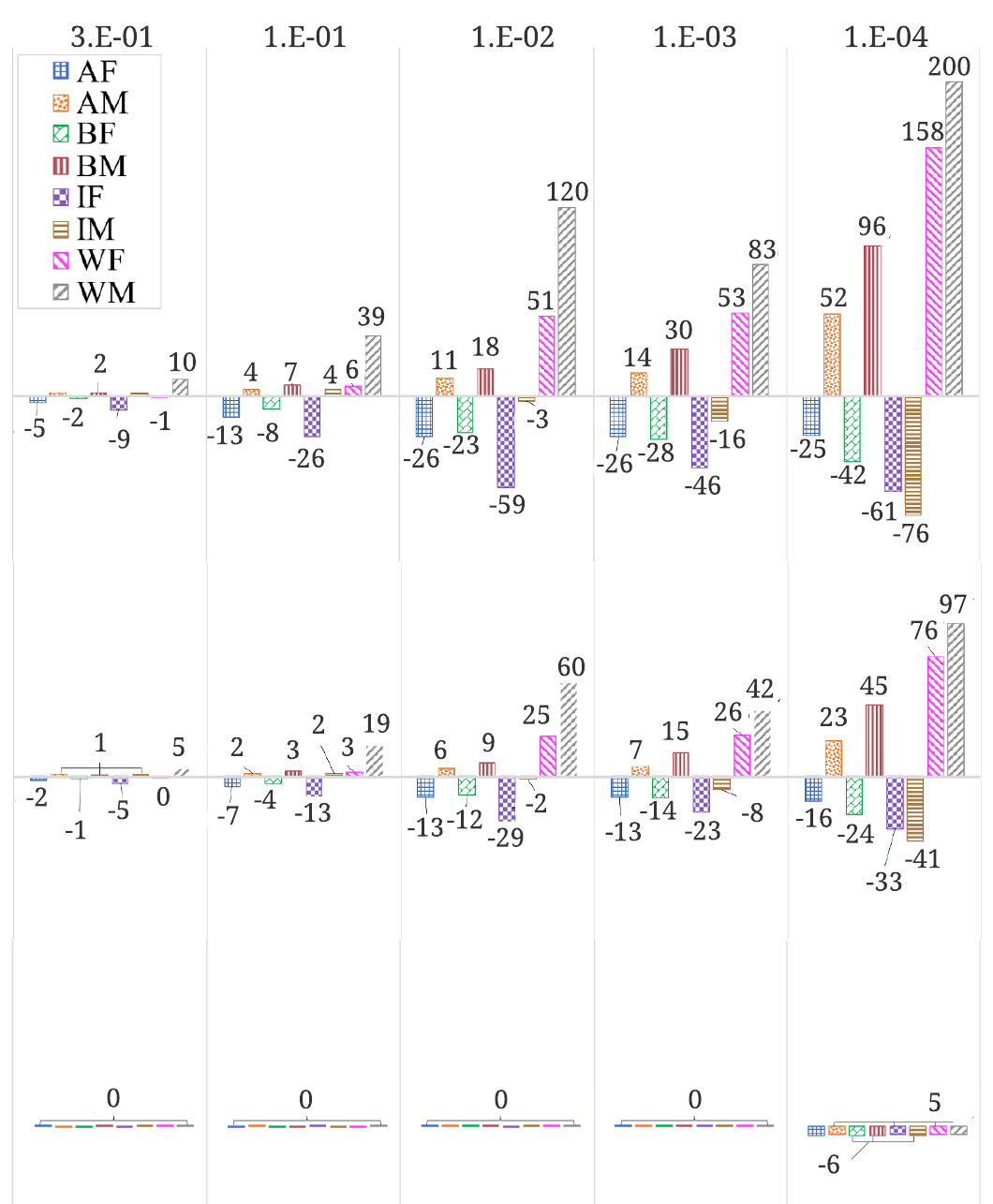}
    \caption{\textbf{Percent difference from intended FPR per subgroup}. \emph{Top:} global threshold ($t_g$) yields a FPR that spans up to 200\% the intended (\ie WM for 1e-4); the F subgroups tend to perform worse than intended for all, while M's overshoot the intended performances besides IM at FPR=1e-4. \emph{Bottom:} Subgroup-specific (or optimal) thresholds $t_o$ reduces the difference closer to zero. Furthermore, the proposed method  (\emph{middle}), which does not assume knowledge of attribute information at inference like for $t_o$, clearly mitigates the issue of the inconsistencies in the true versus reported FPR. Similar to the results in \tabref{chap:bias:tab:ethnicy-far}, the variations are nearly halved: the percent difference for subgroups is more balanced using the adapted features versus the baseline.}
    \label{chap:bias:fig:page1-teaser-barplot}
\end{figure}

To further demonstrate the extent of the problem, we follow settings typical for systems in practice. We set the desired \gls{fpr}, and then determine the percent difference, \ie desired versus actual (\figref{chap:bias:fig:page1-teaser-barplot}, \emph{top}). Furthermore, we mitigate this highly skewed phenomenon by applying subgroup-specific thresholds (\emph{bottom}) - by this, minimal error from the desired \gls{fpr}. Besides where there is a small error, the offset is balanced across subgroups.
\begin{table}[t!]
\centering
    \caption{\textbf{Human assessment (quantitative).} Subgroups listed per row (\ie human) and column (\ie image). Note, most do the best intra-subgroup (\textcolor{blue}{blue}), and second-best (\textcolor{red}{red}) intra-subgroup but inter-gender. WF performs the best; WF pairs are most correctly matched.}
    \label{chap:bias:tab:humsn-eval-results} 

\scalebox{0.94}{
\begin{tabular}{c}
\begin{tabular}{c l c c c cc}
&&\multicolumn{4}{c}{\textbf{Image}}\\
   \multicolumn{2}{c}{(Acc, \%)}& CF  & CM & WF &WM& Avg\\
\end{tabular}\\
\begin{tabular}{c l| r r r r| r}
\cline{3-7}
       &CF &\  \textbf{\textcolor{blue}{52.9}}&   \textbf{\textcolor{red}{48.0}}&43.8&44.7 &47.4 \\
       
        \parbox[t]{2mm}{\multirow{3}{*}{\rotatebox[origin=c]{90}{\textbf{Human}}}}  \hspace{-5mm}
        
        &CM &  \textbf{\textcolor{red}{45.6}} & \textbf{\textcolor{blue}{50.4}}  & 44.4 &36.2 &44.1 \\
       
        &WF & 44.7 &43.8 & \textbf{\textcolor{blue}{57.3}}& \textbf{\textcolor{red}{48.0}} & \textbf{48.5} \\
        &WM & 30.1&\textbf{\textcolor{red}{47.4}} &  45.3 & \textbf{\textcolor{blue}{56.1}} & 44.7\\\cline{3-7}
        &Avg &  43.3& 47.4&\textbf{47.7} &46.3 &46.2\\
 \end{tabular}
 \end{tabular}
 }
\end{table} 

\noindent\paragraph{Model analysis.}
Variations in optimal threshold exist across models (\figref{chap:bias:fig:sdm-appendix-a}). Like in \figref{chap:bias:fig:detcurves}, the \gls{det} curves for three \gls{cnn}-based models, each trained on VGG2 with softmax but with different backbones.\footnote{\href{https://github.com/rcmalli/keras-vggface}{https://github.com/rcmalli/keras-vggface}} Notice similar trends across subgroups and models, which is consistent with  Sphereface as well (\figref{chap:bias:fig:detcurves}). For example, the plots generated with Sphereface and VggFace2 all have the \gls{wm} curve at the bottom (\ie best) and \gls{af} on top (\ie worst). Thus, the additional \gls{cnn}-based models demonstrate the same phenomena: proportional to the overall model performance, exact in the ordering of curves for each subgroup.

\noindent\paragraph{Verification threshold analysis.}
We seek to reduce the bias between subgroups. Such that an operating point (\ie \gls{fpr}) is constant across subgroups. To accomplish that, we used a per subgroup threshold. In \gls{fv}, we consider one image as the query, and all others as the test. For this, the ethnicity of the query image is assumed. We can then examine the \gls{det} curves and pick the best threshold per subgroup for a certain \gls{fpr}.

We evaluated \gls{tar} for specific \gls{far} values. As described in Section~\ref{subsec:pf}, the verification experiments were 5-fold, with no overlap in subject ID between folds. Results reported are averaged across folds in all cases and are shown in \tabref{chap:bias:tab:ethnicy-far}. For each subgroup, the \gls{tar} of using a global threshold is reported (upper row), as well as using the optimal per subgroup threshold (lower row). 

Even for lower \gls{far}, there are notable improvements, often of the order of 1\%, which can be challenging to achieve when \gls{far} is near $\geq$90\%. More importantly, each subgroup has the desired \gls{fpr}, so that substantial differences in \gls{fpr} will remain unfounded. We experimented with ethnicity estimators on both the query and test image, which yielded similar results to those reported here.


\noindent\paragraph{Human evaluation analysis.}
Subjects of a subgroup likely have mostly been exposed to others of the same (\tabref{chap:bias:tab:humsn-eval-results} and \figref{chap:bias:fig:human-eval}). Therefore, it is expected they would be best at labeling their own, similar to the same ethnicity, but another gender. Our findings concur. Each subgroup is best at labeling their type, and then second best at labeling the same ethnicity but opposite sex. 

Interestingly, each group of images is best tagged by the corresponding subgroup, with the second-best having the same ethnicity and opposite gender. On average, subgroups are comparable at labeling images. Analogous to the \gls{fr} system, performance ratings differ when analyzing within and across subgroups. In other words, performance on \gls{bfw} improved with subgroup-specific thresholds. Similarly, humans tend to better recognize individuals by facial cues of the same or similar demographics. Put differently, as the recognition performances drop with a global threshold optimized for one subgroup and deployed on another, human performance tends to fall when across subgroups (\ie performances drop for less familiar subgroups).

\begin{table}
\glsunset{rfw}\glsunset{dp}
\centering
    \caption{\textbf{\gls{bfw} features compared to related resources.} Note, the balance across identity (ID), gender (G), and ethnicity (E). Compared with \gls{dp}, \gls{bfw} provides more samples per subject and subgroups per set. Also, \gls{bfw} uses a single resource, VGG2. \gls{rfw}; on the other hand, supports a different task (\ie subgroup classification). Furthermore, \gls{rfw} and FairFace focus on race-distribution without the support of identity labels.}\label{chap:bias:tab:bfw:attributes} 
         \resizebox{\textwidth}{!}{%
        \begin{tabular}{rccccccc}\toprule
    \multicolumn{2}{c}{Database} & \multicolumn{3}{c}{Number of}& \multicolumn{3}{c}{Balanced Labels}\\
    \cmidrule(lr){1-2}	\cmidrule(lr){3-5} \cmidrule(lr){6-8}
    Name & Source & Faces &  IDs & Subgroups & ID & E & G\\\midrule
    \gls{rfw}~\cite{wang2018racial}     &  MS-Celeb-1M &$\approx$80,000&$\approx$12,000& 4 & \xmark & \checkc &\xmark \\
    \gls{dp}~\cite{hupont2019demogpairs}     & CASIA-W, VGG (+2) & 10,800& 600 & 6 &\checkc& \checkc &\checkc \\

    FairFace~\cite{karkkainen2019fairface} & Flickr, Twitter, Web & 108,000 & -- & 10 &\xmark& \checkc &\checkc\\

    \gls{bfw} (ours)~\cite{robinson2020face} & VGG2 & 20,000 & 800 &8 & \checkc & \checkc &\checkc \\\bottomrule
    \end{tabular}}
\glsreset{rfw}\glsreset{dp}

\end{table}

\section{Experiments}\label{chap:bias:sec:experimental}
Two sets of experiments are done to demonstrate the effectiveness of the proposed using our balanced BFW~\cite{robinson2020face}. First, we evaluate verification performance using debiasing. Specifically, we compare the reported results compared to the actual results per subgroup.
Then, for the privacy preserving claim, we compare the performance of models trained on top of debiased features $f_{deb}$ with those of the original features $f_{in}$. For each, we present the problem statement, metrics and settings, and analysis. Finally, we do an ablation study to show the performance on the renowned LFW benchmark~\cite{LFWTech}.

\subsection{Common settings}
The baseline (\ie $f_{in}$) were encoded using Arcface~\cite{wang2018additive} (\ie ResNet-34).\footnote{\href{https://github.com/deepinsight/insightface}{https://github.com/deepinsight/insightface}} MS1M~\cite{guo2016ms} was the train set, providing about 5.8M faces for 85k subjects. We prepared the faces using MTCNN~\cite{zhang2016joint} to detect five facial landmarks. A similarity transformation then was applied to align the face by the five detected landmarks, from which we cropped and resized each to 96$\times$112. The RGB (\ie pixel values of [0, 255]) were normalized by centering about 0 (\ie subtracting 127.5) and then standardizing (\ie dividing by 128); encodings were later L2 normalized~\cite{wang2017normface}. The batch size was 200, and a stochastic gradient descent optimizer with a momentum 0.9, weight decay 5e-4, and learning rate started at 0.1 and decreased by a factor of 10 two times when the error leveled. The choice in these settings was made based on Arcface being among the best performing \gls{fr} deep models to date, and as it has become a popular choice for an \emph{off-the-shelf} option for face recognition technology and applications. 

For all experiments we used our \gls{bfw} dataset (\secref{chapter:fiw:fig:allpairs}): the debias and privacy-based experiments use the predefined five-folds\footnote{\href{https://github.com/visionjo/facerec-bias-bfw}{https://github.com/visionjo/facerec-bias-bfw}}; the ablation study on LFW uses all of BFW to train M (\figref{chap:bias:fig:framework}). As mentioned, \gls{bfw} was built using data of VGG2, and there exist no overlap with CASIA-Webface and LFW used to train the face encoder and for the ablation, respectfully.

\subsection{Debias experiment}
Typical \gls{fr} systems are graded by the percent error - whether to a customer, prospective staff, and so forth. In other words, specialized curves, confusion matrices, and other metrics are not always the best way to communicate system performances to non-technical audiences. It is better to discuss ratings in a manner that is more globally understood, and more comprehensible with respect to the use-case. A prime example is to share the error rate per number correctly. For instance, claiming that a system predicts 1 \gls{fp} per 10,000 predictions. However, such an approximation can be hazardous, for it is inherently misleading. To demonstrate this, we ask the following questions. \emph{Does this claim hold true for different demographics?} \emph{Does this rating depend on the types of faces - does it hold for all males or all females?} Thus, if we set our system according to a desired \gls{far}, would the claim be fair regardless of demographics (\ie subgroups) of population.

The aforementioned questions were central to our previous work~\cite{robinson2020face}. We found the answer to these questions to be clear - \emph{No, the report FAR is not true when analyzed per subgroup}. We found when comparing the \gls{far} values (\ie the reported-to-the-actual), the values drastically deviate from the reported average when the score threshold is fixed for all subgroups. Furthermore, demographic-specific thresholds, meaning an assumption that demographic information is known prior, proved to mitigate the problem. However, prior knowledge of demographic, although plausible (\eg identifying a known subject on a \emph{black list}), it is a strong assumption that limits the practical uses for which it could be deployed. To extend our prior work, we propose a debiasing scheme to reduce the differences between the reported and actual. In other words, we set out to make the claim in reported false rates to be fair across all involved demographics.

\glsunset{tar} \glsunset{far}
\subsubsection{Metrics and settings}
\gls{tar} and \gls{far} are used to examine the trade-off in the confusions that is dependent on the choice of threshold discussed earlier. Specifically, we look at actual \gls{tar} scores as a function of desired \gls{far}. Furthermore, we compute the following metric, the percent difference of the true and reported \gls{far} values (\ie an average score is targeted). So we ask, \emph{how well do the different subgroups compare to that average?} Specifically,

\begin{equation}
    \text{\% Error}(l) = (\frac{\text{\gls{far}}(l)_{\text{reported}} - \text{\gls{far}}(l)_{\text{actual}}}{\text{\gls{far}}(l)_{\text{reported}}})*100\%
\end{equation}

\subsubsection{Analysis}
The proposed balances results in a way that significantly boosts the \gls{tar} at a given \gls{far}. The percent difference between in reported to actual \gls{far} score implies a more fair representation has been acquired \figref{chap:bias:fig:page1-teaser-barplot}: left-most are the percent differences using a single threshold and the right-most is using a variable threshold (\ie results of~\cite{robinson2020face}). The proposed adaptation scheme did not only preserve performances, and with a slight boost (\tabref{chap:bias:tab:ethnicy-far}), but comparing the actual to the shows a smoothing out in deviations.

\begin{figure}[t!]
    \centering
        \includegraphics[width=.3\linewidth]{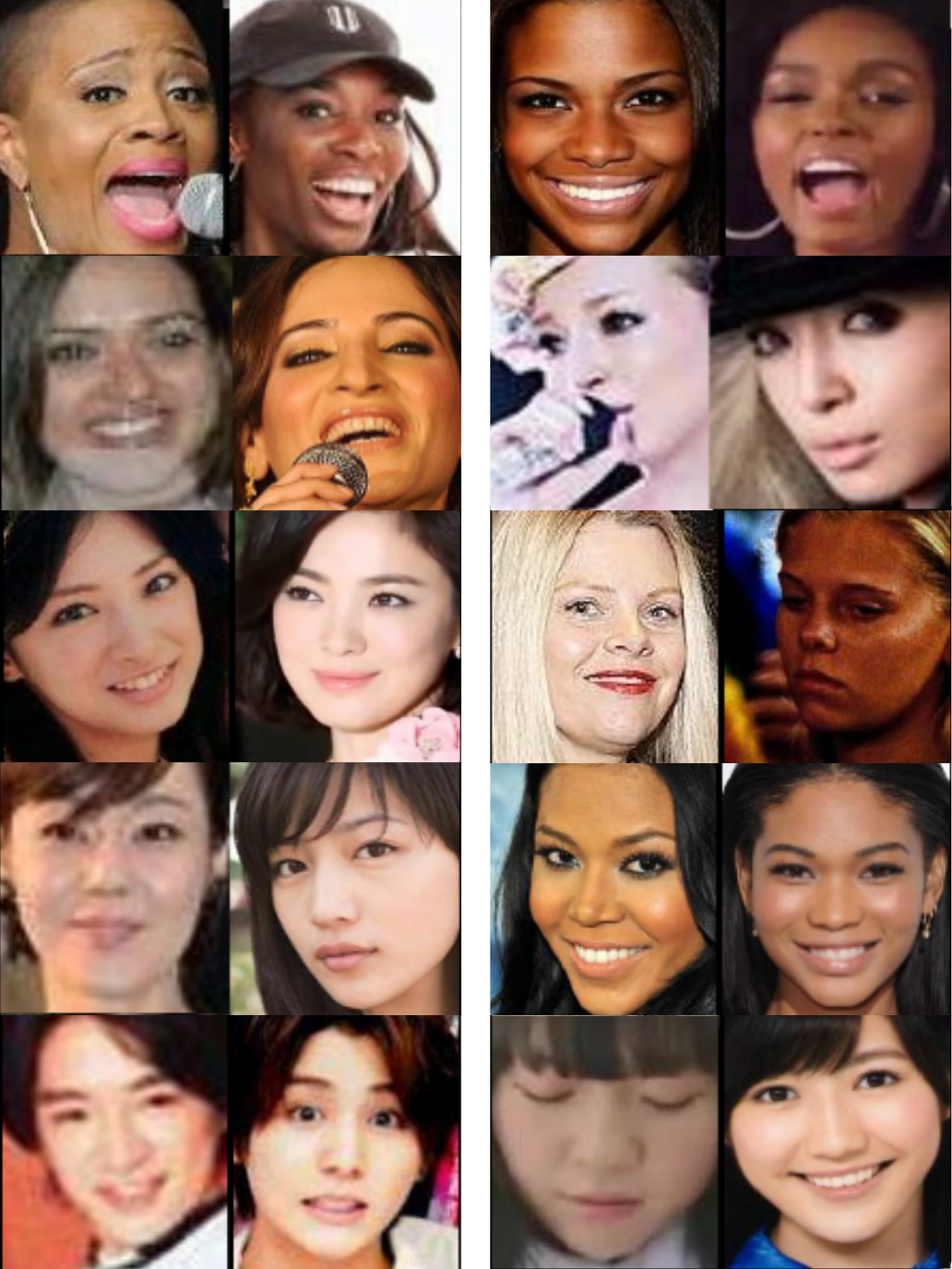}
    \caption{\textbf{Sample pairs of hard positives.} Pairs incorrectly classified by the baseline and correctly matched by the proposed.}\label{chap:bias:fig:hardpositives}
\end{figure}

Several hard positives that were incorrectly classified by the baseline but correctly identified by the proposed are shown in \figref{chap:bias:fig:hardpositives}. The set was selected for having scores closest to the global threshold of the baseline. Also, the sample pairs shown were correctly matched by the proposed. Notice the quality of one or more of the faces in each pairs is often low-resolution; additionally, extreme pose differences between faces of each pair also is common. These challenges are overcome by the proposed scheme - mitigating the issue of bias boosts results, and there displayed are several of the pairs that went from falsely being rejected to correctly being accepted.

\begin{table}[t!]
\glsunset{tar}
\glsunset{far}
\centering
\caption{\textbf{True Acceptance Rate (FAR) for various False Acceptance Rate (FAR)}. \gls{tar} scores for a global threshold (top), the proposed debiasing transformation (middle), optimal threshold (bottom). Higher is better. The standard deviation from the average is shown to demonstrate the standard error comparing the reported (\ie average) to the subgroup-specific scores. The proposed recovers most of the loss from using a global threshold rather than a per-subgroup threshold.}\label{chap:bias:tab:ethnicy-far} 
\scalebox{0.65}{
\begin{tabular}{l c c c c c}
     \gls{far} & 0.3 & 0.1 & 0.01 & 0.001 & 0.0001\\\midrule
    \multirow{3}{.1mm}{\textbf{\gls{af}}} &0.990 & 0.867 & 0.516 & 0.470 & 0.465\\[-4pt]
        &0.996 & 0.874 & 0.521 & 0.475 & 0.470\\[-4pt]
        &1.000 & 0.882 & 0.524 & 0.478 & 0.474\\[-1pt]
    \multirow{3}{3mm}{\textbf{\gls{am}}} &0.994 & 0.883 & 0.529 & 0.482 & 0.477\\[-4pt]
        &0.996 & 0.886 & 0.531 & 0.484 & 0.479\\[-4pt]
        &1.000 & 0.890 & 0.533 & 0.486 & 0.482\\[-1pt]
    \multirow{3}{3mm}{\textbf{\gls{bf}}} &0.991 & 0.870 & 0.524 & 0.479 & 0.473\\[-4pt]
        &0.995 & 0.875 & 0.527 & 0.481 & 0.476\\[-4pt]
        &1.000 & 0.879 & 0.530 & 0.484 & 0.480\\[-1pt]
    \multirow{3}{3mm}{\textbf{\gls{bm}}} &0.992 & 0.881 & 0.526 & 0.480 & 0.474\\[-4pt]
        &0.995 & 0.886 & 0.529 & 0.483 & 0.478\\[-4pt]
        &1.000 & 0.891 & 0.532 & 0.485 & 0.480\\[-1pt]
    \multirow{3}{3mm}{\textbf{\gls{if}}} &0.996 & 0.881 & 0.532 & 0.486 & 0.481\\[-4pt]
        &0.998 & 0.883 & 0.533 & 0.487 & 0.483\\[-4pt]
        &1.000 & 0.884 & 0.534 & 0.488 & 0.484\\[-1pt]
    \multirow{3}{3mm}{\textbf{\gls{im}}} &0.997 & 0.895 & 0.533 & 0.485 & 0.479\\[-4pt]
        &0.998 & 0.897 & 0.534 & 0.486 & 0.480\\[-4pt]
        &1.000 & 0.898 & 0.535 & 0.486 & 0.481\\[-1pt]
    \multirow{3}{3mm}{\textbf{\gls{wf}}} &0.988 & 0.878 & 0.517 & 0.469 & 0.464\\[-4pt]
        &0.992 & 0.884 & 0.522 & 0.472 & 0.468\\[-4pt]
        &1.000 & 0.894 & 0.526 & 0.478 & 0.474\\[-1pt]
    \multirow{3}{3mm}{\textbf{\gls{wm}}} &0.989 & 0.896 & 0.527 & 0.476 & 0.470\\[-4pt]
        &0.996 & 0.901 & 0.530 & 0.479 & 0.474\\[-4pt]
        &1.000 & 0.910 & 0.535 & 0.483 & 0.478\\[-1pt]
        \midrule
    \multirow{3}{3mm}{\textbf{Std. Dev.}} &0.030	&0.010	&0.006	&0.006&	0.006\\[-4pt]
        &0.002&	0.009&	0.005&	0.005&	0.005\\[-4pt]
        &0.000&	0.010&	0.004&	0.004 &	0.004\\[-1pt]
    \midrule
    \multirow{3}{3mm}{\textbf{Avg.}} &0.992 & 0.881 & 0.526 & 0.478 & 0.473\\[-4pt]
        &0.998 & 0.886 & 0.528 & 0.481 & 0.476\\[-4pt]
        &1.000 & 0.891 & 0.531 & 0.483 & 0.479\\[-1pt]
\end{tabular}}
\glsreset{tar}
\glsreset{far}
\end{table}

\subsection{Privacy preserving experiment} 
As mentioned, our objective was two-fold. First, we aimed to preserve identity information while debiasing the facial features, as demonstrated in the prior experiment. But then, secondly, our objective function injected the reverse gradient as part of the loss to force the embedding to be unable to classify subgroups (\ie demographics). In other words, another benefit of the proposed debiasing scheme is that it rids the facial encodings of demographic information. This, in itself, is of interest in problems of privacy and protection - ideally, face encodings, which often are the only representation of face information available at the system level, will not include attribute information, like gender or ethnicity. Hence, we aim for the inability to learn subgroup classifiers on top of the features as a means to show the demographic information has been reduced.

To show the effectiveness of the proposed in removing subgroup information, we train a \gls{mlp} to classify subgroups on top of the features. We can then measure the amount of information present in the face representation~\cite{acien2018measuring}.

The \gls{mlp} designed in Keras as follows: three fully-connected layers of size 512, 512, and 256 fed into the output fully-connected layer (\ie size 8 for the 8 subgroup classes). The first three layers were separated by ReLU activation and drop-out~\cite{srivastava2014dropout} (\ie probability of 0.5) while only dropout (again, 0.5) was place prior to the output softmax layer. A categorical crossentropy loss with Adam~\cite{kingma2014adam} set with a learning rate of 1e-3 was used to train.

\subsubsection{Metrics and settings}
We will examine the overall accuracy of the subgroup classifiers by use of a confusion matrix. Specifically, we will look at how often each subgroup was predicted correct and, when incorrect, the percentage it was mistaken for the others. The confusion was generated by averaging across the five folds.

Besides accuracy and confusions, we examine the precision and recall for each of the subgroups, along with the overall average. Precision, a measure of time correct when the prediction assumed to be true, is calculated as follows:
\begin{equation}
\text{P}(l)=\frac{\text{TP}}{\text{TP}+\text{FP}},
\end{equation}
\noindent where the \gls{ap} is the mean of all subgroups $l\in L$ (\ie $|L| = {P_L}$) for a given \gls{tpr}.

The recall R, the ratio of the number of predicted-to-actual positive samples, is found as
\begin{equation}
\text{R}(l)=\frac{\text{TP}}{\text{TP}+\text{FN}}.
\end{equation}

This compliments the confusion by allowing the specificity and sensitivity of the subgroups to also be examined. Nonetheless, there are inherent trade-offs between P and R. This motivates the $F_{1}$-score~\cite{jeni2013facing}, which fuses P and R as the harmonic mean, 

\begin{equation}
    \text{F}_1 = 2*\frac{P*R}{P+R}.
\end{equation}

\begin{table}[t!]
\centering
    \caption{\textbf{Subgroup classification results.} The baseline and proposed are on the left and right columns, respectively. Note that the columns on the right have lower scores as intended.}
    \label{table:precrec}
    
 \begin{tabular}{r|cccccccc}
     &\multicolumn{2}{c}{\textbf{Precision}}&& \multicolumn{2}{c}{\textbf{Recall}}&& \multicolumn{2}{c}{\textbf{F1}} \\
     \cline{2-3}\cline{5-6}\cline{8-9}
     AF &  0.962 &  0.734 &&  0.927 &  0.852&&0.943&0.788\\
{AM} &  0.864 &  0.707 &&  0.974 &  0.730 &&0.915&	0.717\\
BF &  0.940 &  0.655 &&  0.924 &  0.644 &&0.932&	0.647\\
BM &  0.961 &  0.644 &&  0.962 &  0.668 &&0.961&	0.653\\
IF &  0.961 &  0.641 & & 0.935 &  0.649 &&0.948&	0.644\\
IM &  0.898 &  0.519 &&  0.902 &  0.589 &&0.898&	0.550\\
WF &  0.934 &  0.554 &&  0.970 &  0.547 &&0.951&	0.549\\
WM &  0.943 &  0.524 & & 0.848 &  0.317&&0.892&	0.392\\\midrule
Average &  0.933 &  0.622 &&  0.930 &  0.624 &&0.930&	0.617\\
\end{tabular}
\end{table}

\begin{figure}[t!]
    \centering
    \begin{subfigure}[t]{.55\linewidth}
        \includegraphics[width=\linewidth]{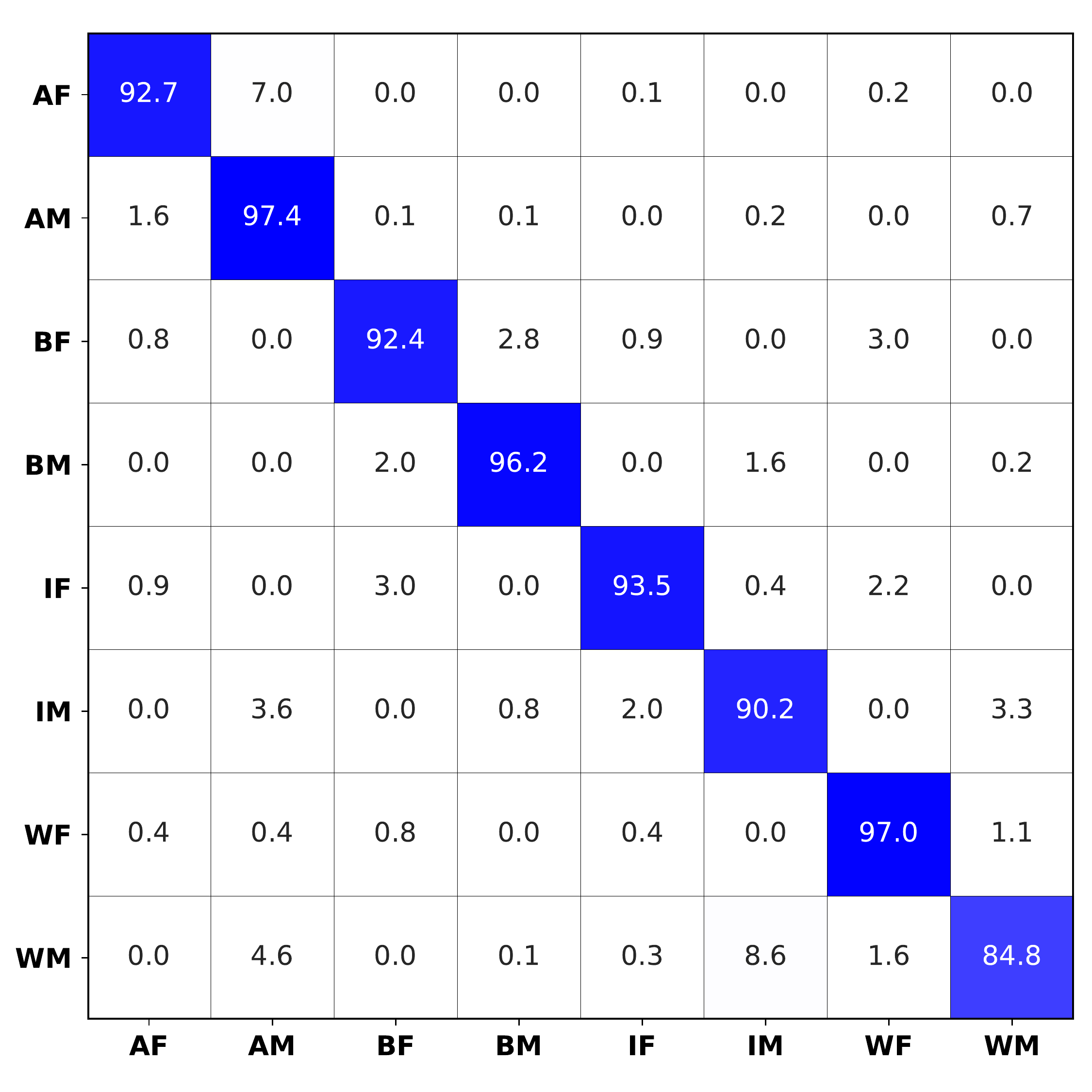}
        \caption{Baseline subgroup classifier.}
        \label{chap:bias:fig:a}
    \end{subfigure}%
    \\
    \begin{subfigure}[t]{.55\linewidth}
        \includegraphics[width=\linewidth]{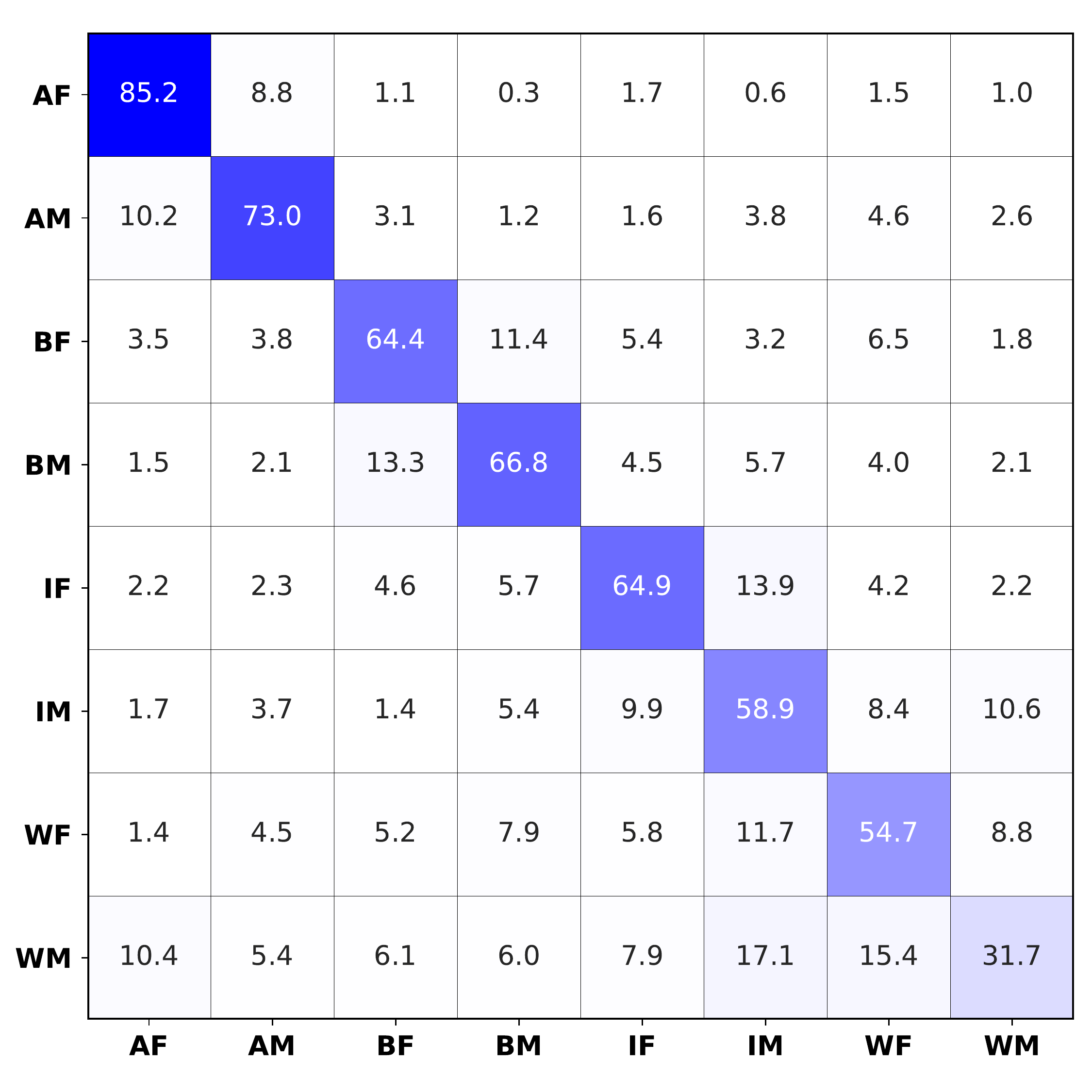}
        \caption{Our subgroup classifier.}
         \label{chap:bias:fig:b}
    \end{subfigure}
    \caption{\textbf{Subgroup confusion matrix.} Comparison of accuracy in classifying and misclassifying the subgroups. Notice the (b) performs significantly worse than (a) as intended.}
    \label{chap:bias:fig:confusion}
\end{figure}

\subsubsection{Analysis}
We demonstrated that identity knowledge is preserved (\tabref{chap:bias:tab:ethnicy-far}), and now we show the other benefits in privacy. The results clearly show that the privacy preserving claim is accurate, leading to a 30\% drop in the ability to predict gender and ethnicity from the encodings (\tabref{table:precrec}). 
Hence, predictive power of all subgroups dropped significantly. Furthermore, the drop in performance is sufficient enough to make the claim that the predictions are now unreliable. Honing in on the specifics, it is interesting to note that the subgroups for which the baseline were most in favor of are hindered the most from the debias scheme. In other words, WM and WF drop the most, while the AM and AF drop the least. All the while the same trends in confusions propagate from the baseline to the proposed results. For instance, WM are mostly confused as IM originally, and then again in the case for the debiased features. The same holds for the opposite sex in all cases.

Next, we examine the confusions for the different subgroups before and after debiasing the face features (\figref{chap:bias:fig:confusion}). As established, the baseline contains more subgroup knowledge-- a model can learn on top. When trained and evaluated on \gls{bfw}, the baseline performs better on F subgroups. This differs from the norm where M are a majority of the data. To the contrary, the WM are inferior in performances to all subgroups in either case.

\subsection{Ablation study}
\begin{figure}[t!]
    \centering
        \includegraphics[width=.5\linewidth]{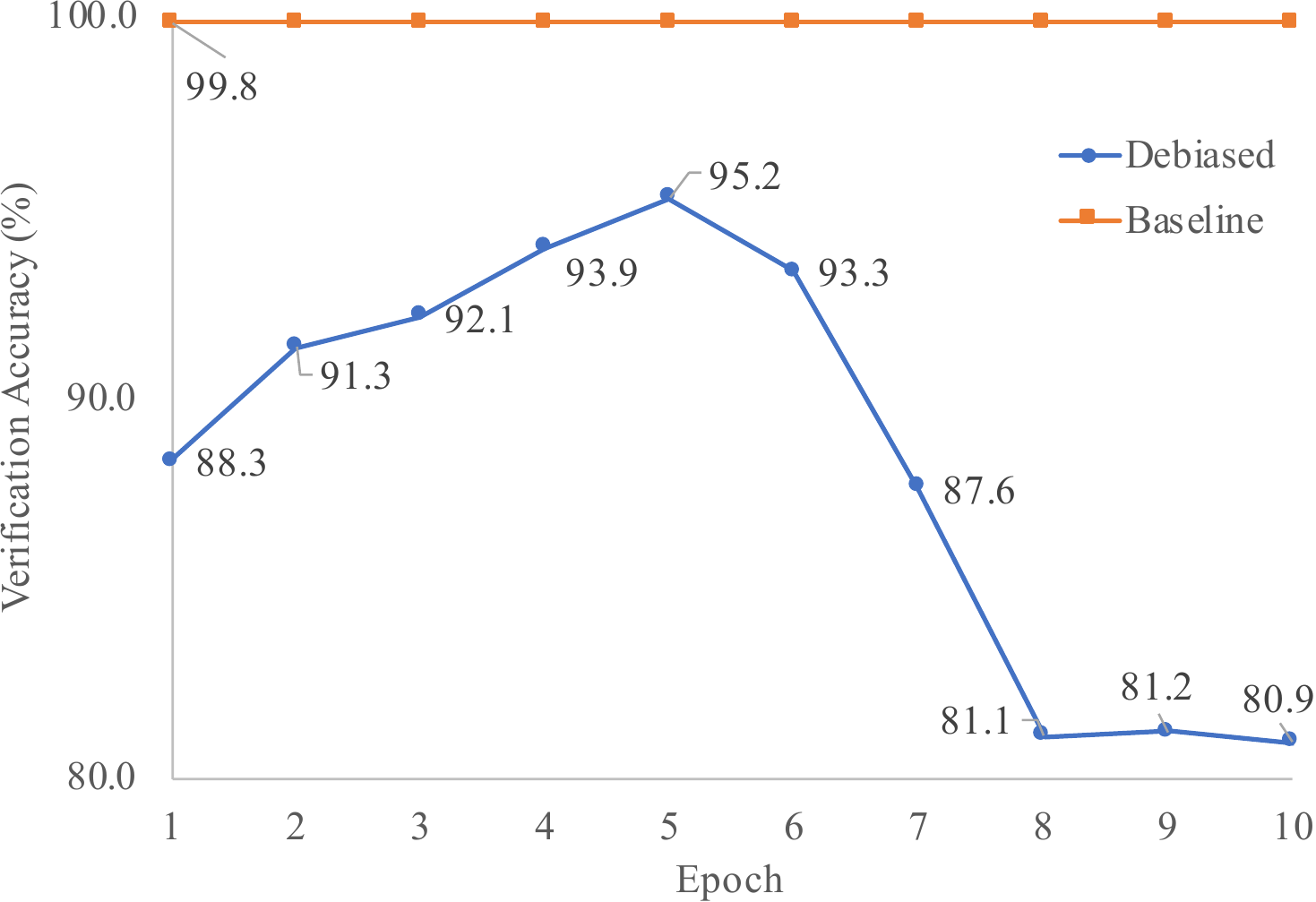}
    \caption{\textbf{Accuracy on LFW benchmark.} The proposed approaches the performance of the baseline before dropping off.}\label{chap:bias:fig:lfwvalidation}
\end{figure}

To check the effectiveness of the proposed scheme we train M using the entire \gls{bfw} dataset and deploy on the well-known LFW benchmark. We note that the training dataset that we employ is significantly smaller than that used by \gls{soa} networks trained to achieve high performance on LFW, which employ the MS1MV2 dataset, which contains 5.8M images of 85k identities. By contrast, even though we initialize our network starting with features learnt on MS1MV2, we train on a small dataset of 20k images of 800 subjects, which is two orders of magnitude smaller. Although we train  The current \gls{soa} with 99.8\% verification accuracy, while the proposed scheme reaches its best score of 95.2\% after 5 epochs before dropping off and then leveling out around 81\% (\figref{chap:bias:fig:lfwvalidation}). Clearly, the benefits of privacy and debiasing are hindered on unbalanced data (\ie LFW is made up of about 85\% WM). Furthermore, we optimized M by choosing the epoch with the best performance prior to the drop off. Future steps could be improving the proposed when transferring to unbalanced sets with a way to detect the optimal training settings.

\glsreset{bfw}	
\section{Discussion}\label{chap:bias:sec:conclusions}
We introduced the \gls{bfw} dataset with eight subgroups (\ie different gender and ethnicity) for which data is split evenly across. With this, we provide evidence that the subgroups we chose and formed is meaningful, \ie the FR algorithm rarely makes mistakes across subgroups. We used \textit{off-the-shelf} CNNs, hypothesizing that these SOTA CNN suffers from bias because of the imbalanced train-set. Once established that the results do suffer from problems of bias, we observed that the same threshold across ethnic and gender subgroups leads to differences in the \gls{fpr} up to a factor of two. Also, we clearly showed notable percent differences in ratings across subgroups. Furthermore, we ameliorate these differences with a per-subgroup threshold, leveling out \gls{fpr}, and achieving a higher \gls{tpr}. We hypothesized that most humans grew among more than their own demographic and, therefore, effectively learn from imbalanced datasets. In essence, a human evaluation validated that humans are biased, as most recognized their personal demographic best. This research, along with the data and resources, are extendable in vast ways. Thus, this is just a sliver of the larger problem of bias in ML.

We show a bias for subgroups in \gls{fr} caused by the selection of a single threshold. Previously, a subgroup-specific threshold was proposed as a solution for the case in which such knowledge (\ie subgroup information) is accessible prior. Inspired by works in feature alignment and domain adaptation we propose a scheme to mitigate the bias problem. We learn a lower dimensional mapping that preserves identity and removes knowledge of subgroup. The encodings balance the performance across subgroups, while boosting the overall accuracy. Using a single threshold, the difference between actual and average performance across subgroups is reduced. Furthermore, the resulting encodings hold reduced knowledge of subgroups, increasing privacy. 



\chapter{Discussion}\label{chapter:discussion}
We now wrap-up the dissertation with a discussion on the broader impacts, future directions in practice and in research, and, finally, a concluding section. To keep things focused on the main topic, the closing discussion is on topics of automatic kinship understanding and modeling. Hence, the other subtopics were handled in a more open and close manner (\ie provided as preliminary or secondary knowledge in nearly standalone chapters), we chose to omit from this final chapter as to appropriately conclude the majority of the research tailored to the kin-based topics.

\section{Broader impacts}
We believe that, collectively, our greatest contribution for automatic kinship recognition was the labeled data, \ie the \gls{fiw} dataset.\footnote{More information, downloads, and publications are on the project page, \href{https://web.northeastern.edu/smilelab/fiw/}{https://web.northeastern.edu/smilelab/fiw/}.} Also, the task evaluations released with clearly defined problem statements, task protocols, and benchmark reports provide additional structure for researchers to follow as we continue to push the envelope of capabilities and technology developed as solutions for the problems. Thus, by laying the groundwork for others to get started and incorporate modern-day, data driven modeling ideas, we hope the trend continues in that attention for this problem continues to climb as our automatic face analysis and modeling capabilities progress as well.

The fourth \gls{rfiw} gained fair attention--
T1 (verification) saw the most; T2 (tri-subject) and T3 (search and retrieval), both supported for the first time, are more complex and practically motivated than the classic T1. The broader impact spans greater than current tasks in application (\eg generative-based tasks~\cite{gao2019will, ozkan2018kinshipgan}) and experimental settings (\eg with privacy a concern~\cite{mingaaai2020}). \gls{rfiw} met the difficulty and practicality of today; the question how best to formulate the problem remains an open research question. As such, this survey aims to provide a stronghold on the laboratory-style evaluations as seen appropriate in the modern day.

\section{Future work}\label{sec:next-steps}
It is an exciting and opportune time for kin-based problems for researchers and practitioners alike. For starters, there is a lot of room for improving \gls{sota}, and even the experiments (\ie design, purpose, and extent). For instance, incorporating additional label types (\ie other soft attributes like expression, age, and ethnicity), different data splits and protocols (\eg given a father, daughter, and grandparents from the side of the mother, determine the mother), and practical use-cases (\eg automate family photo-album creation). Generative-based tasks also hold promise in directions to take next: whether improved predictive capability of a child's face - provided a pair of parents, or a more fine-grained view of predicting any node in a family tree - provided samples of all other family members - then the room for improvement and potential for growth is furthered.

\paragraph{Fairness.} A few recent attempts have been made by researchers to address fair AI and transparency in kinship understanding. For instance, the latest version of \gls{rfiw}, supplemental to this survey, \gls{fiw} is now supported with a \emph{datasheet}: ``datasheets for datasets''~\cite{gebru2018datasheets}.\footnote{The datasheet for \gls{fiw} is available online, \href{https://web.northeastern.edu/smilelab//fiw/fiw_ds.pdf}{https://web.northeastern.edu/smilelab/fiw/fiw\_ds.pdf}.} The motivation of datasheets is to promote transparency and, thus, to minimize the doubt from unknown biases that come and are inherited by publicly available data resources.
Specifically, \emph{datasheets} completely spec-out the data (\eg motivation, composition, collection process, preprocessing, updates, legal and ethical considerations). There are other methods for transparency that have been recently proposed with a similar motivation as ``datasheets for datasets'', such as \emph{fact sheets}~\cite{arnold2019factsheets} and \emph{model cards}~\cite{DBLP:journals/corr/abs-1810-03993}. Nonetheless, we found that the format and motivation of ``datasheets for datasets'' as the best for \gls{fiw}. So, this was used to record and archive data specifications. 

\paragraph{Privacy.} 
As is the case for many \gls{ml} tasks, privacy has motivated researchers. Recently, Kumar~\etal proposed using a \gls{gnn} to first achieve \gls{sota} in family classification, and to then add noise to encrypt the data, and demonstrating that a variant of the model safely encapsulates the learned knowledge (\ie an ability to accurately deceiver)~\cite{mingaaai2020}.

\paragraph{Social and cultural.}
A near radical piece of its time, Goode~\cite{goode1963world} surveyed family structure as more of a complex system than the `conjugal family form' of many traditional cultures (\eg Western, Chinese, Arab). Hence, we currently look at sets of persons as being either related or not related-- this does not account for the realistic setting that would be faced in the modern world. For instance, tri-subject pairs are structured by using a parent pair and a child (\ie using evidence of both parents)-- a scenario that is certainly a step in the right direction, as parents are often inferred through records on marriage and offspring. However, families are dynamic in many modern cultures-- step siblings and parents are common. Considering the setting of tri-subject: what happens if the father is true, but the mother is not, or vice versa-- a concept that propagates to all levels of the problem, and especially when considering complete family trees with connections to in-laws. Thus, the problem remains: how to best weight (\ie fuse~\cite{zhang2019feature}) different relationship types. Even simple questions have soft, varying solutions ~\cite{BREDART1999129} like \emph{Do we look more like our father?} 

\paragraph{Feature fusion.} Still today, the underlying question remains. \emph{How to best fuse prior knowledge?} For instance, in tri-subject verification, the fusion of the features from the two parents. Flipping this very problem around (\ie given parents, generate the child's face), the question of feature fusion is still prominent. Looking ahead at attempts to solve the fine-grained problem of populating family trees, regardless if viewed as discriminate or generative, the question remains: \emph{how to best leverage prior knowledge of additional family members relatively of different types and degrees?}

Although the number of methods is great-- whether metric-learning, deep features, a variant of both-- most recent attempts only differ in the broad sense. Bottom-line, successes in all tasks have been tributes of systems based on a Siamese network(s) that encodes inputs from image-to-feature space. The feature space learned typically differs in the point and method of fusion. Specifically, paired samples are usually split evenly (\ie the number of pair-types of type \emph{KIN} and \emph{NON-KIN} for each relationship type is split \emph{fifty-fifty}). Provided a Siamese network, often pre-trained on auxiliary face recognition dataset, act as face encoders. In order to transform from feature-to-score space, either a metric, fusion technique, or both are applied-- this tends to be where methods differ, yet the same conventional coarse system holds (\ie \figref{fig:siamese}). In summary, it is the Siamese net to encode faces, followed by some means of feature-fusion that are stove-piped to a metric or learning objective. Hence, some relevant aspects of such a system produce current \gls{sota} from which we had drawn conclusions, and especially in identifying research trends and open issues. We consider the most relevant among aspects for achieving effective systems as follows: (1) effective method for fusion; (2) representation that considers the relationship's direction; (3) detecting other attributes (\eg age and gender) and knowledge of the higher-level scene (\eg face detected in picture with car styles that hint the picture was taken in the 1970s).

\paragraph{Multimodal data.} Let us consider other signals that can define visual data; let us consider other label types for faces that could also enhance performance. For instance, expressions and mannerisms are often similar for parent and child (\eg \emph{they have the mother's smile}). More complex dynamics for individual expressions and mannerisms can be effectively captured in video data~\cite{kollias2020analysing}. Hence, added knowledge that complements the visual information has proven useful for boosting kin-based recognition ratings~(\eg 3D facial images\cite{vijayan2011twins}, voice~\cite{wu2019audio}, MM~\cite{robinson2020familiesinMM}).

\paragraph{Family synthesis.}
The existing technology for synthesizing family members is still immature and generalization remains unsolved. A system should accept one-to-many members of a family tree and synthesize the appearances of the desired relative type. Still, many questions remain: how to fuse, handle dynamic inputs, the optimal way to reason about a family tree, and more.

\section{Conclusion}
Upon completing the dissertation, and looking back at years of research, we see decent progress in automatic kinship recognition, along with an increase in attention, an improvement in the methodology, advancement in available data and resources, and improved benchmarks (\ie both in protocol and proposed \gls{sota}). This dissertation contributed in each of the aforementioned ways: we improved models for kin-specific face-based recognition tasks; we attracted researchers by organizing several workshops, challenges, tutorials, \etc; we released our large-scale \gls{fiw} data collection, along with the data splits and benchmark scores for public use. Besides discriminate tasks (\ie recognition), for which we supported various tasks to span different use-cases, we also explored generative-based problems. Specifically, we were the first to attempt the appearance of off-spring from a pair of parents as the input (\ie not a single parent, but with two-to-one mapping so more information was available for inference). Also, we teased the use of multimedia in kinship recognition with our \gls{fiwmm}--the benefits of the added modalities were clearly demonstrated. Finally, we recently delivered a comprehensive survey on the topic as means of establishing a stronghold on the state of technology after a decade of research-- a single resource to reference for experimental protocols, benchmark ratings, and a high-level view of the latest-and-greatest methods. We believe the research for this dissertation was well spent for the advancing of automatic kinship understanding.
\bibliographystyle{IEEEtran}

\bibliography{bib/thesis}


\printindex

\end{document}